\documentclass[10pt]{article}
\usepackage[preprint]{tmlr}

\usepackage[utf8]{inputenc}
\usepackage[T1]{fontenc}
\usepackage{hyperref}
\usepackage{url}
\usepackage{booktabs}
\usepackage{amsfonts}
\usepackage{nicefrac}
\usepackage{microtype}
\usepackage{xcolor}
\usepackage{graphicx}
\usepackage{subcaption}
\usepackage{amssymb}
\usepackage{wrapfig}
\usepackage{enumitem}
\usepackage{placeins}

\usepackage{amsmath,amsfonts,bm}

\def\eqref#1{equation~\ref{#1}}

\def\ceil#1{\lceil #1 \rceil}

\def\1{\bm{1}}

\def\vh{{\bm{h}}}

\def\vx{{\bm{x}}}

\def\vz{{\bm{z}}}

\DeclareMathAlphabet{\mathsfit}{\encodingdefault}{\sfdefault}{m}{sl}
\SetMathAlphabet{\mathsfit}{bold}{\encodingdefault}{\sfdefault}{bx}{n}

\def\sD{{\mathbb{D}}}

\def\sX{{\mathbb{X}}}
\def\sY{{\mathbb{Y}}}

\usepackage{tcolorbox}
\tcbuselibrary{skins}
\definecolor{takeawayblue}{RGB}{210, 235, 255}
\definecolor{takeawayborder}{RGB}{100, 160, 220}
\newtcolorbox{takeawaybox}[1][]{
  colback=takeawayblue,
  colframe=takeawayborder,
  fonttitle=\bfseries,
  title=Takeaway,
  rounded corners,
  #1
}
\newtheorem{definition}{Definition}

\title{Sometimes less is more: Investigating inductive biases of MLPs and Transformers on noisy k-parity features}

\author{\name Alasdair Paren \\
      \addr Chalmers University of Technology}

\begin{document}

\maketitle

\begin{abstract}
We empirically investigate how MLPs and Transformers differ when trained on data sets containing one or more features encoded by noisy k-parity.
Noisy k-parity gives us hard synthetic data sets in which we can control the predictivity and availability of every feature.
Across a range of data distributions, data scales and model sizes, we find large differences in the training dynamics and inductive biases of the two architectures: Transformers are more robust to noisy features but struggle to learn complex, low-availability features, whereas MLPs show the opposite trend and improve steadily with more data.
We also find that, under a fixed training budget, increasing the number of unique training examples can slow Transformer optimisation enough to reduce final accuracy, so that a smaller data set performs better.
Training-accuracy results show this is an optimisation-speed effect; it shrinks, but does not vanish, when the budget is scaled with the data set size.
By giving precise control over feature predictivity and availability, this testbed offers a clear starting point for understanding when and why Transformers and MLPs learn differently, and for designing data and training budgets that exploit those differences.
\end{abstract}

\section{Introduction}\label{sec:introduction}

Neural network (NN) architectures exhibit different training dynamics and inductive biases \citep{battaglia2018relational, Goodfellow2016}, which influence the types of functions learnt throughout training \citep{teney2024neural,mingard2025characterising}.
While most NNs can approximate most functions \citep{cybenko1989approximation, HORNIK1989359, lu2017expressive}, these biases can affect how many optimiser iterations a model requires to achieve a good performance on a given training set \citep{neyshabur2017exploring, zhang2016understanding, zhang2021understanding}, as well as the generalisation performance away from the training data \citep{soudry2018implicit}.
Better understanding these biases is key to guide future research and design choices in ML systems \citep{goyal2022inductive, battaglia2018relational}.

One of the best understood inductive biases is the Simplicity Bias (SB) where, due to common architectural choices \citep{teney2025we} NN architectures are biased toward learning ``simple'' functions. 
This phenomenon acts like Occam's razor and can be a benefit or drawback depending on the relative complexity of the desired function to be learnt from the data. 
The SB can be highly beneficial if the desired function is the simplest that minimises the loss over the training data. 
However, if the problem is underspecified and there are simpler predictive features\footnote{We use ``feature'' to mean a characteristic of the input.}, commonly named ``shortcuts'', that do not generalise to the test environment, the SB can result in poor performance. 
Indeed, there are many examples of NN trained using ``shortcuts'' \citep{geirhos2020shortcut} to make predictions instead of some desired attributes or ``labelling function'' \citep{puli2023don, kirichenko2022last}. 
Moreover, it has been shown that parameter settings corresponding to ``simple functions'' occupy the majority of the solution space\footnote{The subset of parameter space which achieves near optimal training loss.} \citep{scimeca2021shortcut}.
However, there are many differing definitions over what features are ``simple'' \footnote{In Section \ref{subsec:predictivity} we discuss the differing prior notions of simplicity in detail.}. 
To sidestep this issue \citet{hermannfoundations} introduced the notion of availability which is a measure of how easily a feature can be learnt by an NN, where more available features are preferred.
Due to this implied preference ordering, if a data set contains multiple equally predictive features it is likely that the encoded function at test time will make most use of the most available feature. 
\begin{wrapfigure}[15]{r}{0.35\textwidth}
  \vspace{-4pt}
  \centering
  \includegraphics[width=0.33\textwidth]{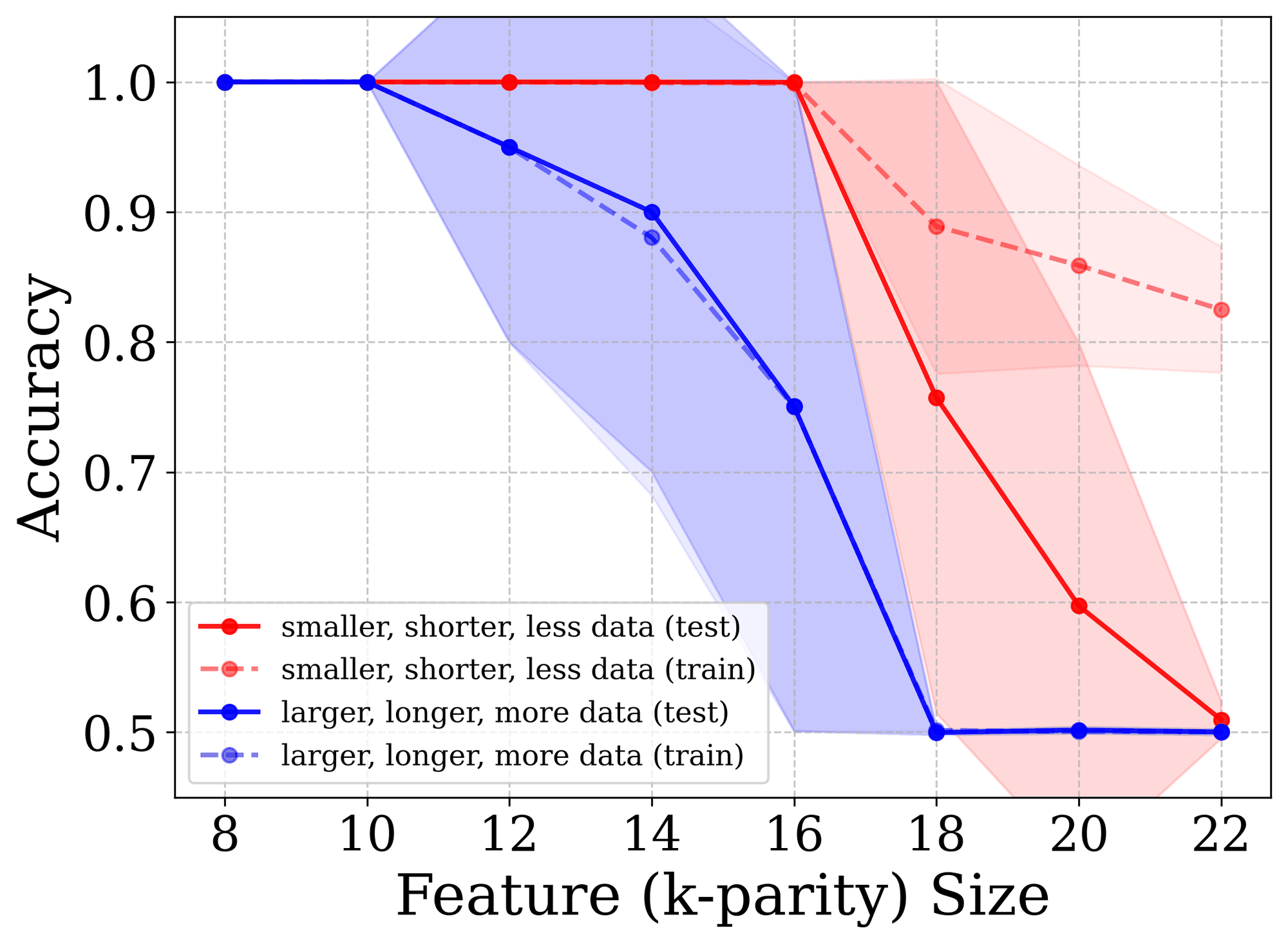}
  \caption{\small Test accuracy vs.\ k-parity size $d$.
  \textbf{Red}: $m=10$, $N=8000$, $T=10^4$.
  \textbf{Blue}: $m=20$, $N=240000$, $T=2\times10^4$.
  Scaling up all three gives worse results (Section \ref{sec:k_parity_experiments}).}
  \label{fig:pull}
  \vspace{-4pt}
\end{wrapfigure}
Thus, any features of interest with higher complexity, or lower availabilities, may not be learnt during training, as more available features are sufficient to minimise the loss.
In this way more available features can be seen as distractors, that prevent the learning of more complex features \citep{geirhos2020shortcut}.

In this manuscript we empirically investigate how the differing biases of MLPs and Transformers \citep{vaswani2017attention}, affects the learning dynamics of these two architectures, on data sets containing multiple features with specified predictabilities and availabilities.
We do this by constructing data sets that contain multiple noisy k-parity features. 
K-parity is useful as a proxy for real problems requiring axis-aligned feature learning \citep{edelman2023pareto}.
The goal of this investigation is to provide new insights on the differing inductive biases of these models on a well studied class of hard non-linear feature, showing where theory matches with practice and where it does not. 
Specifically, our contributions are as follows:
\vspace{0.6\baselineskip}
\begin{enumerate}[leftmargin=10pt]
    \item A simple framework for creating hard non-linear data sets with any number of features of specified availability and predictivity, for investigating the simplicity bias.
    Using this framework we conduct extensive experiments to explore the inductive biases of MLPs and Transformers.
    \item We show that, for data sets whose features are k-parities and under a fixed training budget, training on more i.i.d.\ samples can slow the optimisation of Transformers enough to reduce final accuracy (Figure \ref{fig:pull}).
    Training accuracy and test accuracy side by side (Table \ref{tab:main_train_test}) show that this is an optimisation-speed effect rather than a change in the achievable generalisation: at large data set sizes the training accuracy itself stays low within the budget.
    The effect persists, in weaker form, when the iteration budget is scaled with the data set size, where more data extends the frontier of learnable features while still optimising more slowly in the partially learned regime. 
    \item We show the stronger SB of Transformers means they are less prone to over-fitting and better at learning from noisy data.
    However, this also means they have difficulty in learning complex yet highly predictive features even with large amounts of training data and iterations.
    Conversely, for highly complex features MLPs with a lower SB scale much better when presented with more data.
\end{enumerate}

\section{Related Work}
\paragraph{Simplicity Bias.}
In their work Learning Qualitatively Diverse and Interpretable Rules for Classification, \citet{ross2018learning} introduced a way to identify a maximal set of distinct but accurate models for a dataset, demonstrating empirically that SGD tends to recover simpler, more interpretable classifiers. 
\citet{valle2018deep} suggested the good generalisation performance of neural networks is because the parameter-function map is biased towards simple functions. 
\citet{nakkiran2019sgd} showed that SGD-trained NNs learn functions of increasing complexity, starting with a linear model retained even to convergence. 
\citet{scimeca2021shortcut} show that Kolmogorov-simple solutions are abundant in parameter space and thus more likely found by NNs. 
More recently, \citet{bell2022simplicity} showed that SB can lead to larger performance disparities between subsets of the input distribution, and \citet{mingard2023deep} claim NNs have an inbuilt Occam's razor with a strong bias toward low Kolmogorov complexity functions. 
\citet{teney2024neural} proposed that SB depends on architectural components such as non-linearities, residual connections, and layer normalizations, and in follow-up work \citet{teney2025we} design NNs with specialised activations that instead induce a complexity bias.
\citet{shah2020pitfalls} was the first to focus on the negative consequences of SB, showing that: (i) SB can be extreme, causing networks to rely solely on the simplest feature; (ii) extreme SB exacerbates degradation under distribution shifts and adversarial attacks; (iii) SB can hinder generalisation even when simpler features are less predictive; and (iv) common mitigations like ensembles and adversarial training struggle to address it. 
Theoretically, \citet{soudry2018implicit, morwani2023simplicity} showed that ERM with the CE loss finds a maximum margin model when one exists, and \citet{puli2023don} showed this implicit bias can prevent linear models from using the label when it appears in the input, proposing losses to address this.

\paragraph{Repetition and Data Diversity.}
\citet{charton2024emergent} show that, for a fixed training budget, transformers trained on algorithmic tasks can learn faster and perform better when trained on smaller data sets with repeated examples than on larger single-use data sets. \citet{liu2026lessdata} show that small data sets can speed up training and analyse the role of repetition in detail. Our fixed-budget results are consistent with this line of work. The main differences are the setting (k-parity, where statistical-query hardness results give a precise notion of difficulty), the comparison with MLPs, and our scaled-budget experiments, which keep the expected number of passes per example constant.

\paragraph{Learning Parities with Neural Networks.}
The k-parity function has long served as a canonical hard case for gradient-based learning, with its difficulty rooted in the SQ framework of \citet{Kearns1998}, which establishes that any SQ algorithm requires exponentially many queries to learn a worst-case parity. 
The staircase property of \citep{abbe2021staircase, abbe2022merged} provides a necessary and nearly sufficient condition for SGD learnability of sparse functions on depth-2 networks, revealing that k-parity's key obstacle is its inability to decompose hierarchically into lower-order correlations. 
\citet{abbe2023sgd} characterised the saddle-to-saddle dynamics governing sequential feature acquisition, showing SGD complexity for parity matches Correlational SQ lower bounds. 
On the positive side, \citet{daniely2020learning} showed gradient-trained networks can learn parities under non-uniform input distributions that break the SQ barrier, and \citet{kou2024matching} showed two-layer networks trained with sign-SGD can solve k-sparse parity with sample complexity matching the SQ lower bound.
\citet{shoshani2025hardness} formalised the hardness of fixed parities for one-hidden-layer ReLU networks, proving exponential-in-k failure guarantees. 
\citet{edelman2023pareto} reframed the SQ lower bound as a multi-resource trade-off over data, width, compute, and randomness. 
\citet{han2025attention} showed theoretically that a single attention head can efficiently identify the relevant k coordinates of a k-parity where an equivalently parametrised MLP cannot.
\citet{barak2022hidden} demonstrated that standard architectures can learn sparse parities near the SQ limit, exhibiting sharp phase transitions at approximately $n^{O(k)}$ iterations, driven by gradual amplification of a ``Fourier gap'' invisible to the loss curve. 
This behaviour was corroborated by \citet{merrill2023tale}, which attributed the transition to rapid norm growth in a small neuron subset. 
Our work is most similar to \citet{peters2026trapped}, who investigate the simplicity bias of Transformers and LSTMs on sparse Boolean functions including k-parity.
They show that the extreme simplicity bias of Transformers makes it possible to construct data sets where they fail to learn from noisy data.
We build on this by studying how data set size and training budget interact with feature complexity, and by contrasting Transformers with MLPs under identical conditions.
Our findings on noise robustness are consistent with theirs; the dependence of Transformer optimisation speed on data diversity is, to our knowledge, not covered by their study.
Orthogonal to this body of work, we ask \emph{where} k-parity features are learnt rather than \emph{how}, examining datasets containing one or more k-parity-encoded features across different data and model scales.

\section{Preliminaries}\label{sec:preliminaries}

\paragraph{Scope.} We follow the long line of research that investigates NN training on synthetic data sets \citep{hermann2020shapes, power2022grokking, nanda2023progress, elhage2022toy, vasudeva2024transformers, tong2024mlps}.
We design data sets with a number of different features so they each have a desired predictivity (see Section \ref{subsec:predictivity}) and availability \citep{hermannfoundations}. 
In order to simplify our analysis and let us precisely control the relative availability of different features we follow in the large body of work that restricts their analyses to training of NNs on classes of boolean (or ternary functions) \citep{raz2018fast, malach2019learning,daniely2020learning,hermann2020shapes, barak2022hidden, pmlr-v180-bronstein22a, bhattamishra2023simplicity}.
Concretely we only consider binary classification data sets, where the input space is $\sX \triangleq \{-1,0,1\}^{d_{in}}$, and the label space is $\sY \triangleq  \{-1,1\}$.\footnote{In the implementation, labels are stored as $\{0,1\}$.
We use $y\in\{-1,1\}$ in the text for convenience; the two encodings are related by $y = 2g-1$.} 

\paragraph{Features.}
To formalise a precise notion of features we use the definition presented in \citet{Ilyas2019}.
According to their definition, a feature is some function of the input space to a single scalar that represents the presence or lack of a given attribute or concept.
Formally:
\begin{equation}
\mathcal{F}_{all} = \{f : \sX \rightarrow \mathbb{R}\},
\end{equation}
where $\mathcal{F}_{all}$ is the space of all features.
In practice data sets contain a continuum of different features rather than a discrete set, however, to promote understanding we construct data sets that contain a finite set $F$ of one or more carefully designed features.

\subsection{Predictivity and Availability}\label{subsec:predictivity}
We adopt the definitions of \emph{predictivity} and \emph{availability} from
\citet{hermannfoundations}. Predictivity quantifies the correlation of a given feature with the label.
\begin{definition}[Predictivity]
\label{def:predictivity}
A feature $f_i(\vx)$ has a predictivity of $\alpha_i$ over a data set $\sD_N \triangleq \{(\vx_n,y_n)\}_{n=1}^N$ given by the following relation:
\begin{equation}
    \alpha_i \triangleq P_{\{\vx,y\}\sim\sD_N}(y =\text{sign}(f_i(\vx)))
\end{equation}
In words, the predictivity of feature $f_i$ over a data set $\sD_N$ is the fraction of examples for which the sign of $f_i(\vx)$ matches the label $y \in \{-1,1\}$. 
\end{definition}

\begin{definition}[Availability]
More-available features are easier for models to extract and leverage \citep{hermannfoundations}. 
\label{def:availability} 
\end{definition}
Availability is much harder to uniquely define mathematically.
It has been shown by the extensive work on the simplicity bias that neural networks find simpler features easier to learn and thus more available.
However, there is no consensus on the definition of ``simple''.
\citet{shah2020pitfalls} define the simplicity of a function by the minimum number of linear pieces in the decision boundary that achieves optimal classification accuracy using that function.
Unfortunately, this definition is not easy to quantify outside of toy data sets.
\citet{wang2023neural} suggest low frequency features are simple and hence learnt first.
\citet{anonymous2024mitigating} define simple as those that can be learnt by models with fewer parameters.
Conversely, \citet{hermann2020shapes} look at what features convolutional neural networks (CNNs) learn first and suggest simple features are those that can be most easily decoded by linear probing before training.
\citet{morwani2023simplicity} suggest simple features can be characterised by a low dimensional projections of the inputs.
Several works \citep{scimeca2021shortcut,mingard2023deep} have suggested NN are biased toward learning Kolmogorov simple functions.
More recently \citet{fel2024understanding} introduced $\mathcal{V}$-information to quantify the computational transformations required for visual features to be extracted.
We note all these metrics correspond to some measure of the information required in describing the function and thus, in practice, outside of carefully constructed functions are likely to be highly correlated. 
In this work we use the bit-string length of a k-parity to encode the availability of a feature.

\subsection{k-parity}\label{subsec:k_parity}
In this work every feature $f_i \in \mathcal{F}$ encodes a noisy k-parity function with $k = \ceil{d/2}$, where $d$ is the length of the bit string of the domain of $f_i$. 
By increasing the size of $d_i$ we can decrease the availability of the $i^{th}$ feature making it harder to learn.
We use noisy k-parity as our feature of choice as k-parity is simple, well studied theoretically \citep{daniely2020learning, barak2022hidden, kou2024matching}, provably hard to learn \citep{Kearns1998,shalev2017failures, Blum2003, shoshani2025hardness}, and gives us easy control of both predictivity and complexity. 
Before introducing noisy k-parity we first introduce the noiseless version of k-parity.

\begin{definition}[k-parity]
k-parity is a binary function $g_d:\{0,1\}^d \rightarrow \{0,1\}$ given by the following relation:
\begin{equation}
g_d(\vx) \triangleq \langle \vx, \vh \rangle \text{ mod } 2,
\end{equation}\noindent
where $\vh \in \{0,1\}^d$ is an unknown latent vector with $||\vh||_1 = k = \ceil{d/2}$ that must be learnt via samples $(\vx_n, y_n)$ where $y_n \triangleq g_k(\vx_n)$ \citep{raz2018fast,daniely2020learning}.
In this paper we sample $\vh$ uniformly from $\{\vh \in \{0,1\}^d : ||\vh||_1 = \ceil{d/2}\}$. 
\label{def:k_parity}
\end{definition}
In other words, we ensure approximately half the elements of $\vh$ are $1$ and half are $0$.
As $d$ increases the number of possible $\vh$ vectors increases as $\binom{d}{d/2}$.
Note, in this formulation $d$ corresponds to the total number of bits, rather than the bit width of latent vector $\vh$.
One effect of this choice is approximately half of the input encodes the label and the other half is uncorrelated noise.
The k-parity learning task is effectively identifying these two subsets \citep{cornacchia2023mathematical}. 

\begin{definition}[Noisy k-parity]
Noisy k-parity is a stochastic binary function $\hat{g}_d:\{0,1\}^d \rightarrow \{0,1\}$ given by:
\begin{equation}\label{def:noisy_k_parity}
\hat{g}_d^\alpha \triangleq (\langle \vx, \vh \rangle + \text{Bernoulli}(1-\alpha)) \text{ mod } 2,
\end{equation}
where $\vh$ is sampled via the same process for k-parity.
In words, noisy k-parity returns the same label as the noiseless k-parity with probability $\alpha$ and with probability $1-\alpha$ returns the wrong label.
\end{definition}

The Predictivity of noisy k-parity is given exactly by $\alpha$ and a measure of its availability scales with the inverse of the number of bits $d$, this makes it an ideal candidate function for constructing data sets with many features where we want to have precise control of these quantities.
Thus, in this paper all features considered are simply different versions of the noisy k-parity function. 

For data sets with $F$ features; these data sets can be described by a list of $F$ tuples where each tuple contains the number of bits of each feature $d$ and $\alpha$ for a given feature.
In order to have data sets containing multiple features, non-overlapping subsets of the input space $\sX$ form the domain to different noisy k-parity functions.
For example, for a data set with only two features, $f_1, f_2$ the first $d_{in}/2$ dimensions of the input would encode $f_1$ with a noisy k-parity feature with a given $k_1$, $\alpha_1$ and $\vh_1$ and the second $d_{in}/2$ dimension would encode a different noisy k-parity feature with a different  $k_2$, $\alpha_2$ and $\vh_2$.
\footnote{Where any differences in size are accounted by padding by a fixed average value.} To create larger data sets containing even more features we simply concatenate more features together.
For the exact details of the Data Generating Process see Appendix \ref{sec:dgp}.

\section{Experiments}

\subsection{Models and Training Procedure}\label{sec:models_and_training}

We investigate two classes of models, a decoder only Transformer \citep{vaswani2017attention} which we denote (GPT), and a standard multi-layer perceptron (MLP).
These models are three blocks and three layers deep, respectively.
We run experiments at a number of model scales by increasing the width of the model or parameters and attention heads for the Transformer in each layer, by a width factor $m$, with a default value of $m=10$.
\footnote{Scaling the width of the model proved to be a lot more effective than scaling the depth in preliminary experiments.} The GPT like model has approximately 35 times the parameter count of the MLP for the same value of $m$.
We did not match the models on parameter count, FLOPs or any other single axis: no one choice of matching is unambiguously fair between architecture classes, since parameters, FLOPs, memory and wall-clock time all differ.
Instead, the two sizes were chosen so that both models exhibit interesting variation over a similar range of k-parity feature sizes, i.e.\ the regime in which neither model trivially succeeds or fails.
This is a deliberate design decision; the consequences for interpreting the comparison are discussed in Section \ref{sec:limitations}.
The two models also differ in regularisation and learning rate: the GPT model uses weight decay ($10^{-3}$), dropout (0.1), gradient clipping and a learning rate of $10^{-4}$, while the MLP uses no weight decay or dropout and a learning rate of $3\times10^{-4}$ (Appendix \ref{app:hyperparams}).

\paragraph{Input format.} The MLP receives the input vector $\vx\in\{-1,0,1\}^{d_{in}}$ directly.
The GPT model treats the same input as a sequence of $d_{in}$ tokens, one per input coordinate, each taking one of three values (Appendix \ref{app:hyperparams}).
Classification uses the output at the final sequence position, which under the causal mask attends to every earlier token.
Both models therefore have access to the complete input, and the comparison is not confounded by information restrictions from the causal mask.
All models were trained with the same procedure, unless stated otherwise.
The MLP was trained with Adam \citep{Kingma2015} and the GPT model with AdamW, in both cases for $T = 10^4$ iterations (the \emph{fixed budget}) or until the training accuracy hit 99.99\%.
Where stated, the budget is instead \emph{scaled} with the training set size as $T=1.25N$.
Throughout, $N$ in this scaling rule denotes the number of training examples, e.g.\ a data set of $N=8{,}000$ training examples and $T=10^4$ under the scaled budget.
With a batch size of 256 this gives an expected 320 passes over each training example, independent of $N$.
An ``infinite'' data set ($N=\infty$) means every batch is sampled fresh from the DGP. 
A \emph{noiseless} test set is one in which all features are fully predictive.
For all experiments at least 5 runs with different seeds were performed and the results averaged.
Where relevant: shaded areas show one standard deviation calculated across these runs.
For details of the exact models and hyperparameters see Appendix \ref{app:hyperparams}.

\begin{figure*}[ht]
    \centering
    \begin{subfigure}[]{0.24 \textwidth}
        \includegraphics[width=\textwidth]{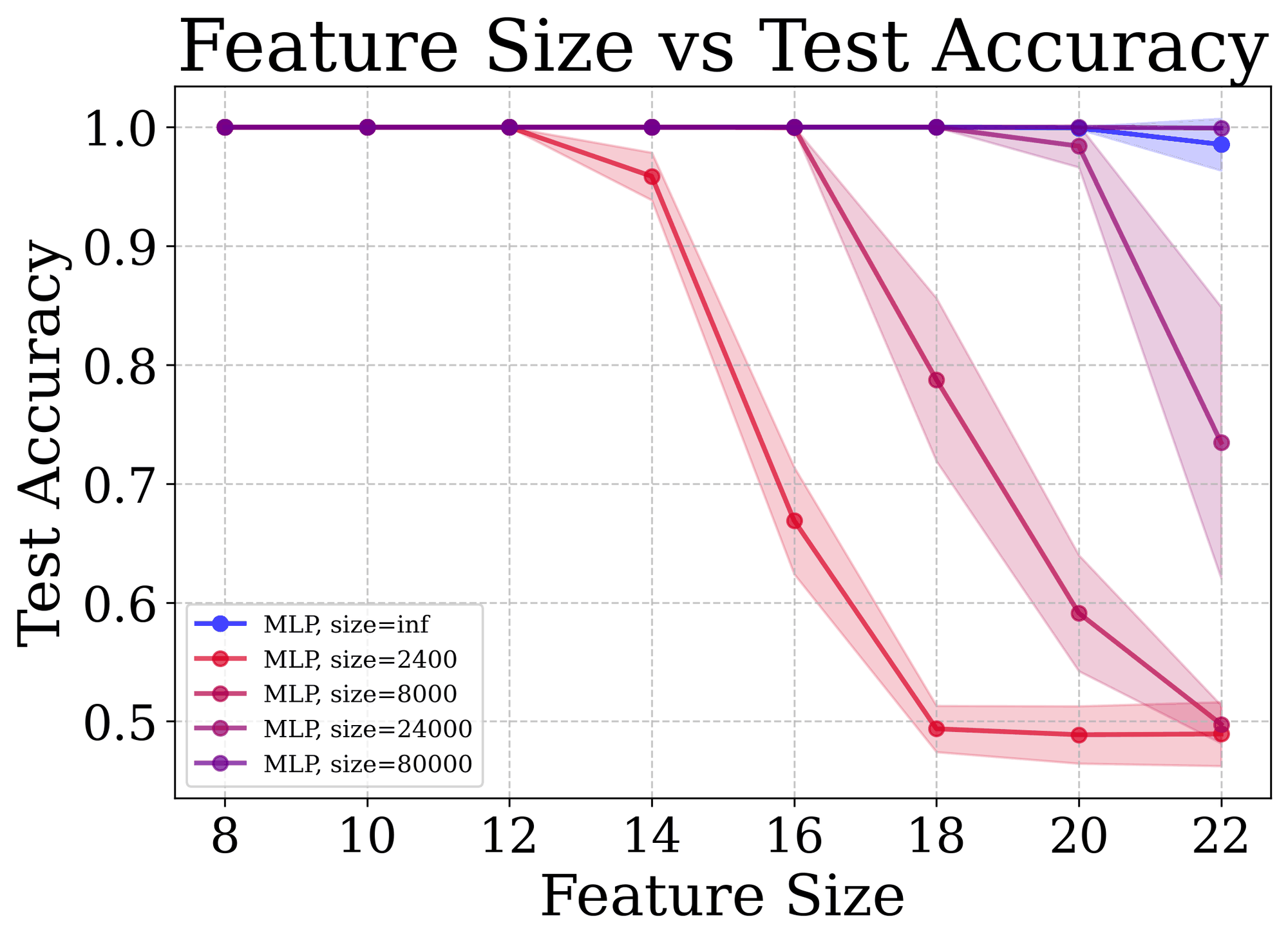}
    \end{subfigure}
    \begin{subfigure}[]{0.24 \textwidth}
        \includegraphics[width=\textwidth]{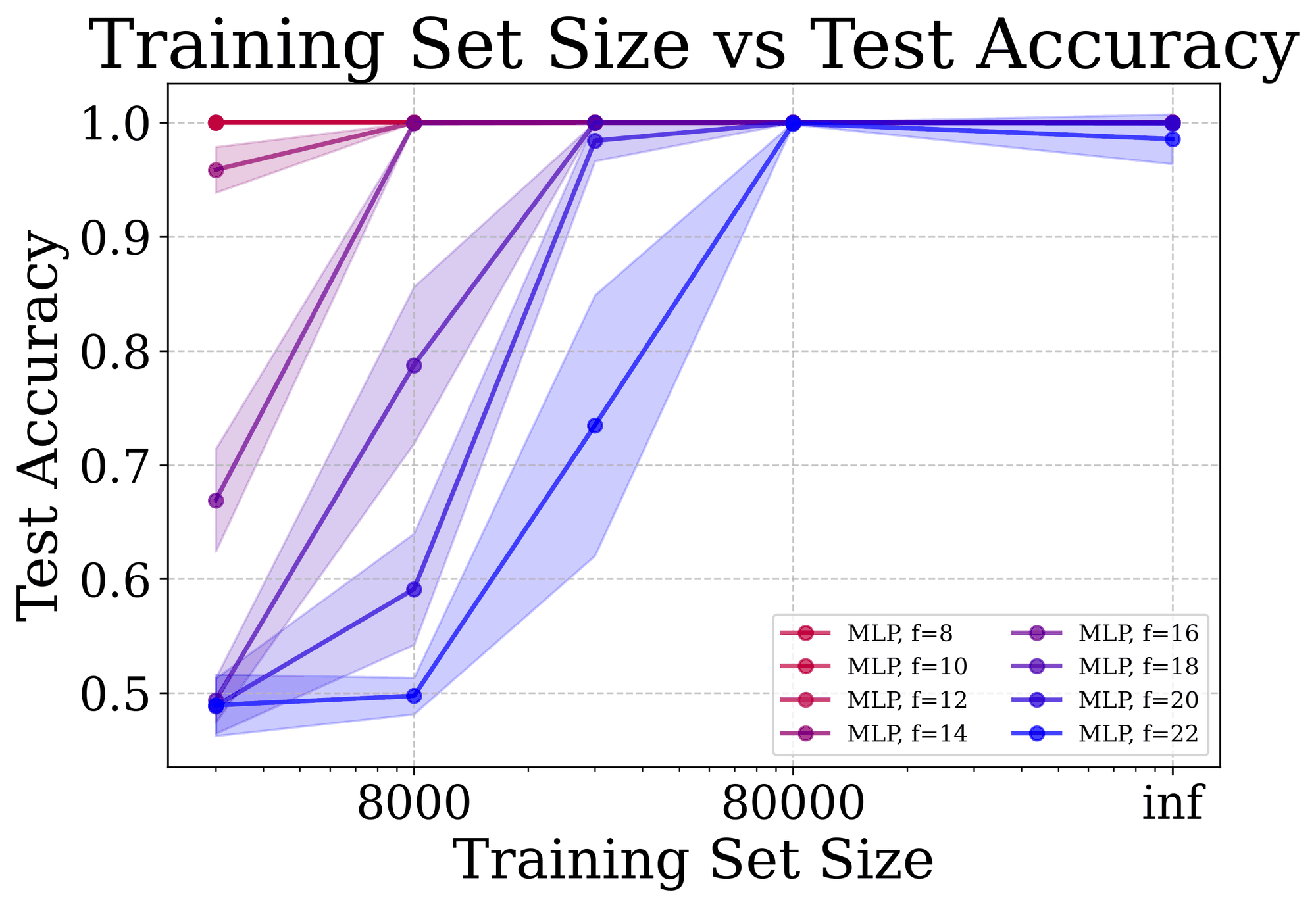}
    \end{subfigure}
    \begin{subfigure}[]{0.24 \textwidth}
        \includegraphics[width=\textwidth]{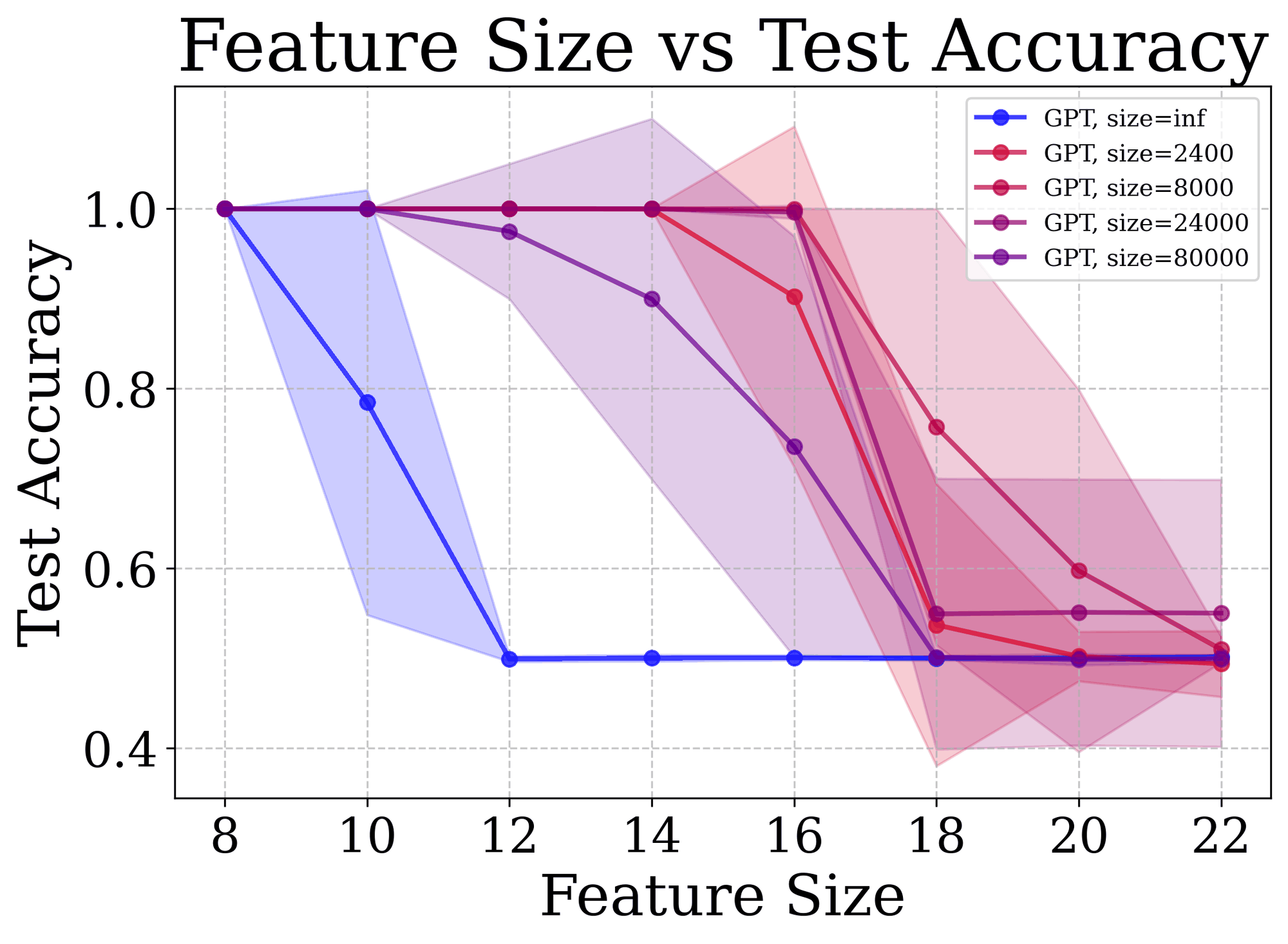}
    \end{subfigure}
    \begin{subfigure}[]{0.24 \textwidth}
        \includegraphics[width=\textwidth]{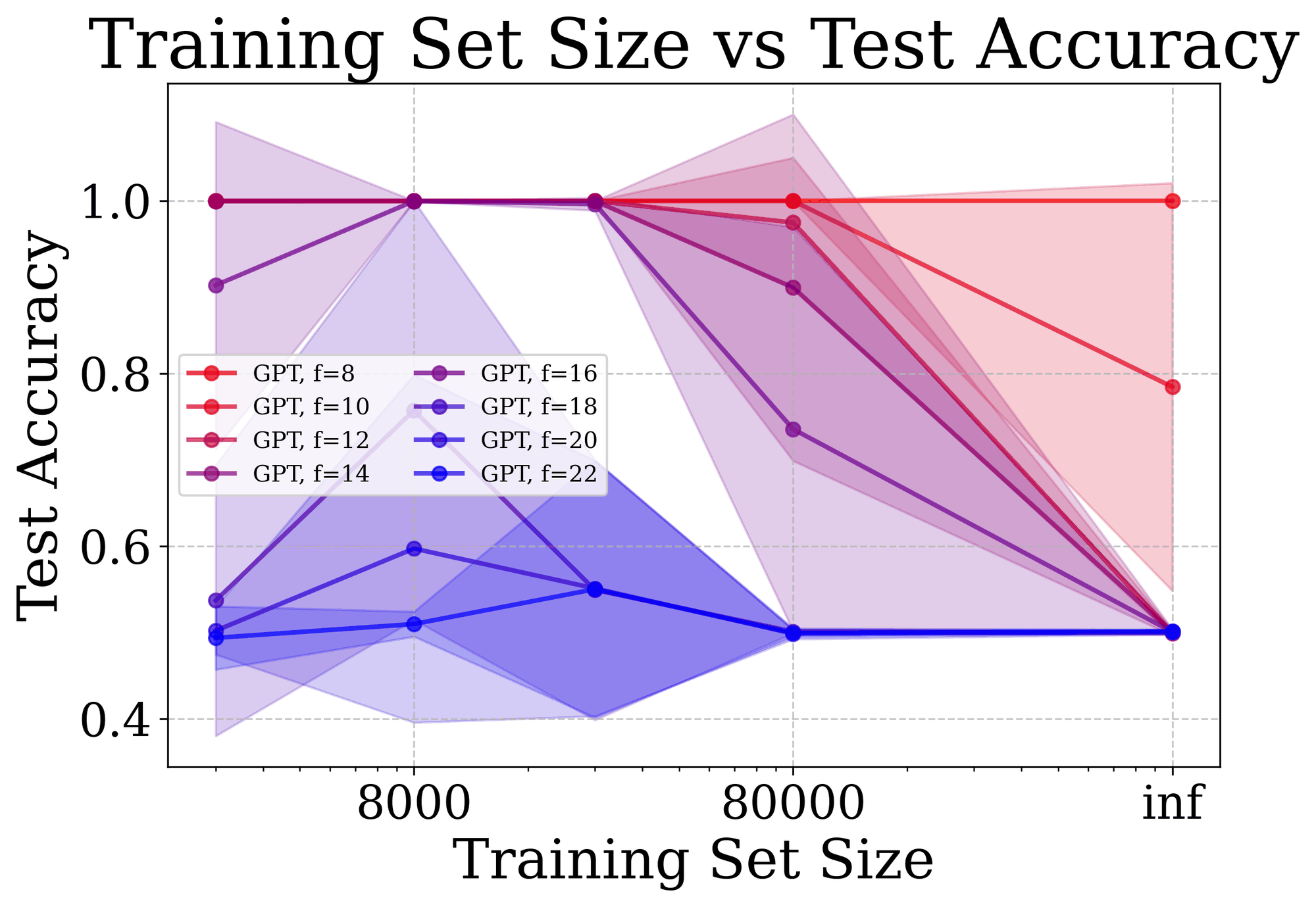}
    \end{subfigure}
    \caption{Test accuracy on a noiseless k-parity feature with a fixed training budget ($T=10^4$, finite training sets and $N=\infty$).
    \textbf{Left two panels:} MLP.
    \textbf{Right two panels:} GPT.
    For each model the first panel shows accuracy against feature size $d$ and the second against training set size $N$.
    Corresponding training accuracies are in Appendix \ref{app:train_tables}.}\vspace{-0.3cm}
    \label{fig:k_parity}
\end{figure*}

\subsection{K-Parity Experiments }\label{sec:k_parity_experiments}

We train both models (MLP \& GPT) on data sets containing a single noiseless k-parity feature for a fixed training budget ($T=10^4$). 
These data sets contain differing numbers of data points and size of k-parity feature. 
The accuracies of the models trained in these experiments are shown in Figure \ref{fig:k_parity}. 
The results for the MLP model show that the larger the k-parity the more difficult it is to learn, As $d$ increases the test accuracy decreases. 
The point at which this decreases from 100\% accuracy on the test set depends on the size of the training set; with larger k-parity features requiring more training examples to learn reliably. 
For a training data set of 8000 examples the MLP can reliably learn a k-parity feature up to $d = 16$, this increases to $d = 22$ when the data set size is increased by a factor of 10. 

For the Transformer model (GPT) we again see the same trend for feature sizes where larger, more complex, less available features are harder to learn from the same amount of data. 
However, the behaviour with regard to data set size is quite different to that of the MLP.
Specifically, as the data set size increases the maximum size of feature that is learnt first increases, but then decreases again. 
The best results for all feature sizes are achieved at a data set size of 8000, for which the GPT model also reliably learns a k-parity feature with $d = 16$. 
Conversely, the GPT model trained directly on samples from the DGP performs worst failing to learn the k-parity features for $d>10$. 
This is a surprising result: for a fixed iteration budget, greater data diversity hurts test performance even when train and test sets are sampled i.i.d.\ from the same DGP.
This does not contradict Transformer scaling laws \citep{kaplan2020scaling}, which describe the benefit of more data when training is scaled appropriately, but it shows that under a fixed budget more data is not always better.
The training curves (Figures \ref{fig:training_curves} and \ref{fig:validation_curves} in the Appendix) and the training accuracies in Appendix \ref{app:train_tables} show what is happening.
Larger data diversity leads to substantially slower training for the Transformer, and two distinct failure modes appear depending on the data set size.
At small $N$ the model fits the training set but fails to generalise (overfitting): for $N=2{,}400$ and $d=18$ the training accuracy is 0.993 but the test accuracy is 0.537.
At large $N$ the model cannot fit the training set within the budget (undertraining): for $N=80{,}000$ and $d=18$ the training accuracy is only 0.506.
The extreme case is $N=\infty$, where the model never sees an example twice and fails to fit the training distribution for $d\geq 12$.
In our fixed-budget experiments an intermediate training set size ($N=8{,}000$) is the sweet spot at which the model both fits the training set and generalises.
Our claim is therefore specifically that, for a fixed iteration budget, increasing data diversity slows Transformer optimisation on k-parity, and that smaller data sets can be preferable as a result.
It is not a claim that more data lowers the final achievable accuracy, nor a challenge to scaling laws in general.
Notably, this slowdown occurs even though a large number of works prove lower bounds on the number of unique queries required to learn k-parity \citep{Kearns1998,peters2026trapped,kou2024matching,shoshani2025hardness}: extra unique examples should not make the problem harder in a statistical sense, so the effect is one of optimisation.
Related repetition effects have been reported for other tasks \citep{liu2026lessdata, charton2024emergent}, and \citet{liu2026lessdata} give a detailed account of how repetition and sampling bias from small data sets can speed up training.
Our results complement this line of work by showing that the same phenomenon arises for Transformers on k-parity, a setting with a precise notion of difficulty, and by contrasting it with the behaviour of MLPs.
In Figure \ref{fig:pull} we exploit this phenomenon to create an example where scaling up data, model size and training iterations leads to worse results, noting that the iteration budget grows far more slowly than the data set.

In Figure \ref{fig:iter_scaling_k_parity_noiseless} (Appendix) we show the results for the transformer model when scaling the number of training iterations proportionally to the data set size, specifically $T=1.25N$, so that with a batch size of 256 each sample is expected to appear in a batch approximately $320$ times regardless of $N$.
Here the results more closely match those of the MLP with a fixed budget, where more samples generally lead to better performance: the frontier of learnable feature sizes moves from $d=14$ at $N=2400$ to $d=20$ at $N=240{,}000$ (Appendix \ref{app:train_tables}).
However, the optimisation slowdown does not disappear in the partially learned regime.
At $d=22$, $N=24{,}000$ reaches a training accuracy of 0.904 whereas $N=240{,}000$ reaches only 0.771, even though the latter is given ten times more iterations.
The slowdown with more diverse data is therefore a persistent property of Transformer k-parity learning, not only an artefact of a fixed iteration budget.
We also note that we are using a much larger budget for the GPT model in this setting than in the fixed-budget experiments.

\begin{table}[ht]
\centering\small
\setlength{\tabcolsep}{5pt}
\begin{tabular}{lrrrrrr}
\toprule
 & \multicolumn{2}{c}{Fixed, $d=18$} & \multicolumn{2}{c}{Scaled, $d=18$} & \multicolumn{2}{c}{Scaled, $d=22$} \\
\cmidrule(lr){2-3}\cmidrule(lr){4-5}\cmidrule(lr){6-7}
Training set size $N$ & train & test & train & test & train & test \\
\midrule
2{,}400 & 0.993 & 0.537 & 0.817 & 0.494 & 0.831 & 0.495 \\
8{,}000 & 0.889 & 0.757 & 0.867 & 0.705 & 0.822 & 0.506 \\
24{,}000 & 0.594 & 0.549 & 0.792 & 0.610 & 0.904 & 0.597 \\
80{,}000 & 0.506 & 0.501 & 1.000 & 1.000 & 0.876 & 0.600 \\
240{,}000 & -- & -- & 1.000 & 1.000 & 0.771 & 0.702 \\
$\infty$ (DGP) & 0.500 & 0.500 & -- & -- & -- & -- \\
\bottomrule
\end{tabular}
\caption{GPT train and test accuracy on a noiseless k-parity feature under the fixed budget ($T=10^4$) and the scaled budget ($T=1.25N$).
Rows give the training set size $N$.
Under the fixed budget, large data sets are not fitted even on the training set (undertraining), and under the scaled budget the training accuracy at $d=22$ is lower for the largest data set than for $N=24{,}000$ despite ten times more iterations.
Full tables are in Appendix \ref{app:train_tables}.}
\label{tab:main_train_test}
\end{table}

\paragraph{Separating repetition, undertraining and diversity.} The experiments above separate the candidate explanations as follows.
\emph{Repetition}: in the fixed-budget experiments the expected number of passes per example falls as $1/N$ by construction (Table \ref{tab:passes}), so these results alone cannot separate repetition from diversity; the scaled-budget experiments hold the expected passes per example constant.
\emph{Undertraining}: the training accuracies show that the large-$N$ fixed-budget failures are undertraining rather than overfitting.
\emph{Diversity}: the $N=\infty$ runs remove repetition entirely and are the slowest to optimise, and the scaled-budget results show that more diverse data remains harder to optimise at constant passes per example.
These experiments build on the careful analysis of repetition by \citet{liu2026lessdata}.

The observation of more data leading to slower training under a fixed budget suggests an efficient method of learning k-parities with Transformers using a curriculum. 
This agrees with the findings of \citet{cornacchia2023mathematical} that curricula can be used to efficiently learn parities. 
However, we find this to be most effective for Transformers which was not part of their study. 
We investigate the following scheme where only a subset of the full data set is used for training for the first fraction of the training budget, before later switching to the full training set.
We show this is indeed effective by restricting training to a $10^{th}$ of the training set for the first $c\%$ of training, and then using the full data set for the remaining $100-c \%$ of training. 
We experiment with $c \in \{10,50,90\}$ and $N=80000$. 
The result of this experiment is shown in Figure \ref{fig:curriculum}.
The curriculum with a 90/10 split performs best. 
This is further evidence that care is required to efficiently train  Transformers on certain classes of hard functions. 

Finally, it is worth noting there are many examples where both models achieve perfect test accuracy when trained on data set sizes containing far lower then the number of possible input vectors $(2^d)$.
For $d=16$ there exist $2^{16}$ unique strings however both models manage to learn the ``correct'' boolean function and generalise perfectly from $8000\approx2^{13}$ training examples or roughly one eighth of all possible input strings. 
This shows that the inductive biases of both models encourage them to learn the k-parity function over other less structured function that would also give zero training loss, such as memorisation. 
This is an example of the SB working in our favour.

\begin{takeawaybox}[title=Key Finding 1]
  For Transformers trained on k-parity, greater data diversity leads to markedly slower optimisation.
  Under a fixed training budget this can make smaller data sets give better train and test performance, and it is possible to construct cases where scaling up the data, model size and iterations gives worse results (Figure \ref{fig:pull}).
  When the budget is scaled with the data set size, more data extends the frontier of learnable features, but larger data sets still optimise more slowly in the partially learned regime.
\end{takeawaybox}

\subsection{Noisy k-parity Experiments}\label{sec:noisy_k_parity_experiments}
We now investigate the performance of both models on noisy k-parity, that is features where the predictivity is less than $1.0$.
Similar to section \ref{sec:k_parity_experiments} we again create a selection of data sets with a single feature $f_1$ and vary the following key quantities, the data set size $N$, the size of the single feature $d_1$ and the predictivity of said feature $\alpha_1 \in \{0.6,0.7,0.8,0.9,1.0\}$.

\begin{figure*}[ht]
    \centering

    \begin{subfigure}[width=\textwidth]{0.24 \textwidth}
        \includegraphics[width=\textwidth]{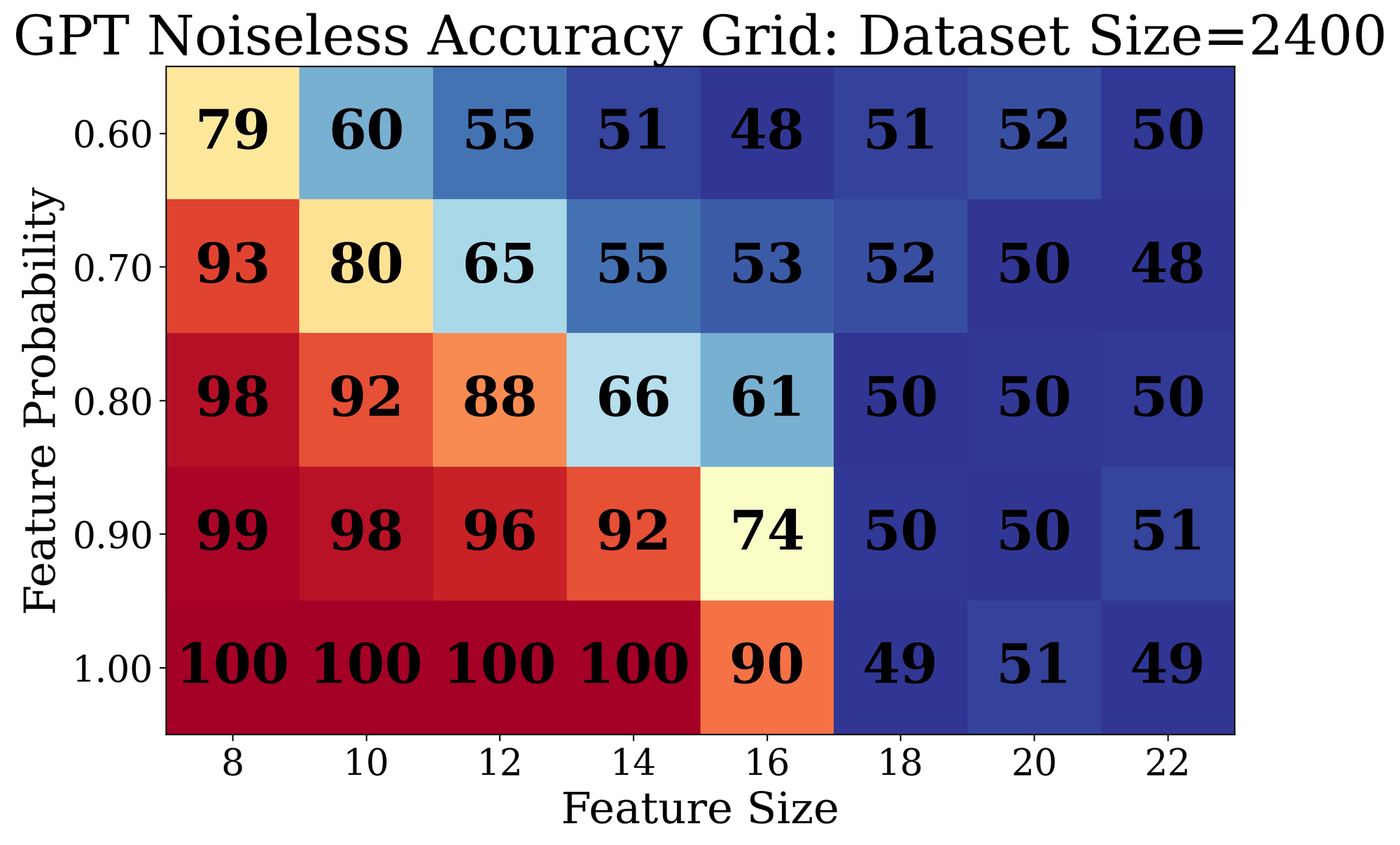}
    \end{subfigure}
    \begin{subfigure}[width=\textwidth]{0.24 \textwidth}
        \includegraphics[width=\textwidth]{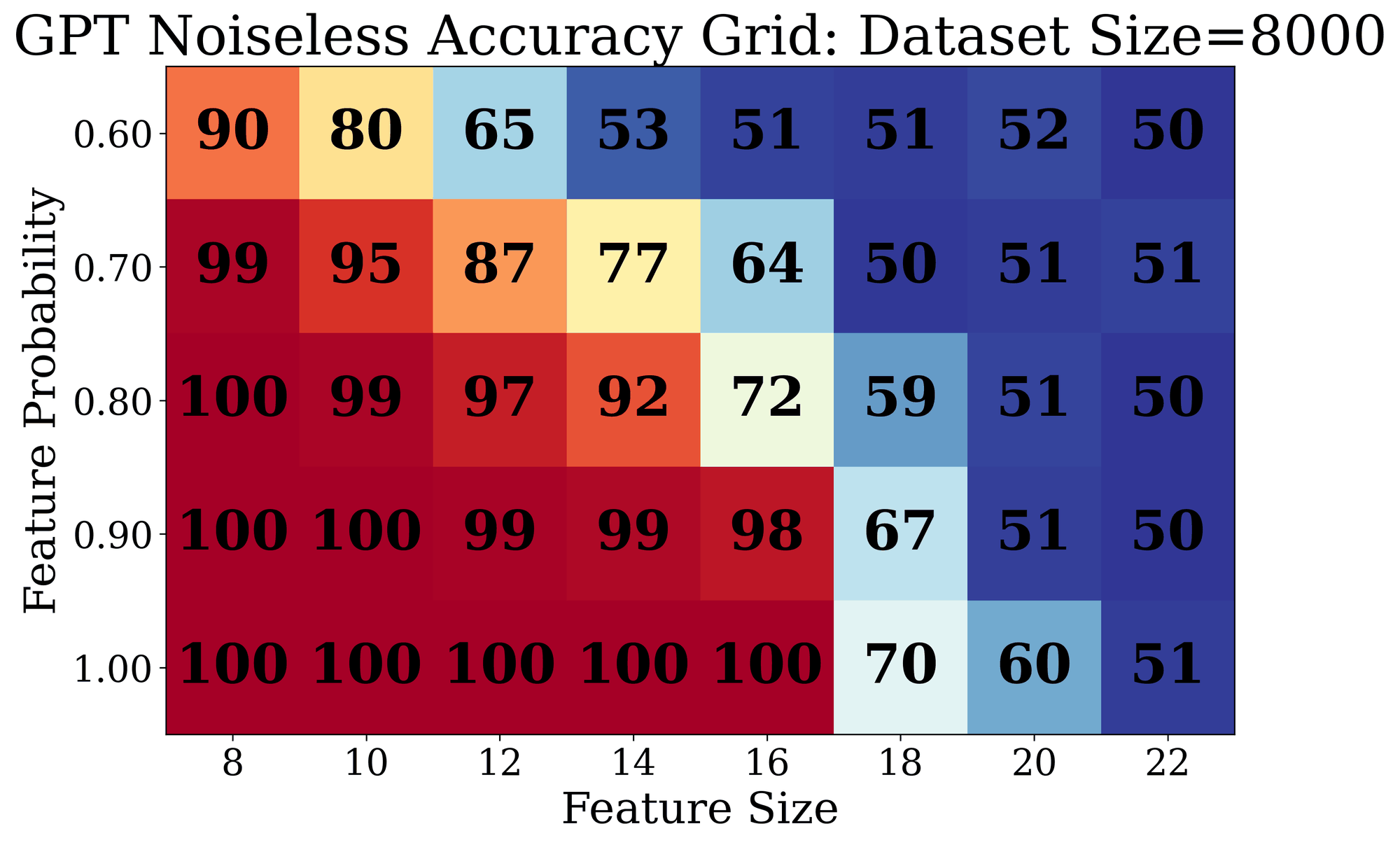}
    \end{subfigure}
    \begin{subfigure}[width=\textwidth]{0.24 \textwidth}
        \includegraphics[width=\textwidth]{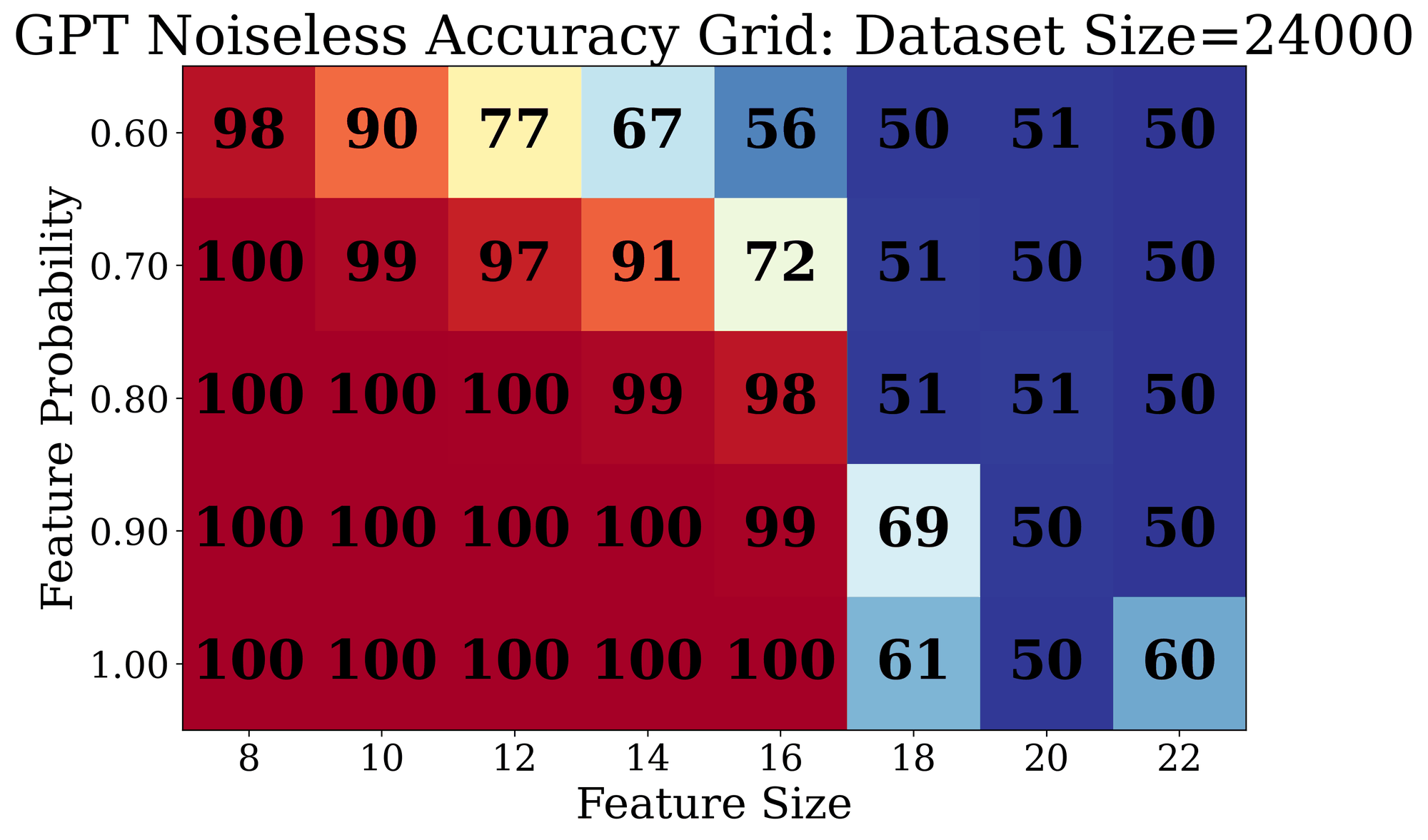}
    \end{subfigure}
    \begin{subfigure}[width=\textwidth]{0.24 \textwidth}
        \includegraphics[width=\textwidth]{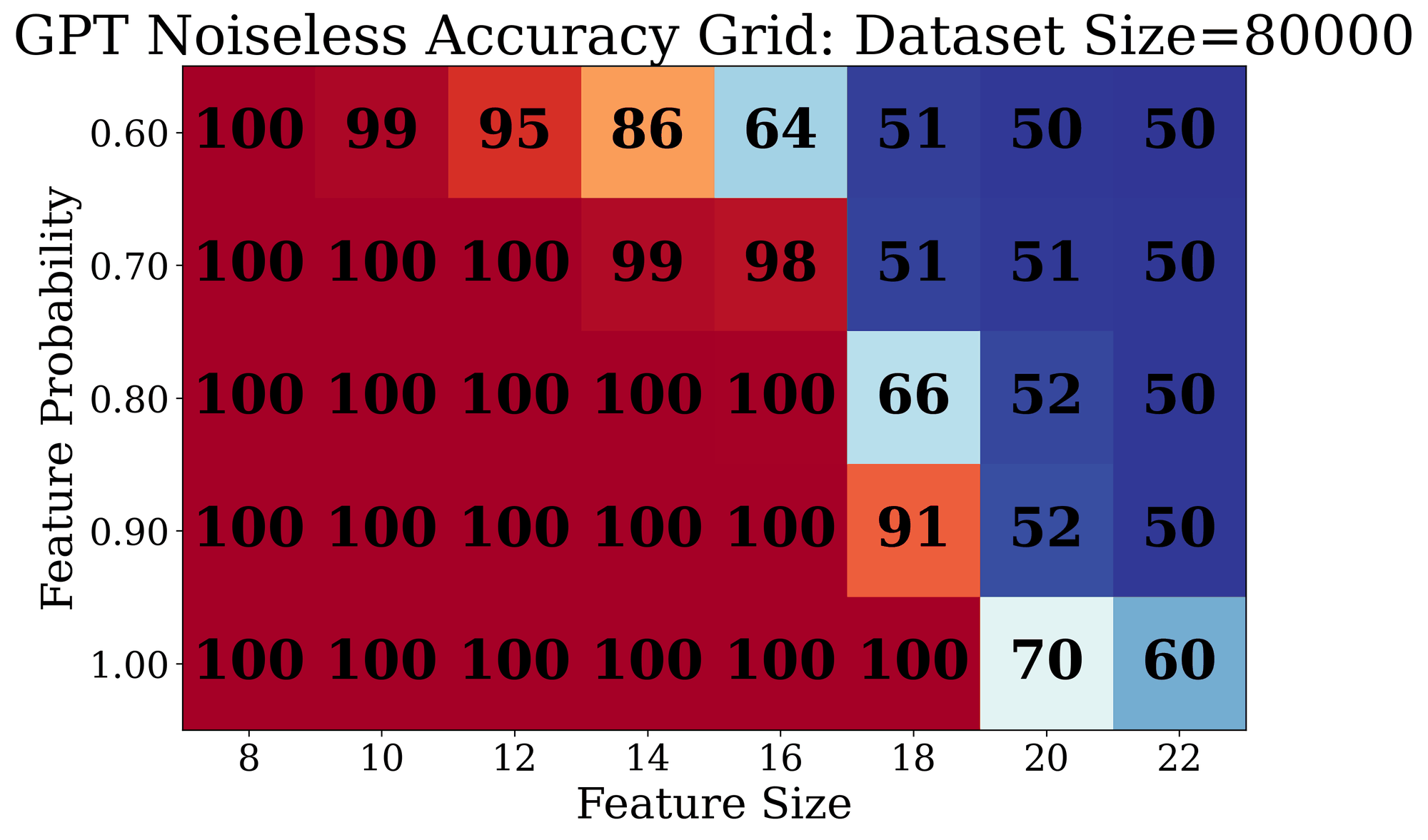}
    \end{subfigure}
    
    \begin{subfigure}[width=\textwidth]{0.24 \textwidth}
        \includegraphics[width=\textwidth]{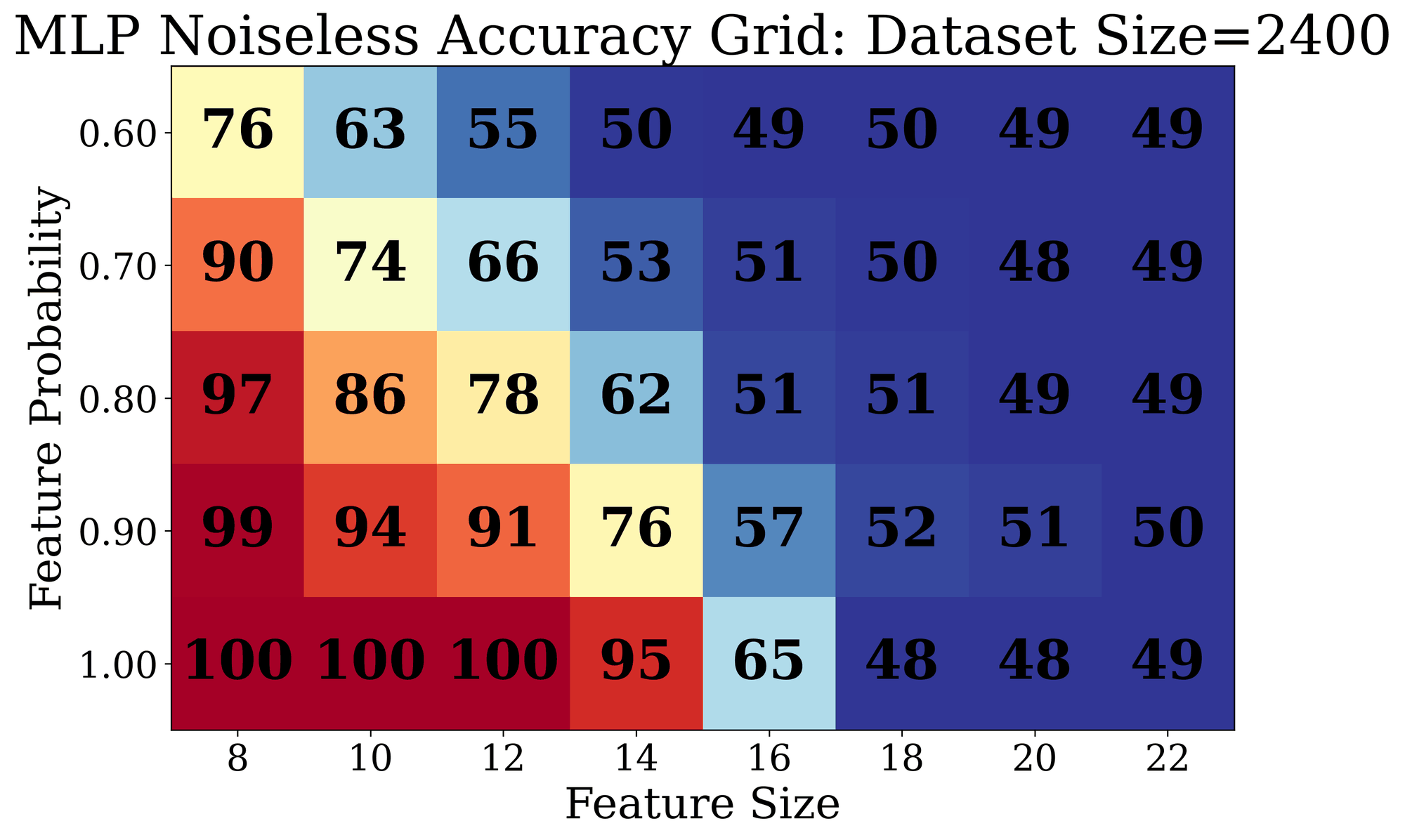}
    \end{subfigure}
    \begin{subfigure}[width=\textwidth]{0.24 \textwidth}
        \includegraphics[width=\textwidth]{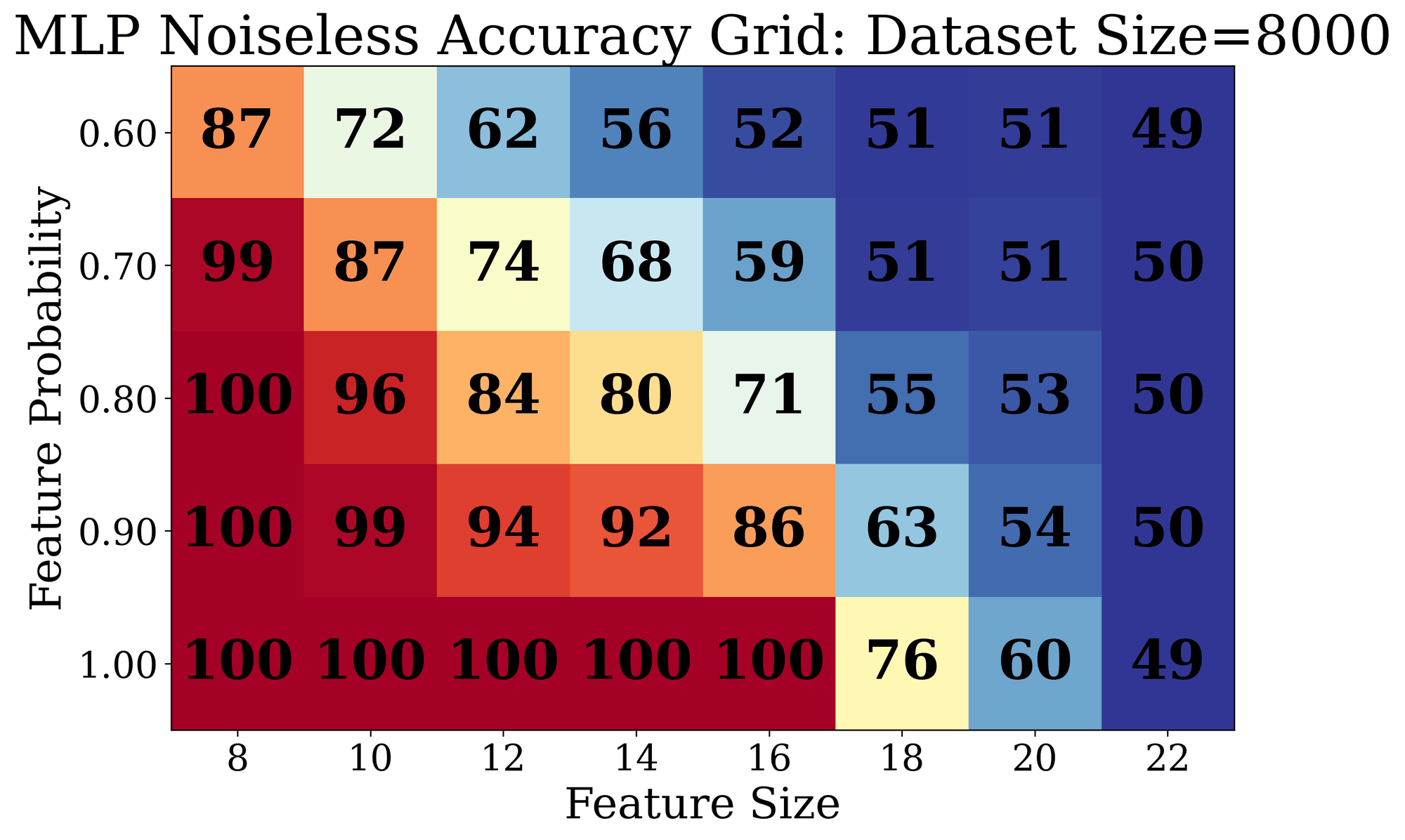}
    \end{subfigure}
    \begin{subfigure}[width=\textwidth]{0.24 \textwidth}
        \includegraphics[width=\textwidth]{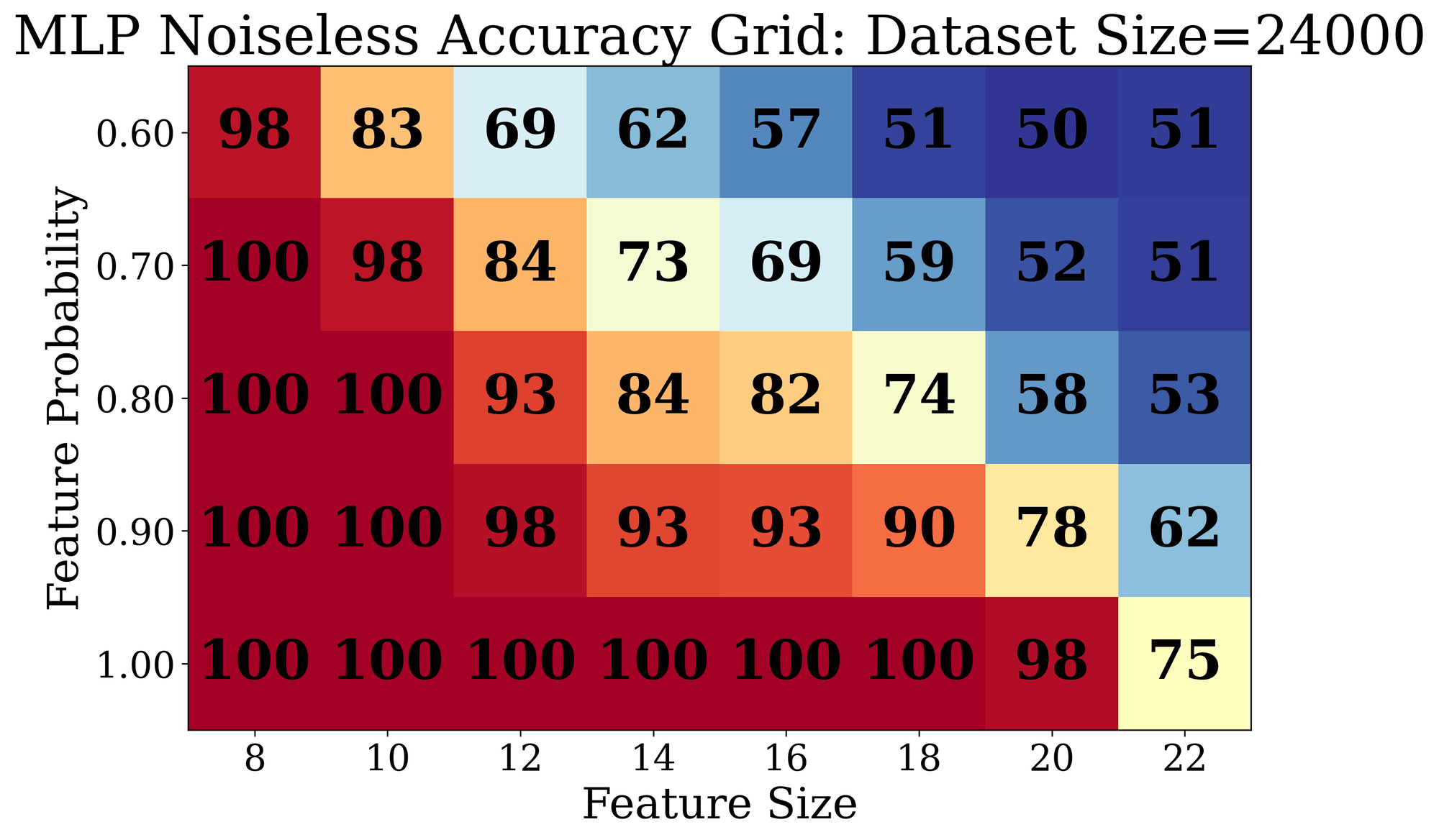}
    \end{subfigure}
    \begin{subfigure}[width=\textwidth]{0.24 \textwidth}
        \includegraphics[width=\textwidth]{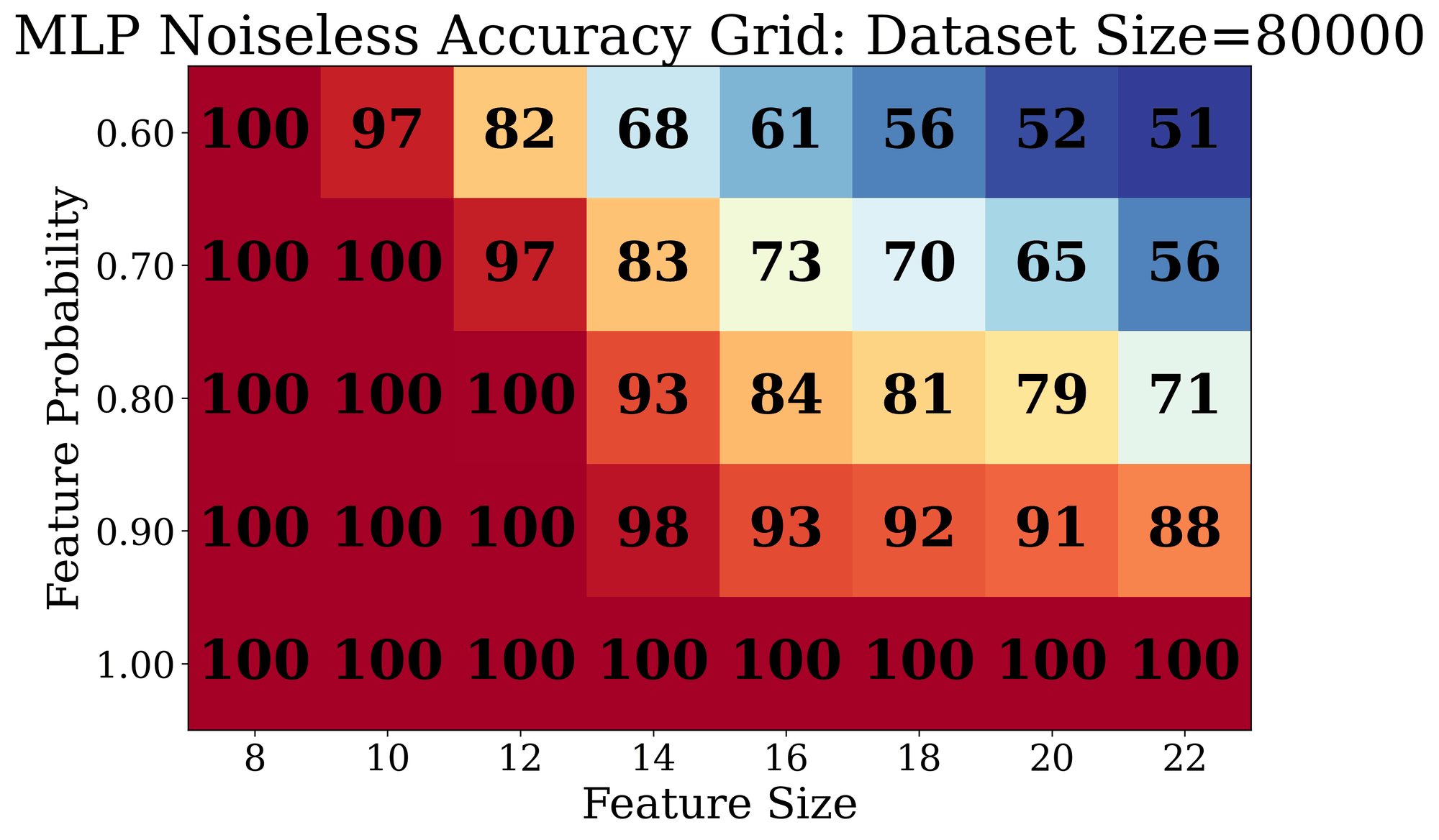}
    \end{subfigure}

    \caption{Test accuracy on a noiseless test set for models trained on a noisy k-parity feature with a scaled iteration budget ($T=1.25N$).
    Each panel shows accuracy as a function of feature size $d_1$ and predictivity $\alpha_1$.
    \textbf{Top row:} GPT.
    \textbf{Bottom row:} MLP.
    Columns correspond to $N=2400, 8000, 24000, 80000$.}
    \label{fig:noisy_iter_scaling_noiseless}
\end{figure*}

\begin{figure*}[ht]
    \centering
    
    \begin{subfigure}[width=\textwidth]{0.24 \textwidth}
        \includegraphics[width=\textwidth]{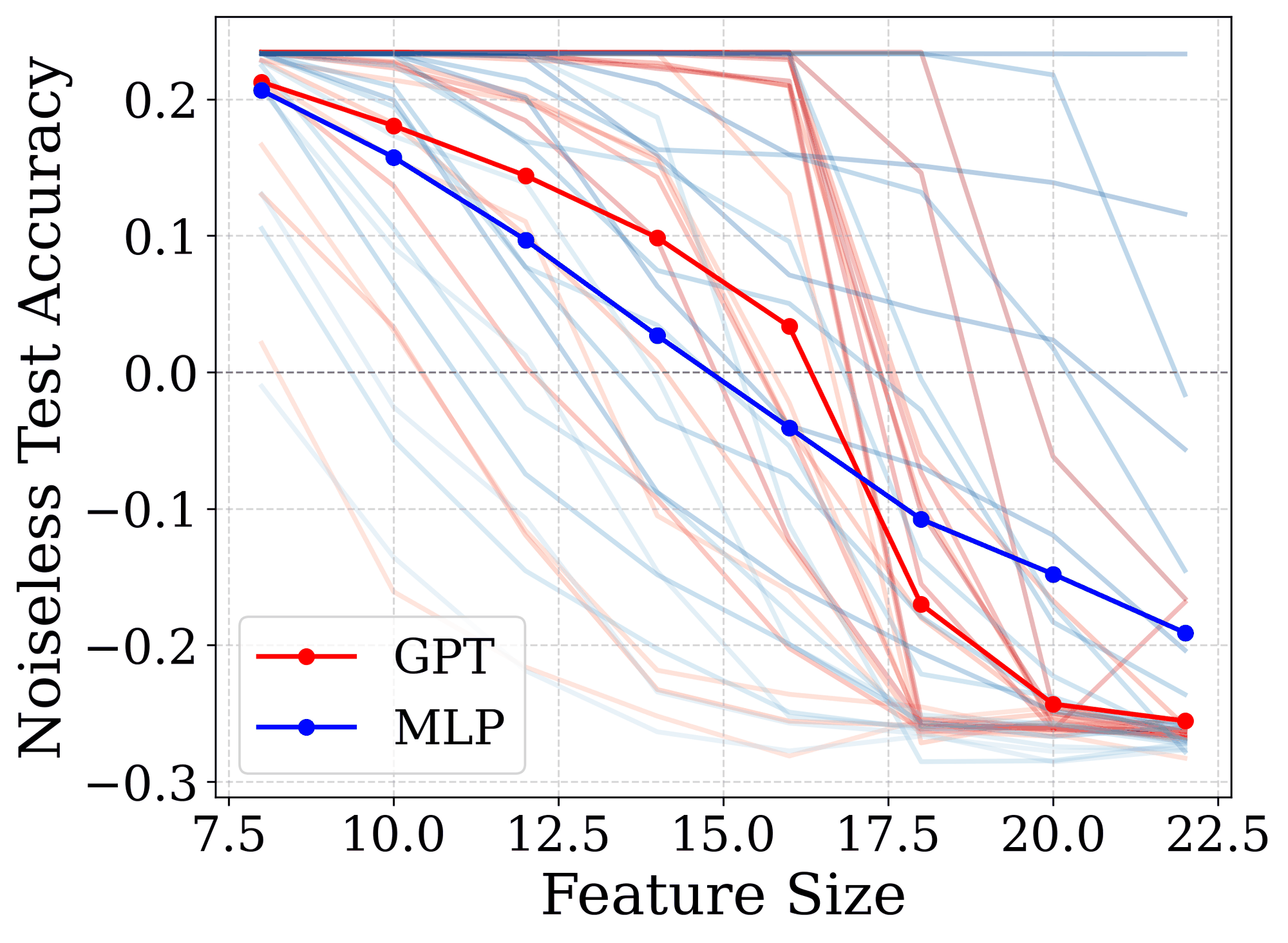}
    \end{subfigure}
    \begin{subfigure}[width=\textwidth]{0.24 \textwidth}
        \includegraphics[width=\textwidth]{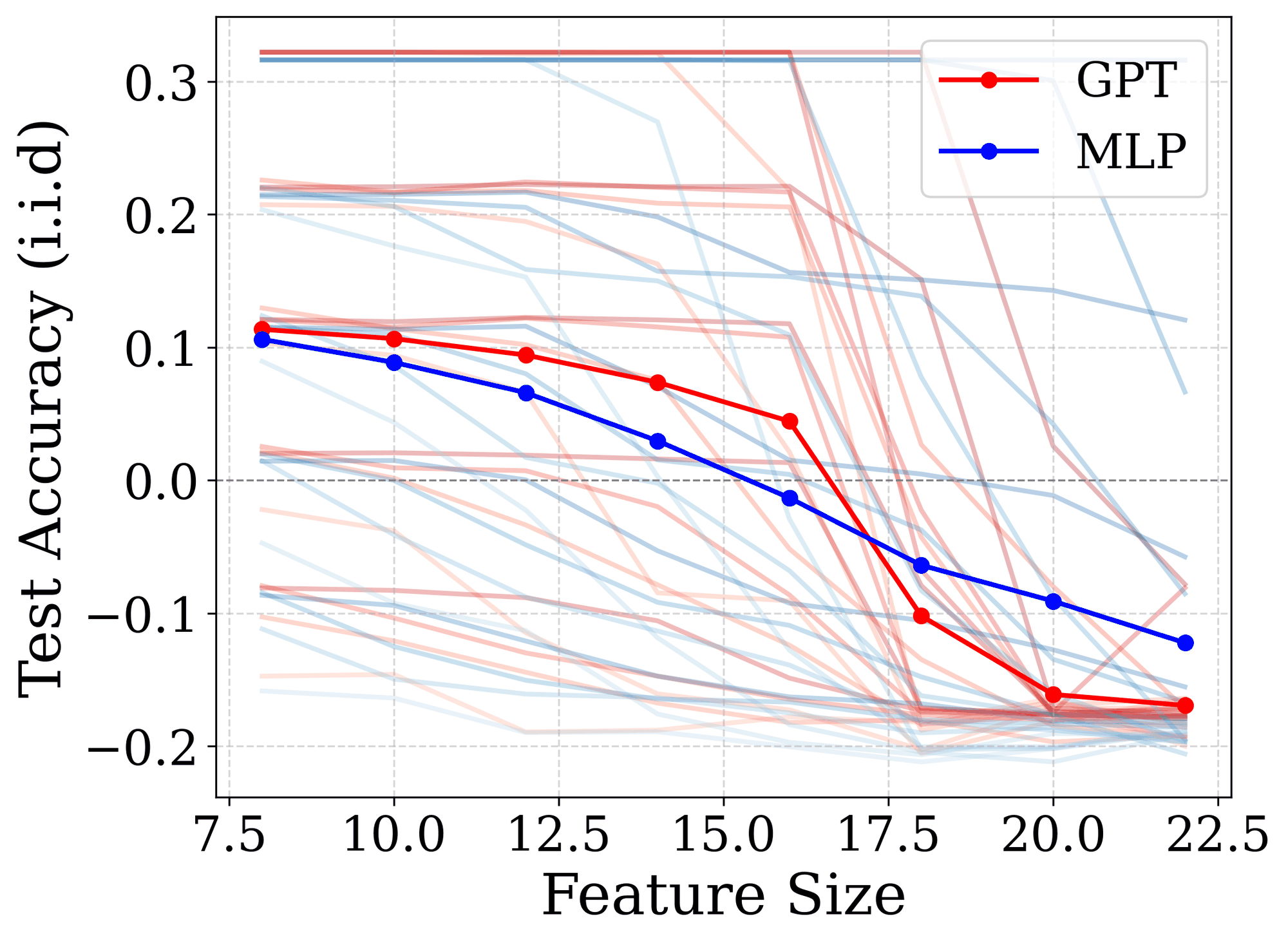}
    \end{subfigure}
    \begin{subfigure}[width=\textwidth]{0.24 \textwidth}
        \includegraphics[width=\textwidth]{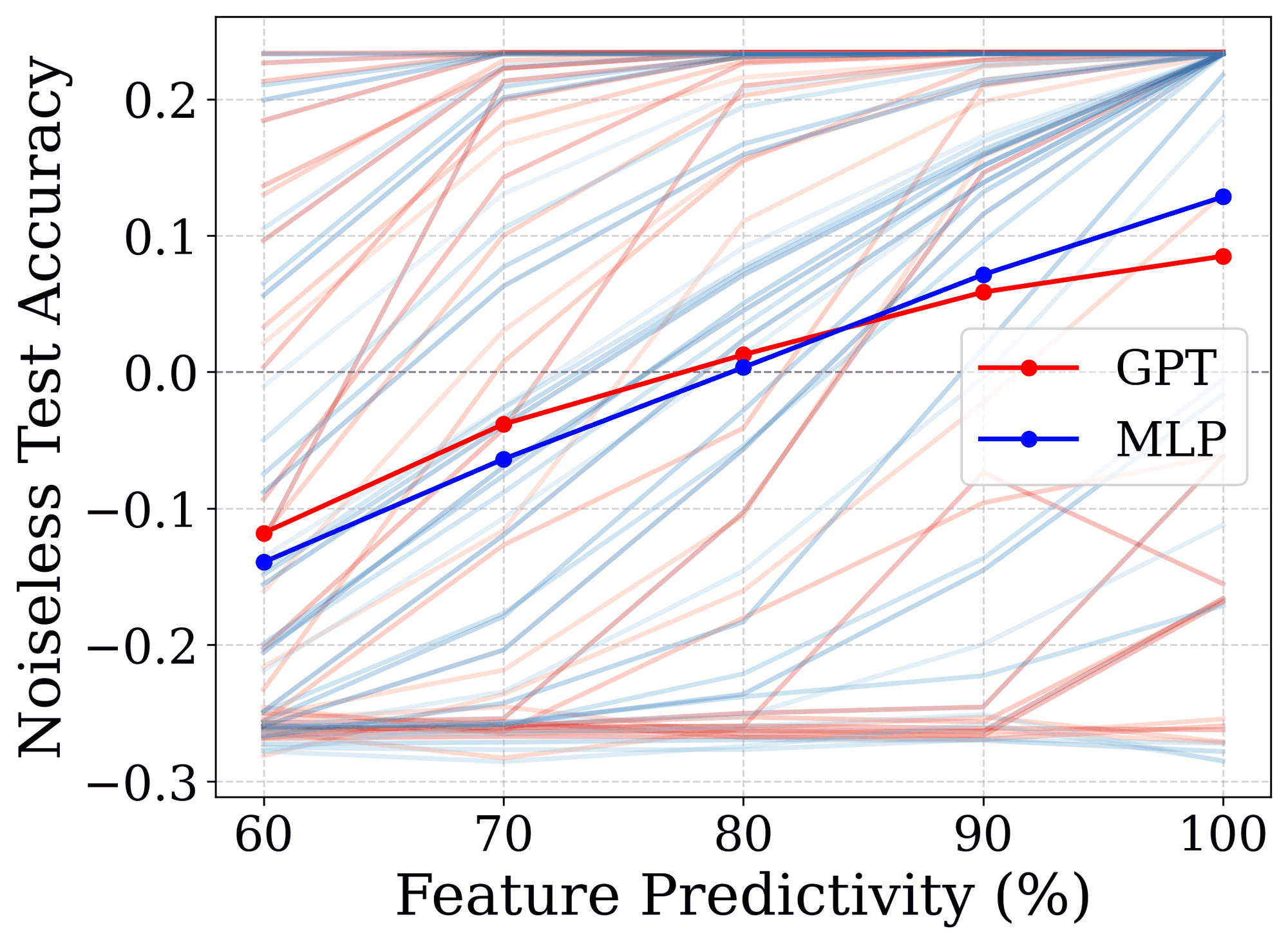}
    \end{subfigure}
    \begin{subfigure}[width=\textwidth]{0.24 \textwidth}
        \includegraphics[width=\textwidth]{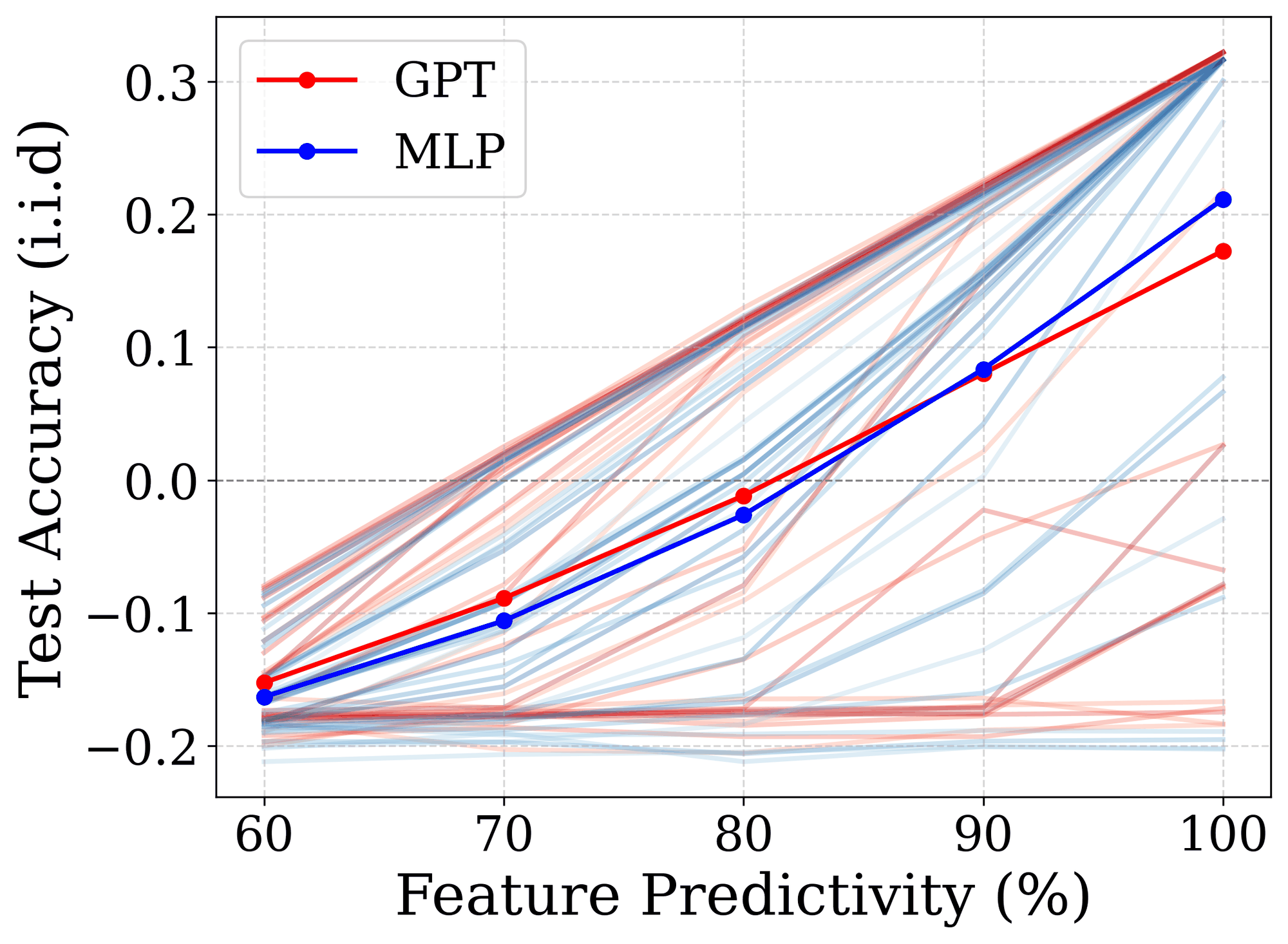}
    \end{subfigure}

    \caption{Comparison of MLP and GPT inductive biases toward predictivity and availability, with a scaled iteration budget ($T=1.25N$), averaged over data set sizes.
    \textbf{Left two panels:} average accuracy against feature availability (larger features are less available), on a noiseless test set and on an i.i.d.\ test set respectively.
    \textbf{Right two panels:} average accuracy against feature predictivity, on a noiseless and an i.i.d.\ test set respectively.
    Here Transformers are relatively less sensitive to noise (lower predictivity) but are more sensitive to feature size (lower availability) than MLPs.}\vspace{-0.3cm}
    \label{fig:noise_and_size}
\end{figure*}

Figure \ref{fig:noisy_iter_scaling_noiseless} shows the test accuracies on a noiseless test set for the MLP and GPT models when the budget is scaled with the data set size ($T=1.25N$).
Results for $\alpha_1=0.8$ are shown in more detail in Figures \ref{fig:iter_scaling_k_parity} (GPT, scaled budget) and \ref{fig:noisy_k_parity_appendix} (both models, fixed budget) in the appendix. 
Train accuracies and test accuracies on i.i.d test set are additionally shown in Figure \ref{fig:noisy_k_parity_test_scaling} within the appendix.
We observe a global trend that holds for both models and all data set sizes, as the predictivity decreases, the test accuracy decreases.
Features more highly correlated with the output are more easy to disentangle from noise.
Additionally large less available features are harder to learn. 
However, again there are marked differences between the two models particularly when scaling up the data set size. 
The MLP excels at learning low noise features with low availability especially as the data set size is increased, but often fails on noisy features due to overfitting to the training set, as shown in Figures \ref{fig:noisy_iter_scaling_noiseless}, \ref{fig:noisy_k_parity_test_scaling} \& \ref{fig:noisy_k_parity_appendix}.
Conversely, as noted by \citep{vasudeva2024transformers, peters2026trapped} the GPT model is highly biased to learning simple functions, resulting in it preferentially learning k-parity over memorising the training data. 
This corroborates the result of \citet{peters2026trapped} that Transformers are effective at learning from noisy data, but we additionally show how data set size amplifies this ability while doing little to help the learning of large low availability features.
In Figure \ref{fig:noise_and_size} we show how the normalised average accuracy of the models changes with the feature size and predictivity average across all data.
We show results both on a noiseless test set and an i.i.d test set. 
In both cases we observe when learning k-parity features Transformers are relatively less sensitive to noise (lower predictivity) but are more sensitive to feature size (lower availability) than MLPs.

For a fixed training budget we again observe the trend that increasing the data set size for the transformer first increases the test performance but then leads to a sharp drop off where the transformer fails to learn the function within the given training budget (Figures \ref{fig:noisy_k_parity_appendix} \& \ref{fig:noisy_k_parity_test_fixed} in the appendix).

\subsection{Data sets of Two features}\label{sec:two_features}
\begin{figure*}[ht]
    \centering
    \begin{subfigure}[width=\textwidth]{0.16 \textwidth}
        \includegraphics[width=\textwidth]{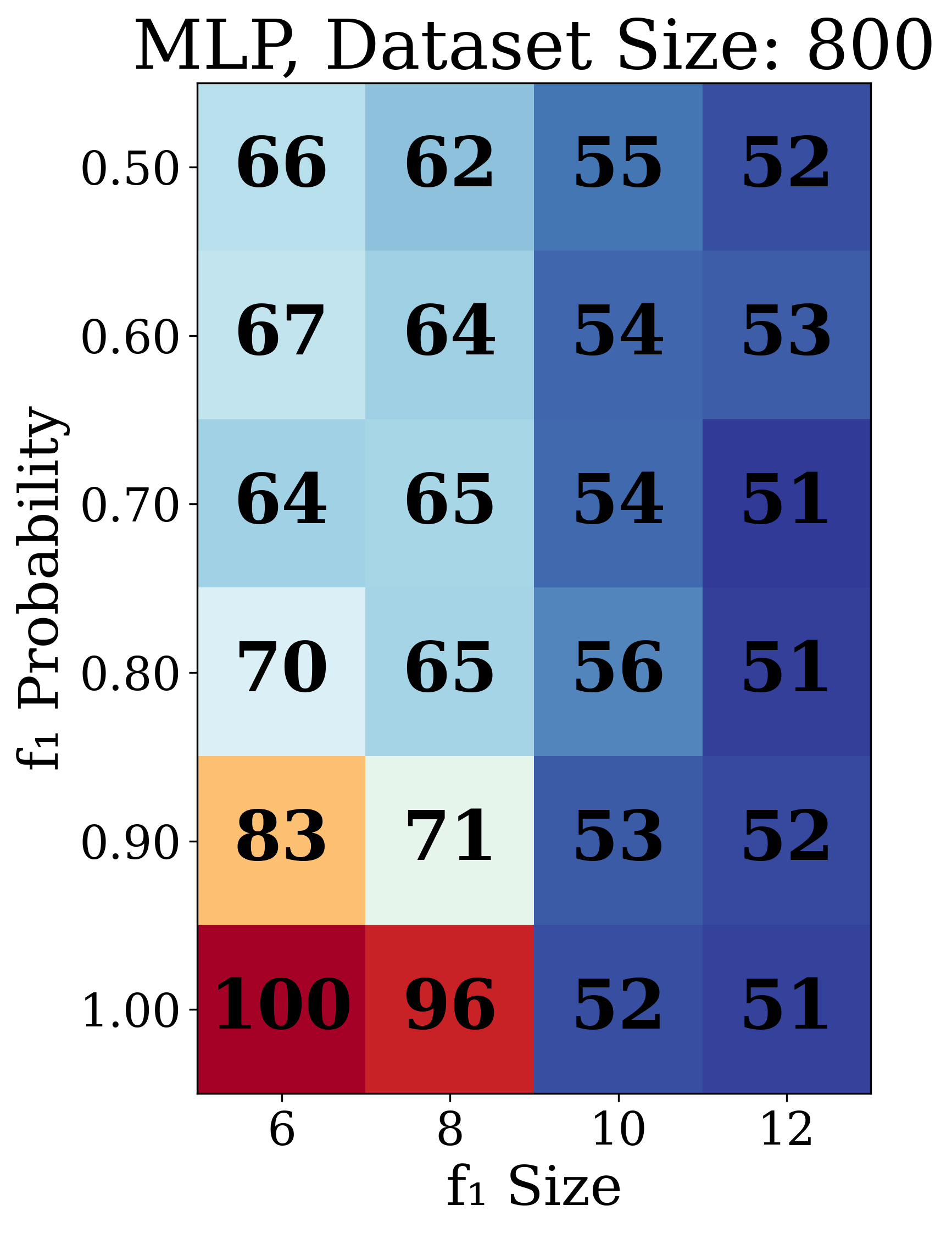}
    \end{subfigure}
    \begin{subfigure}[width=\textwidth]{0.16 \textwidth}
        \includegraphics[width=\textwidth]{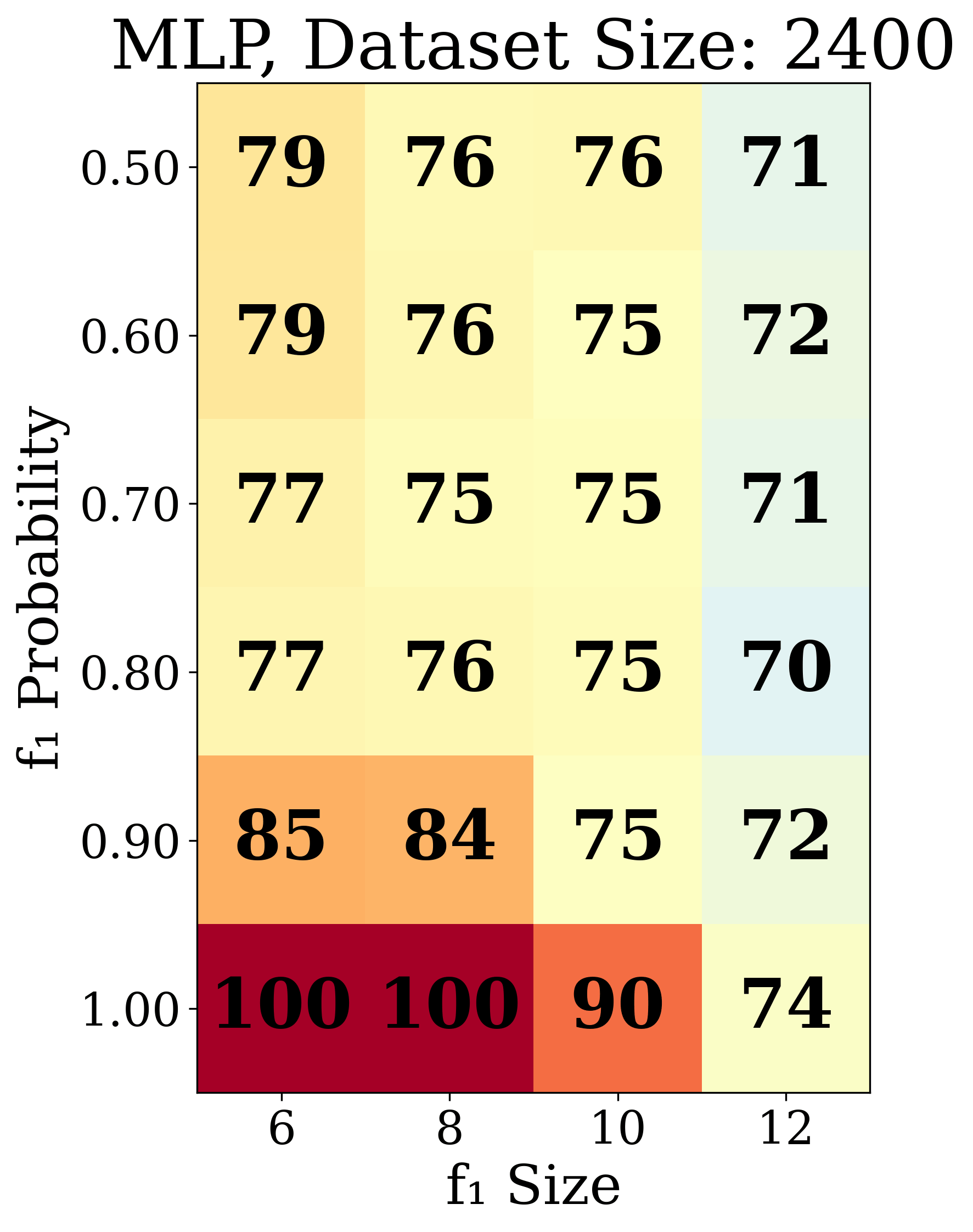}
    \end{subfigure}
    \begin{subfigure}[width=\textwidth]{0.16 \textwidth}
        \includegraphics[width=\textwidth]{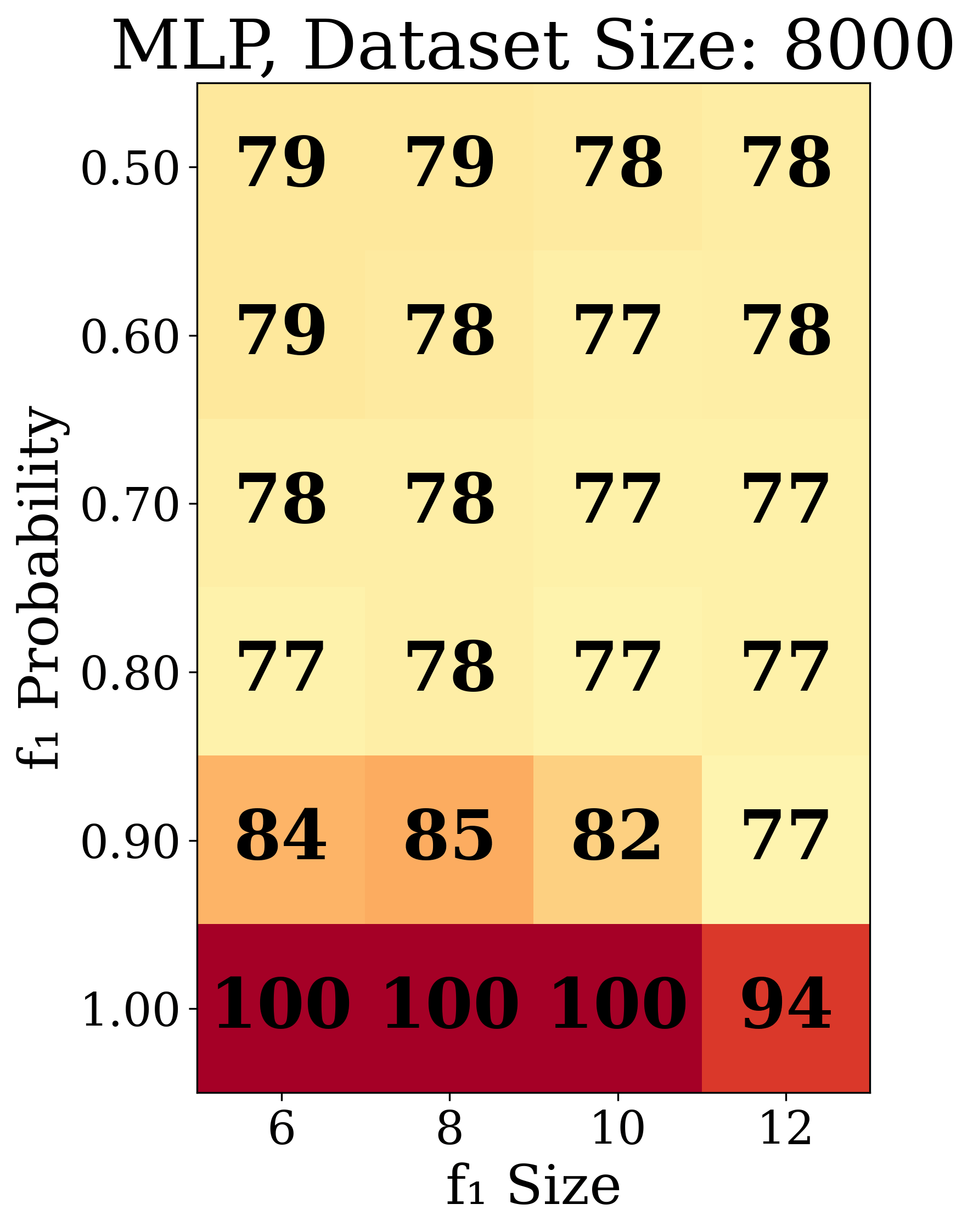}
    \end{subfigure}
    \begin{subfigure}[width=\textwidth]{0.16 \textwidth}
        \includegraphics[width=\textwidth]{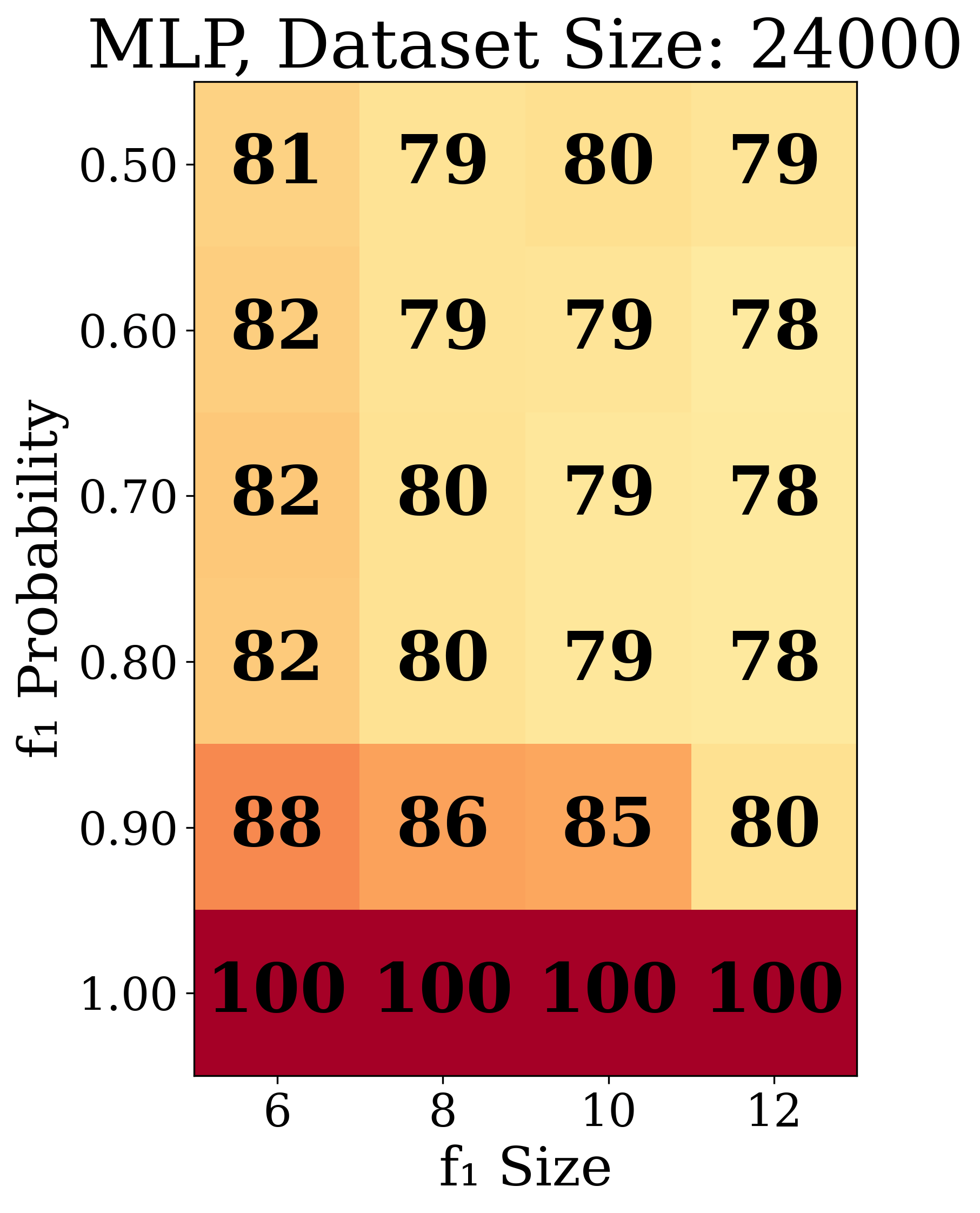}
    \end{subfigure}
    \begin{subfigure}[width=\textwidth]{0.16 \textwidth}
        \includegraphics[width=\textwidth]{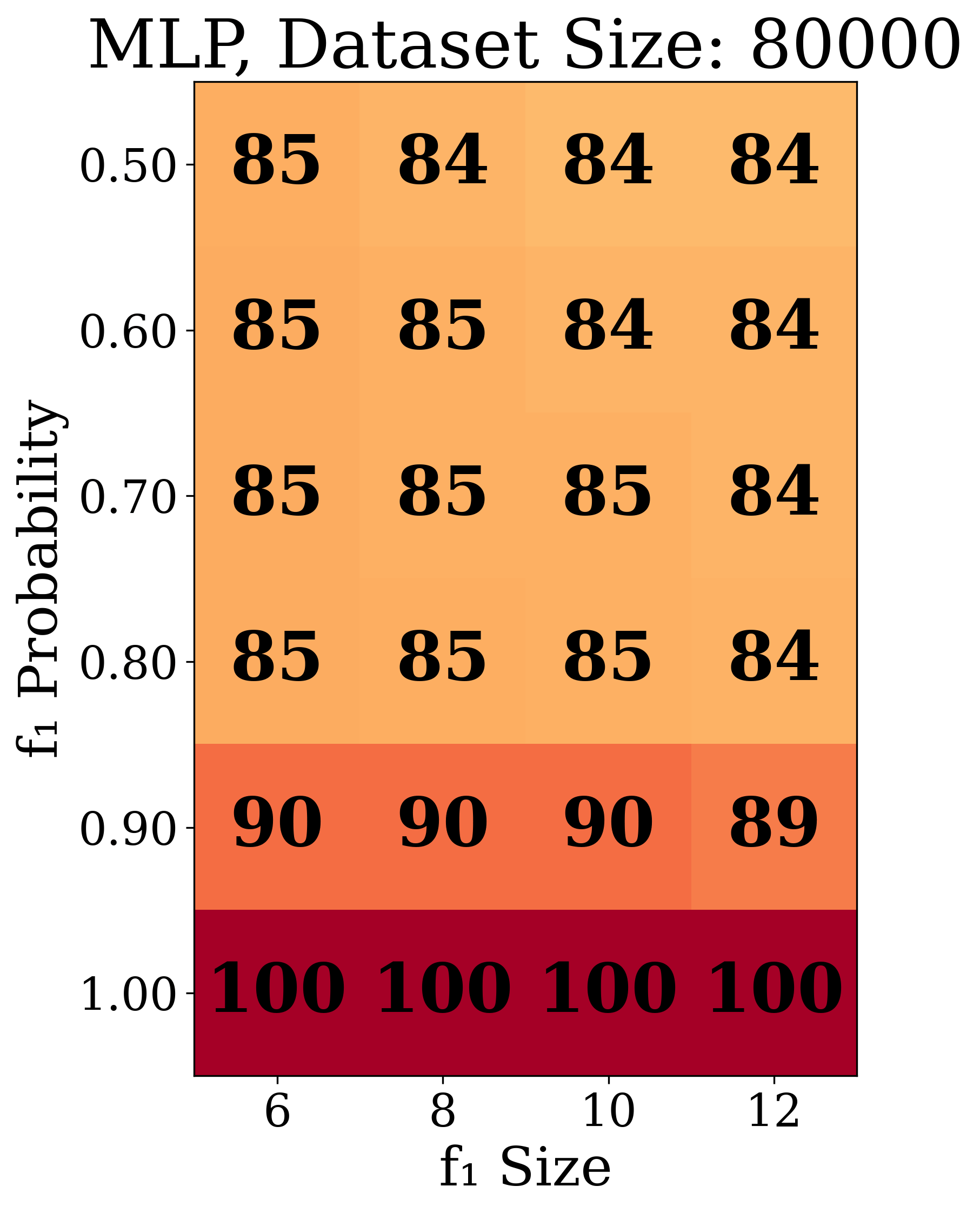}
    \end{subfigure}
    \begin{subfigure}[width=\textwidth]{0.16 \textwidth}
        \includegraphics[width=\textwidth]{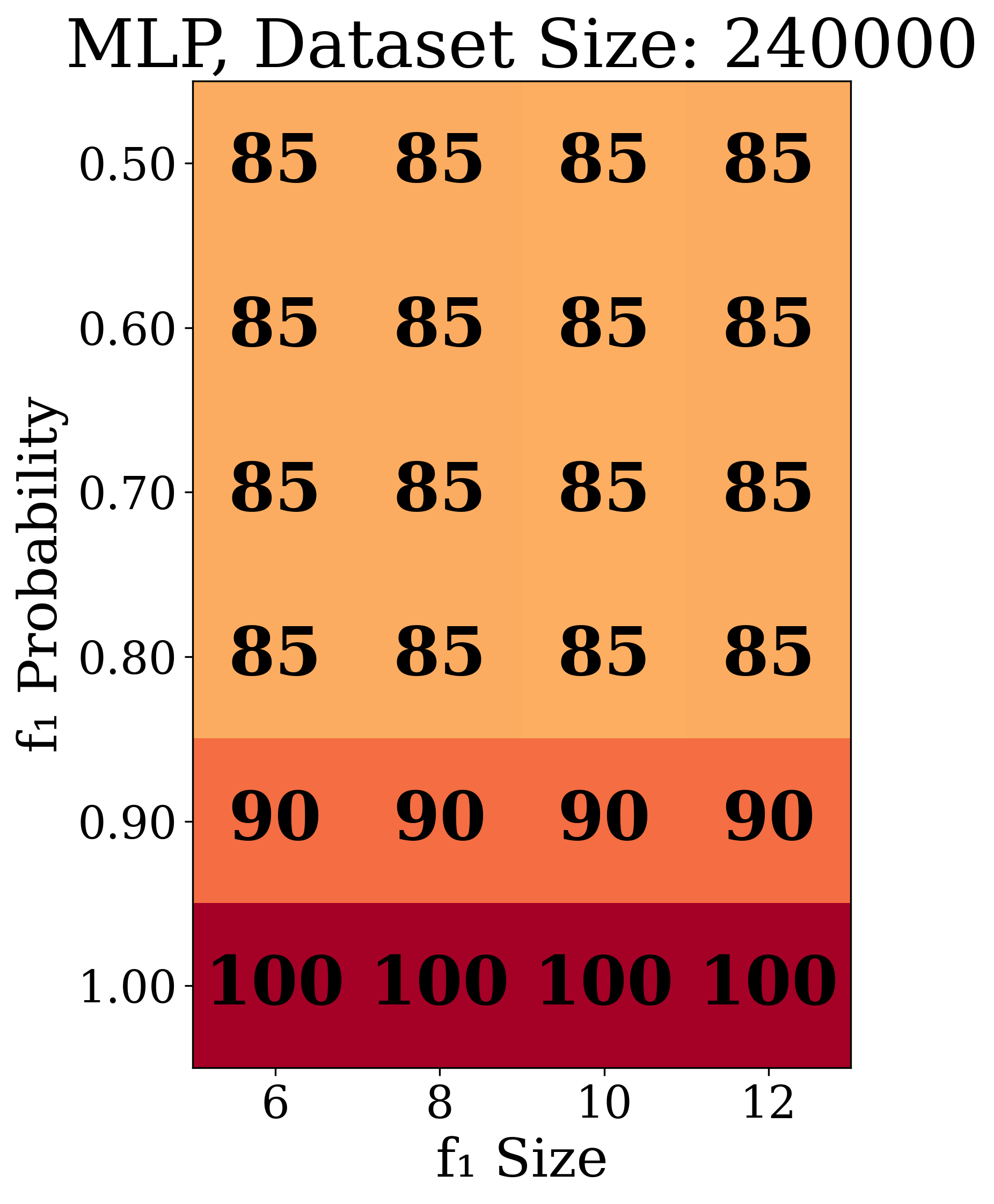}
    \end{subfigure}
  
    \begin{subfigure}[width=\textwidth]{0.16 \textwidth}
        \includegraphics[width=\textwidth]{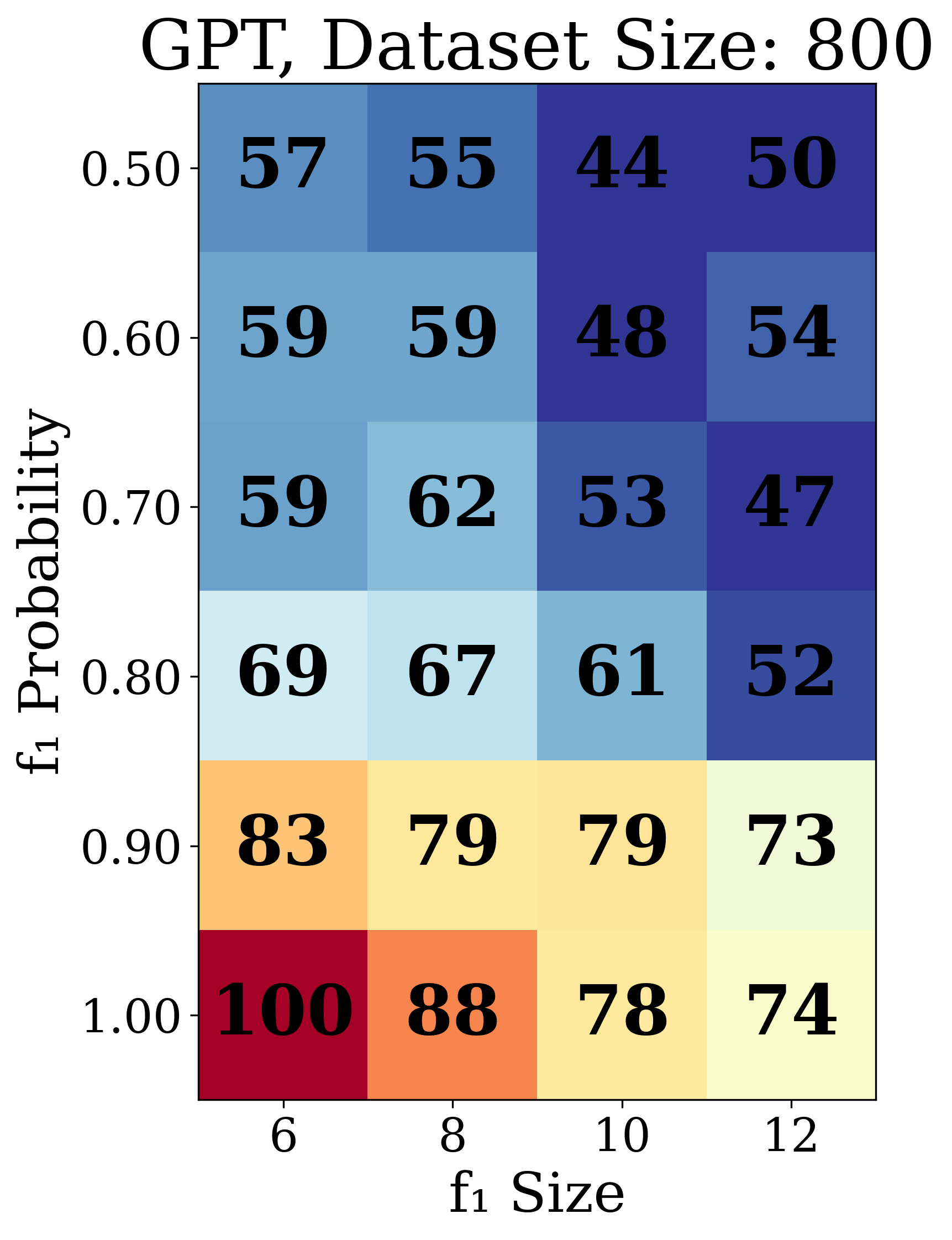}
    \end{subfigure}
    \begin{subfigure}[width=\textwidth]{0.16 \textwidth}
        \includegraphics[width=\textwidth]{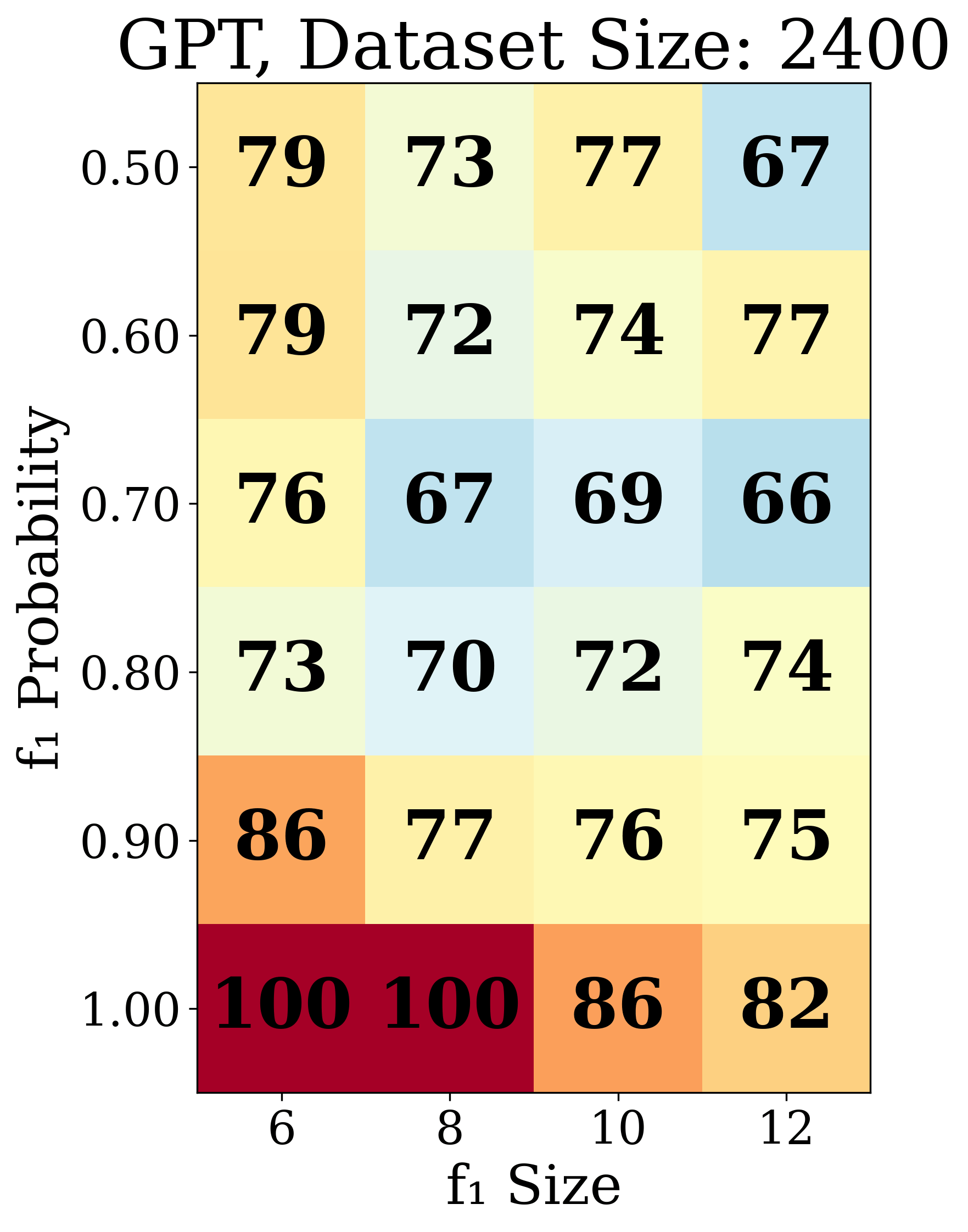}
    \end{subfigure}
    \begin{subfigure}[width=\textwidth]{0.16 \textwidth}
        \includegraphics[width=\textwidth]{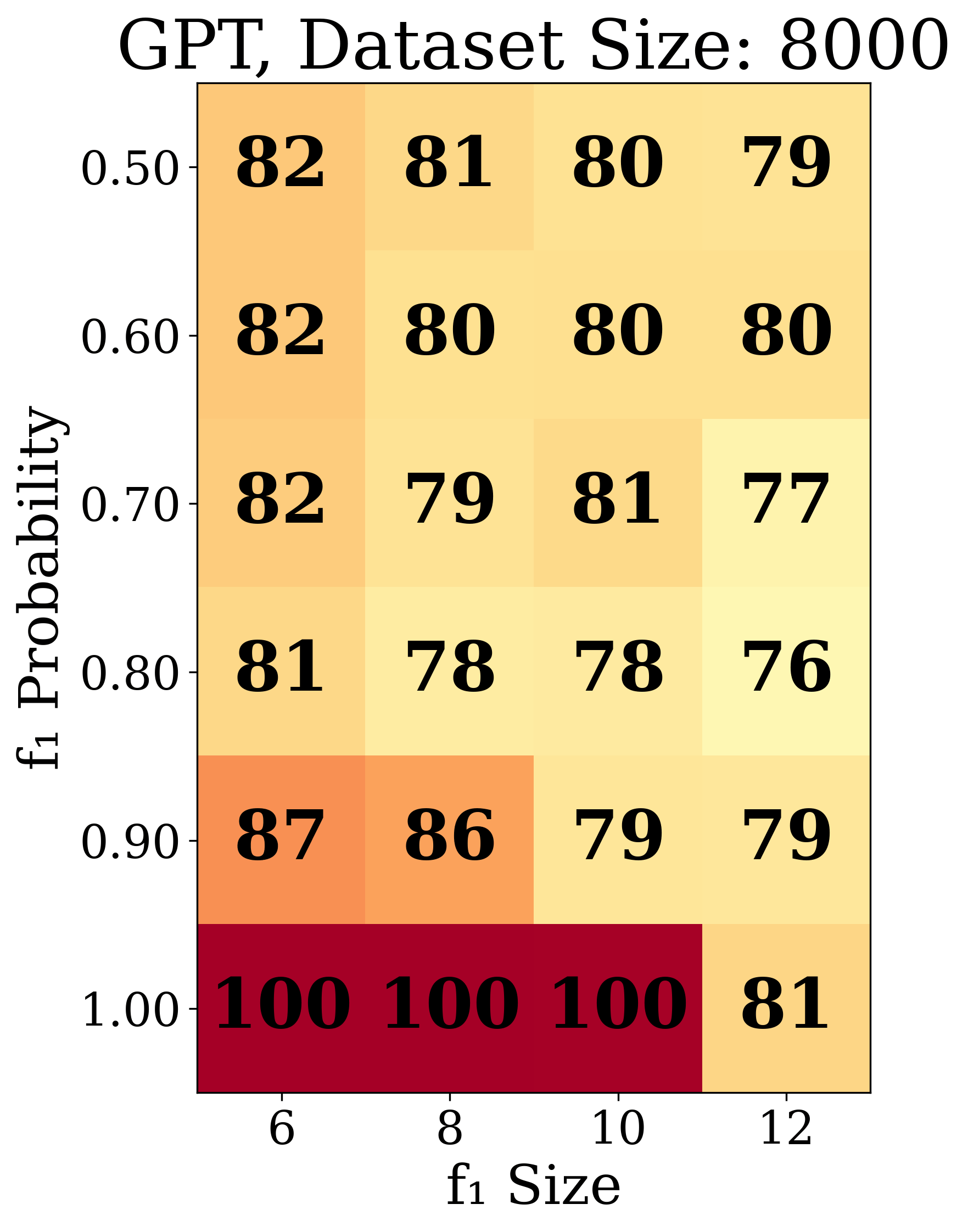}
    \end{subfigure}
    \begin{subfigure}[width=\textwidth]{0.16 \textwidth}
        \includegraphics[width=\textwidth]{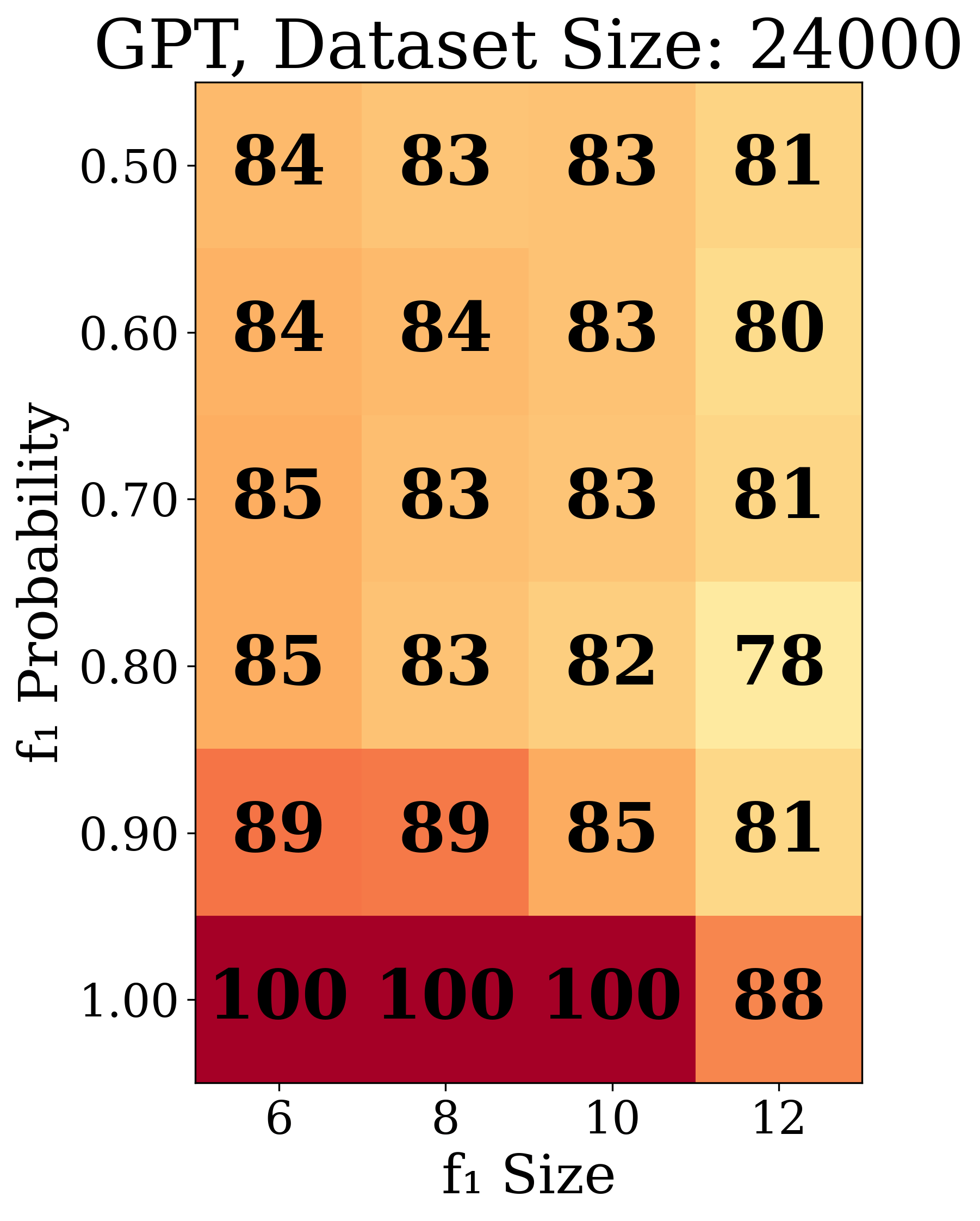}
    \end{subfigure}
    \begin{subfigure}[width=\textwidth]{0.16 \textwidth}
        \includegraphics[width=\textwidth]{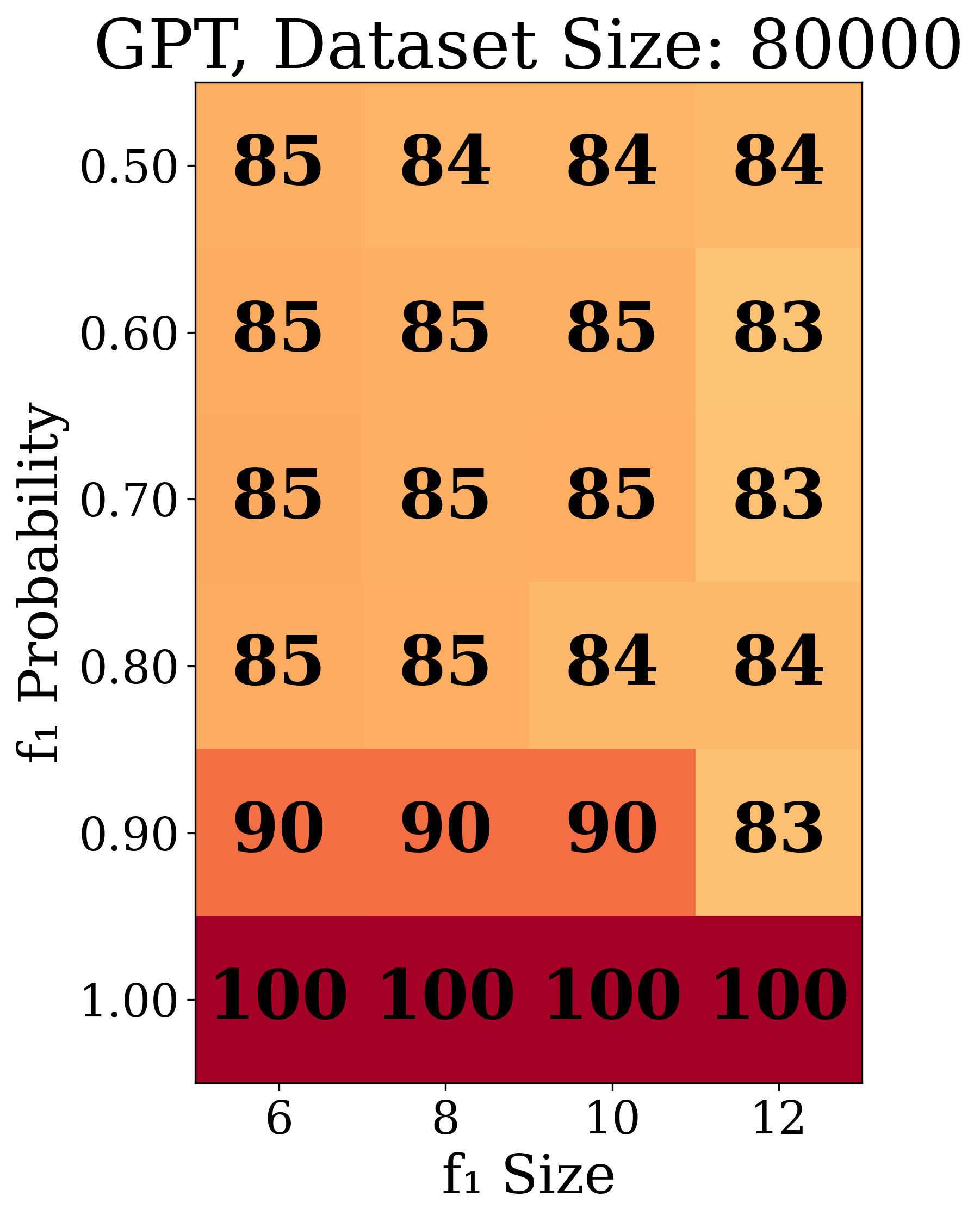}
    \end{subfigure}
    \begin{subfigure}[width=\textwidth]{0.16 \textwidth}
        \includegraphics[width=\textwidth]{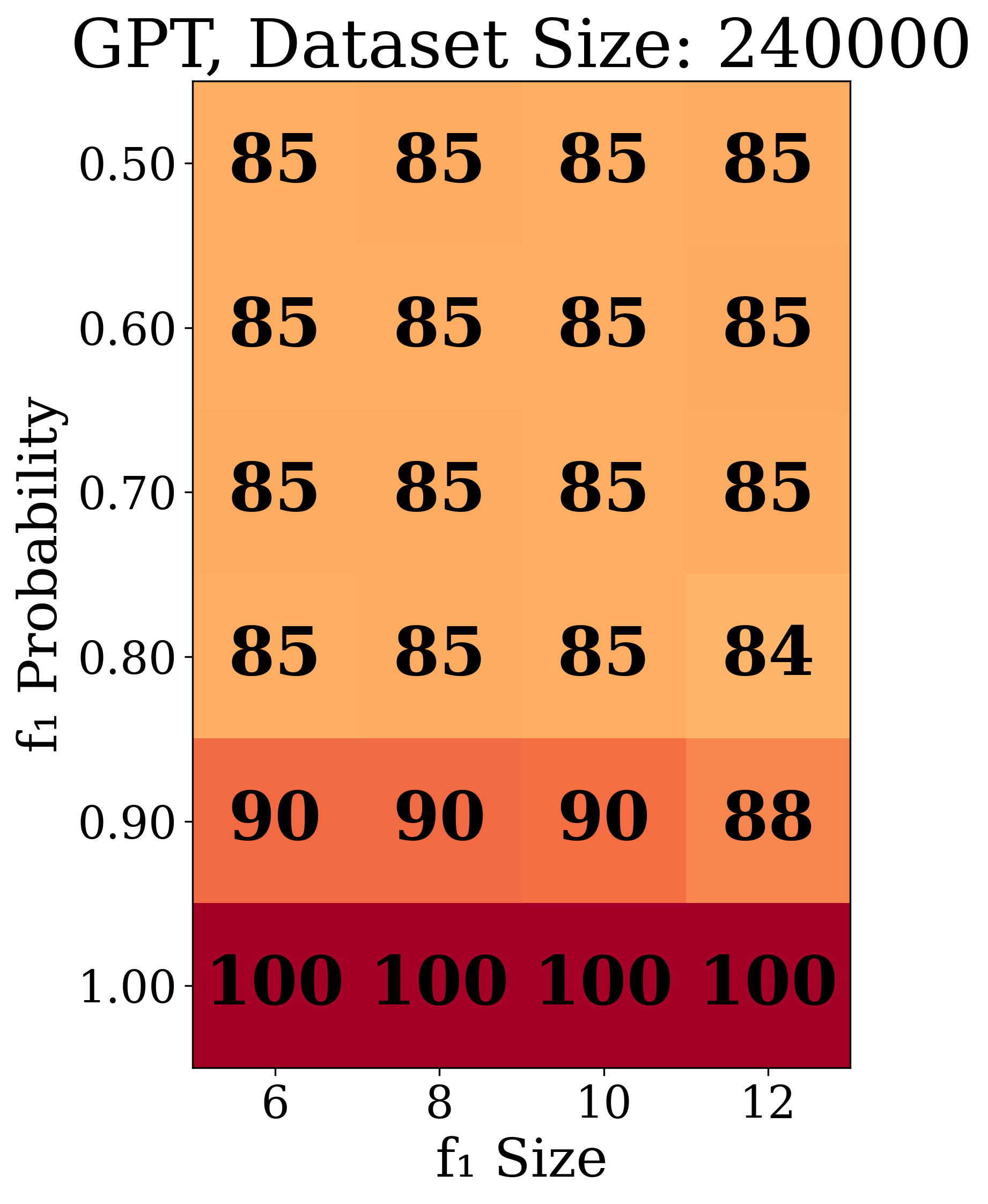}
    \end{subfigure}

    \caption{Two Feature Experiments \textbf{Top:} MLP with constant compute budget $(T=10000)$, \textbf{Bottom:} Transformer with compute budget proportional to data set size ($T=1.25N$).
    Each panel shows test accuracy as a function of $d_1$ and $\alpha_1$.
    Under these different regimes the models exhibit similar learning to use the optimal feature in all cases for $N\geq80,000$.}
    \label{fig:two_features}
\end{figure*}

In this section we perform experiments on data sets that contain exactly two features.
These features are designed to form a strict hierarchy see Sec.
\ref{sec:strict_hierarchical}. 
Similar to \citet{hermann2020shapes}, this lets us probe when the two models learn the most predictive feature.
We construct a data set with a feature $f_1$ which we vary the size and predictivity of. 
Specifically we consider $d_1\in \{6,8,10,12\}$ and $\alpha_1\in \{0.5,0.6,0.7,0.8,0.9,1.0\}$. 
The other feature, $f_2$ has a fixed size of $d_2=8$ bits and a predictivity of $\alpha_2=0.85$.
The test accuracy of the resulting two models are shown in Figure \ref{fig:two_features} for a variety of sizes of the training set.
Here we only scale the training iterations of the GPT model in line with the data set size; the MLP uses the fixed budget $T=10^4$. 

Under this regime the models exhibit similar biases resulting in similar test performances at different scales and predictivities and availabilities for $f_1$. 
While slight, the GPT model typically performs better at learning $f_2$ when $f_1$ has lower predictivity, whereas the MLP typically does beResults on data set containing five weak features and one strong featuretter at learning $f_1$ when it has high predictivity but low availability.
\footnote{For the results when $\alpha_1=0.5$, $f_1$ is indifferent from random noise, and hence we can effectively observe the results of learning noisy k-parities with different degrees of sparsity. Specifically, with $k=4$ and $d = \{14,16,18,20\}$ bits. In these results we show that the sparsity factor typically has less pronounced effect on the results, it being most pronounced on the limited data regimes. This observation in part motivated our choice to avoid further investigations on the effect of sparsity.} 
Results for the GPT model with a fixed training budget are shown in Figure \ref{fig:two_features_gpt_fixed} within the Appendix.
With a fixed budget we see more training data slows the GPT training to a point where a smaller data set size is optimal.

\begin{takeawaybox}[title=Key Finding 2]
When learning noisy k-parity features Transformers are less sensitive to noise (lower predictivity) but are more sensitive to feature size (lower availability) relative to MLPs.
Increasing the data set size amplifies these differences.  
\end{takeawaybox}

\subsection{Data Sets with Larger Numbers of Features}
We now turn our attention to data sets with more than two features.
These data sets are designed to highlight difference in the SB of the two models and be more representative of hard data sets that one might encounter in practice.

\paragraph{One Strong vs Many Weak Features.}
Here we construct data sets containing six features $\mathcal{F} = \{f_1, \dots, f_6\}$, five ``weak'' features and one ``strong'' feature.
The features $f_1,\dots,f_5$ are the ``weak'' features and have low predictivities, $\alpha_i = 0.6$ for $ i \in \{1,\dots,5\}$, but high availability.
This high availability, is achieved by using a low number of bits.
We run experiments with $d_i \in \{2,4,6\}$ for $i \in \{1,\dots,5\}$.
We additionally have one ``strong'' feature with perfect predictivity $\alpha_6 = 1.0$ whose availability we vary, specifically $d_6 \in \{2,4,\dots,14\}.$ We train models for a longer $10^5$ iterations, at two data scales $N=8000$ and sampling batches directly from the DGP. 

\begin{figure*}[ht]
    \centering
    \begin{subfigure}[width=\textwidth]{0.24\textwidth}
        \includegraphics[width=\textwidth]{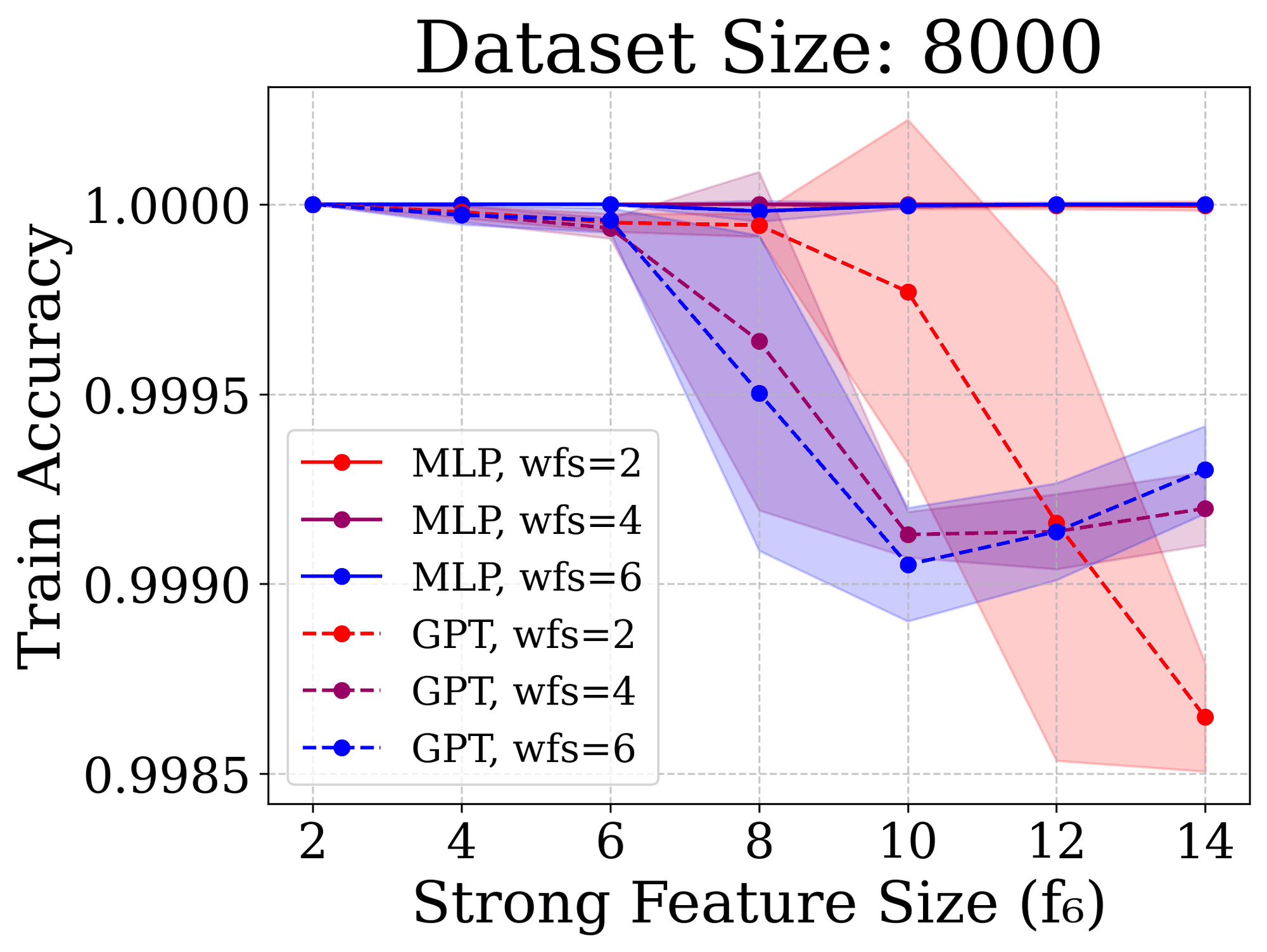}
    \end{subfigure}
        \begin{subfigure}[width=\textwidth]{0.24\textwidth}
        \includegraphics[width=\textwidth]{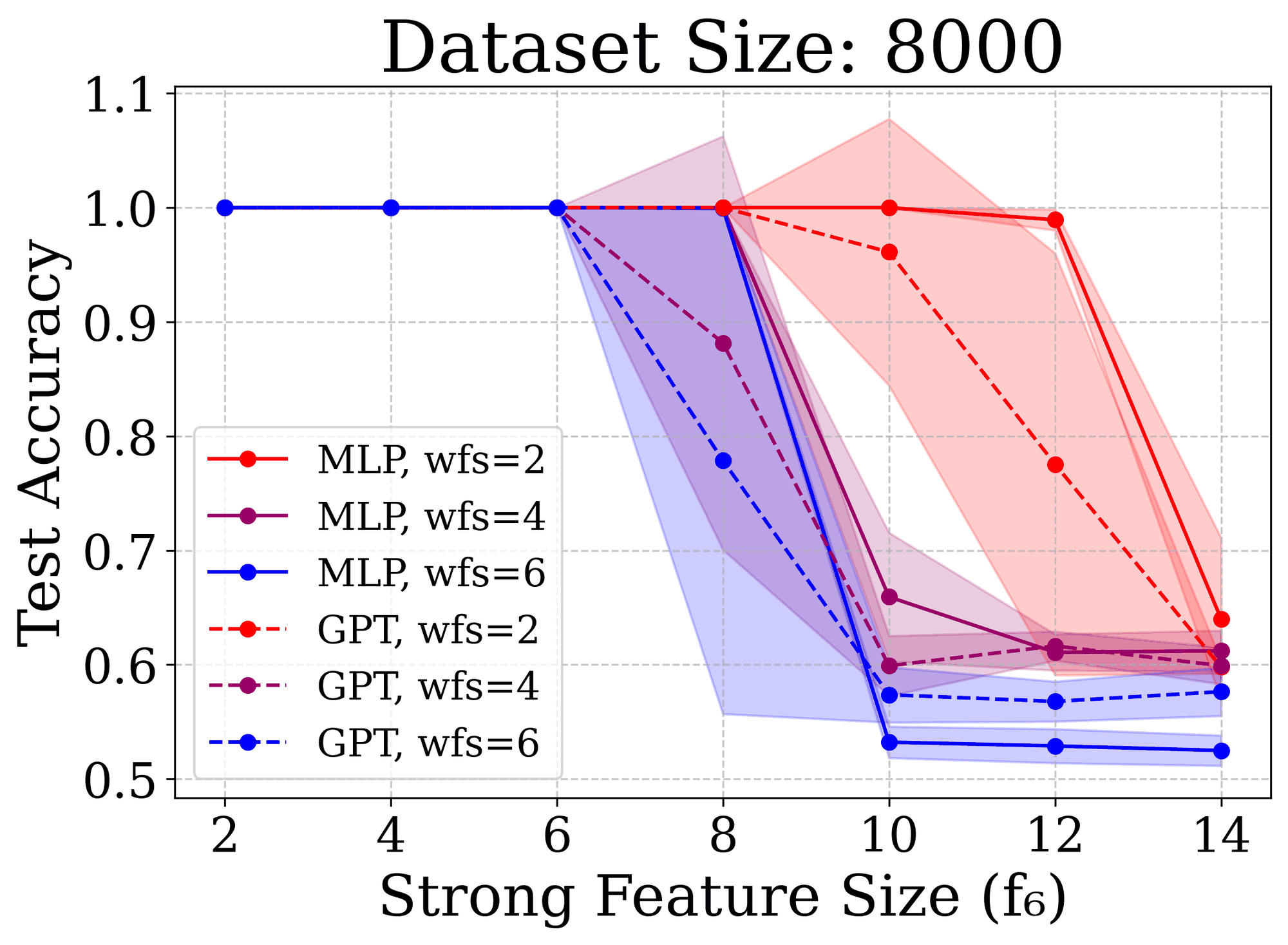}
    \end{subfigure}
    \begin{subfigure}[width=\textwidth]{0.24 \textwidth}
        \includegraphics[width=\textwidth]{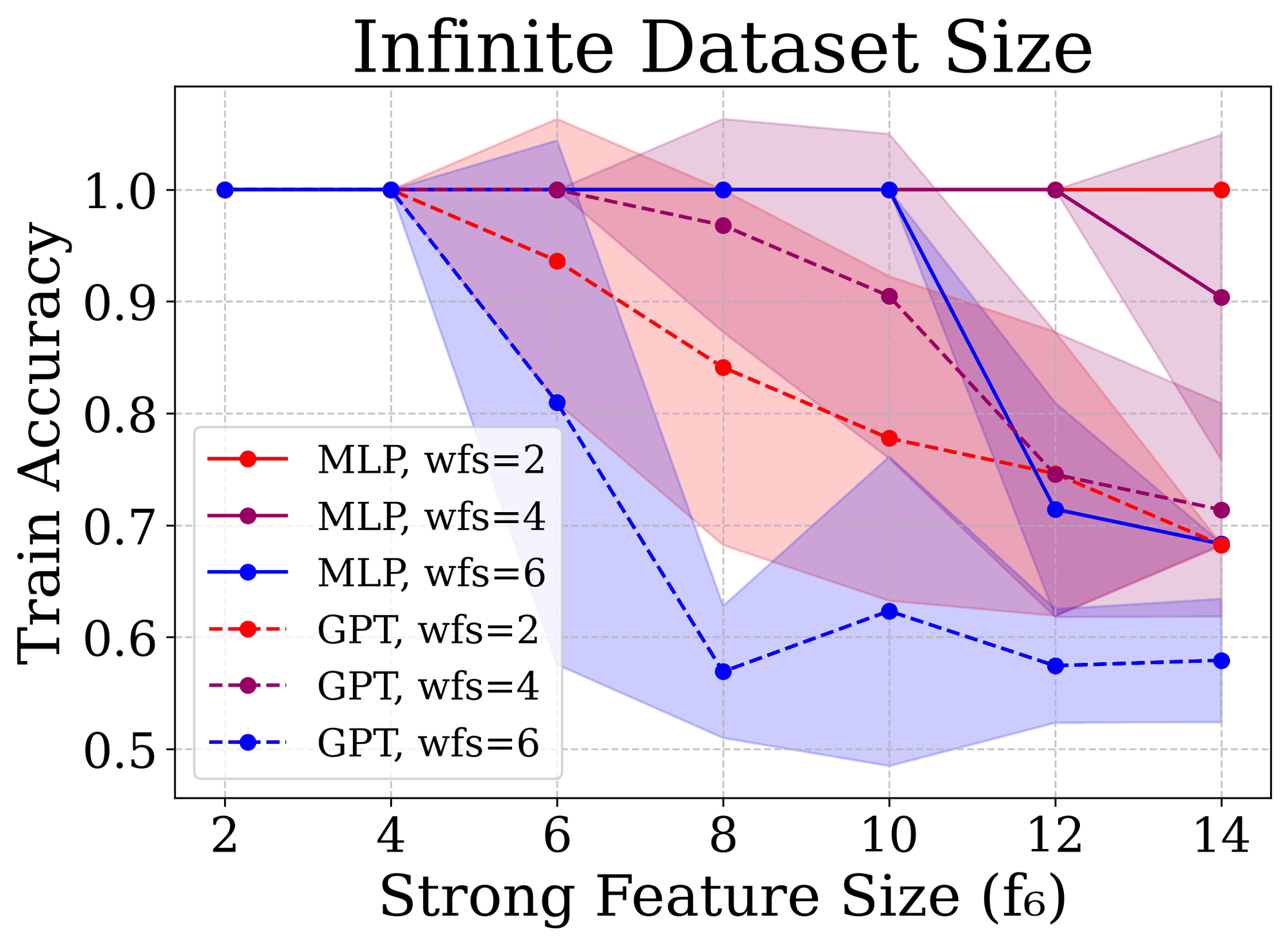}
    \end{subfigure}
        \begin{subfigure}[width=\textwidth]{0.24 \textwidth}
        \includegraphics[width=\textwidth]{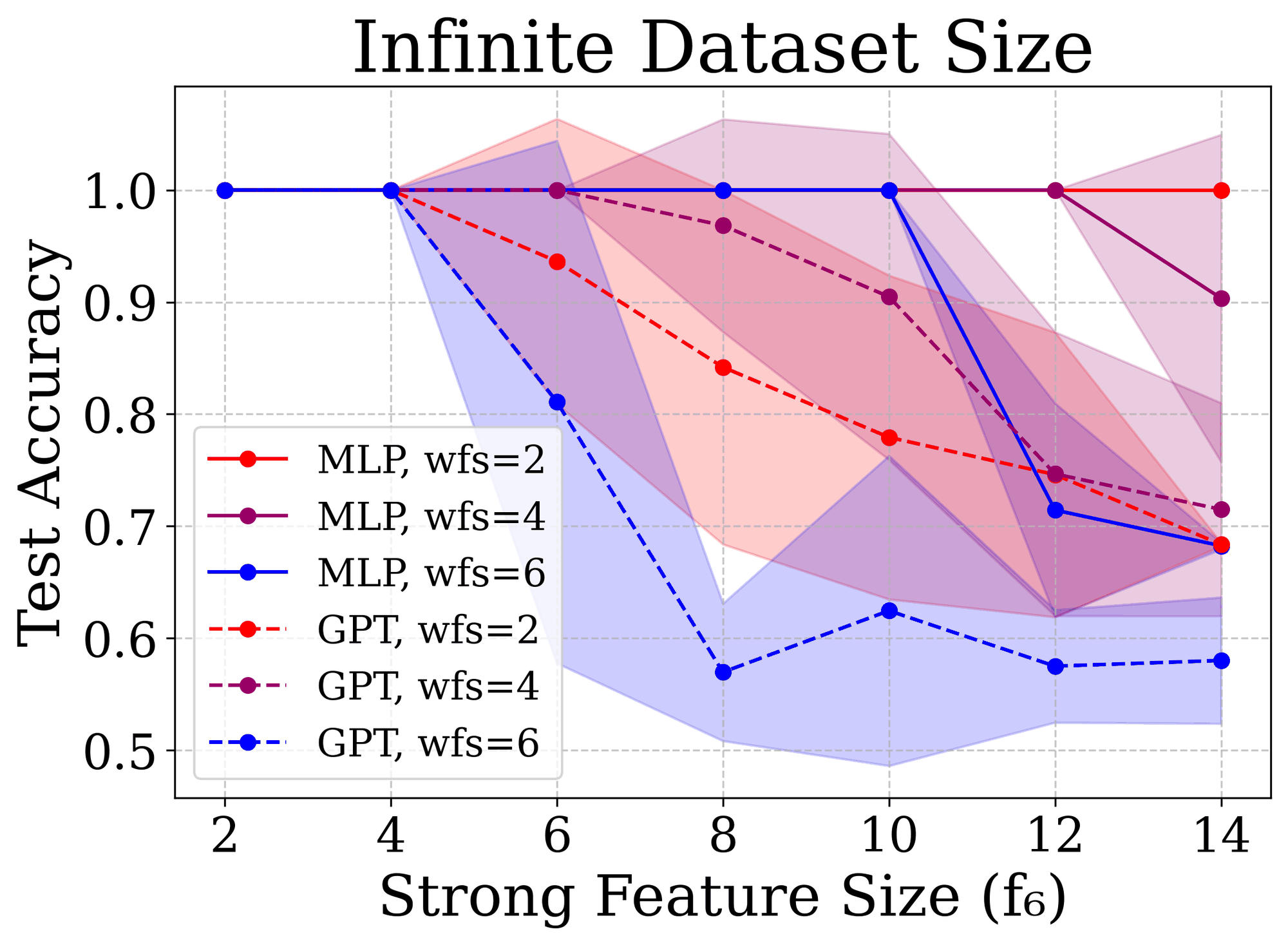}
    \end{subfigure}
    \caption{Results on data settings containing five weak features and one strong feature ($T=10^5$, fixed budget).
    \textbf{Left two panels:} finite training set.
    \textbf{Right two panels:} batches sampled directly from the DGP ($N=\infty$).
    When highly complex the transformer fails to learn the strong feature at all data scales.
    \emph{``wfs'': weak feature size.}}
    \label{fig:weak_vs_strong}\vspace{-0.3cm}
\end{figure*}

We report the test accuracies in Figure \ref{fig:weak_vs_strong}. 
For $(N=8000)$, despite both models achieving $\sim 99.9\%$ for all three data sets they struggled to learn to use $f_6$ for $d_6\geq 10$, except in the case where $d_1,\dots,d_5 = 2$ where both models only fail to learn $f_6$ for $d_6\geq 14$. 
The models simply do not have enough data to learn to correctly use $f_6$ which is required to have good generalisation performance. 
When training directly on the DGP we see the MLPs accuracy improve across the board.
However, the Transformer often performs worse, despite the greater data diversity and larger training budget.
The strong bias for learning simple functions means in the cases when there are no predictive simple functions, such as when memorisation is not possible, transformers struggle to learn anything at all.

\paragraph{Hierarchical Data Set.}
Finally, we investigate the models on a data set where the features form a strict hierarchy see Sec.
\ref{sec:strict_hierarchical}, that is, there are five features that become more predictive but less available.
In order to correctly classify all examples the models must rely on the most predictive features, as for certain fractions of the data set the weaker features are totally uncorrelated with the label.
Table \ref{tab:data_lin} in the Appendix shows the Availability and Predictability of the five features.

\begin{figure*}[ht]
    \centering
    \begin{subfigure}[]{0.24 \textwidth}
        \includegraphics[width=\textwidth]{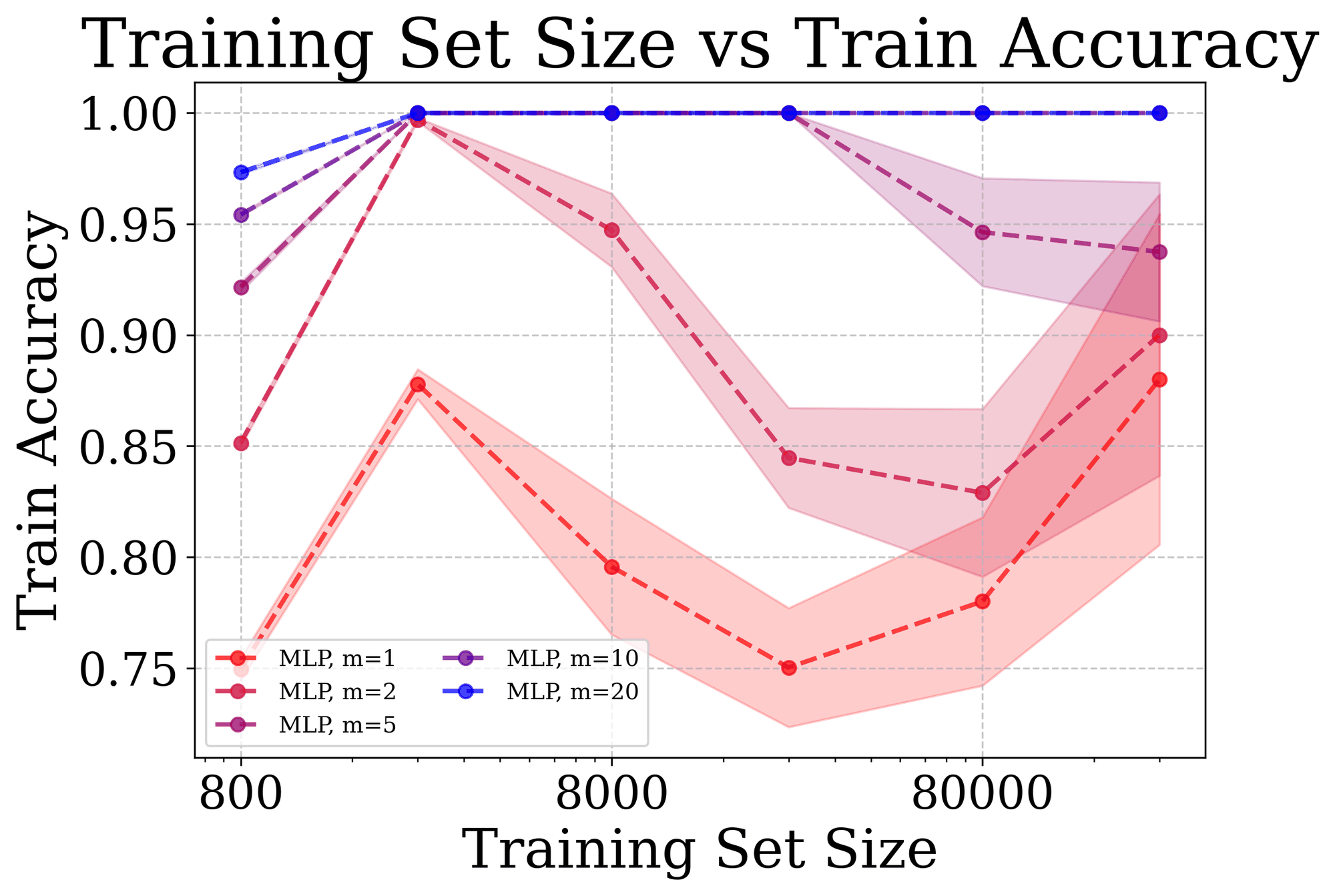}
    \end{subfigure}
    \begin{subfigure}[]{0.24 \textwidth}
        \includegraphics[width=\textwidth]{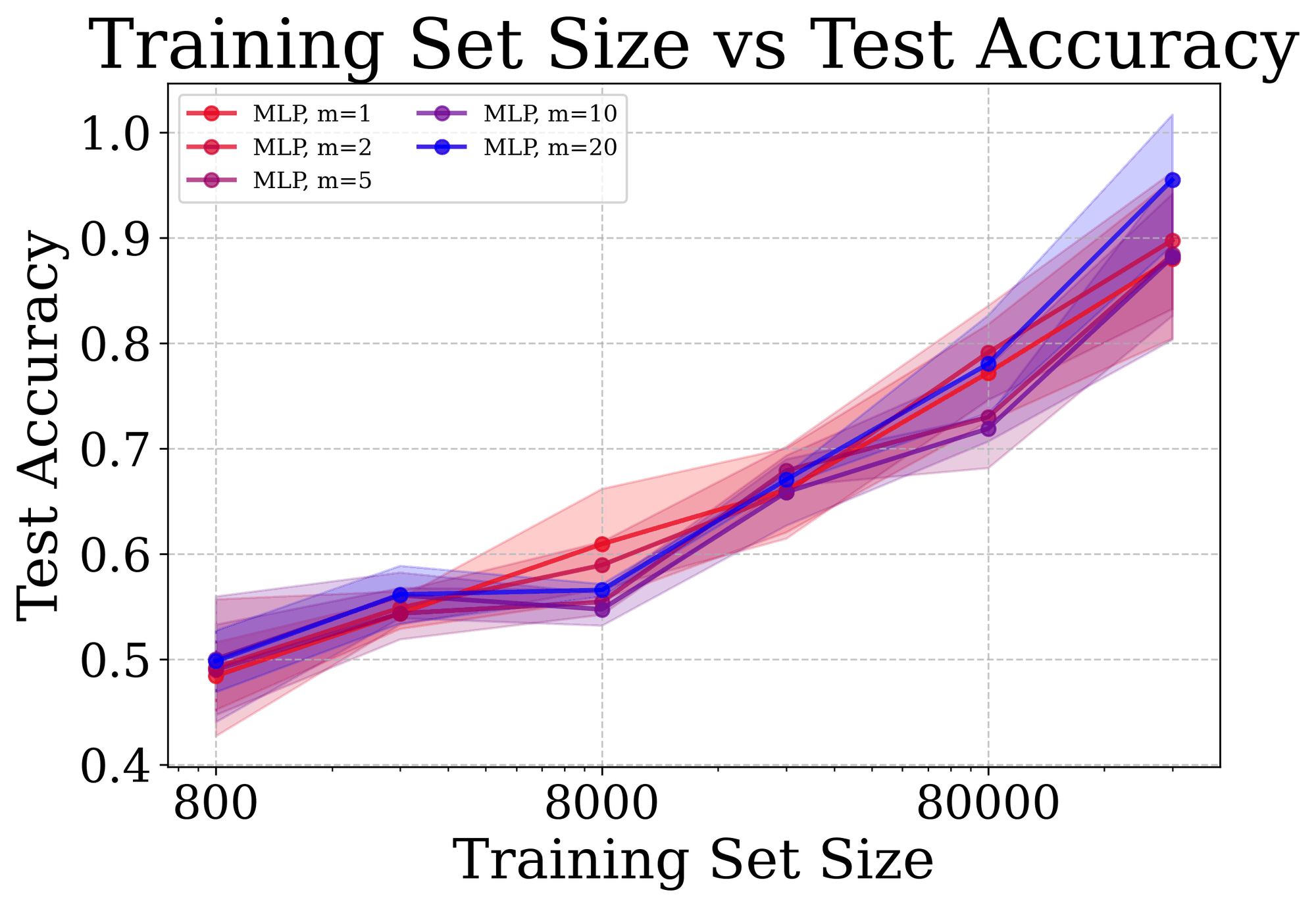}
    \end{subfigure}
    \begin{subfigure}[]{0.24 \textwidth}
        \includegraphics[width=\textwidth]{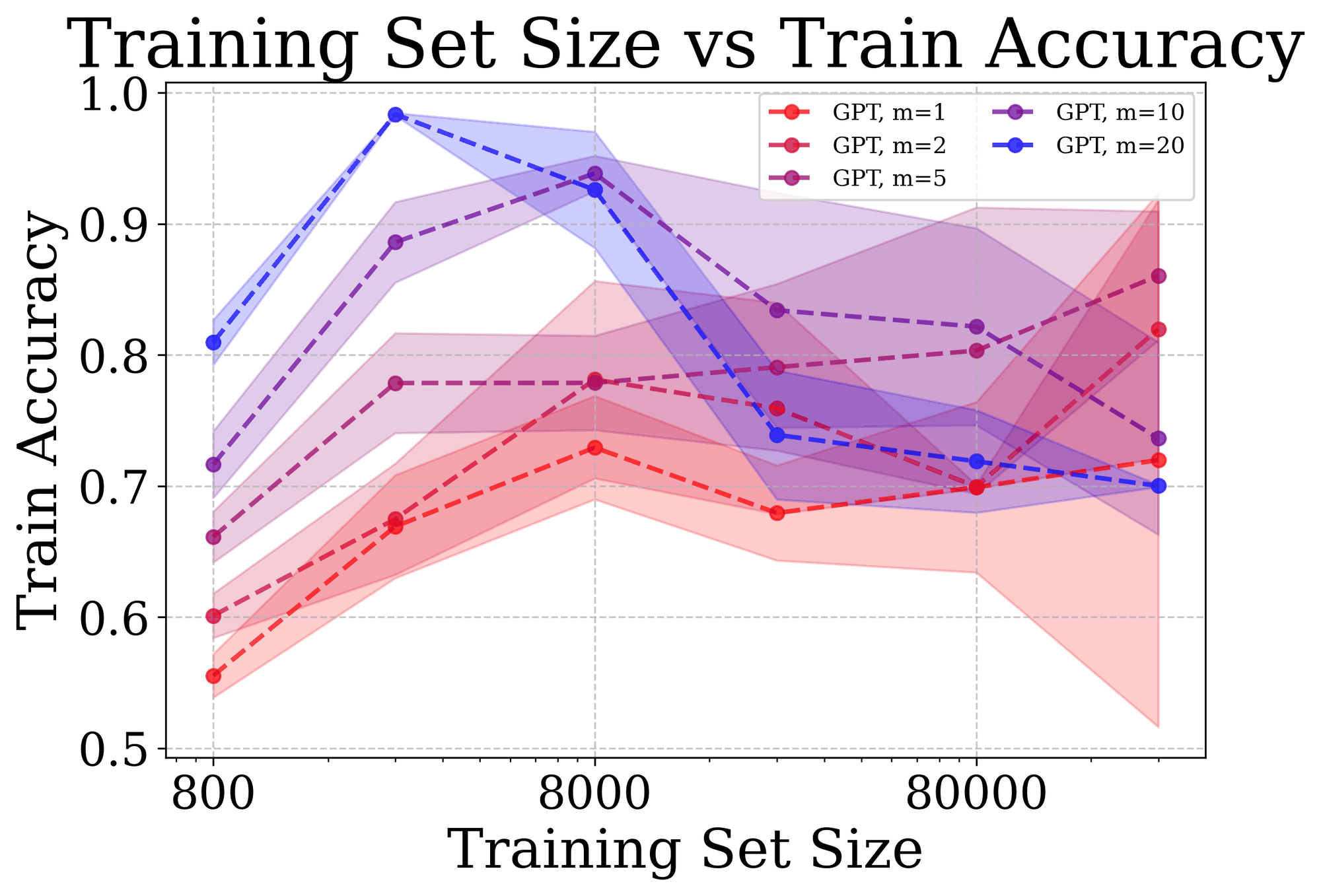}
    \end{subfigure}
    \begin{subfigure}[]{0.24 \textwidth}
        \includegraphics[width=\textwidth]{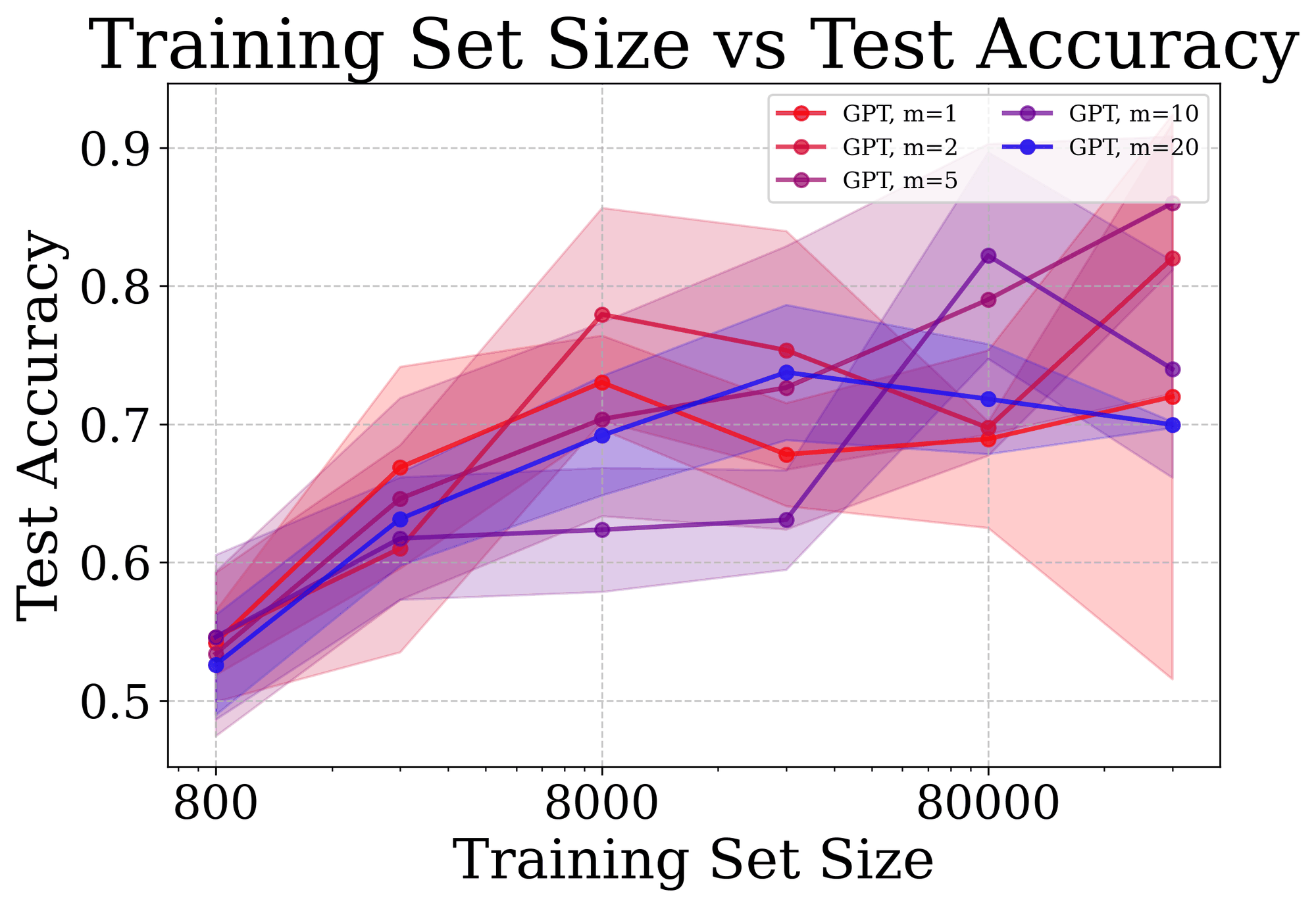}
    \end{subfigure}
    \caption{Train (first and third panels) and test (second and fourth panels) accuracy on the tiered data set with a scaled iteration budget ($T=1.25N$).
    \textbf{Left two panels:} MLP.
    \textbf{Right two panels:} GPT.}
    \label{fig:tiered_body}
\end{figure*}

The results for models at various scales trained on this data set are shown in Figure \ref{fig:tiered_body} with a scaling iteration budget.
The full results can be found in Figures
\ref{fig:tiered_full} and \ref{fig:tiered_iter_scaling_full} within the Appendix. 
Data sets of this form are very difficult for models to learn as by definition the weaker features act as shortcuts. 
This is confirmed by no model achieving close to $100\%$ generalisation performance despite the train and test data sets being i.i.d. and the data set only having an effective dimensionality of 40. 
We again see marked differences between the two models. 
The GPT model performs better at low data regimes (the $m=2$ model) achieving close to $78\%$ accuracy when trained with 8000 samples, far above the best performing MLP. 
At this low scale the MLP models nearly all overfit to the training data and generalise poorly.
Conversely, for larger data scales the MLP model dominates the Transformer which never interpolates the training set. 
Again the Transformer often has worse training performance with more training data despite the increasing number of training iterations used.
Additionally the optimal size of data set for the transformer depends on the size of model being used with only two of the model sizes performing best on the largest training set.
Lastly, in our experiments MLPs were a more parameter and compute efficient approach for learning k-parities than Transformers, which differs from what the asymptotic theoretical results of \citet{han2025attention} might suggest, see Figure \ref{fig:scatter} within the Appendix.

\vspace{-0.1cm}
\begin{takeawaybox}[title=Key Finding 3]
The same extreme SB that makes Transformers excel at learning noisy data makes them poorly suited for learning certain classes of highly complex non-linear features in the presence of shortcuts, where they will often fail to minimise the loss.
\end{takeawaybox}

\section{Discussion}\label{sec:discussion}
Our experiments separate the effects of repetition, undertraining and diversity (Section \ref{sec:k_parity_experiments}) and point towards a possible mechanism without yet identifying one.
A natural hypothesis, consistent with the theoretical literature on parity learning \citep{barak2022hidden,han2025attention}, is that the attention mechanism lets the Transformer identify the relevant coordinates quickly, but that this process depends on the signal-to-noise ratio of the gradient updates, which falls as the data become more diverse.
Testing this hypothesis through gradient statistics or Fourier-gap measurements is a promising direction that the present testbed is well suited to support.

K-parity is motivated as a proxy for problems that require axis-aligned feature learning \citep{edelman2023pareto}, and our study follows the established practice of using small synthetic tasks to understand phenomena, such as grokking, that are hard to observe under standard training regimes but are informative about the behaviour of these architectures \citep{power2022grokking, nanda2023progress, singh2026explaining}.
We therefore view the results as a controlled study of two architecture families on a hard class of functions.
We speculate similar effects would be most likely to appear in tabular or structured settings with limited compute budgets; natural data, whose feature distributions are typically long-tailed, and other nonlinear functions (for example cryptographic or other algorithmically generated functions) would be valuable places to look next.

\section{Limitations}\label{sec:limitations}
The scope of our conclusions is set by the experimental design, which we chose to keep tightly controlled.
The ``less data is more'' effect is a statement about optimisation under a fixed iteration budget: smaller data sets are preferable because they are optimised faster rather than because they yield a better final model, and with a scaled budget larger data sets extend the frontier of learnable features.
In several settings, for example large $N$ or large $d$, the models do not fit the training set within the budget (Appendix \ref{app:train_tables}), so conclusions in these regimes describe optimisation speed under our budgets rather than ultimate capacity.
Our scaled-budget experiments hold the expected number of passes per example constant, and a fixed-FLOPs comparison that accounts for differences in per-step cost would complement them.

The comparison between architectures is intended to highlight the differences between these models rather than be taken as a head to head evaluation.
The GPT model has about 35 times more parameters than the MLP at the same width factor, and the two differ in input format, regularisation and learning rate, so the differences we observe reflect the architecture class together with these choices.
Similarly, all experiments use Adam (AdamW for the GPT model).
We avoided SGD because it is highly sensitive to the learning rate \citep{schaul2013no} and is known to perform poorly on parity learning, so a failure to learn would not cleanly isolate the effect of data diversity, and extending the study to SGD with momentum or sign-based optimisers would show how general the effect is, but we leave this to future work.

Finally, we study one decoder-only Transformer family (a minGPT variant) and one MLP family at small scale compared with frontier models \citep{bommasani2021opportunities}, on data sets with noisy k-parity features and uniformly sampled inputs.
Other Transformer variants, the effects of batch size, further hyperparameters and k-parity sparsity are natural next steps.
The current investigation comprised over 50,000 training runs, the longest requiring over two GPU days.

\section{Conclusion}\label{sec:conclusion}
We investigate the performance of MLPs and Transformers on data sets where every feature is a carefully chosen noisy k-parity across a number of model sizes and data scales.
We show across multiple data sets that i) under a fixed training budget, increasing data diversity leads to markedly slower training for Transformers, so that smaller data sets can perform better; with a scaled budget more data extends the frontier of learnable features but larger data sets still optimise more slowly; and ii) the same strong simplicity bias of Transformers that makes them well suited for learning from noisy data makes them poorly suited for learning complex low-availability features.
Our findings give further explanation for phenomena where Transformers often encode algorithms using simple heuristics rather than complex algorithms \citep{nikankin2024arithmetic}.

\section*{Broader Impact Statement}
This work is a fundamental empirical study on synthetic data and does not involve personal data, human subjects or deployed systems.
We do not foresee direct negative societal impacts.
More broadly, understanding when and why different architectures prefer simple features may help practitioners anticipate shortcut learning and design more reliable training data, although we have not tested whether our findings transfer to natural data.


\bibliography{refs}
\bibliographystyle{tmlr}

\newpage
\appendix

\section*{Appendix Roadmap}
\begin{figure*}[h]
    \centering
    \begin{subfigure}[width=\textwidth]{0.48 \textwidth}
        \includegraphics[width=\textwidth]{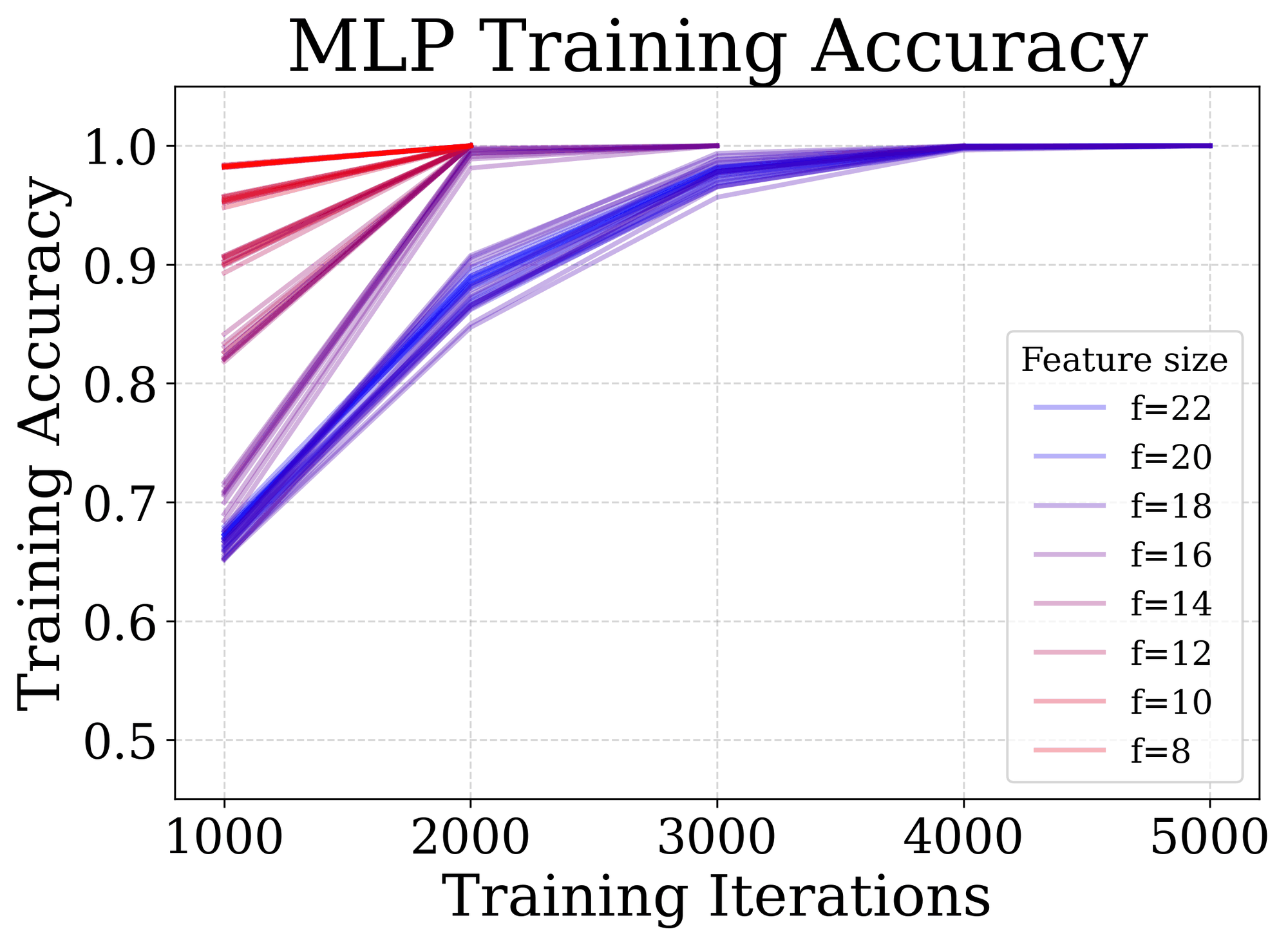}
    \end{subfigure}
    \begin{subfigure}[width=\textwidth]{0.48 \textwidth}
        \includegraphics[width=\textwidth]{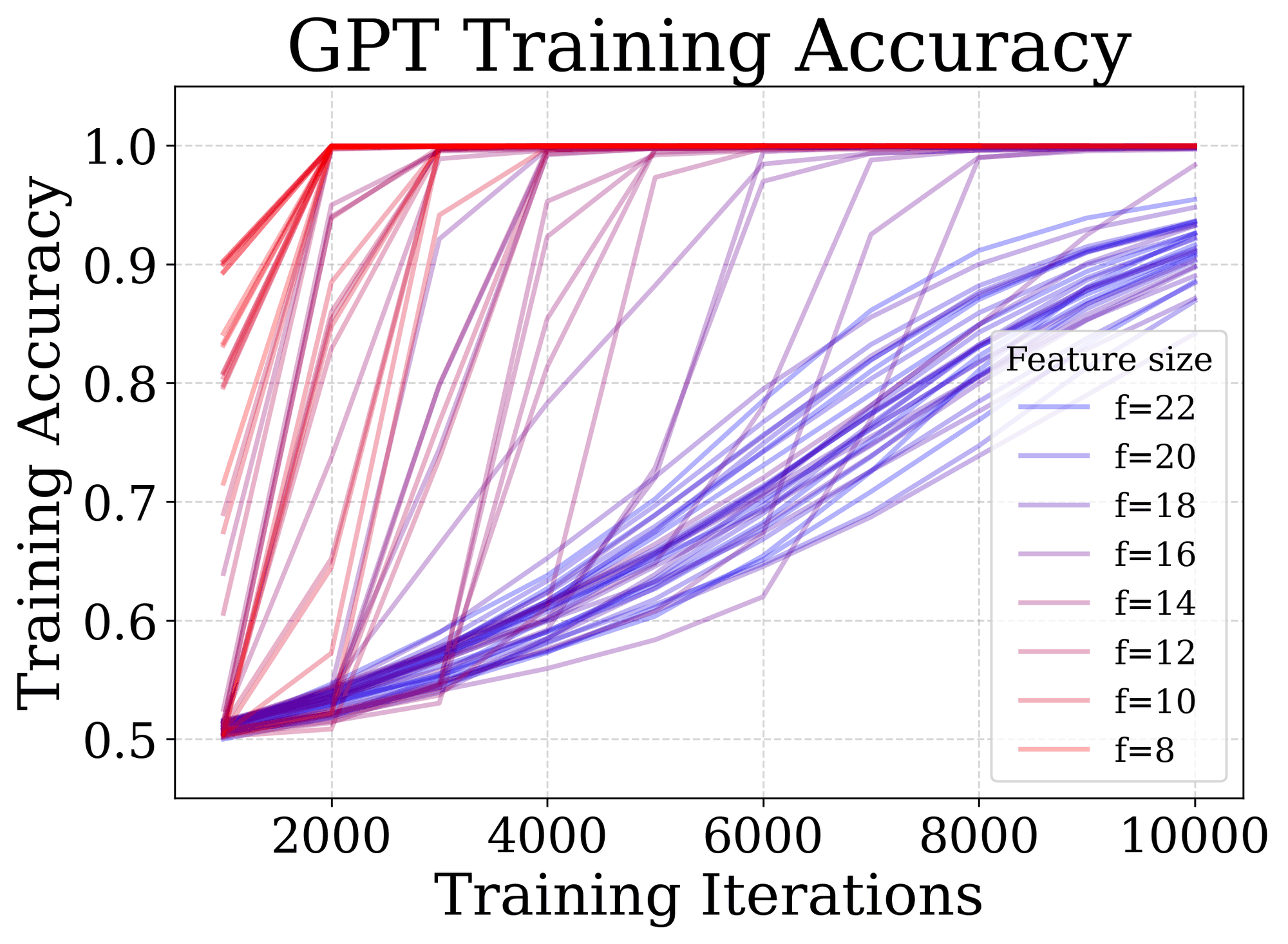}
    \end{subfigure}
    \caption{Training Plots for dataset size of $N=8000$}
    \label{fig:training_curves}
\end{figure*}

\begin{figure*}[h]
    \centering
    \begin{subfigure}[width=\textwidth]{0.48 \textwidth}
        \includegraphics[width=\textwidth]{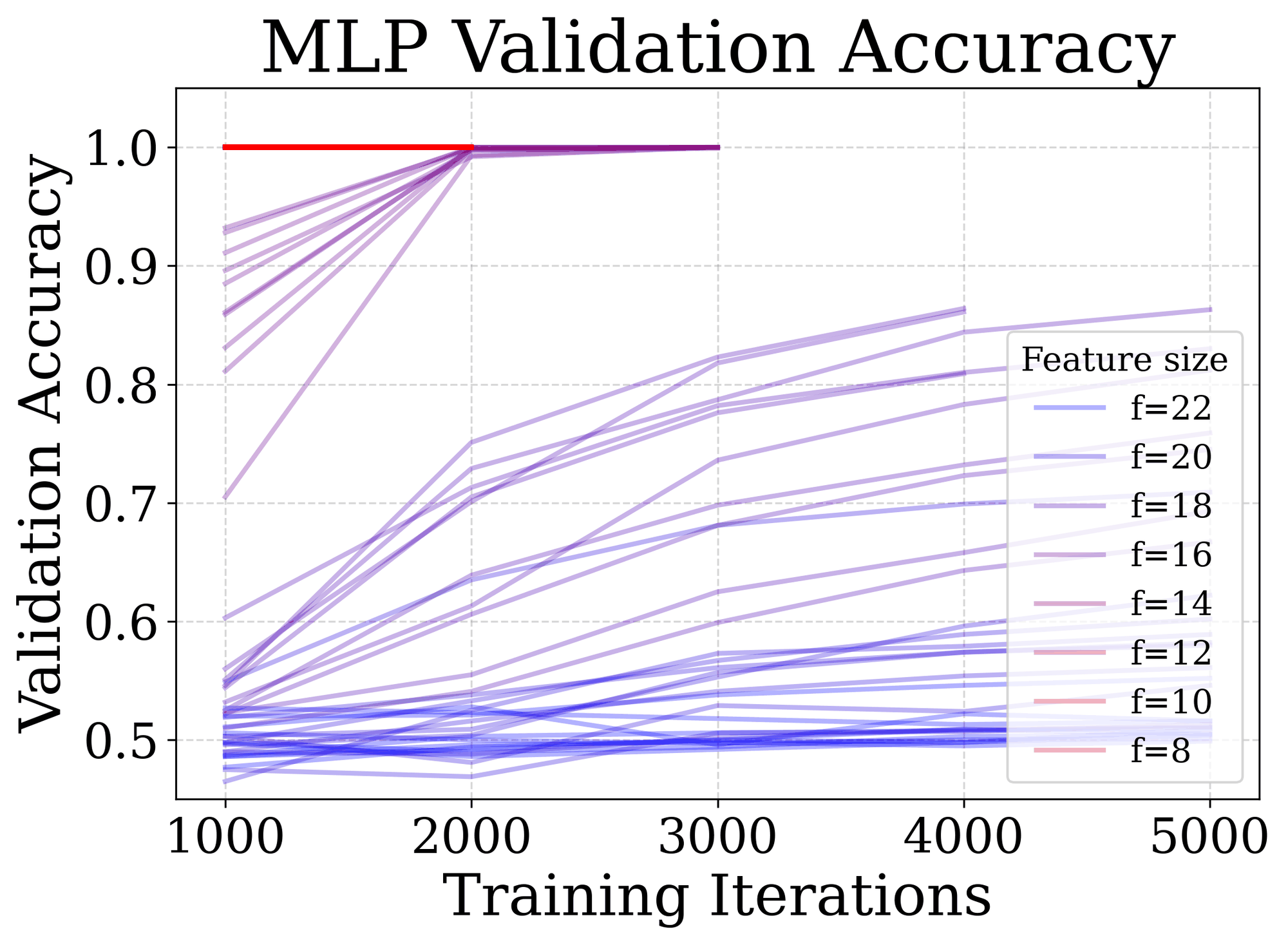}
    \end{subfigure}
    \begin{subfigure}[width=\textwidth]{0.48 \textwidth}
        \includegraphics[width=\textwidth]{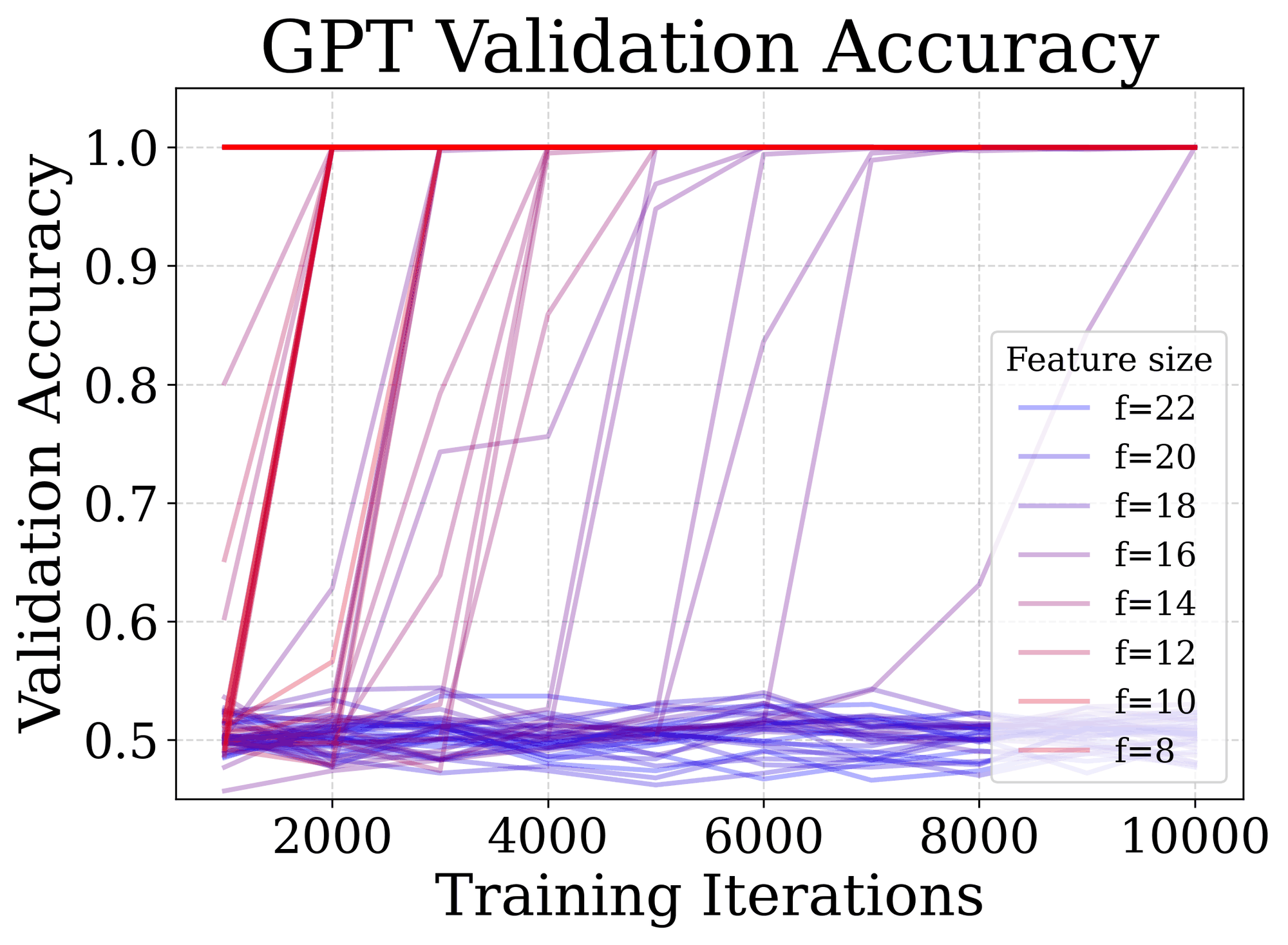}
    \end{subfigure}
    \caption{Validation Plots for dataset size $N=8000$}
    \label{fig:validation_curves}
\end{figure*}

\begin{figure}[h]
    \centering
    \begin{subfigure}[width=\textwidth]{0.48 \textwidth}
        \includegraphics[width=\textwidth]{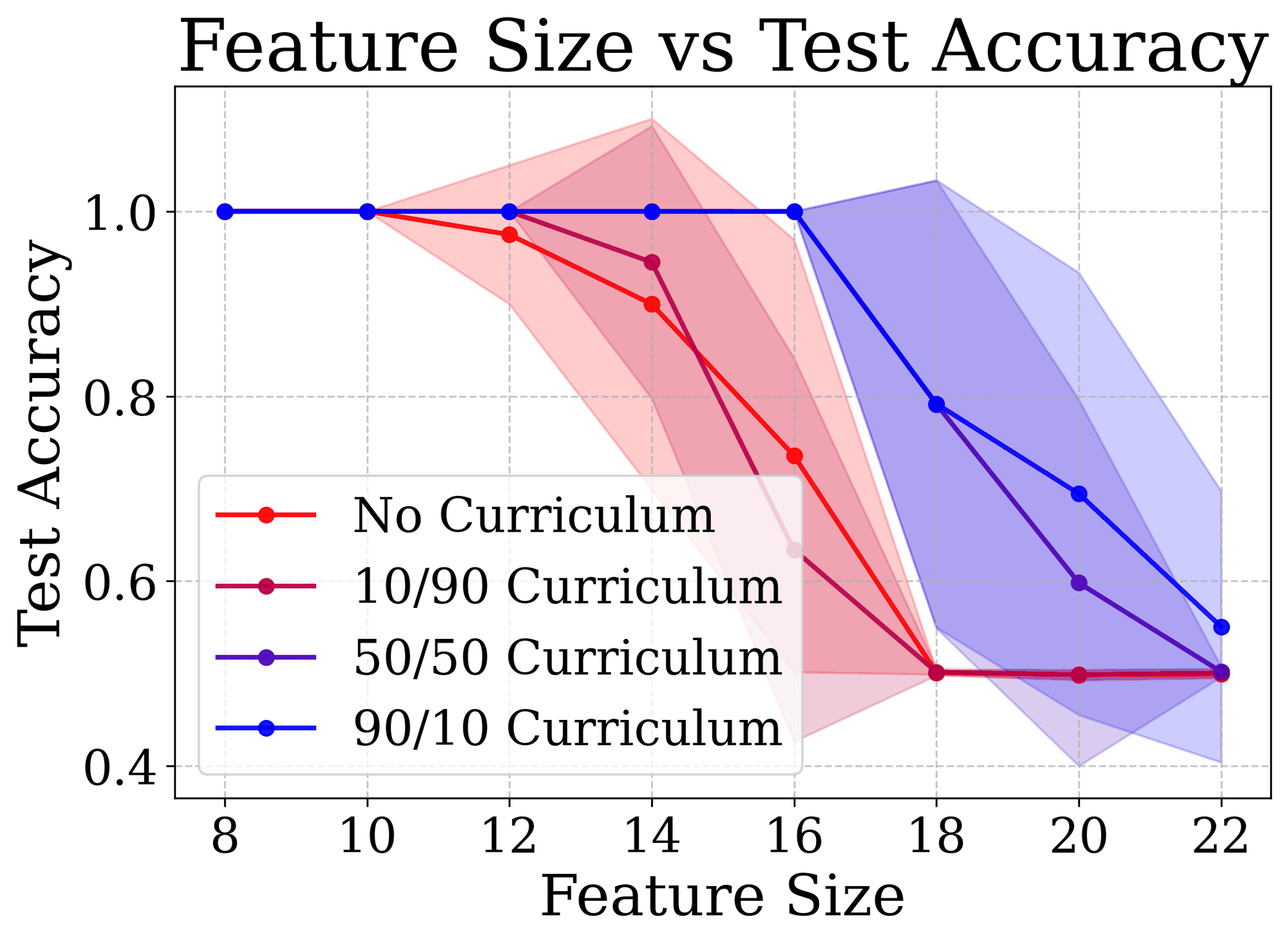}
    \end{subfigure}
    \begin{subfigure}[width=\textwidth]{0.48 \textwidth}
        \includegraphics[width=\textwidth]{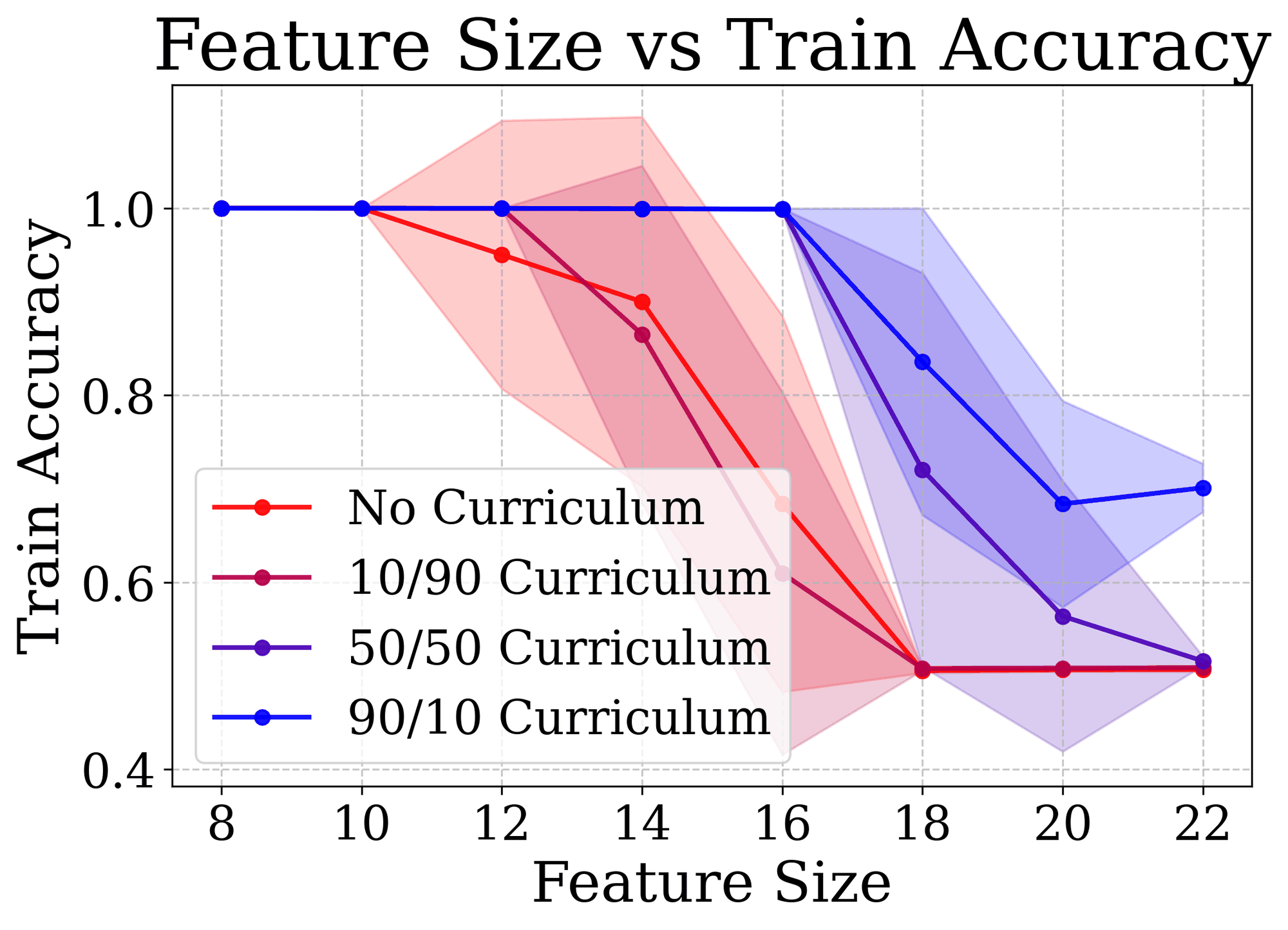}
    \end{subfigure}
    \caption{Effect on train and test accuracy of a Curriculum on the performance on GPT style model at learning k-parity.
    All runs use a fixed iteration budget of $10^4$ and full data set size of $8\times10^4$, the curriculum splits the indicated the fraction of training where only $1/10^{th}$ of the training set is used.}
    \label{fig:curriculum}
\end{figure}

\begin{figure*}[h]
    \centering
    \begin{subfigure}[]{0.24 \textwidth}
        \includegraphics[width=\textwidth]{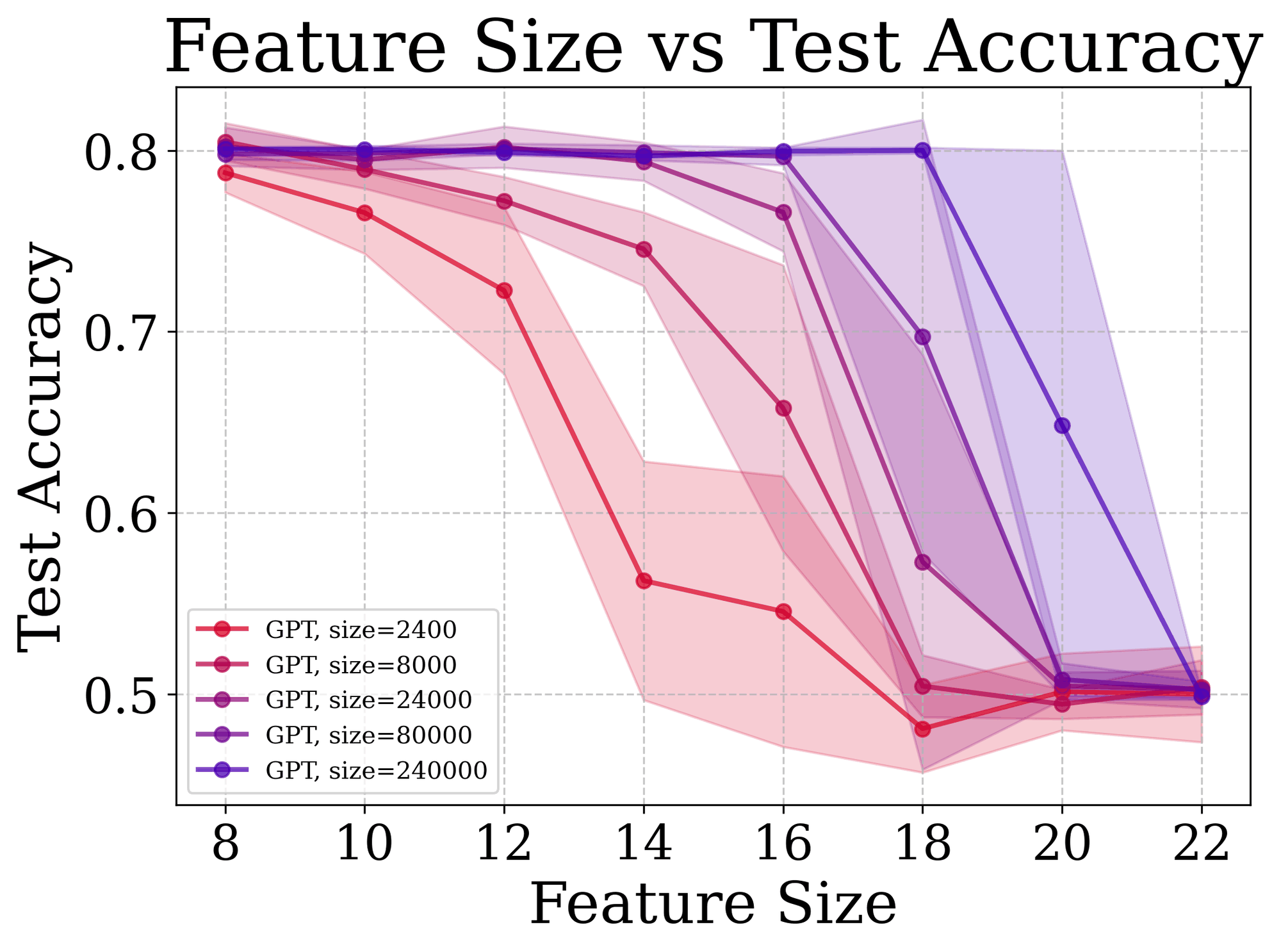}
    \end{subfigure}
    \begin{subfigure}[]{0.24 \textwidth}
        \includegraphics[width=\textwidth]{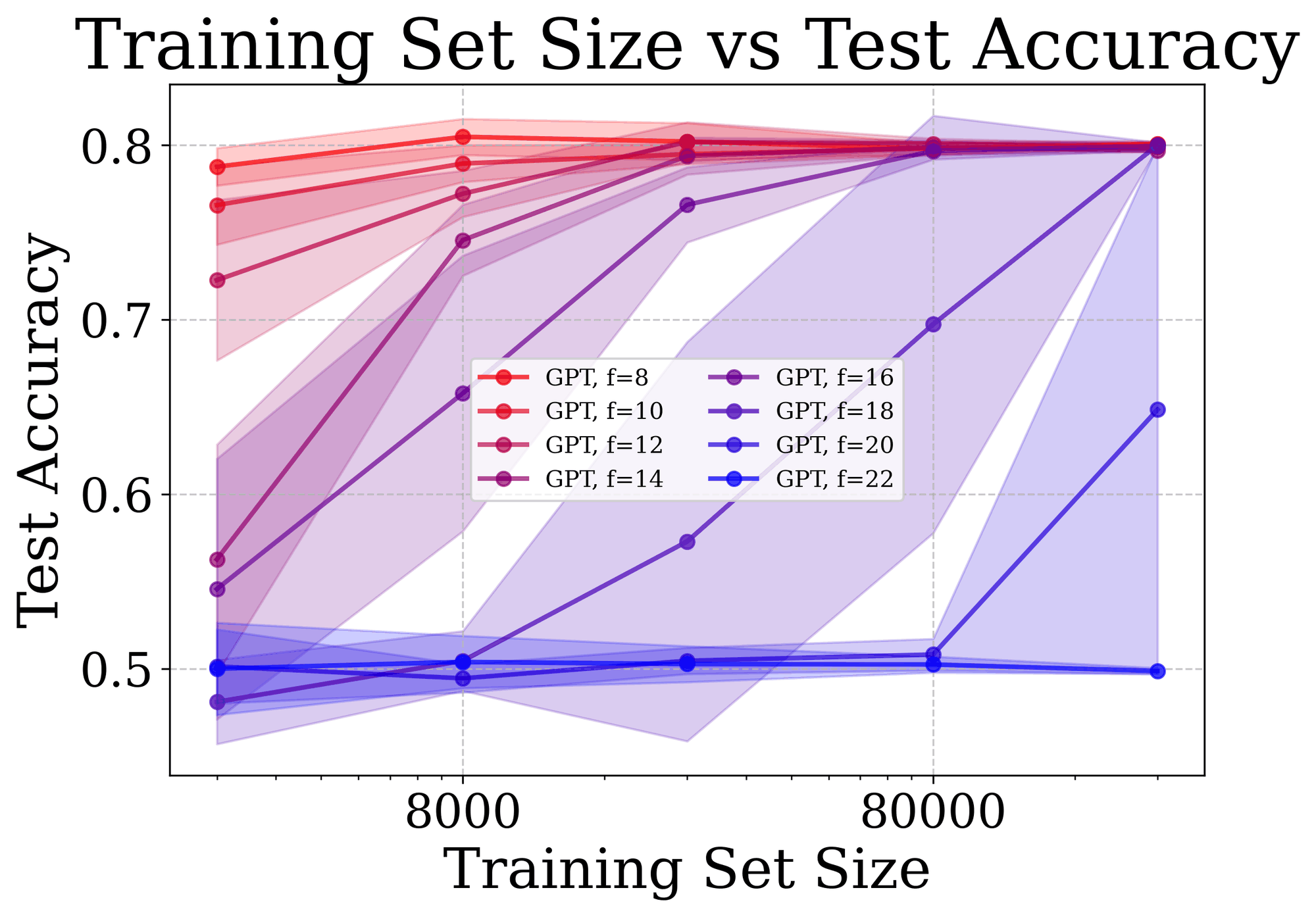}
    \end{subfigure}
    \begin{subfigure}[]{0.24 \textwidth}
        \includegraphics[width=\textwidth]{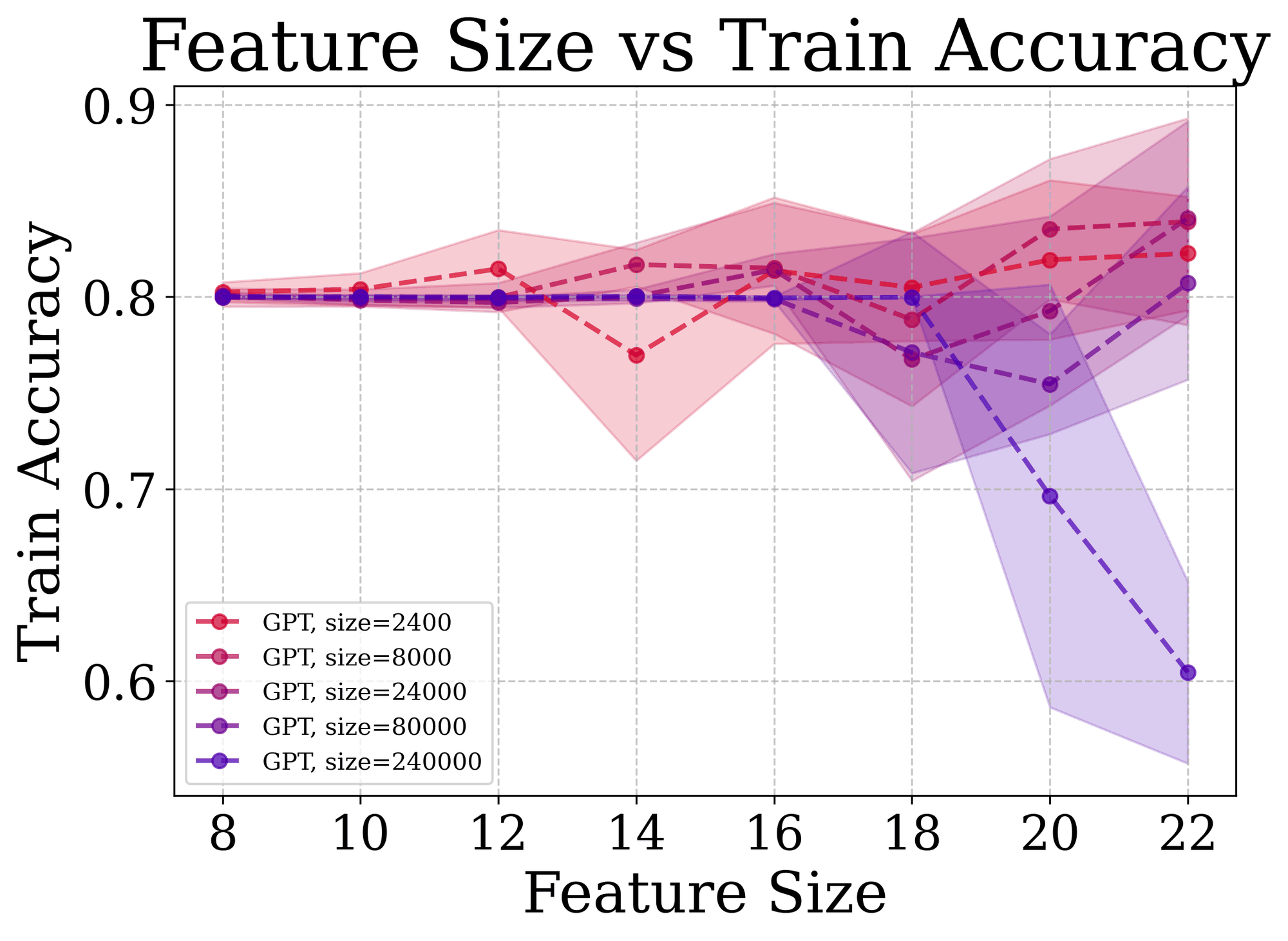}
    \end{subfigure}
    \begin{subfigure}[]{0.24 \textwidth}
        \includegraphics[width=\textwidth]{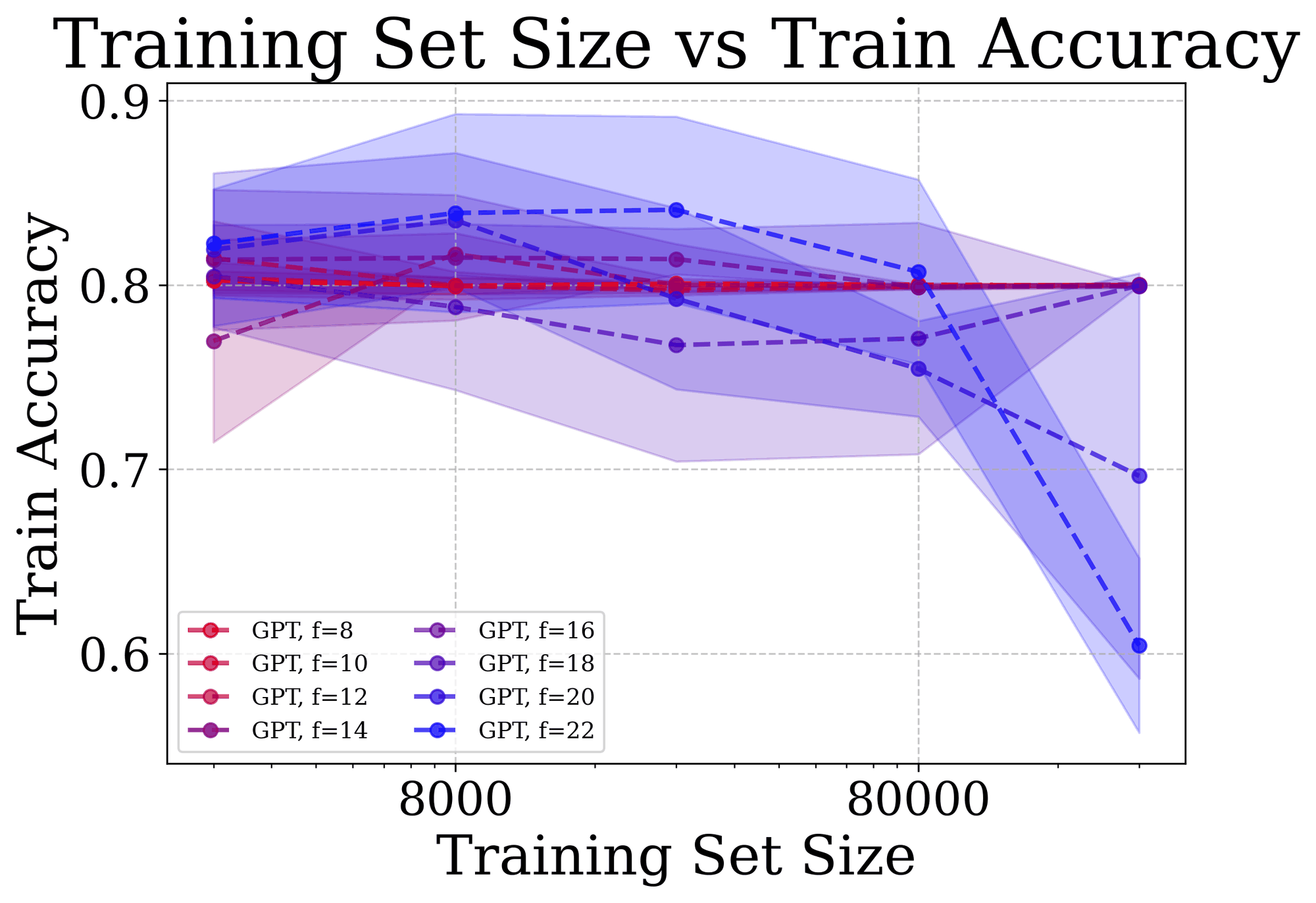}
    \end{subfigure}
    \caption{GPT accuracy on a noisy k-parity feature, $\alpha_1 = 0.8$, with the iteration budget scaled with the data set size ($T=1.25N$).
    \textbf{Left two panels:} test accuracy.
    \textbf{Right two panels:} train accuracy.}
    \label{fig:iter_scaling_k_parity}
\end{figure*}

\begin{figure*}[h]
    \centering
    \begin{subfigure}[]{0.24 \textwidth}
        \includegraphics[width=\textwidth]{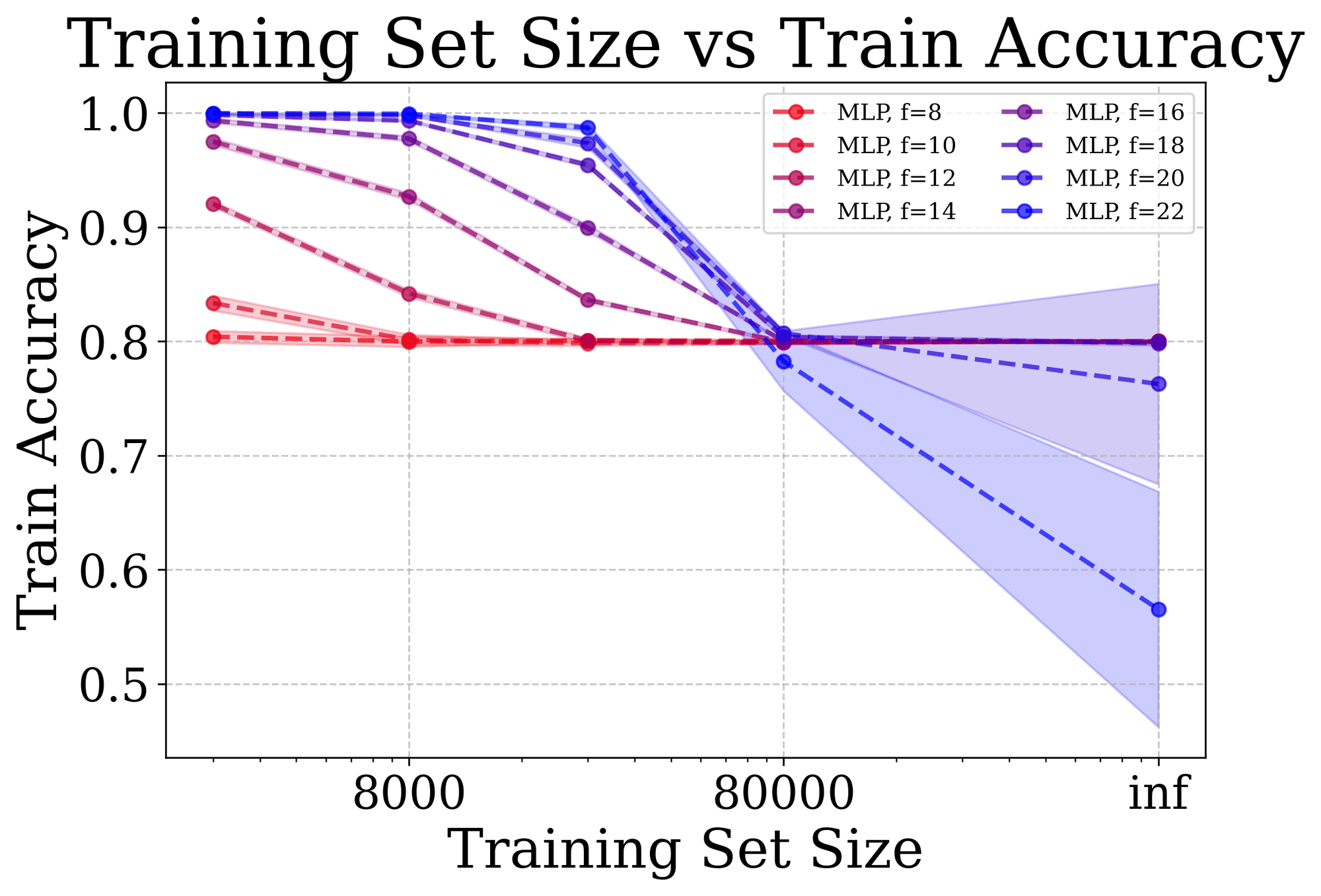}
    \end{subfigure}
    \begin{subfigure}[]{0.24 \textwidth}
        \includegraphics[width=\textwidth]{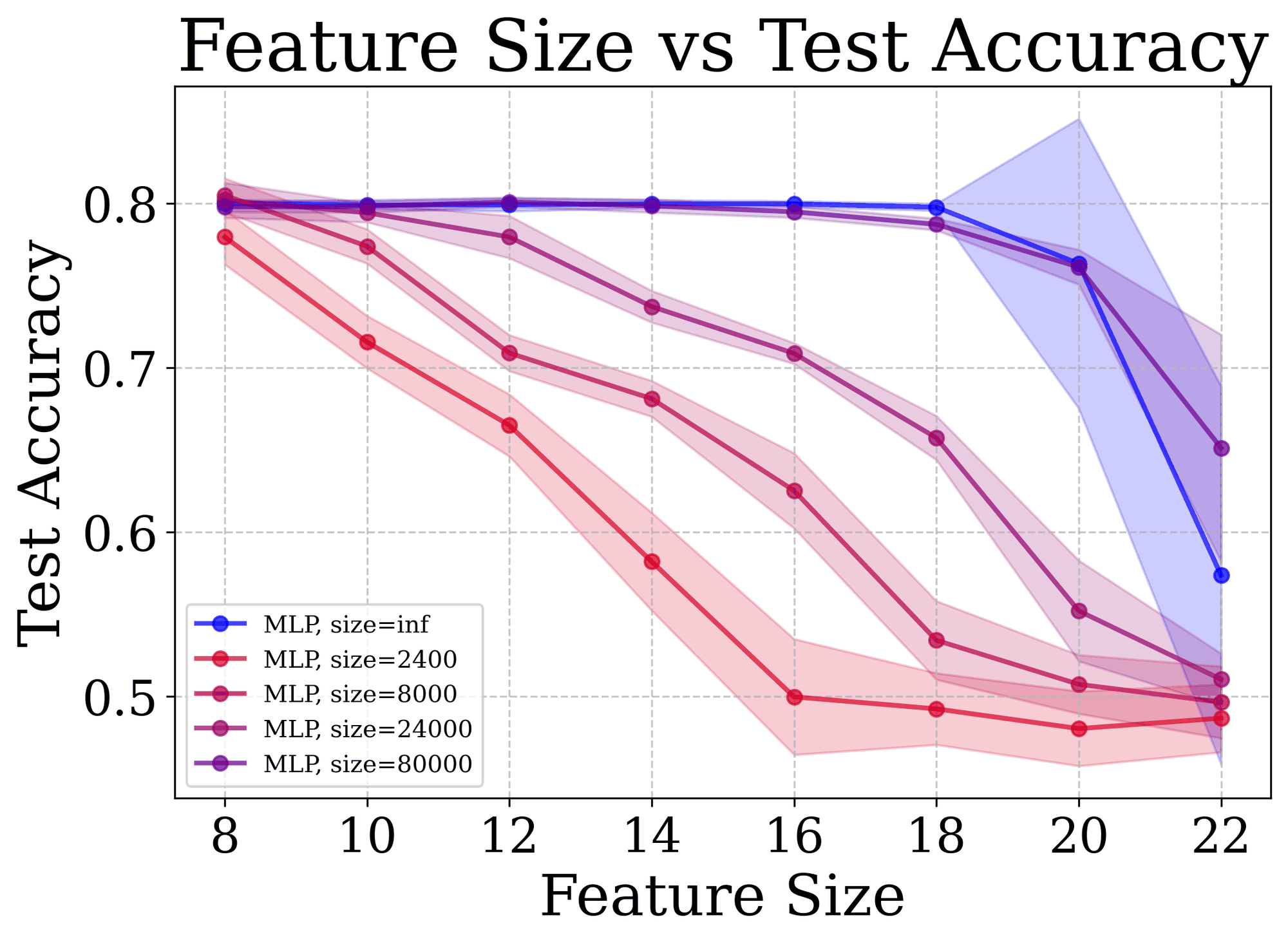}
    \end{subfigure}
    \begin{subfigure}[]{0.24 \textwidth}
        \includegraphics[width=\textwidth]{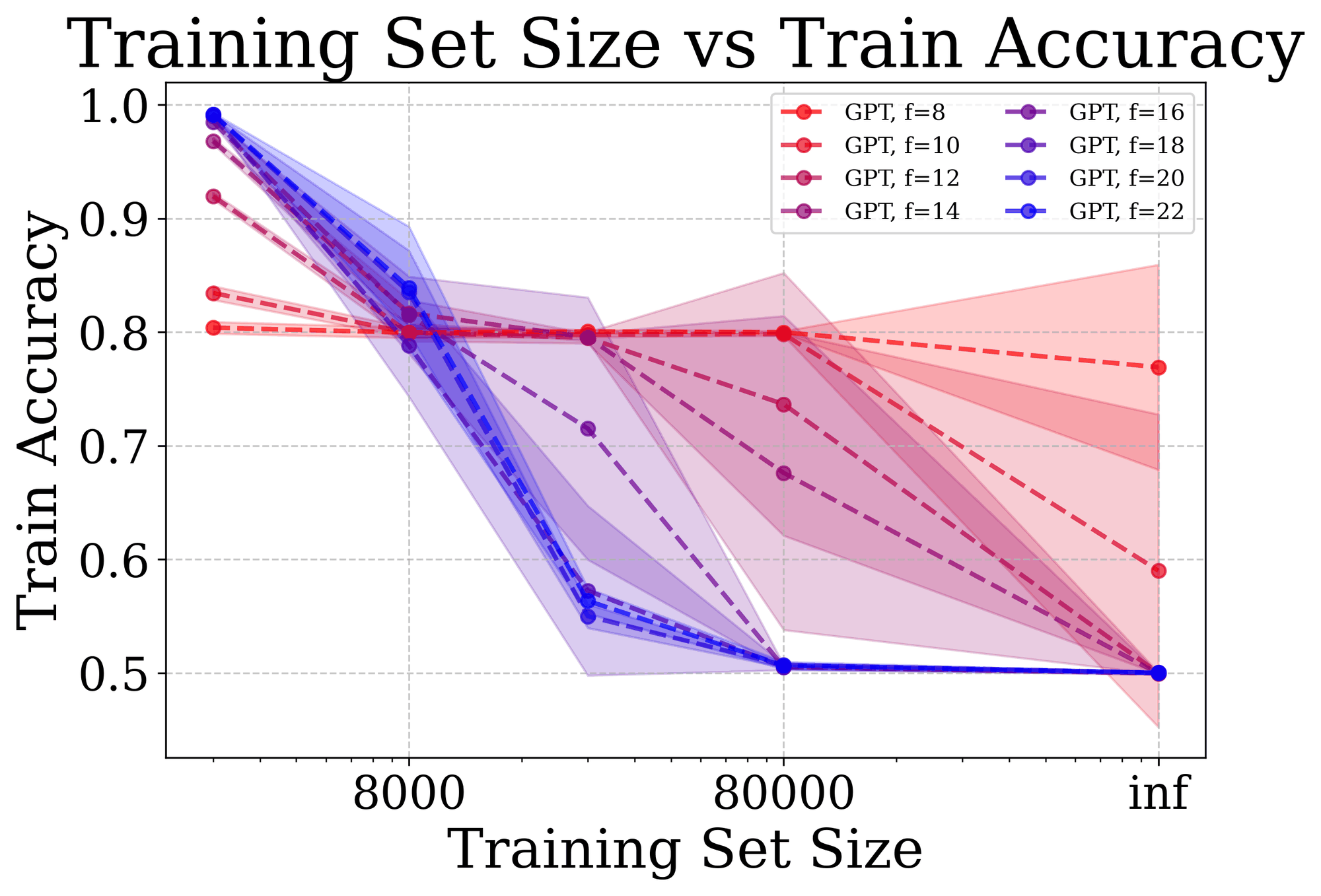}
    \end{subfigure}
    \begin{subfigure}[]{0.24 \textwidth}
        \includegraphics[width=\textwidth]{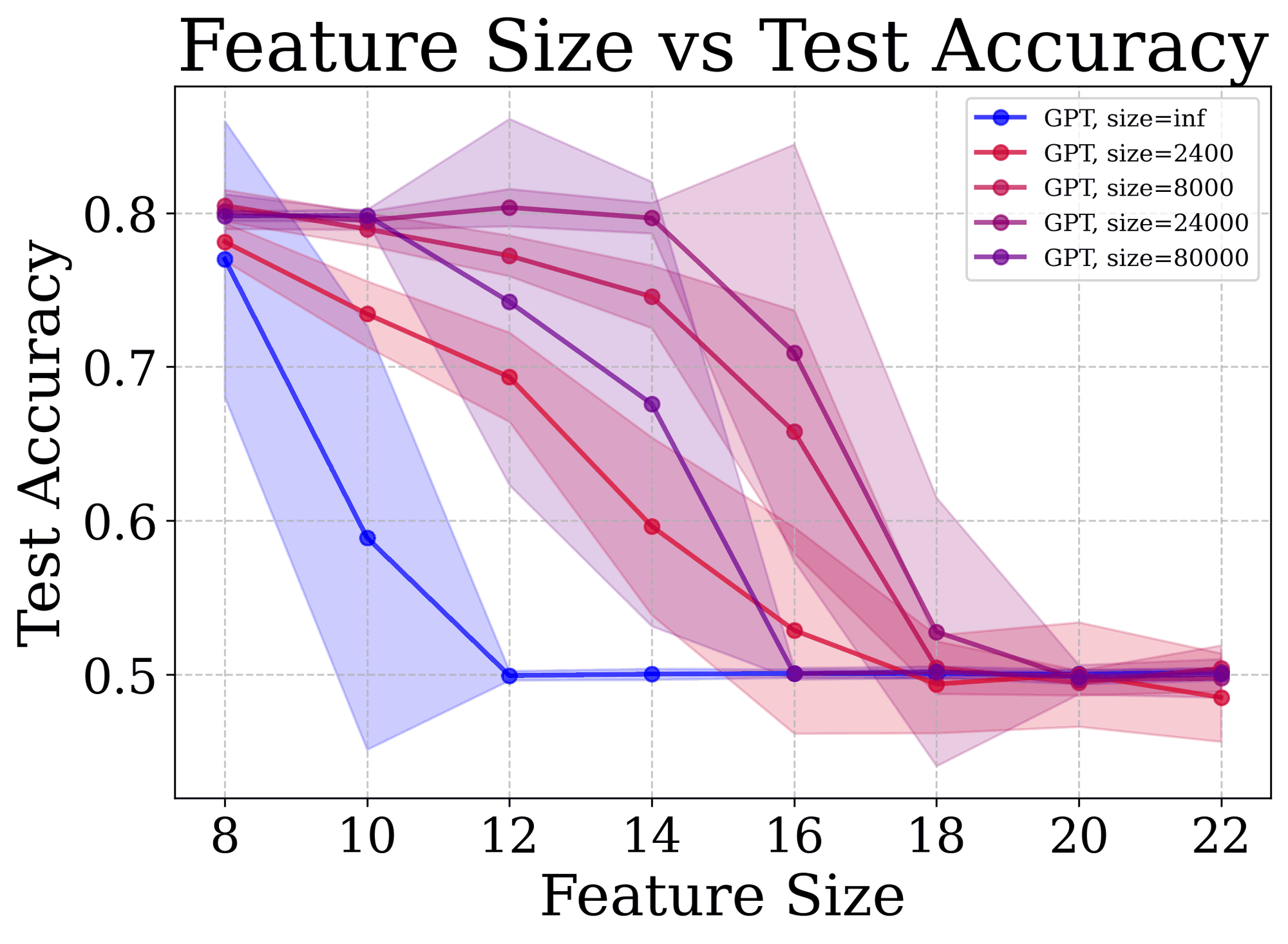}
    \end{subfigure}
    \caption{Accuracy on a noisy k-parity feature, $\alpha_1 = 0.8$, under a fixed budget ($T=10^4$).
    \textbf{Left:} MLP.
    \textbf{Right:} GPT.}
    \label{fig:noisy_k_parity_appendix}
\end{figure*}

\begin{figure*}[h]
    \centering

    \begin{subfigure}[width=\textwidth]{0.24 \textwidth}
        \includegraphics[width=\textwidth]{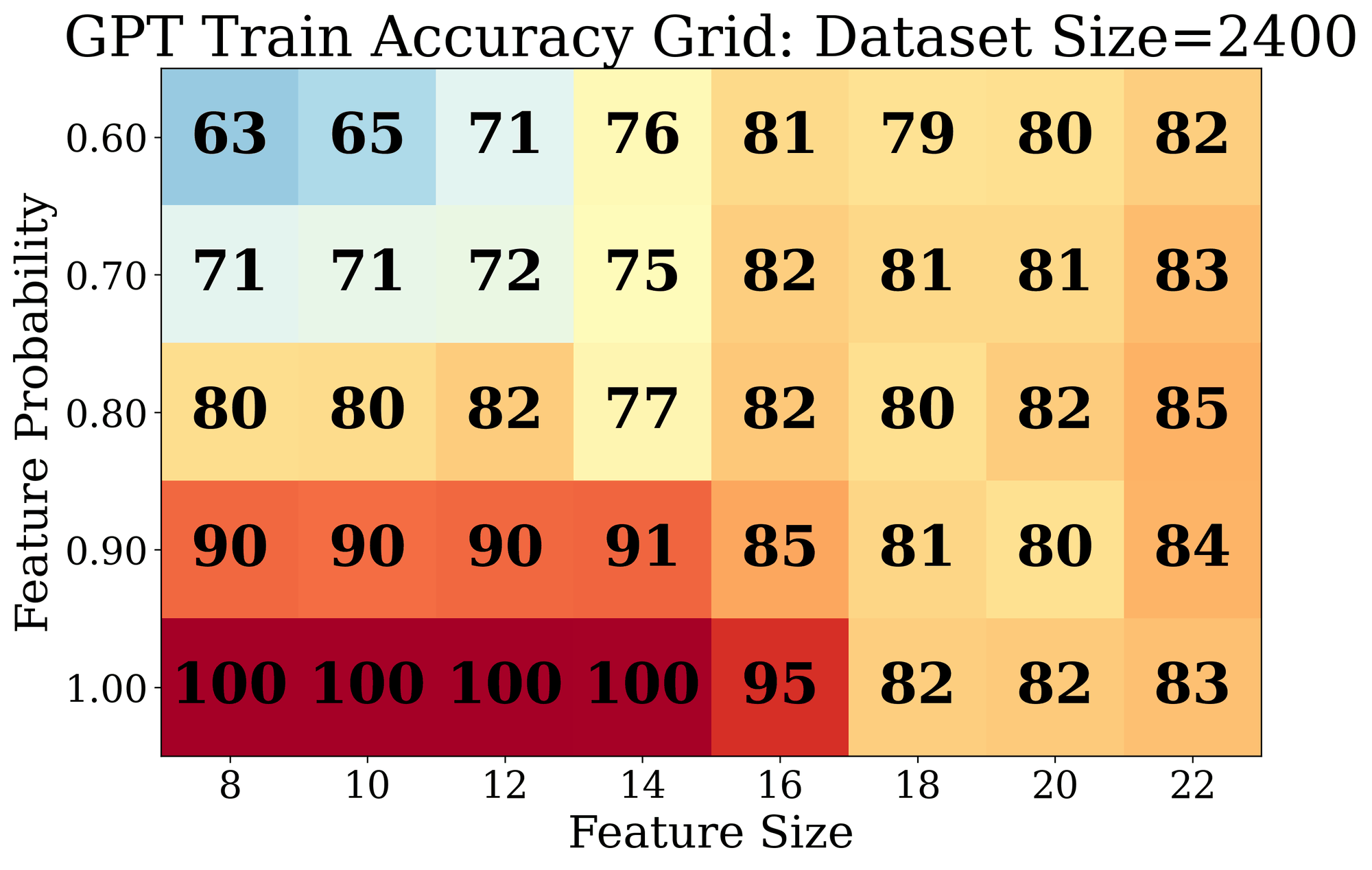}
    \end{subfigure}
    \begin{subfigure}[width=\textwidth]{0.24 \textwidth}
        \includegraphics[width=\textwidth]{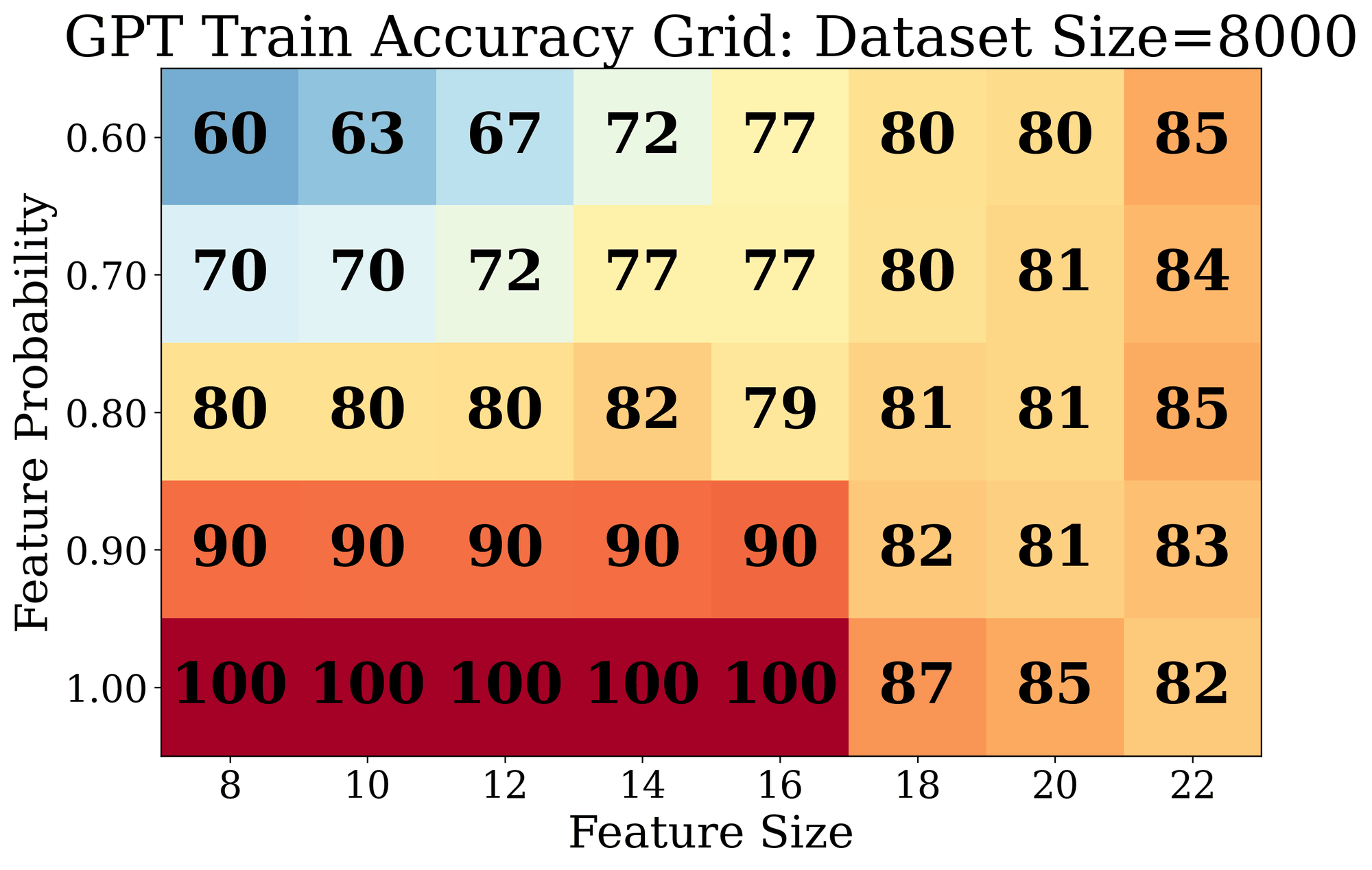}
    \end{subfigure}
    \begin{subfigure}[width=\textwidth]{0.24 \textwidth}
        \includegraphics[width=\textwidth]{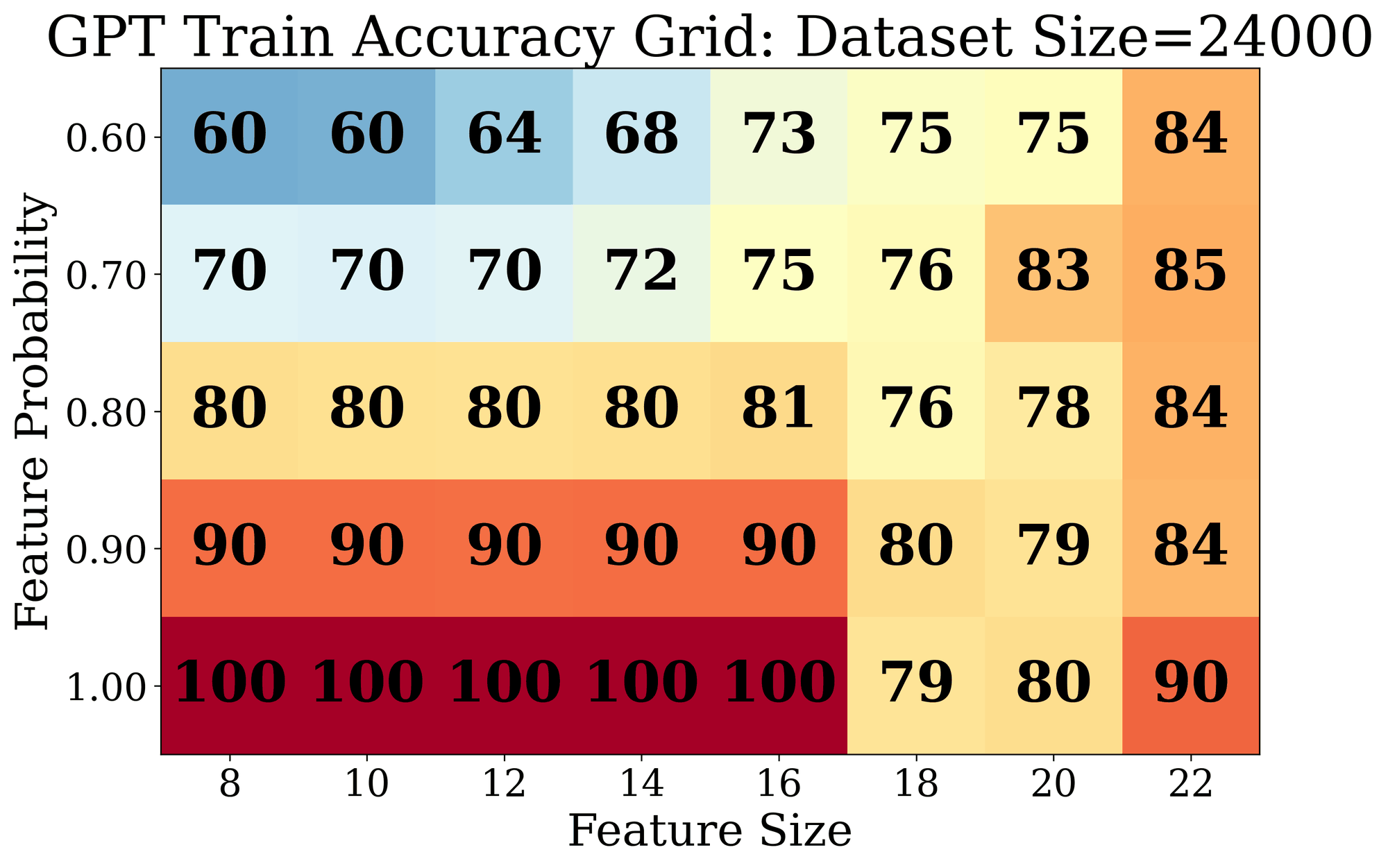}
    \end{subfigure}
    \begin{subfigure}[width=\textwidth]{0.24 \textwidth}
        \includegraphics[width=\textwidth]{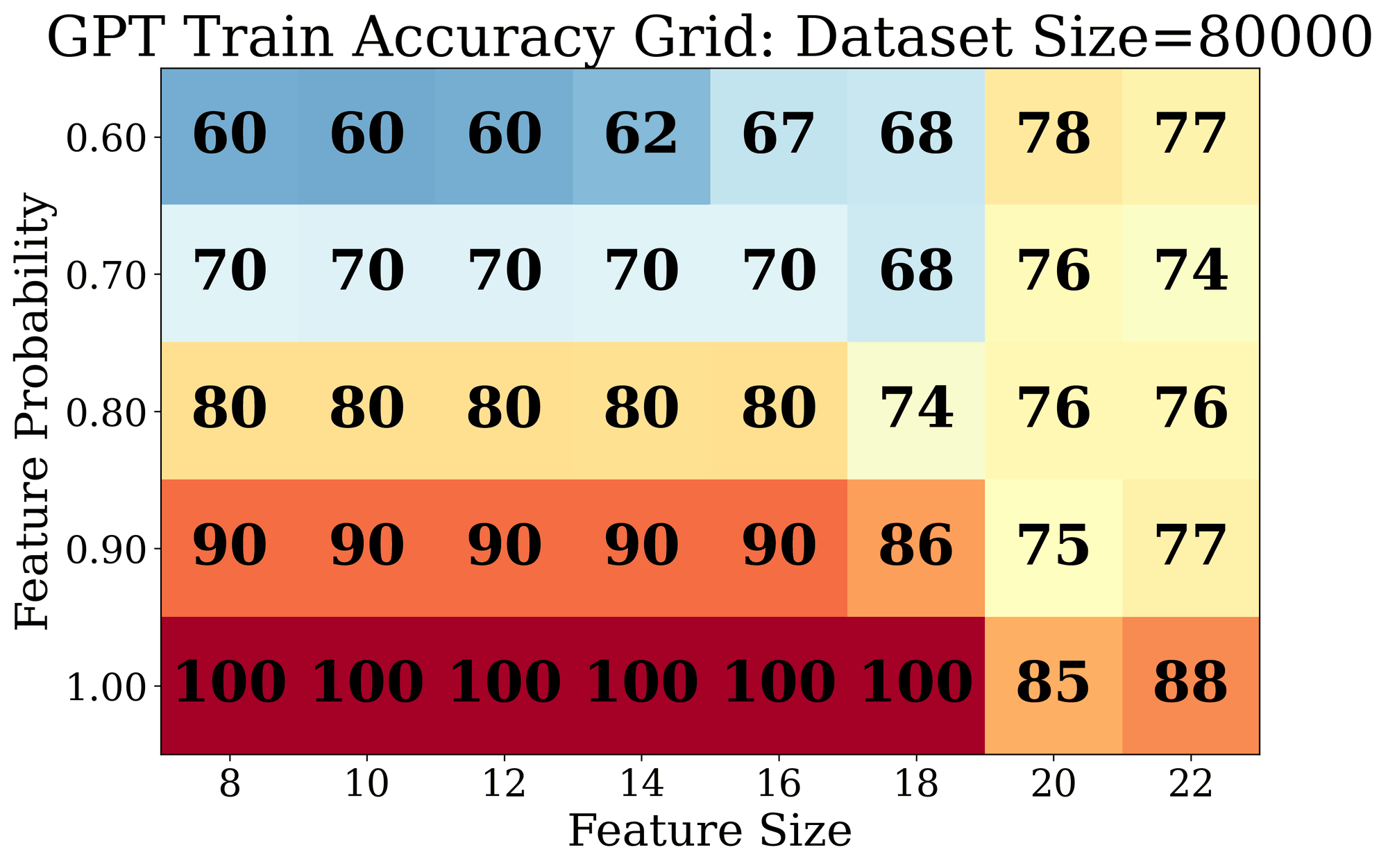}
    \end{subfigure}
    
    \begin{subfigure}[width=\textwidth]{0.24 \textwidth}
        \includegraphics[width=\textwidth]{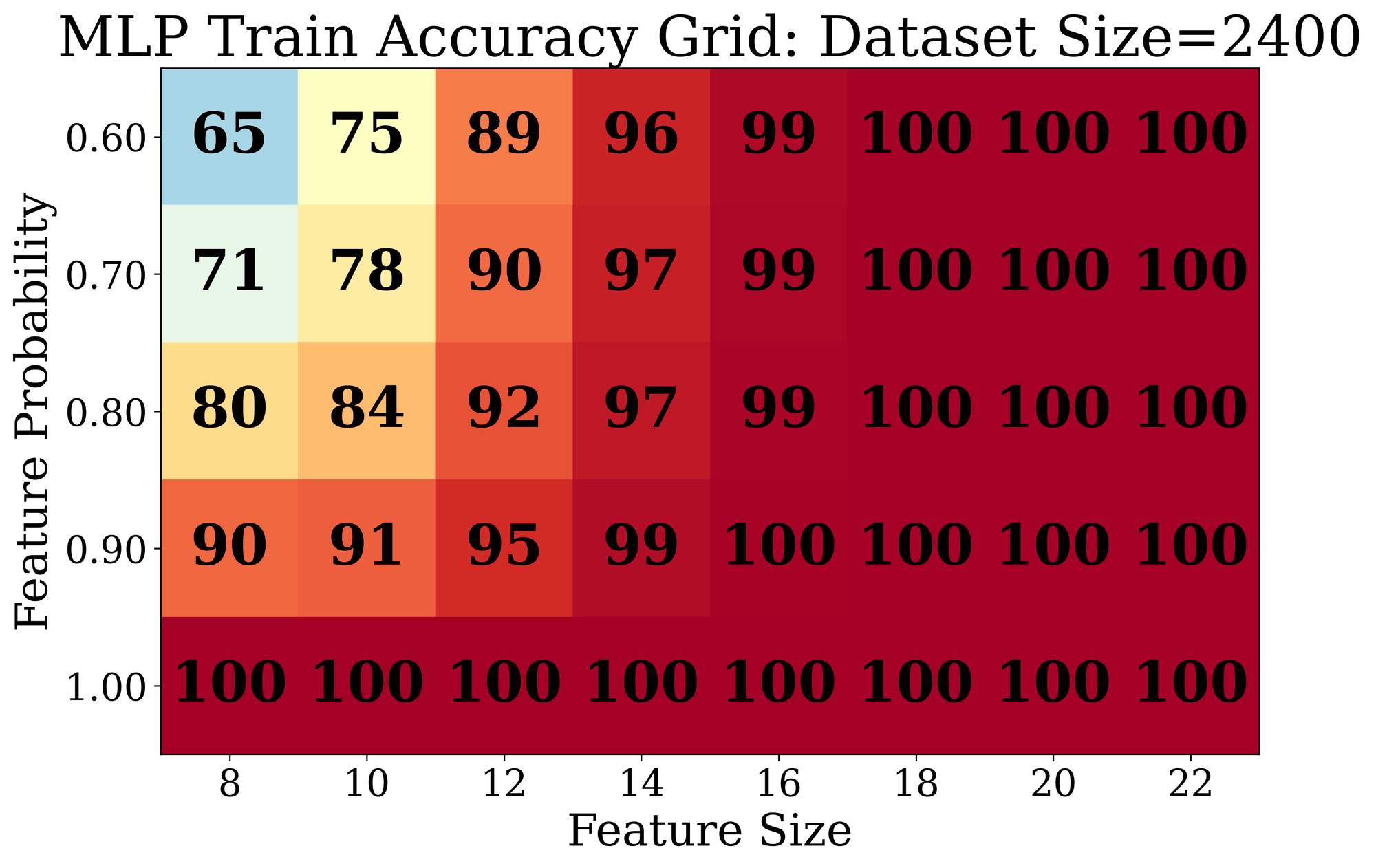}
    \end{subfigure}
    \begin{subfigure}[width=\textwidth]{0.24 \textwidth}
        \includegraphics[width=\textwidth]{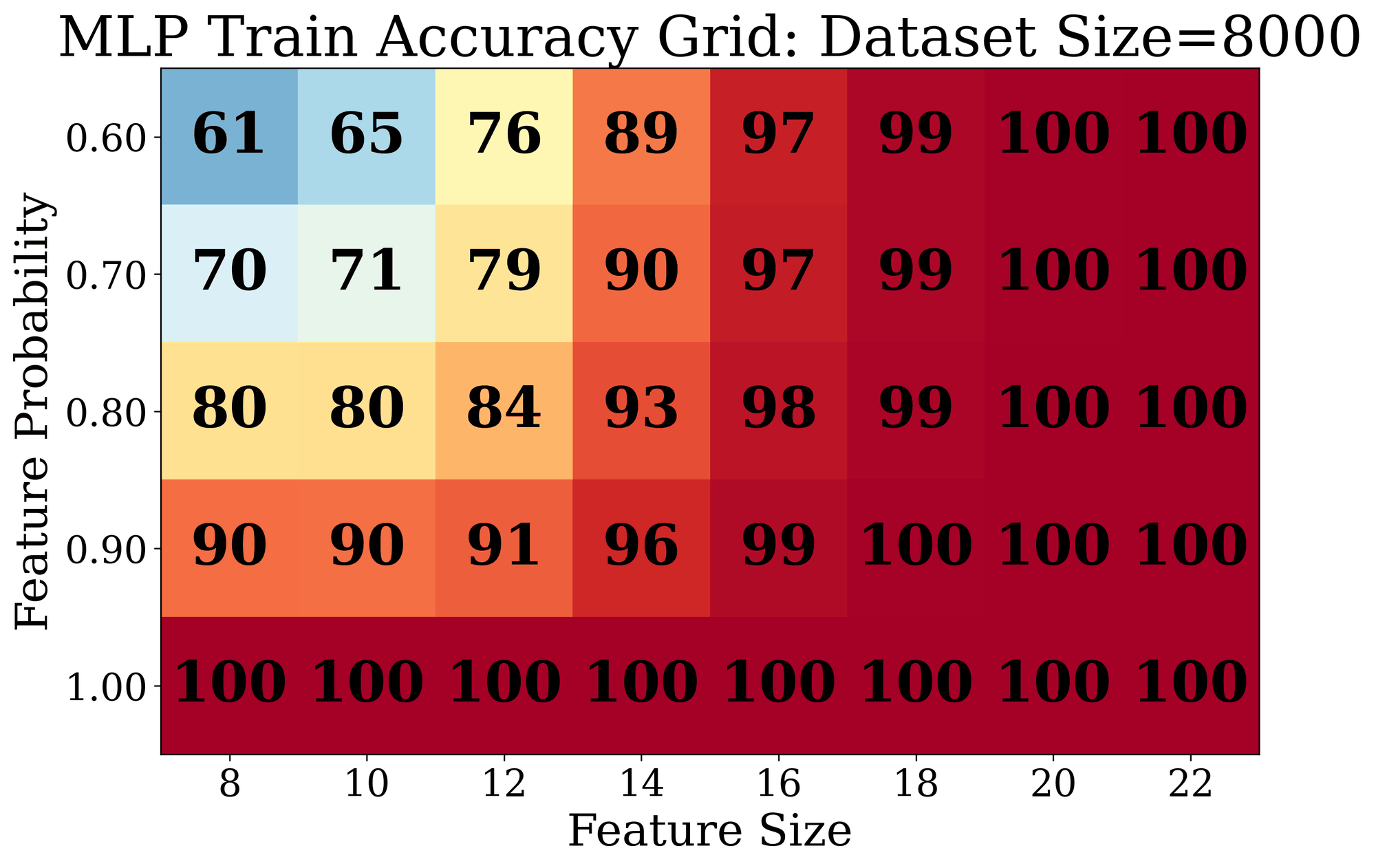}
    \end{subfigure}
    \begin{subfigure}[width=\textwidth]{0.24 \textwidth}
        \includegraphics[width=\textwidth]{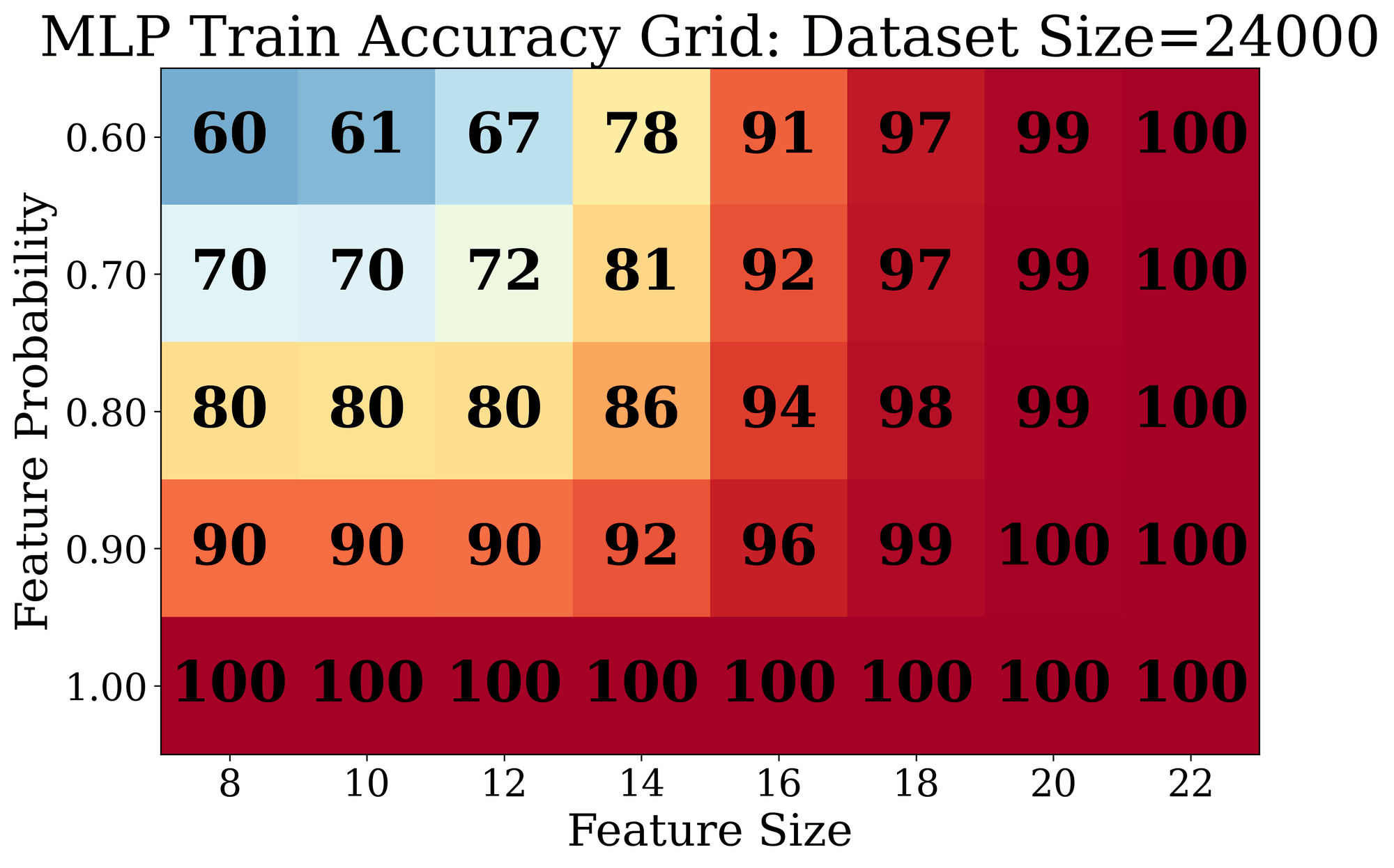}
    \end{subfigure}
    \begin{subfigure}[width=\textwidth]{0.24 \textwidth}
        \includegraphics[width=\textwidth]{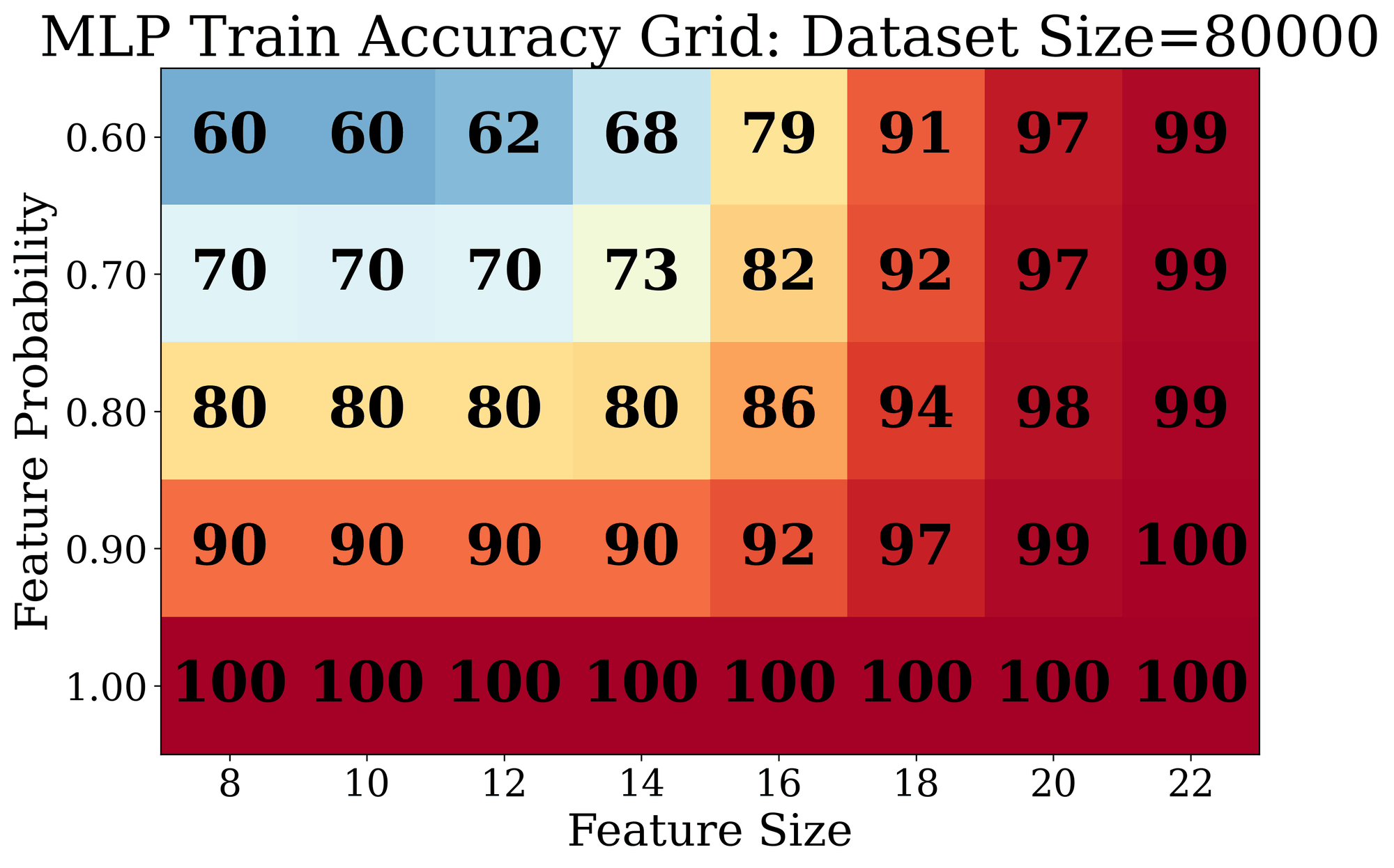}
    \end{subfigure}

    \begin{subfigure}[width=\textwidth]{0.24 \textwidth}
        \includegraphics[width=\textwidth]{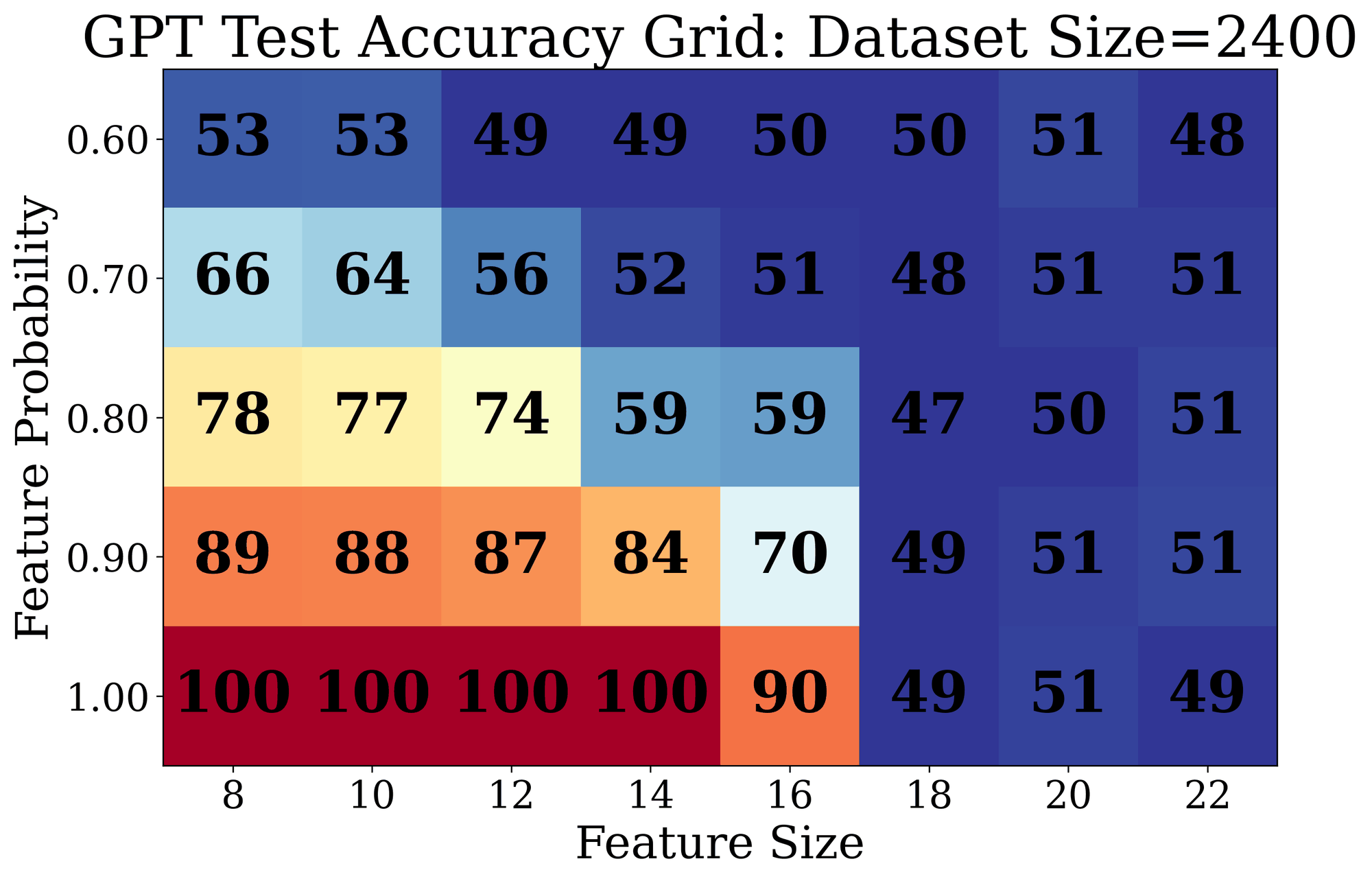}
    \end{subfigure}
    \begin{subfigure}[width=\textwidth]{0.24 \textwidth}
        \includegraphics[width=\textwidth]{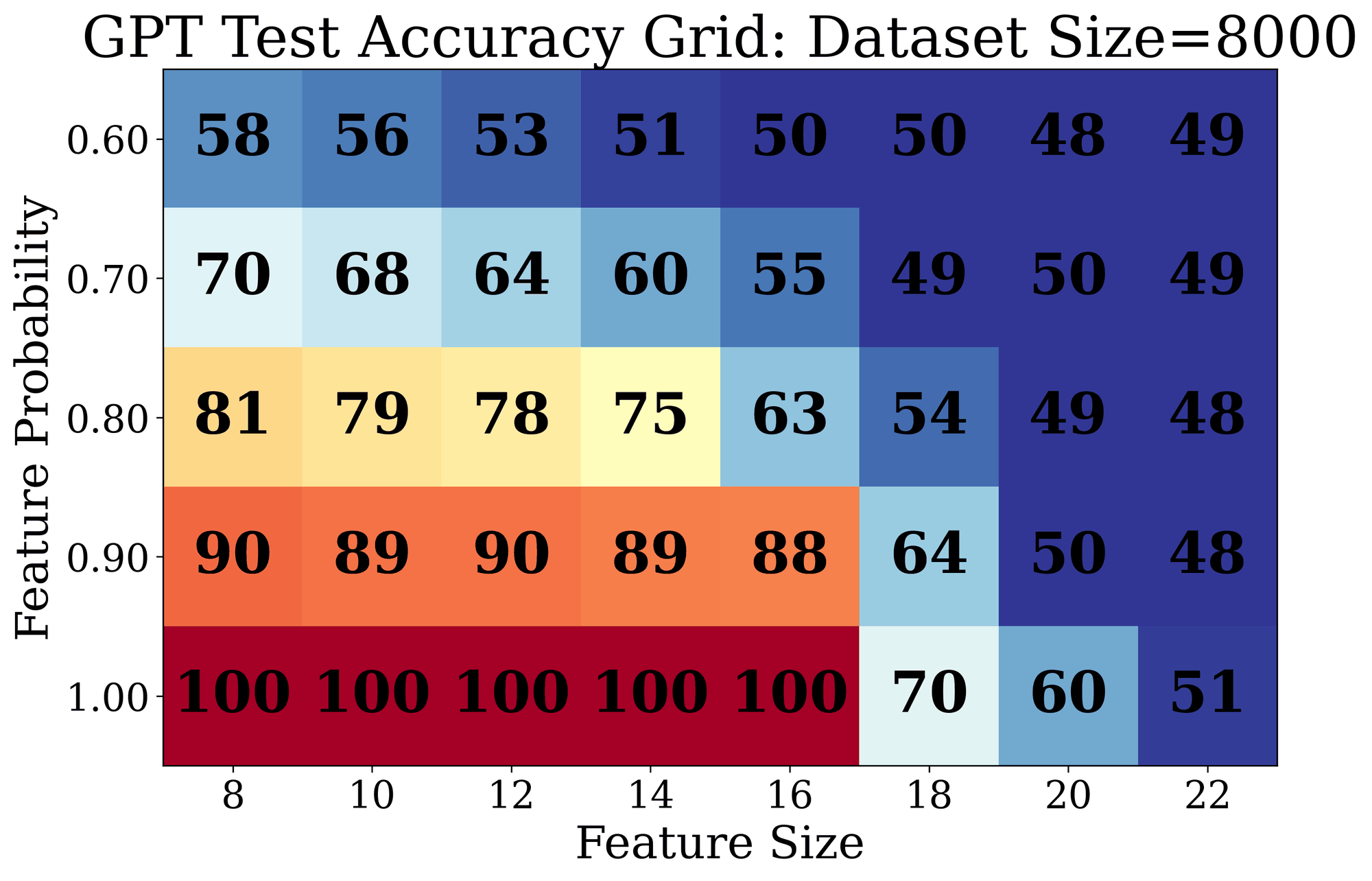}
    \end{subfigure}
    \begin{subfigure}[width=\textwidth]{0.24 \textwidth}
        \includegraphics[width=\textwidth]{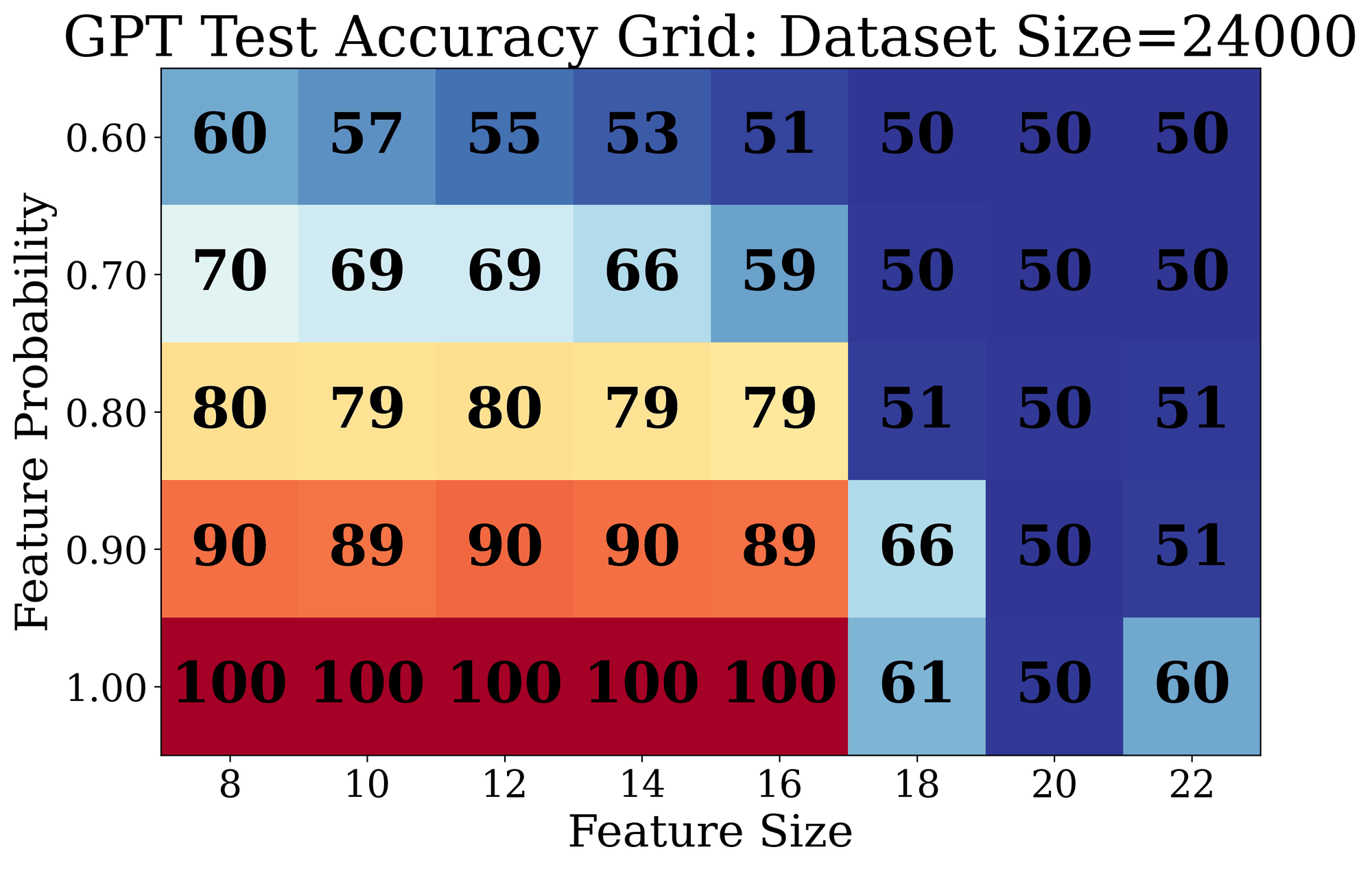}
    \end{subfigure}
    \begin{subfigure}[width=\textwidth]{0.24 \textwidth}
        \includegraphics[width=\textwidth]{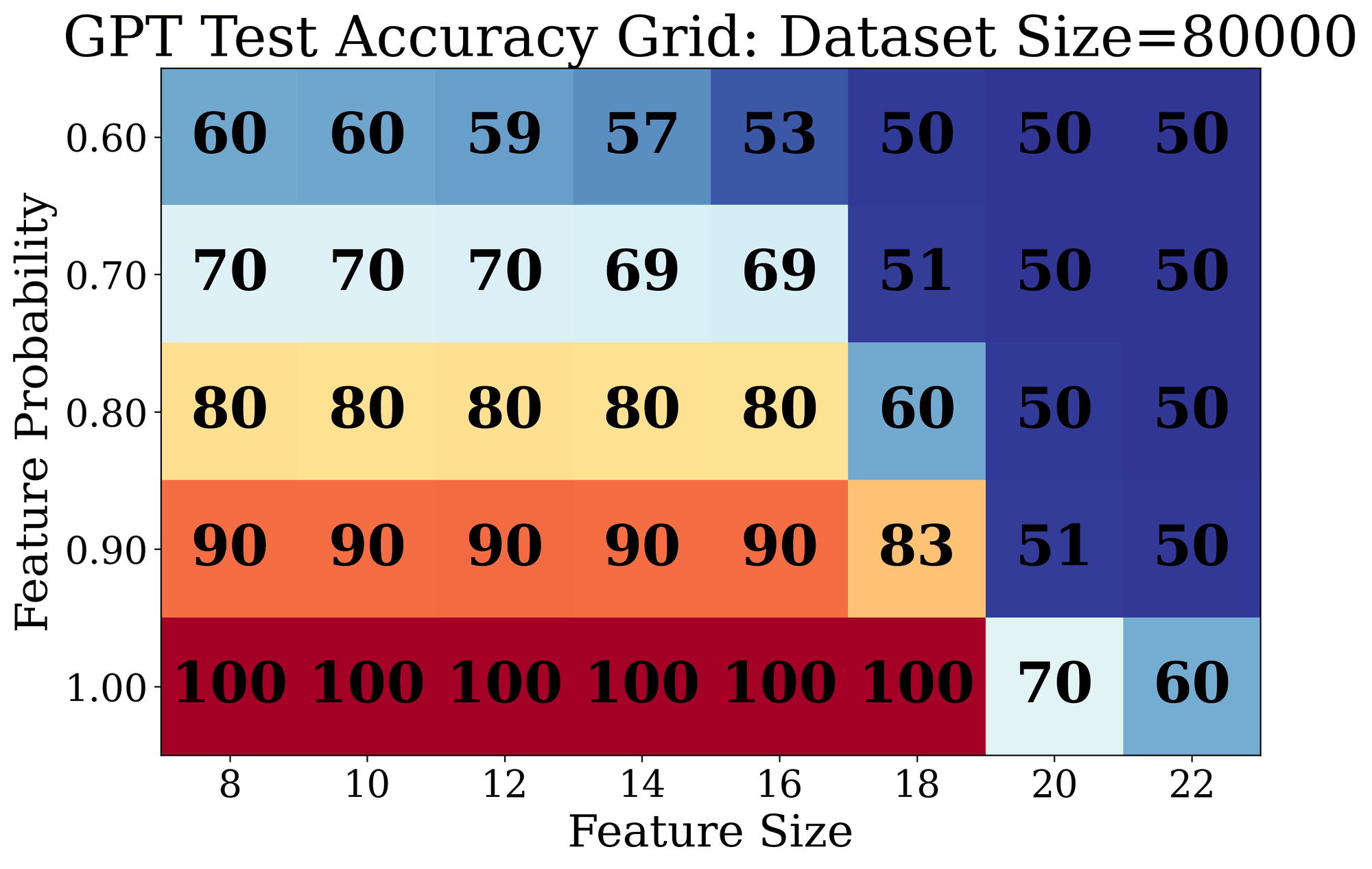}
    \end{subfigure}
    
    \begin{subfigure}[width=\textwidth]{0.24 \textwidth}
        \includegraphics[width=\textwidth]{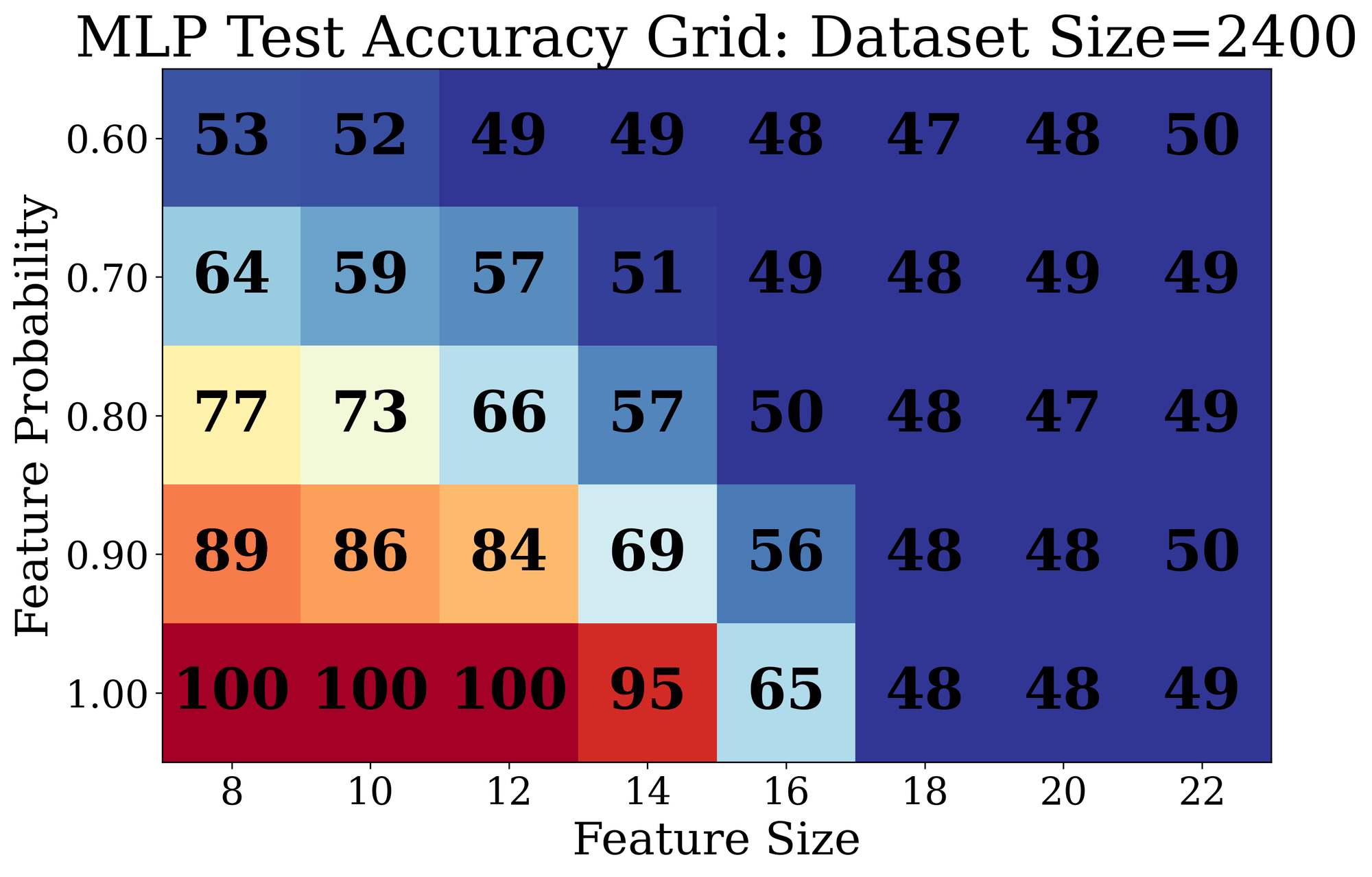}
    \end{subfigure}
    \begin{subfigure}[width=\textwidth]{0.24 \textwidth}
        \includegraphics[width=\textwidth]{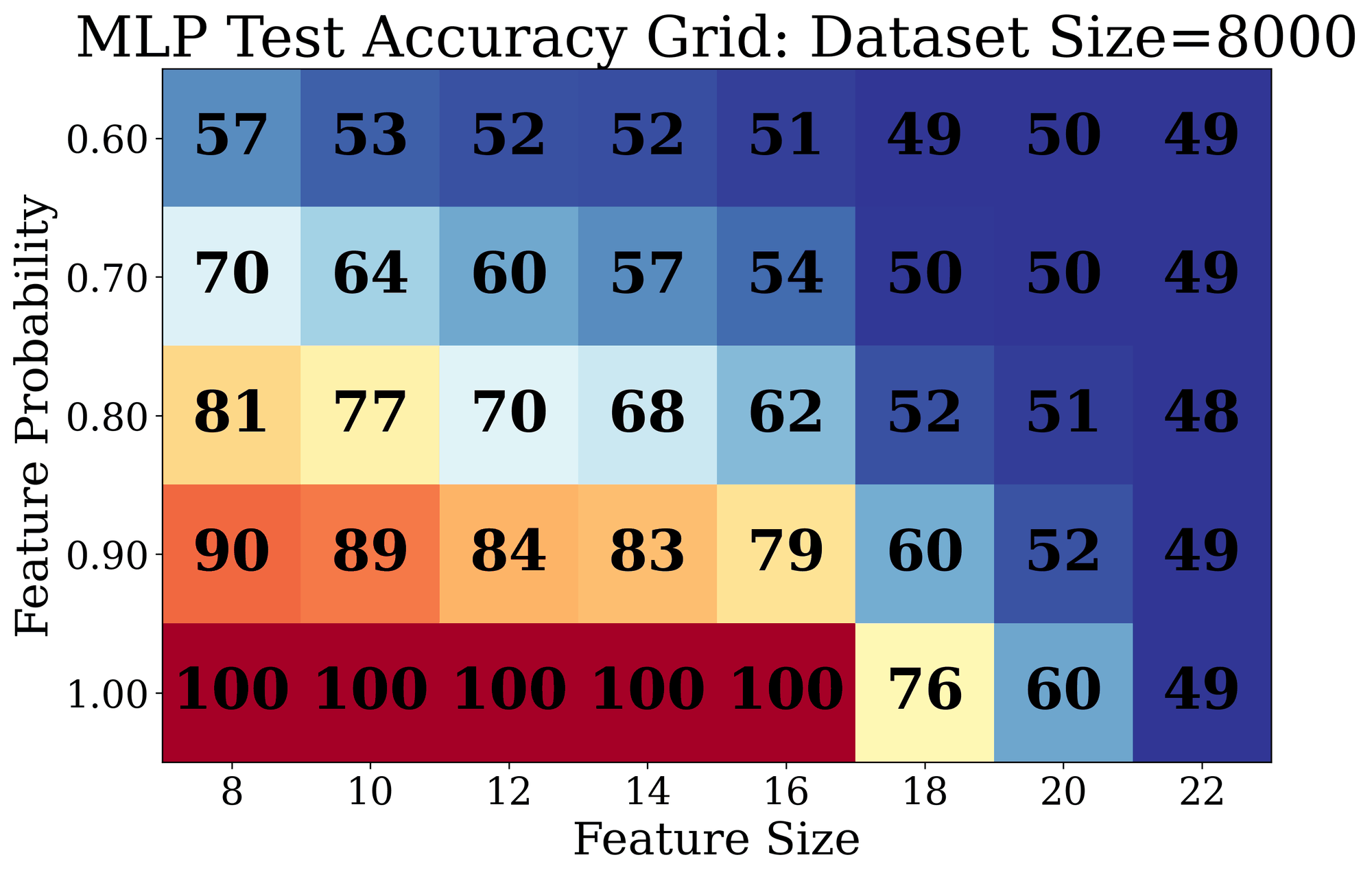}
    \end{subfigure}
    \begin{subfigure}[width=\textwidth]{0.24 \textwidth}
        \includegraphics[width=\textwidth]{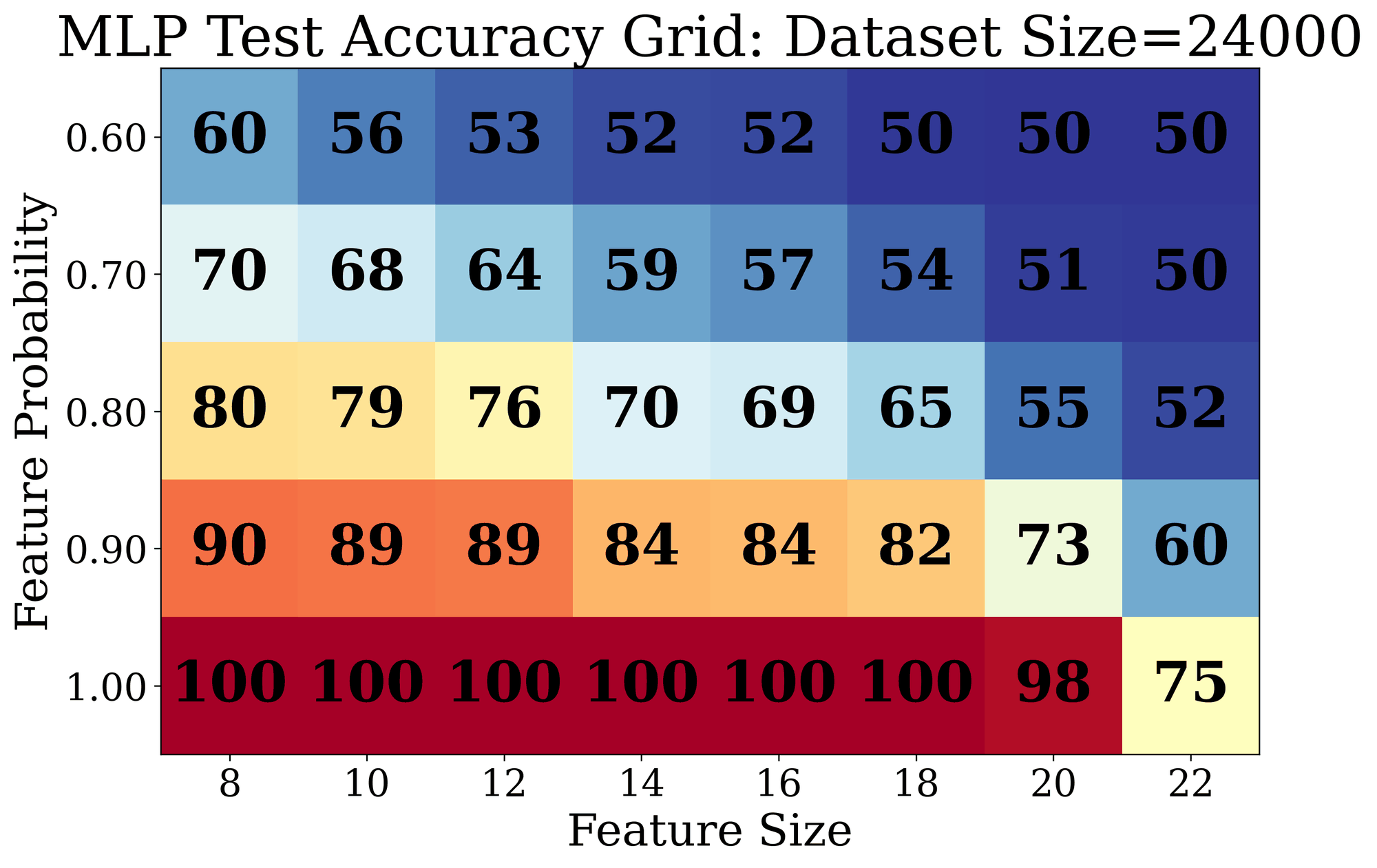}
    \end{subfigure}
    \begin{subfigure}[width=\textwidth]{0.24 \textwidth}
        \includegraphics[width=\textwidth]{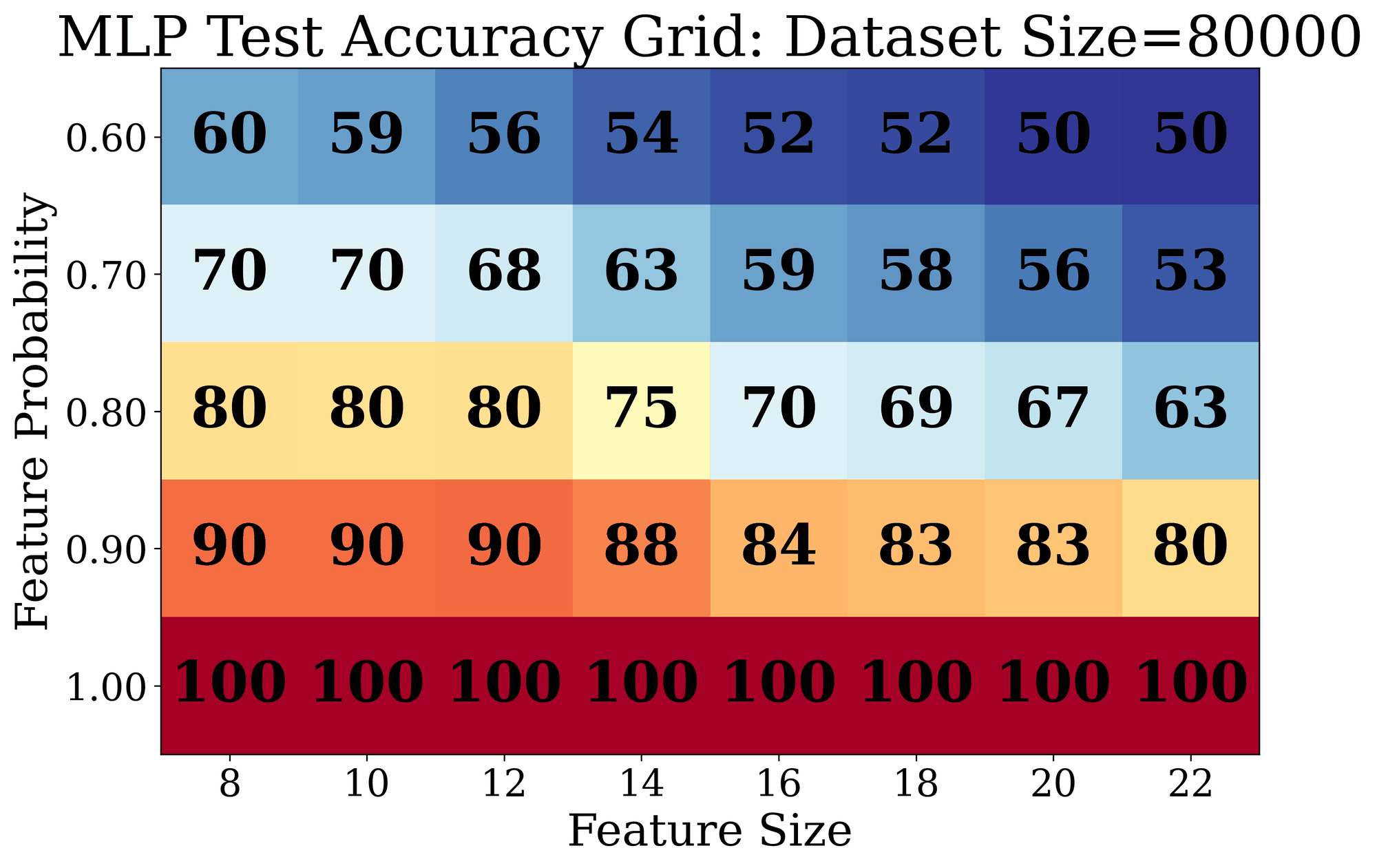}
    \end{subfigure}

        \begin{subfigure}[width=\textwidth]{0.24 \textwidth}
        \includegraphics[width=\textwidth]{figures/noisy_transfer_iter_scaling/noisy_feature_scaling_grid_gpt_size3000_clean_val.png}
    \end{subfigure}
    \begin{subfigure}[width=\textwidth]{0.24 \textwidth}
        \includegraphics[width=\textwidth]{figures/noisy_transfer_iter_scaling/noisy_feature_scaling_grid_gpt_size10000_clean_val.png}
    \end{subfigure}
    \begin{subfigure}[width=\textwidth]{0.24 \textwidth}
        \includegraphics[width=\textwidth]{figures/noisy_transfer_iter_scaling/noisy_feature_scaling_grid_gpt_size30000_clean_val.png}
    \end{subfigure}
    \begin{subfigure}[width=\textwidth]{0.24 \textwidth}
        \includegraphics[width=\textwidth]{figures/noisy_transfer_iter_scaling/noisy_feature_scaling_grid_gpt_size100000_clean_val.png}
    \end{subfigure}
    
    \begin{subfigure}[width=\textwidth]{0.24 \textwidth}
        \includegraphics[width=\textwidth]{figures/noisy_transfer_iter_scaling/noisy_feature_scaling_grid_mlp_size3000_clean_val.png}
    \end{subfigure}
    \begin{subfigure}[width=\textwidth]{0.24 \textwidth}
        \includegraphics[width=\textwidth]{figures/noisy_transfer_iter_scaling/noisy_feature_scaling_grid_mlp_size10000_clean_val.png}
    \end{subfigure}
    \begin{subfigure}[width=\textwidth]{0.24 \textwidth}
        \includegraphics[width=\textwidth]{figures/noisy_transfer_iter_scaling/noisy_feature_scaling_grid_mlp_size30000_clean_val.png}
    \end{subfigure}
    \begin{subfigure}[width=\textwidth]{0.24 \textwidth}
        \includegraphics[width=\textwidth]{figures/noisy_transfer_iter_scaling/noisy_feature_scaling_grid_mlp_size100000_clean_val.png}
    \end{subfigure}

    \caption{Accuracies for MLP and GPT models trained on a noisy k-parity feature with Scaling Iteration Budget $(T=1.25N)$, \textbf{Top Two Rows:} Train Accuracy, \textbf{Middle Two Rows:} Test Accuracy (i.i.d) to train set, \textbf{Bottom Two Rows:} Noiseless Test Accuracy.}
    \label{fig:noisy_k_parity_test_scaling}
\end{figure*}

\begin{figure*}[h]
    \centering

    \begin{subfigure}[width=\textwidth]{0.19 \textwidth}
        \includegraphics[width=\textwidth]{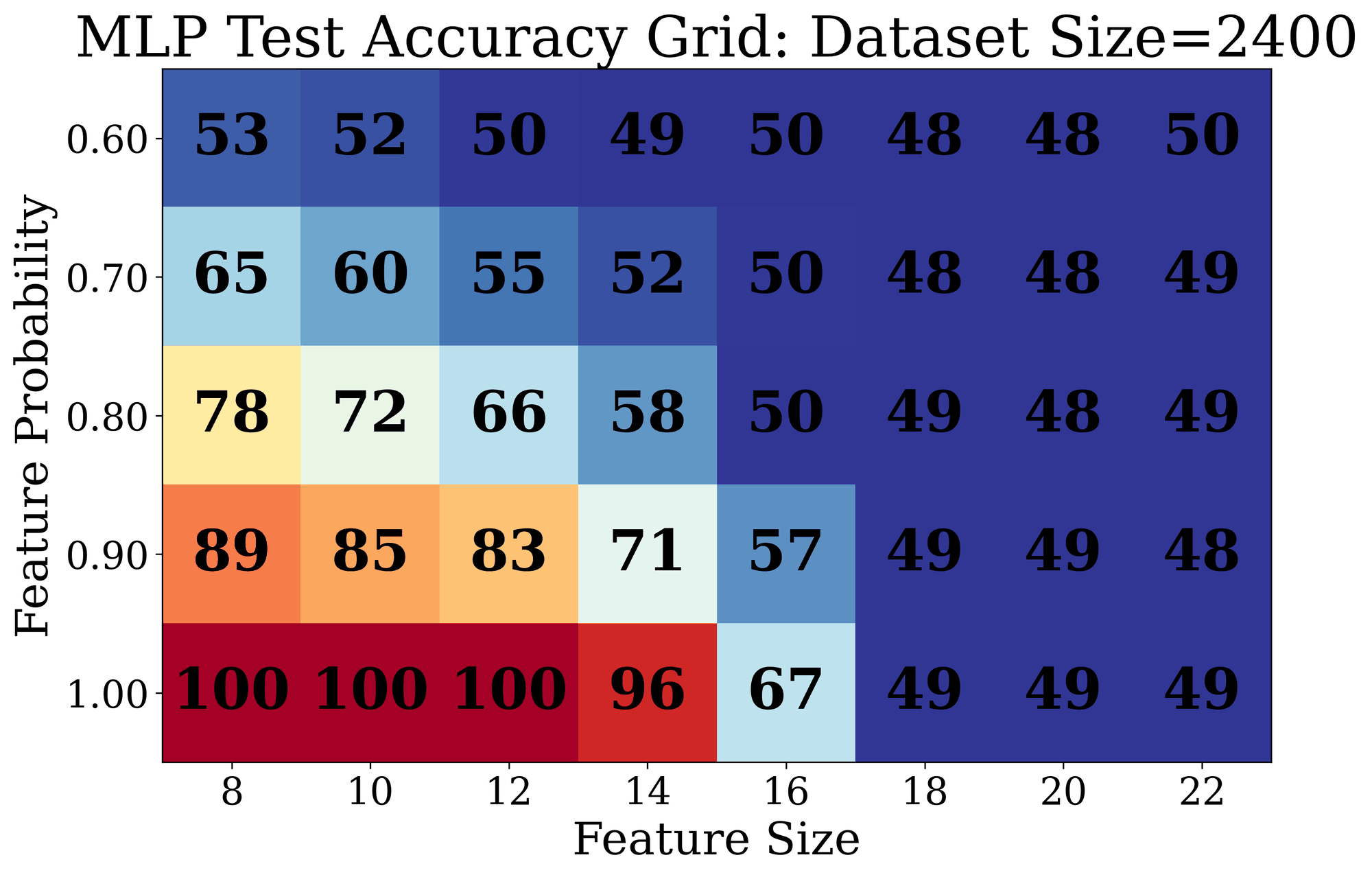}
    \end{subfigure}
    \begin{subfigure}[width=\textwidth]{0.19 \textwidth}
        \includegraphics[width=\textwidth]{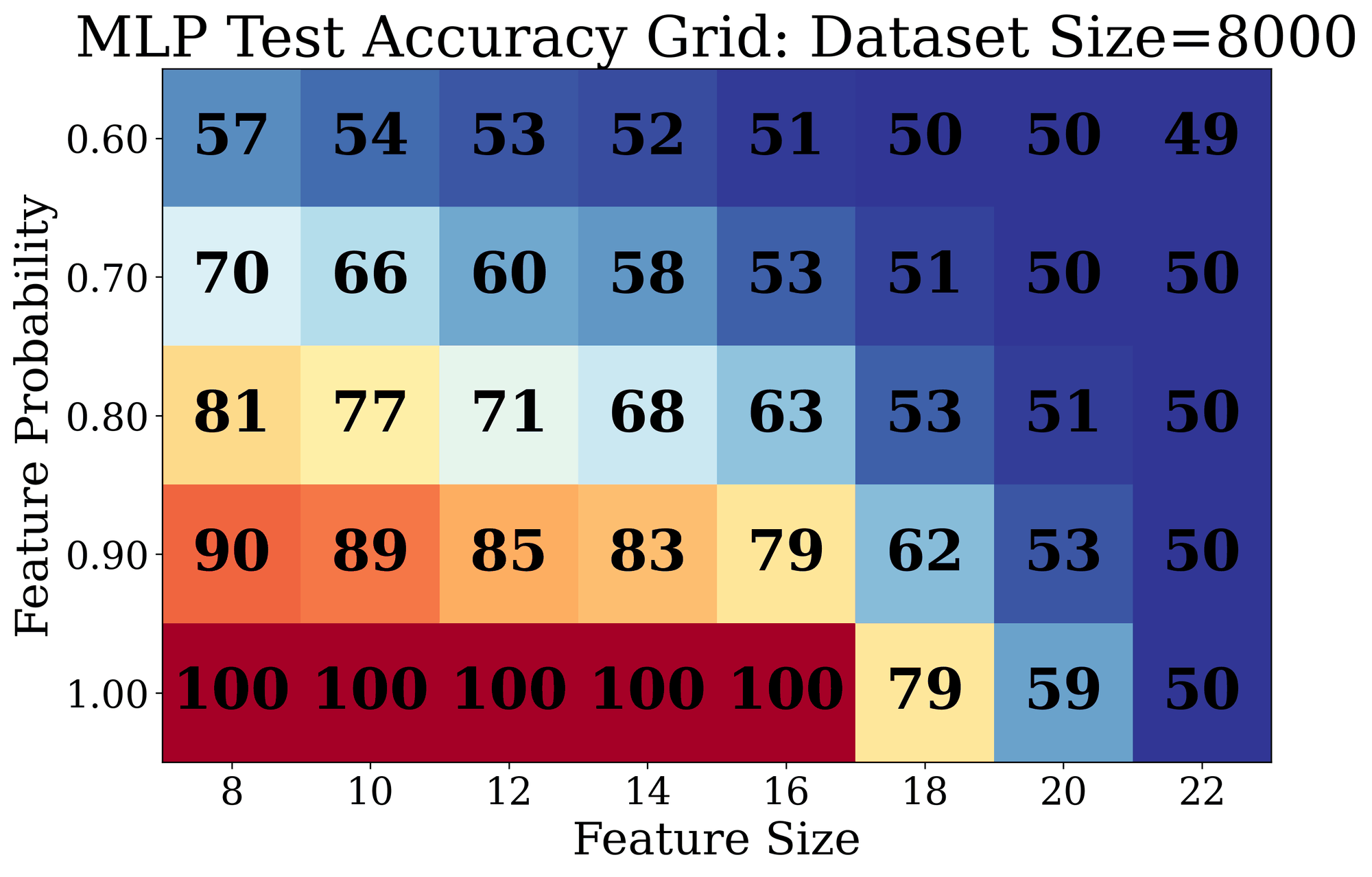}
    \end{subfigure}
    \begin{subfigure}[width=\textwidth]{0.19 \textwidth}
        \includegraphics[width=\textwidth]{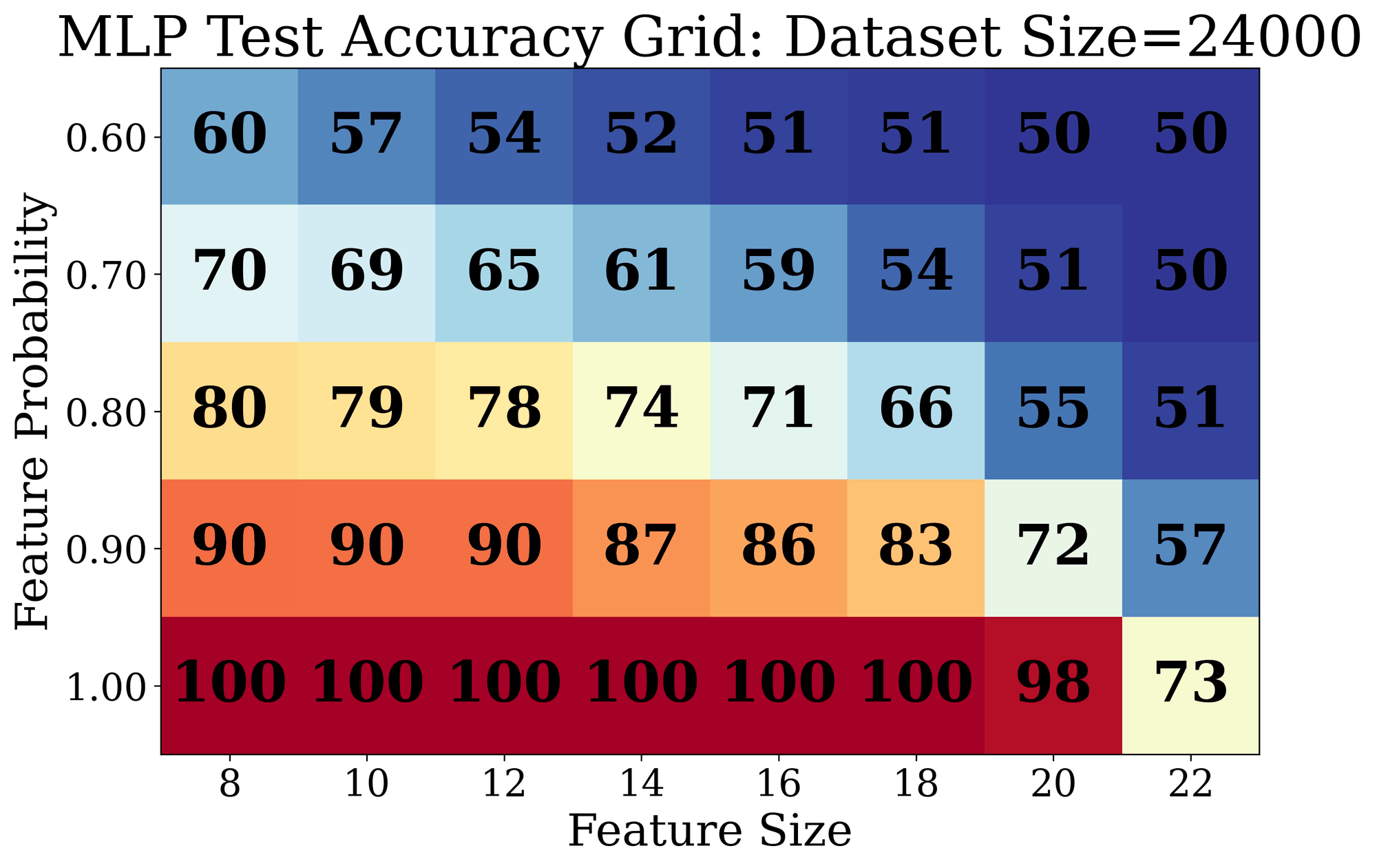}
    \end{subfigure}
    \begin{subfigure}[width=\textwidth]{0.19 \textwidth}
        \includegraphics[width=\textwidth]{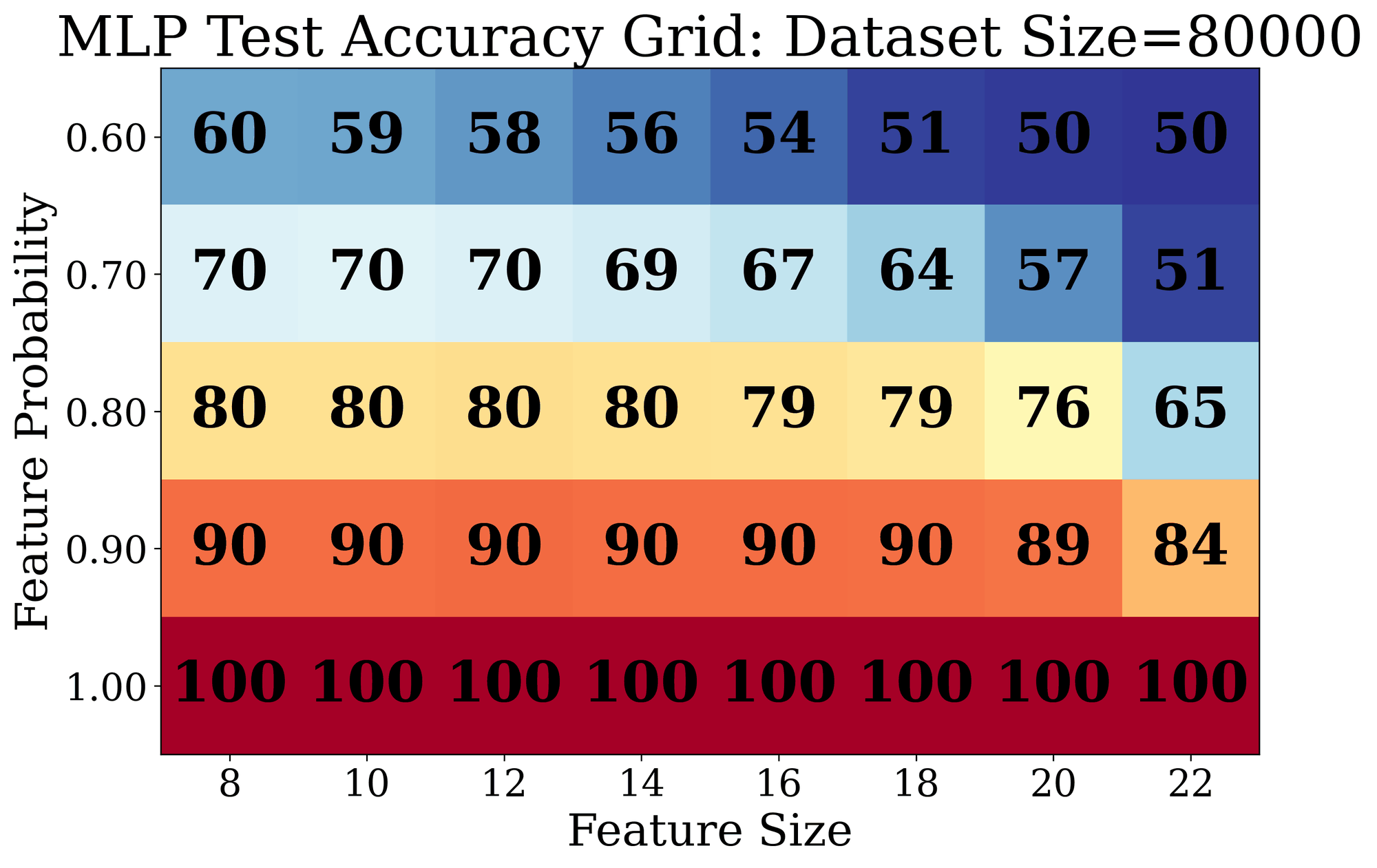}
    \end{subfigure}
    \begin{subfigure}[width=\textwidth]{0.19 \textwidth}
        \includegraphics[width=\textwidth]{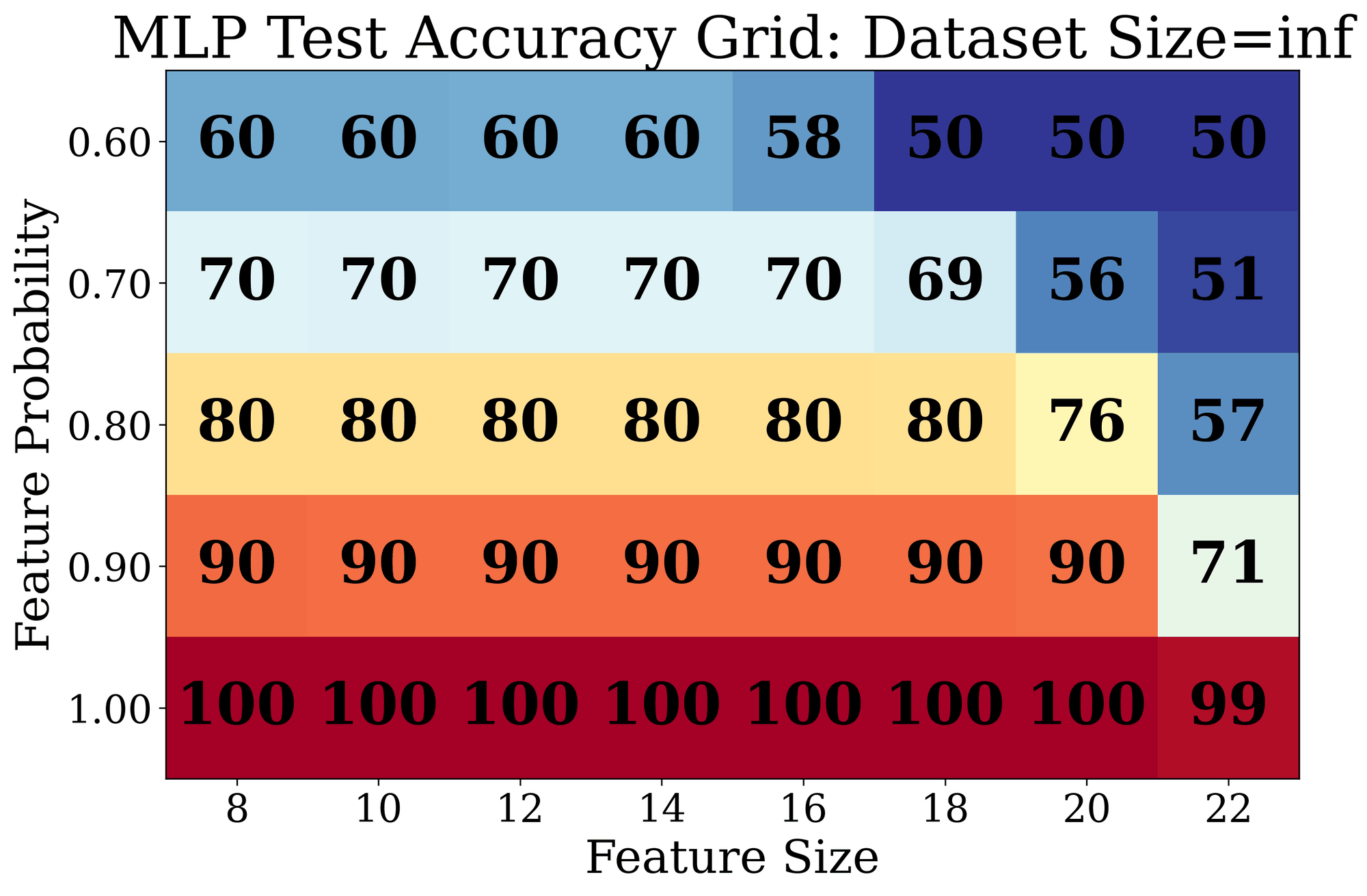}
    \end{subfigure}

    \begin{subfigure}[width=\textwidth]{0.19 \textwidth}
        \includegraphics[width=\textwidth]{figures/noisy/noisy_feature_scaling_w10_d3_lr0.0001/noisy_feature_scaling_grid_gpt_size3000.png}
    \end{subfigure}
    \begin{subfigure}[width=\textwidth]{0.19 \textwidth}
        \includegraphics[width=\textwidth]{figures/noisy/noisy_feature_scaling_w10_d3_lr0.0001/noisy_feature_scaling_grid_gpt_size10000.png}
    \end{subfigure}
    \begin{subfigure}[width=\textwidth]{0.19 \textwidth}
        \includegraphics[width=\textwidth]{figures/noisy/noisy_feature_scaling_w10_d3_lr0.0001/noisy_feature_scaling_grid_gpt_size30000.png}
    \end{subfigure}
    \begin{subfigure}[width=\textwidth]{0.19 \textwidth}
        \includegraphics[width=\textwidth]{figures/noisy/noisy_feature_scaling_w10_d3_lr0.0001/noisy_feature_scaling_grid_gpt_size100000.png}
    \end{subfigure}
    \begin{subfigure}[width=\textwidth]{0.19 \textwidth}
        \includegraphics[width=\textwidth]{figures/noisy/noisy_feature_scaling_w10_d3_lr0.0001/noisy_feature_scaling_grid_gpt_size0.png}
    \end{subfigure}

    \caption{Test Accuracy on i.i.d test set for models trained on a noisy k-parity feature with Constant Iteration Budget ($T=10000$).
    An ``infinite'' data set size is achieved by directly sampling from the DGP.}    \label{fig:noisy_k_parity_test_fixed}
\end{figure*}

\begin{figure*}[h]
    \begin{subfigure}[width=\textwidth]{0.135  \textwidth}
        \includegraphics[width=\textwidth]{figures/2_f/1000_2f_weak_vs_strong_w10_TTrue_lr0.0001/2f_weak_vs_strong_grid_gpt_f6size8_tieredTrue.png}
    \end{subfigure}
    \begin{subfigure}[width=\textwidth]{0.135  \textwidth}
        \includegraphics[width=\textwidth]{figures/2_f/3000_2f_weak_vs_strong_w10_TTrue_lr0.0001/2f_weak_vs_strong_grid_gpt_f6size8_tieredTrue.png}
    \end{subfigure}
    \begin{subfigure}[width=\textwidth]{0.135  \textwidth}
        \includegraphics[width=\textwidth]{figures/2_f/10000_2f_weak_vs_strong_w10_TTrue_lr0.0001/2f_weak_vs_strong_grid_gpt_f6size8_tieredTrue.png}
    \end{subfigure}
    \begin{subfigure}[width=\textwidth]{0.135  \textwidth}
        \includegraphics[width=\textwidth]{figures/2_f/30000_2f_weak_vs_strong_w10_TTrue_lr0.0001/2f_weak_vs_strong_grid_gpt_f6size8_tieredTrue.png}
    \end{subfigure}
    \begin{subfigure}[width=\textwidth]{0.135  \textwidth}
        \includegraphics[width=\textwidth]{figures/2_f/100000_2f_weak_vs_strong_w10_TTrue_lr0.0001/2f_weak_vs_strong_grid_gpt_f6size8_tieredTrue.png}
    \end{subfigure}
    \begin{subfigure}[width=\textwidth]{0.135  \textwidth}
        \includegraphics[width=\textwidth]{figures/2_f/300000_2f_weak_vs_strong_w10_TTrue_lr0.0001/2f_weak_vs_strong_grid_gpt_f6size8_tieredTrue.png}
    \end{subfigure}
    \begin{subfigure}[width=\textwidth]{0.135 \textwidth}
        \includegraphics[width=\textwidth]{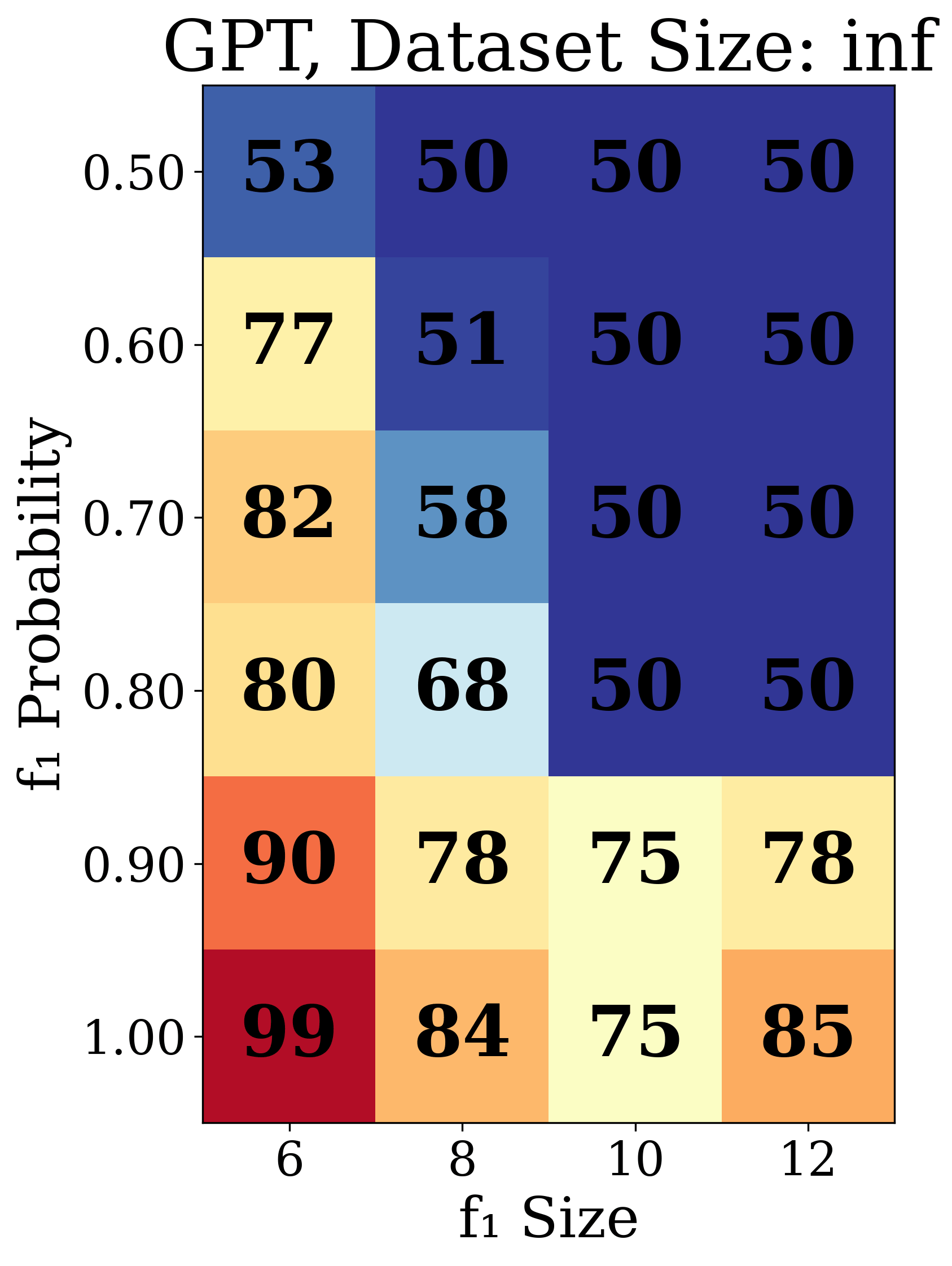}
    \end{subfigure}

    \caption{Test accuracies for Two Feature Experiments Section \ref{sec:two_features} for the GPT Model with constant compute budget $(T=10000)$.
    As the data set increases in size the model fails to learn either feature within the training budget.
    An ``infinite'' data set size is achieved by directly sampling from the DGP.}
    \label{fig:two_features_gpt_fixed}
\end{figure*}

\begin{figure*}[h]
    \centering
    \begin{subfigure}[]{0.24 \textwidth}
        \includegraphics[width=\textwidth]{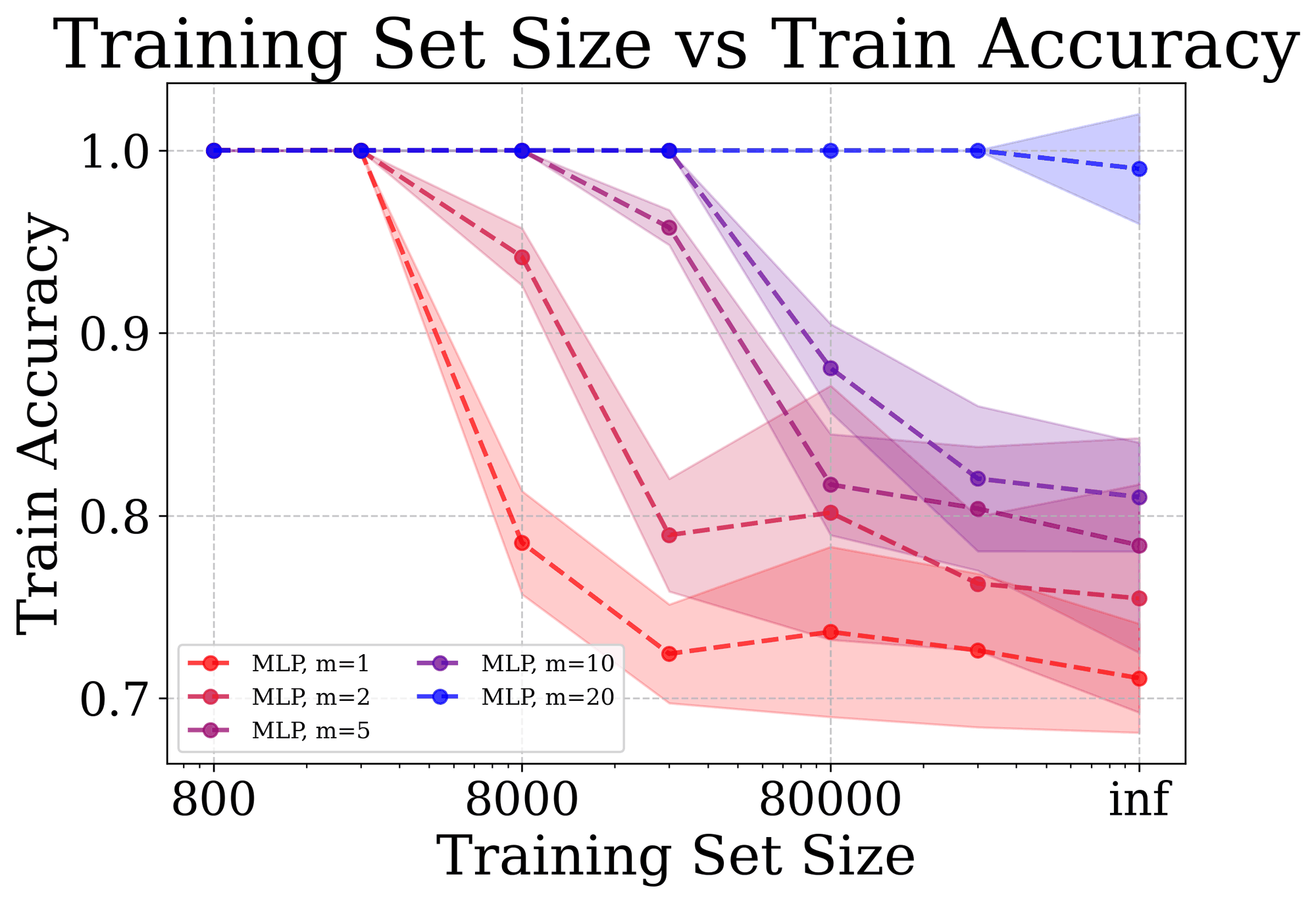}
    \end{subfigure}
    \begin{subfigure}[]{0.24 \textwidth}
        \includegraphics[width=\textwidth]{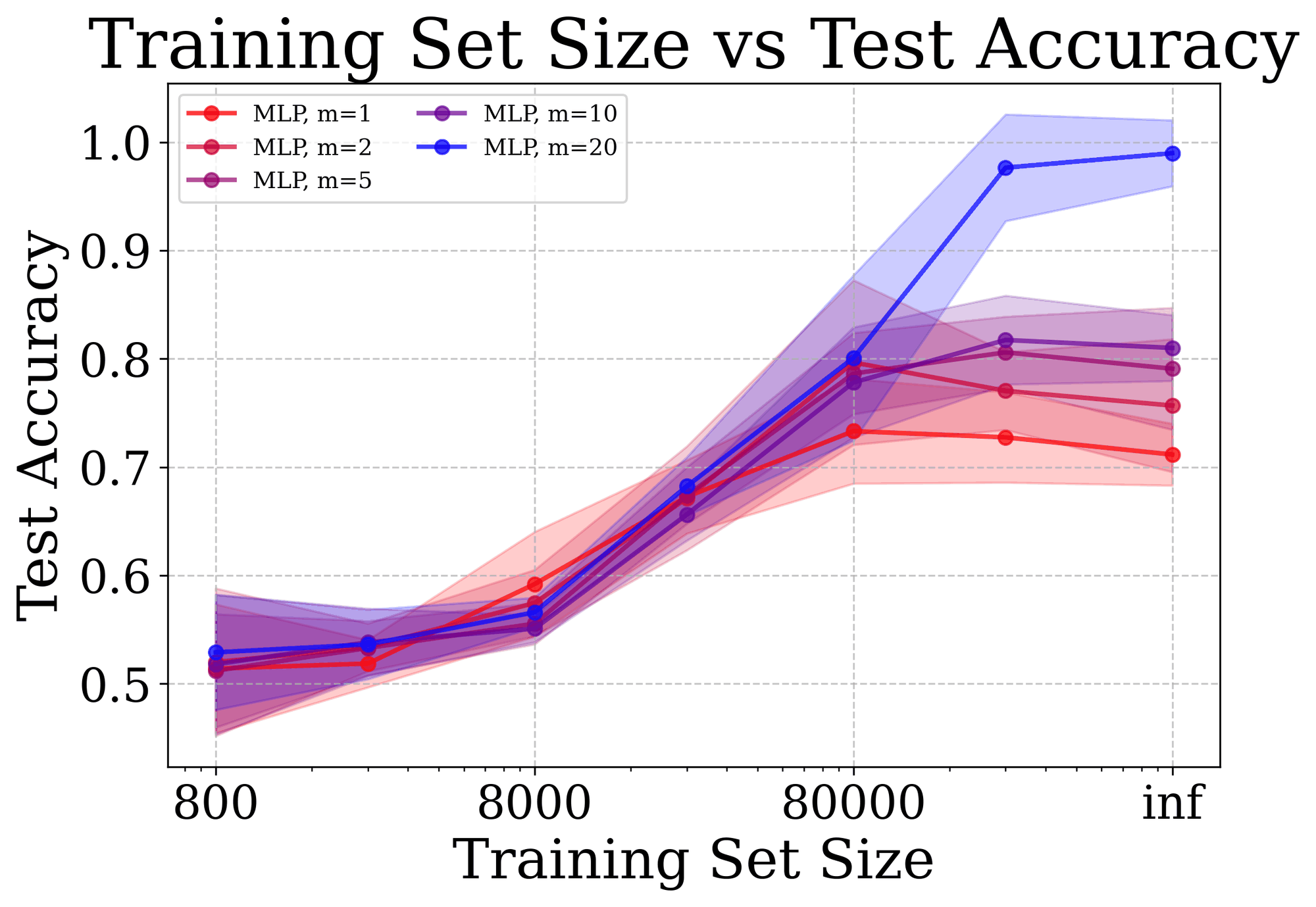}
    \end{subfigure}
    \begin{subfigure}[]{0.24 \textwidth}
        \includegraphics[width=\textwidth]{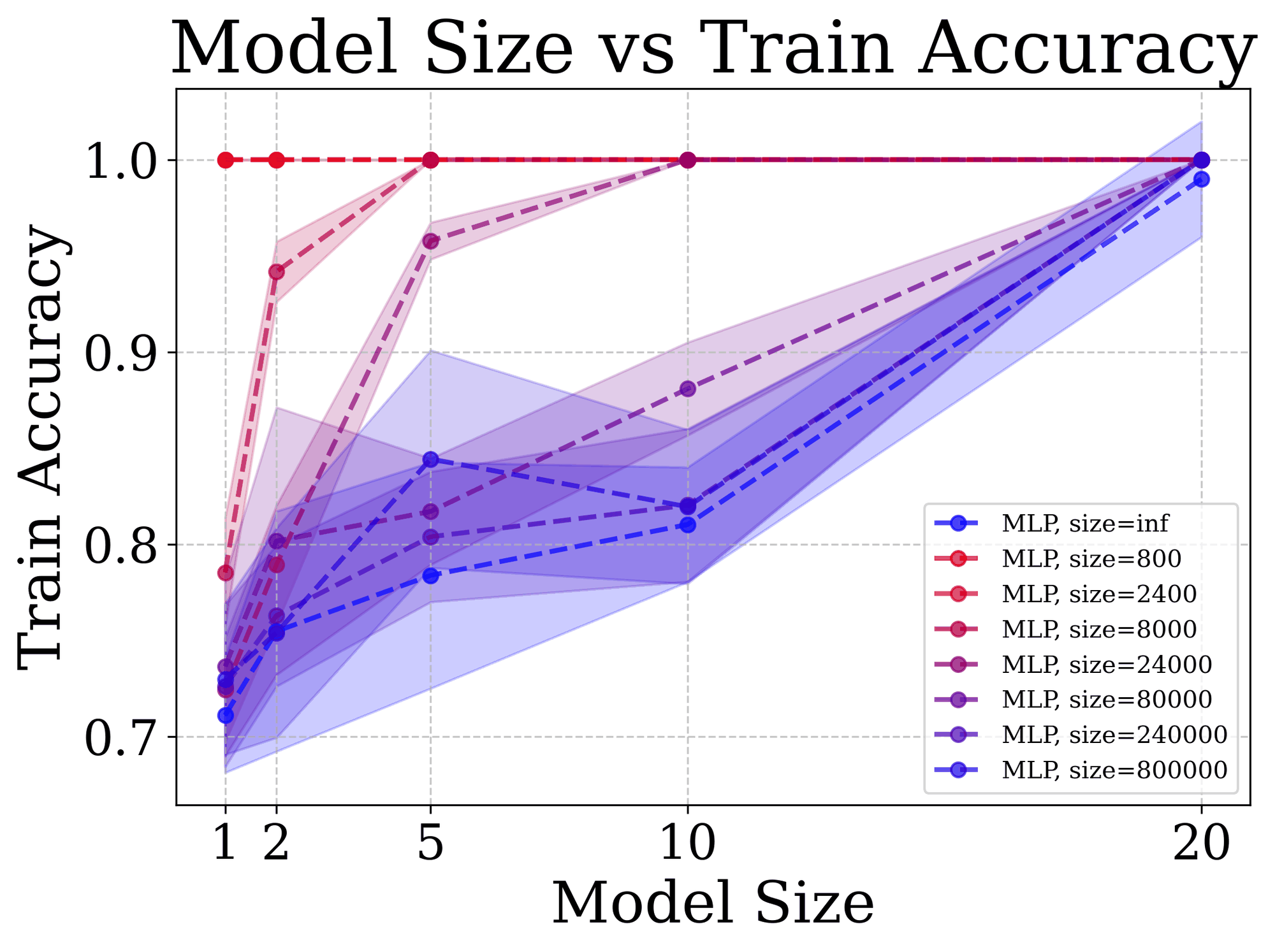}
    \end{subfigure}
    \begin{subfigure}[]{0.24 \textwidth}
        \includegraphics[width=\textwidth]{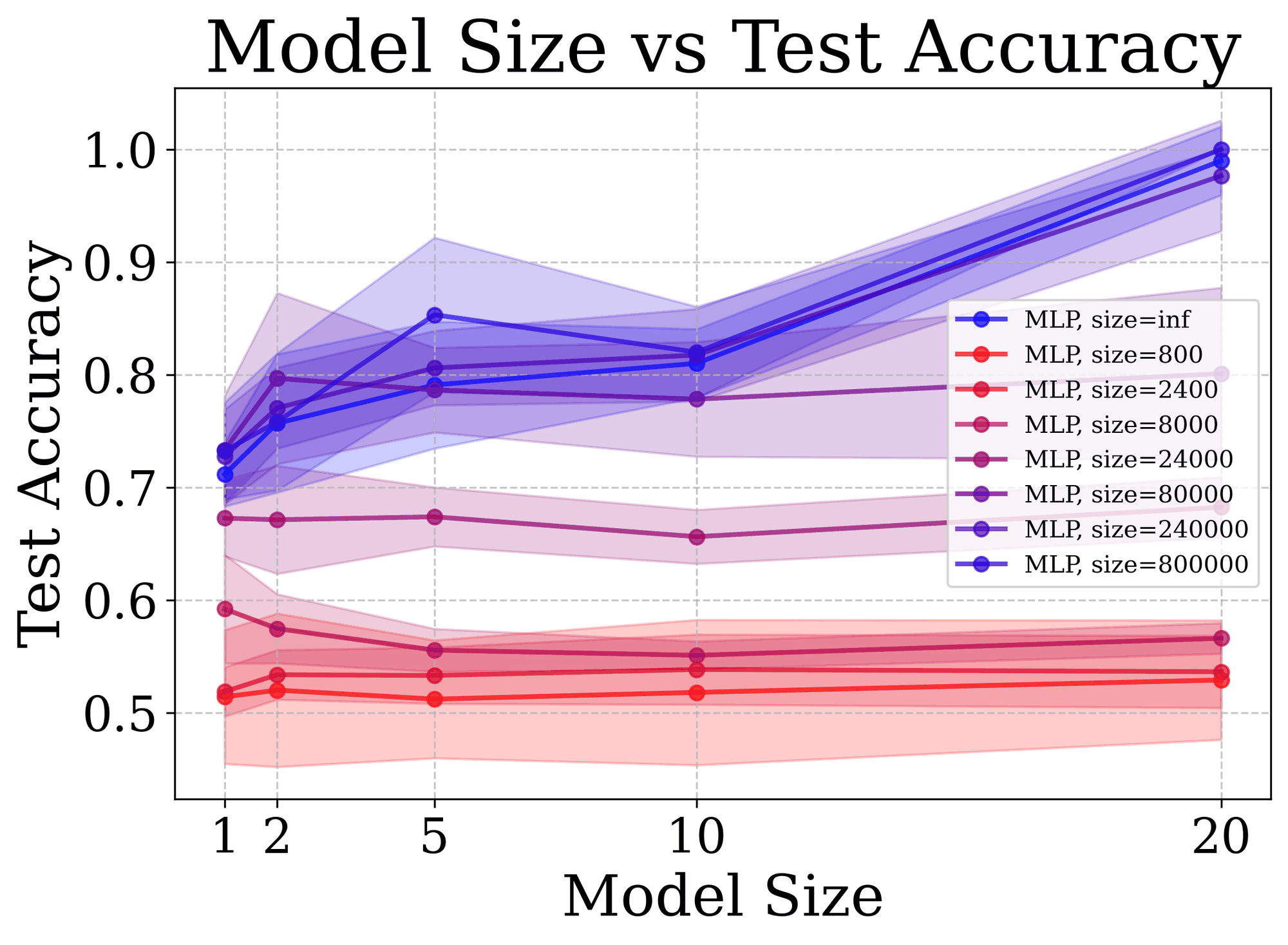}
    \end{subfigure}

    \begin{subfigure}[]{0.24 \textwidth}
        \includegraphics[width=\textwidth]{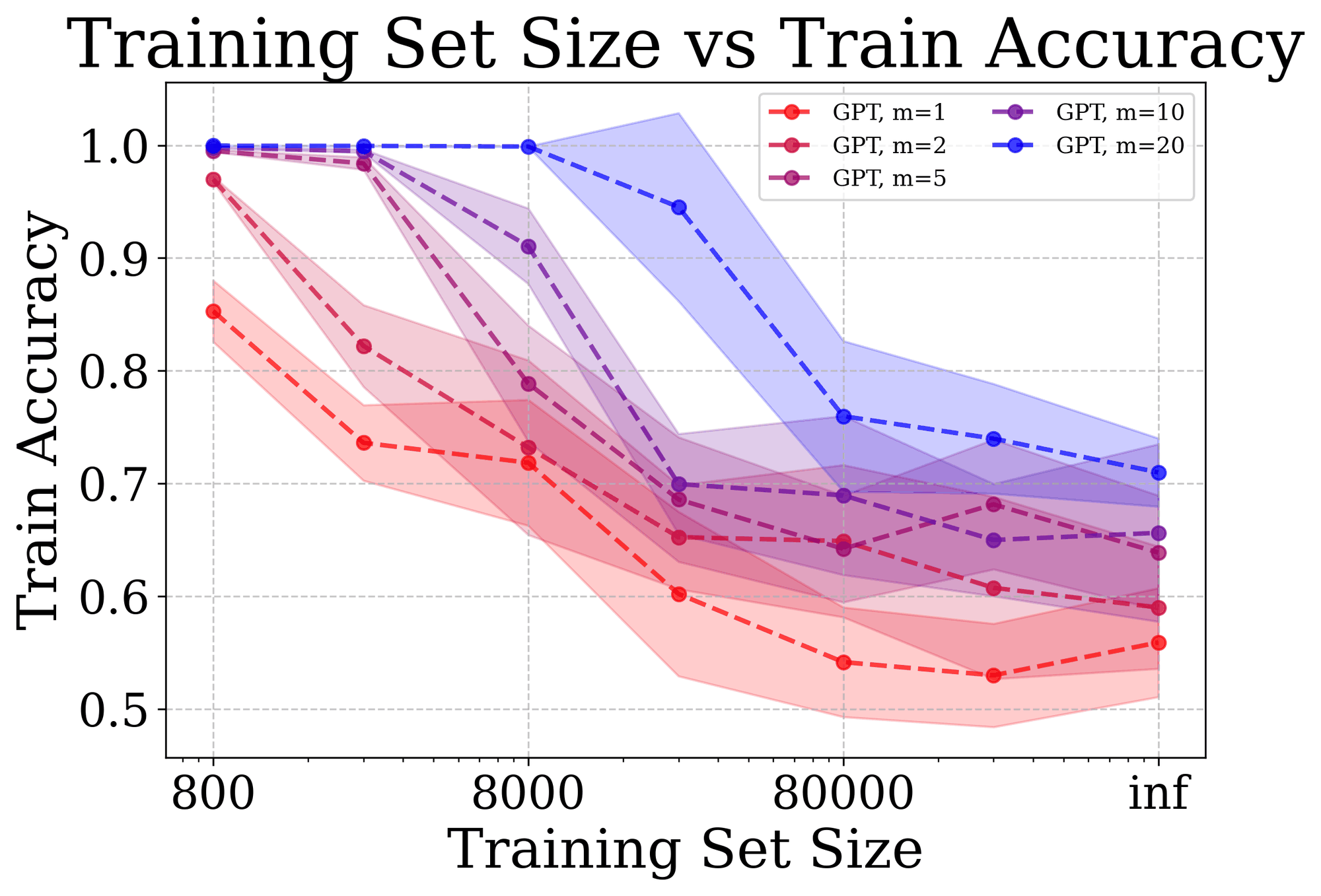}
    \end{subfigure}
    \begin{subfigure}[]{0.24 \textwidth}
        \includegraphics[width=\textwidth]{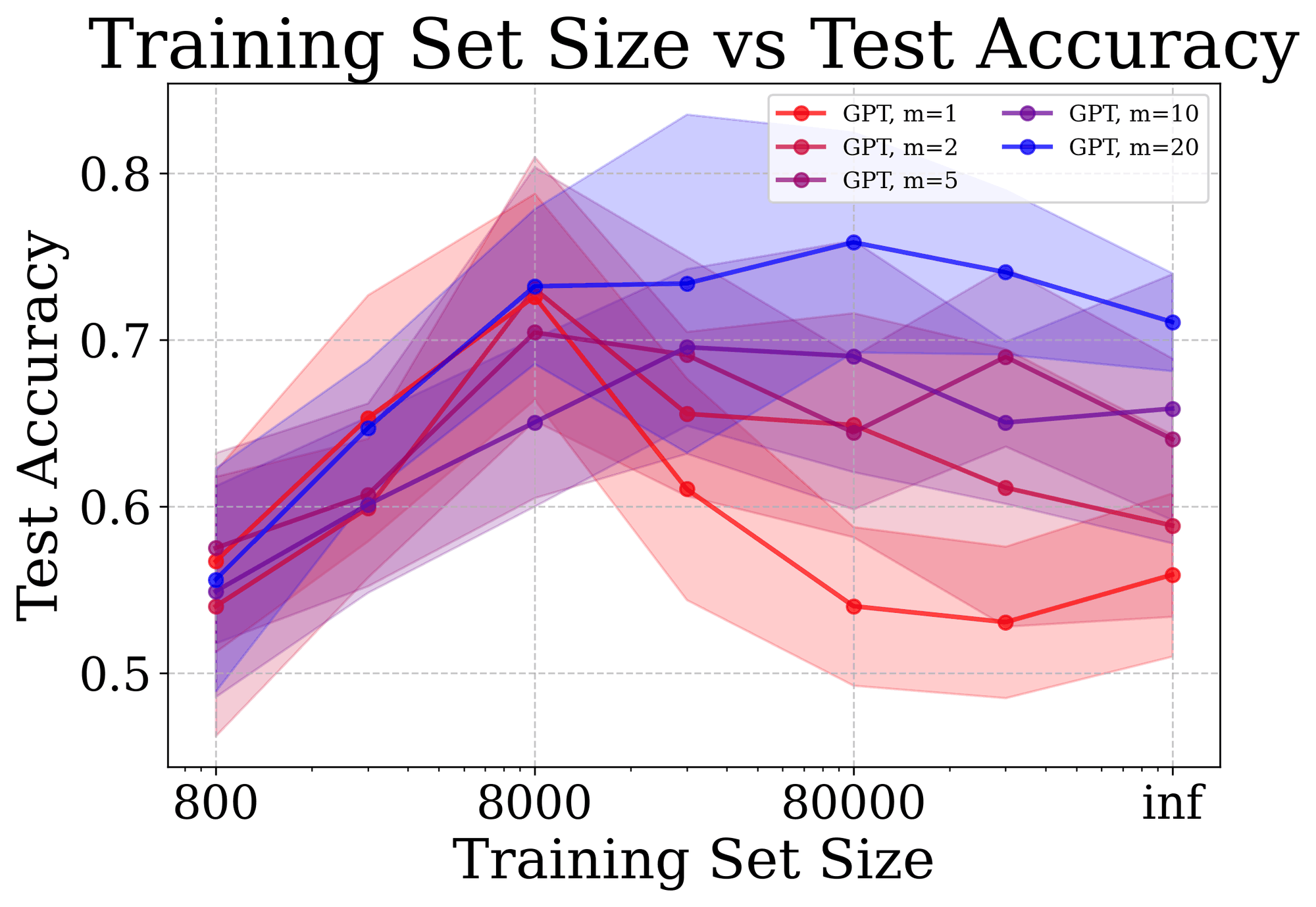}
    \end{subfigure}
    \begin{subfigure}[]{0.24 \textwidth}
        \includegraphics[width=\textwidth]{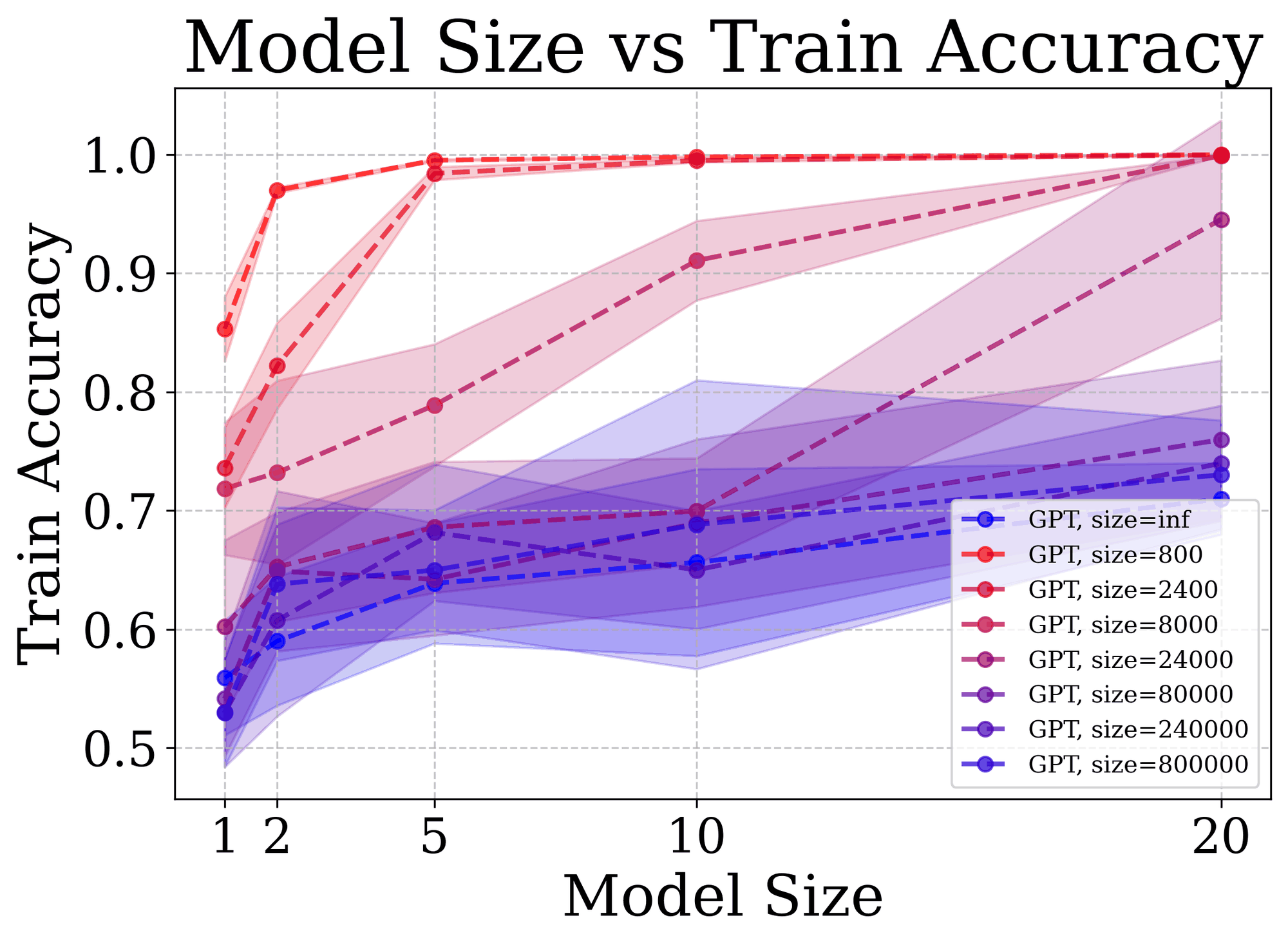}
    \end{subfigure}
    \begin{subfigure}[]{0.24 \textwidth}
        \includegraphics[width=\textwidth]{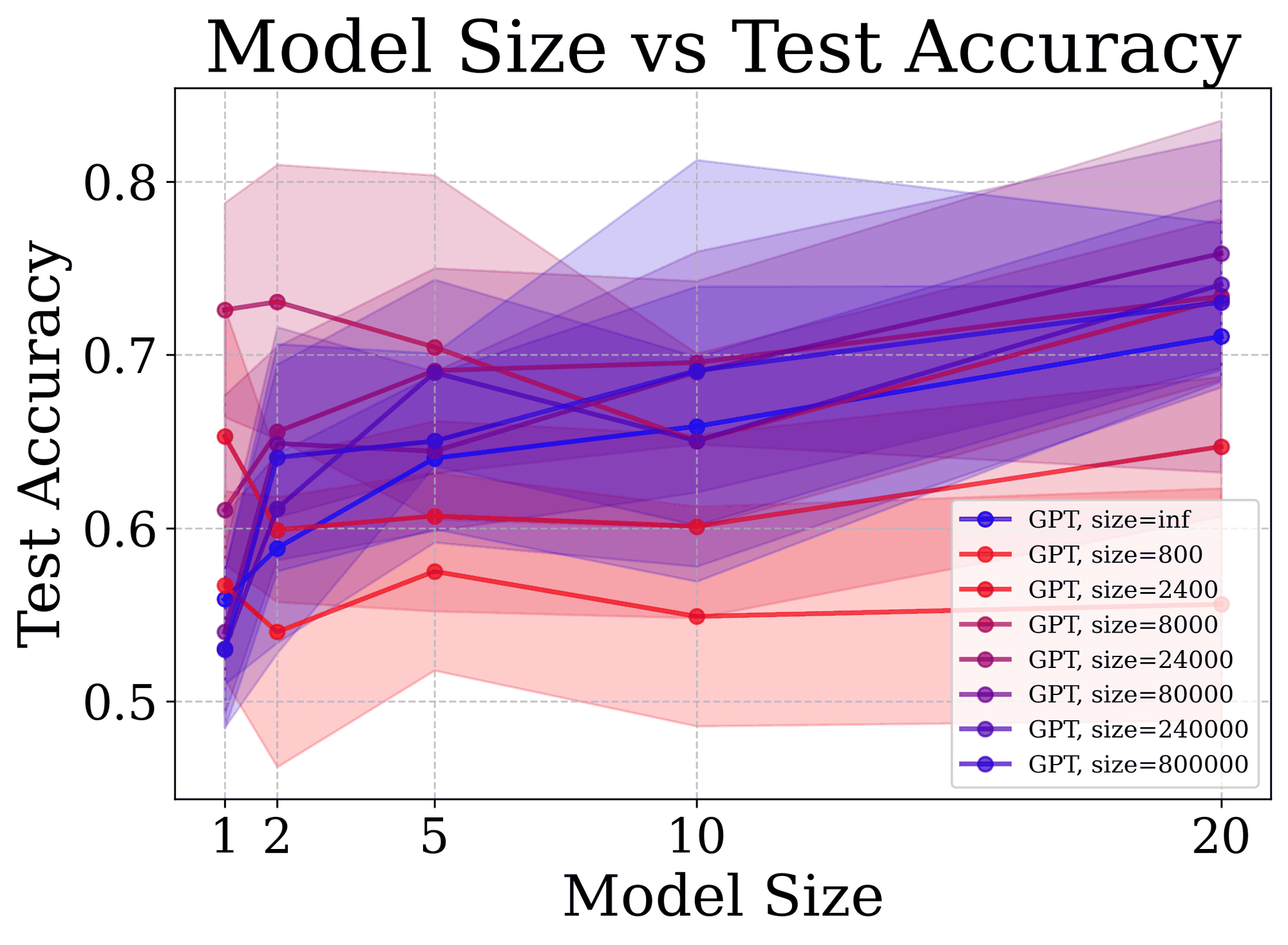}
    \end{subfigure}

    \caption{Test accuracy for Tiered Data Set with a fixed training budget \textbf{Top:} MLP.
    \textbf{Bottom:} GPT.}
    \label{fig:tiered_full}
\end{figure*}

\begin{figure*}[h]
    \centering
    \begin{subfigure}[]{0.24 \textwidth}
        \includegraphics[width=\textwidth]{figures/tiered_all_iter_scaling/mlp_dataset_size_vs_accuracy_set_train.png}
    \end{subfigure}
    \begin{subfigure}[]{0.24 \textwidth}
        \includegraphics[width=\textwidth]{figures/tiered_all_iter_scaling/mlp_dataset_size_vs_accuracy_set.png}
    \end{subfigure}
    \begin{subfigure}[]{0.24 \textwidth}
        \includegraphics[width=\textwidth]{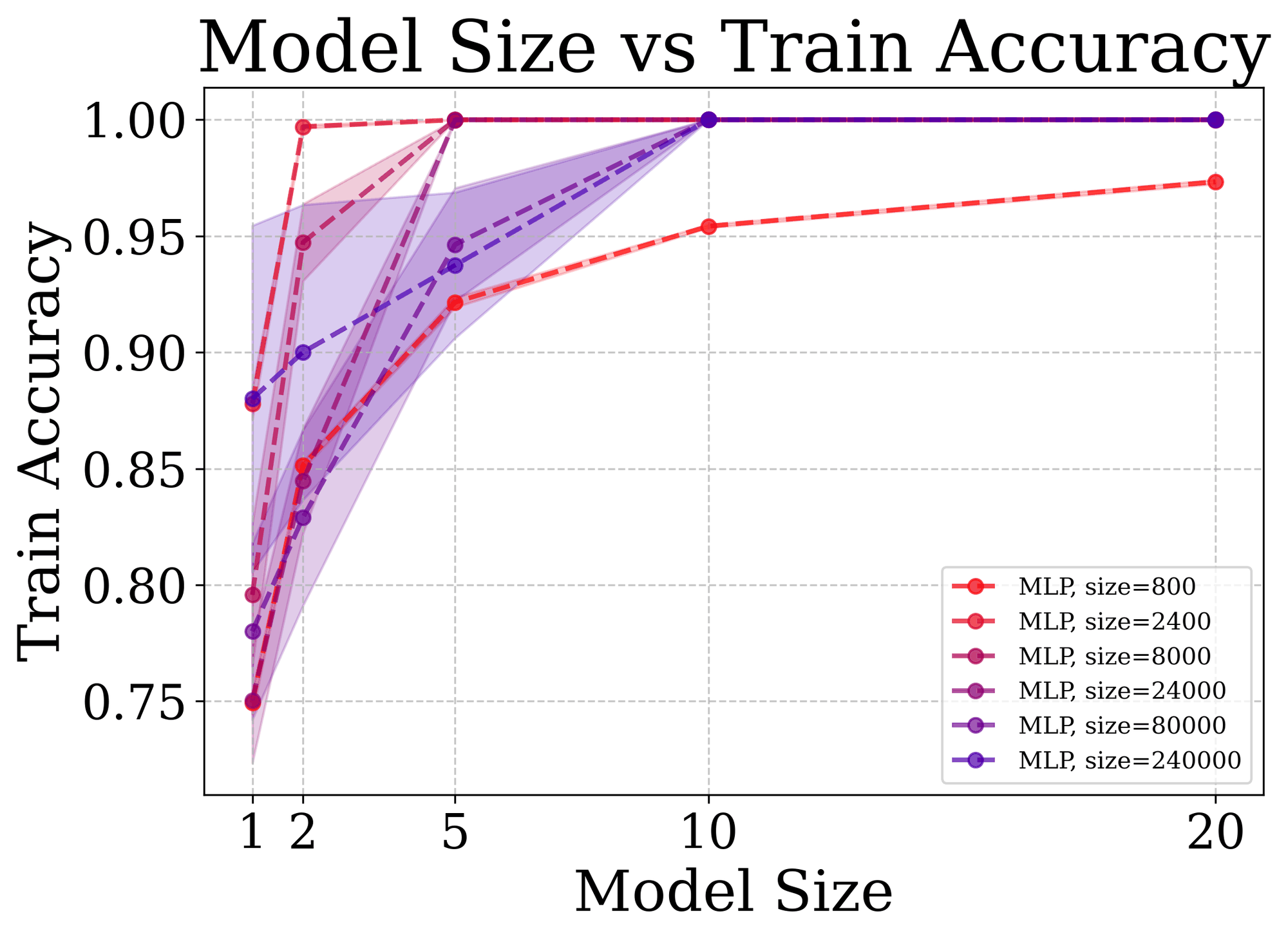}
    \end{subfigure}
    \begin{subfigure}[]{0.24 \textwidth}
        \includegraphics[width=\textwidth]{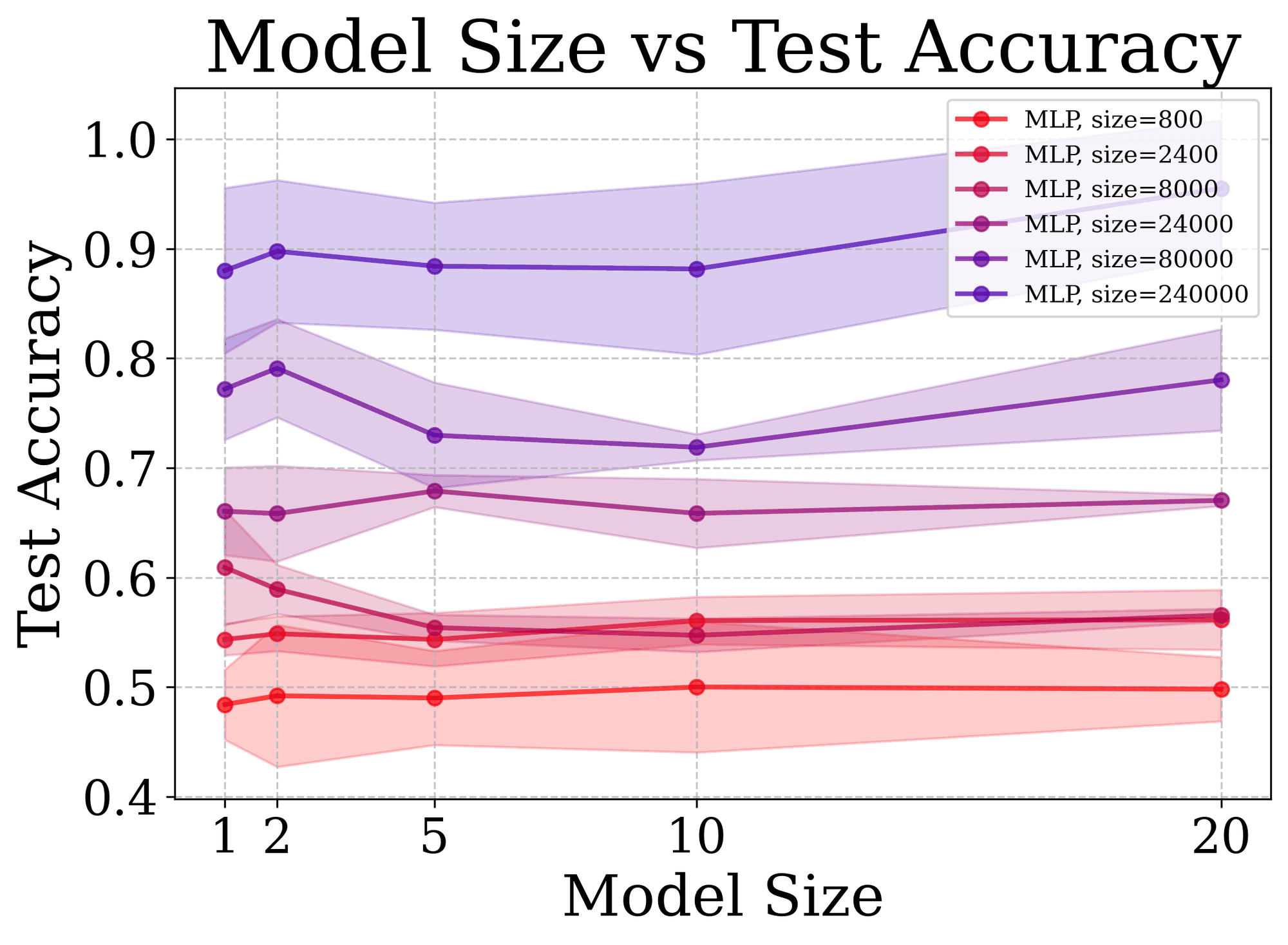}
    \end{subfigure}

    \begin{subfigure}[]{0.24 \textwidth}
        \includegraphics[width=\textwidth]{figures/tiered_all_iter_scaling/gpt_dataset_size_vs_accuracy_set_train.png}
    \end{subfigure}
    \begin{subfigure}[]{0.24 \textwidth}
        \includegraphics[width=\textwidth]{figures/tiered_all_iter_scaling/gpt_dataset_size_vs_accuracy_set.png}
    \end{subfigure}
    \begin{subfigure}[]{0.24 \textwidth}
        \includegraphics[width=\textwidth]{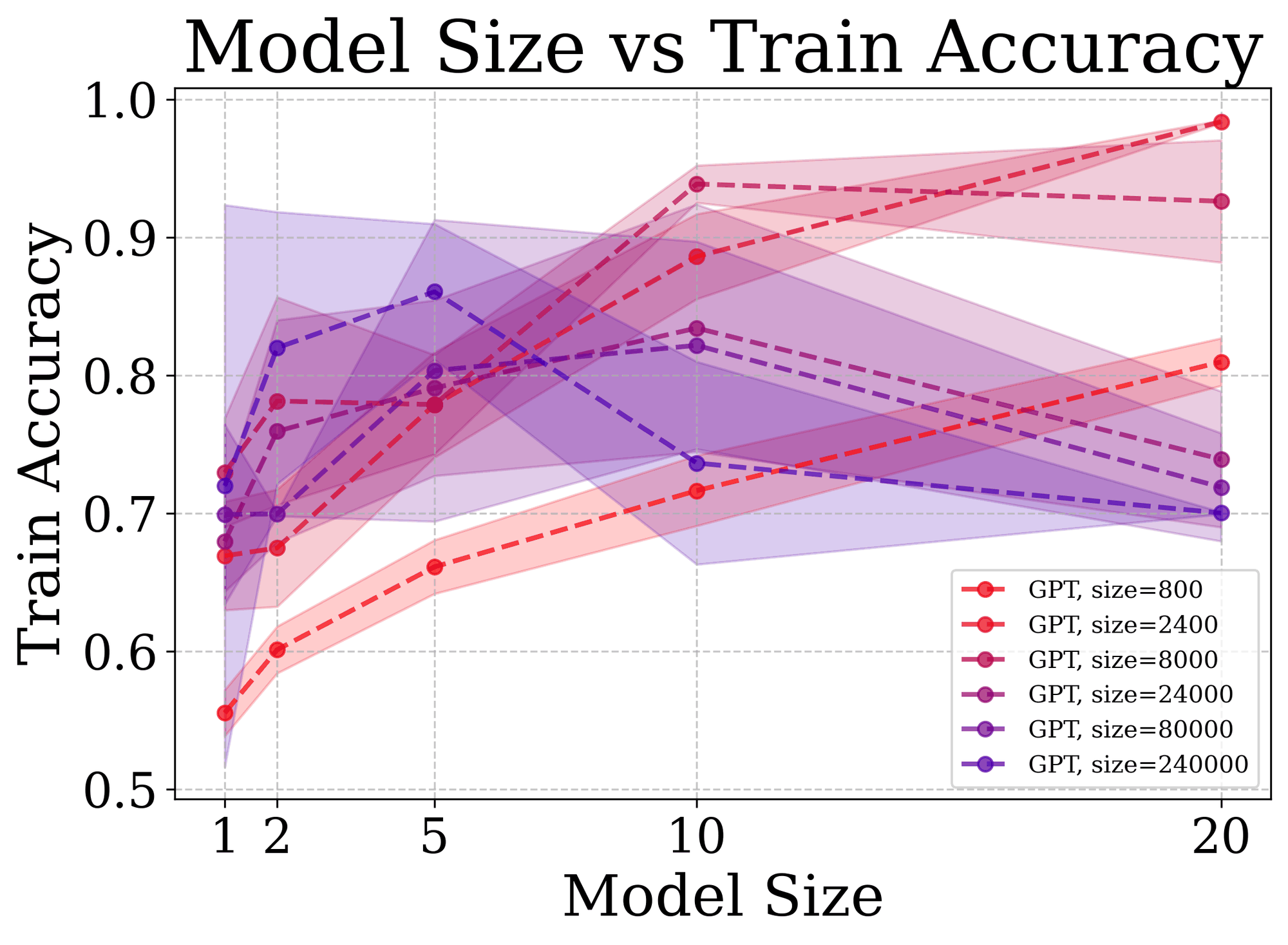}
    \end{subfigure}
    \begin{subfigure}[]{0.24 \textwidth}
        \includegraphics[width=\textwidth]{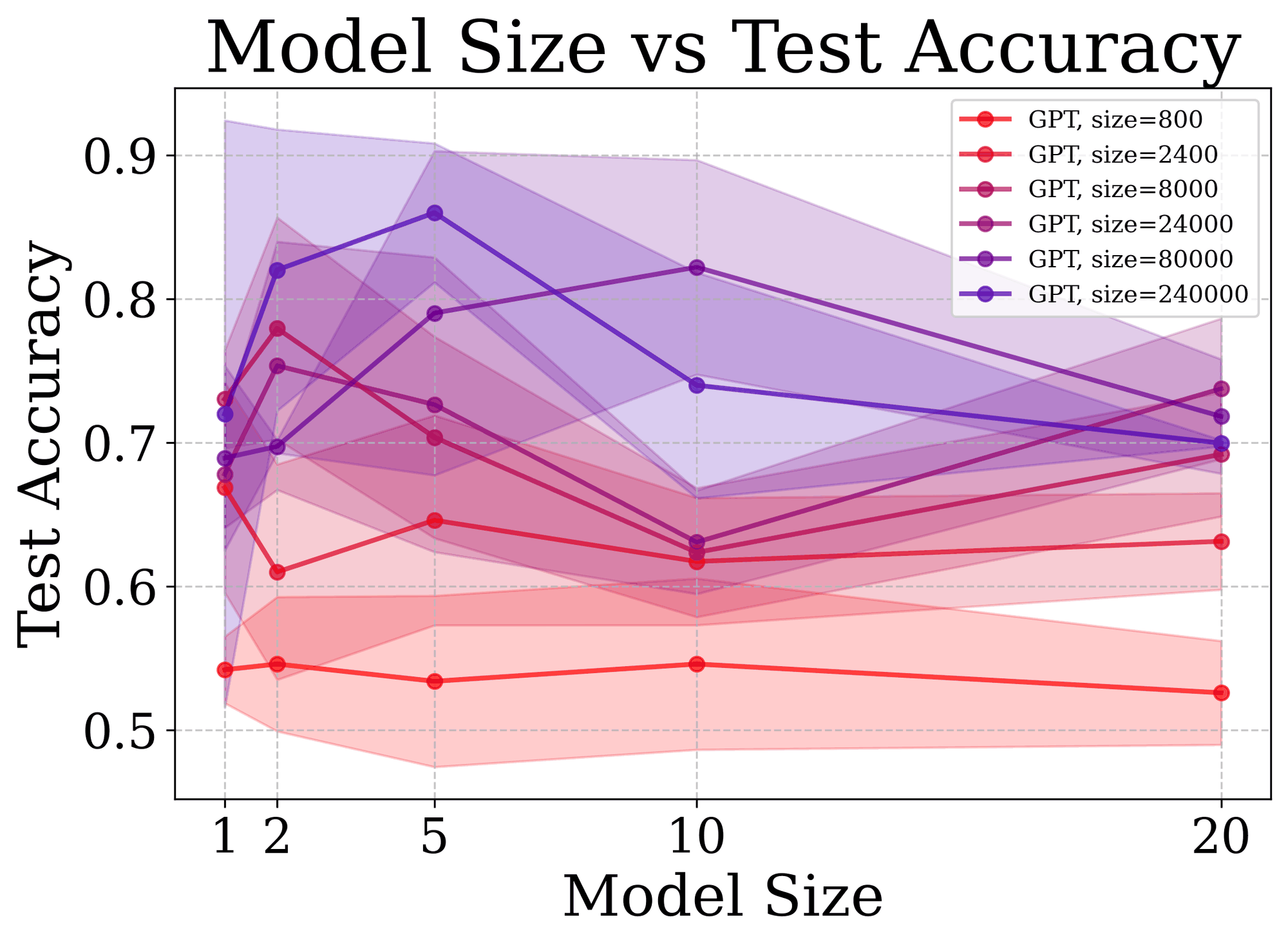}
    \end{subfigure}

    \caption{Test accuracy for Tiered Data Set with a scaling training budget \textbf{Top:} MLP.
    \textbf{Bottom:} GPT.}
    \label{fig:tiered_iter_scaling_full}
\end{figure*}

\begin{figure*}[h]
    \centering
    \begin{subfigure}[width=\textwidth]{0.48 \textwidth}
        \includegraphics[width=\textwidth]{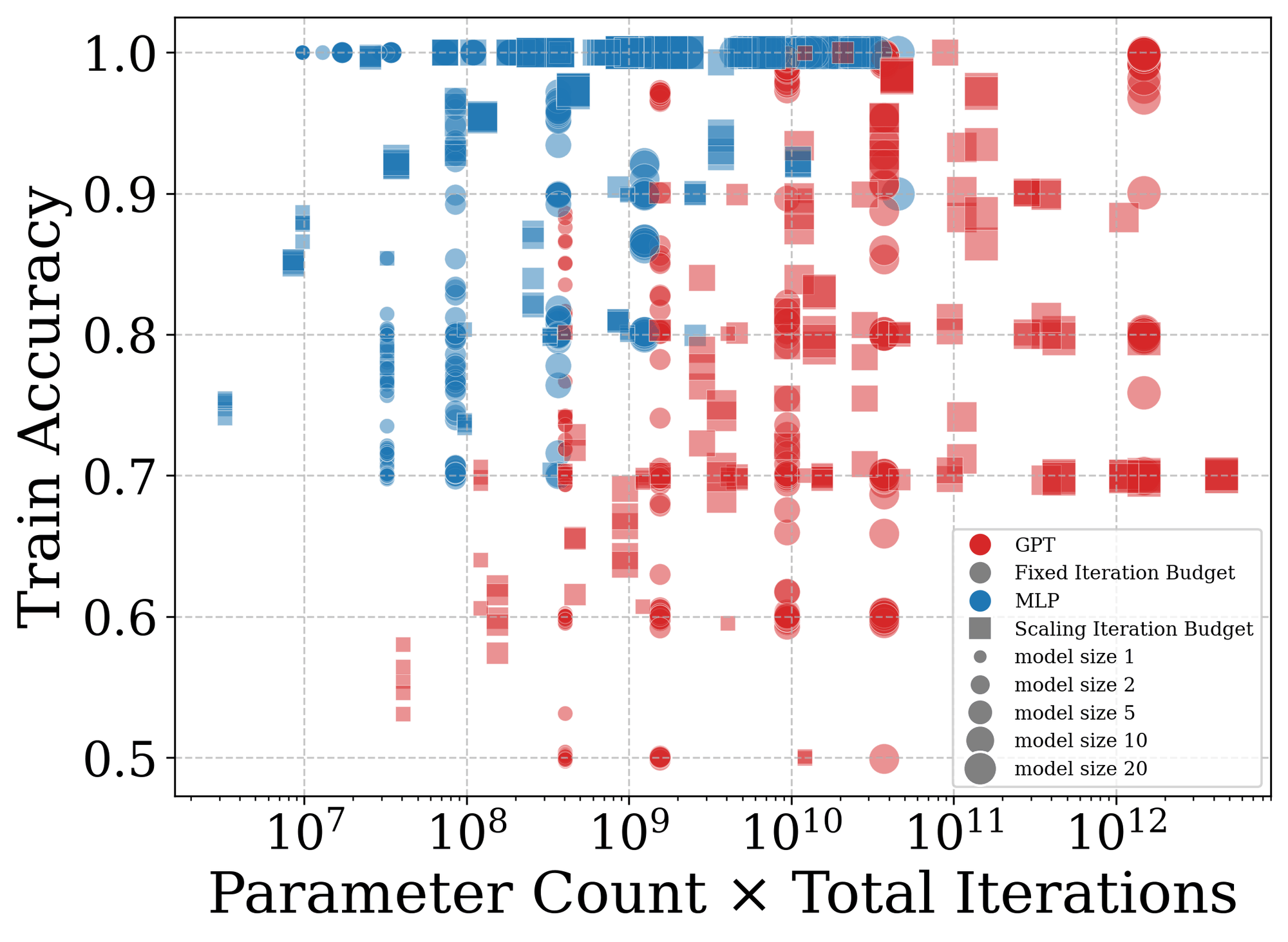}
    \end{subfigure}
    \begin{subfigure}[width=\textwidth]{0.48 \textwidth}
        \includegraphics[width=\textwidth]{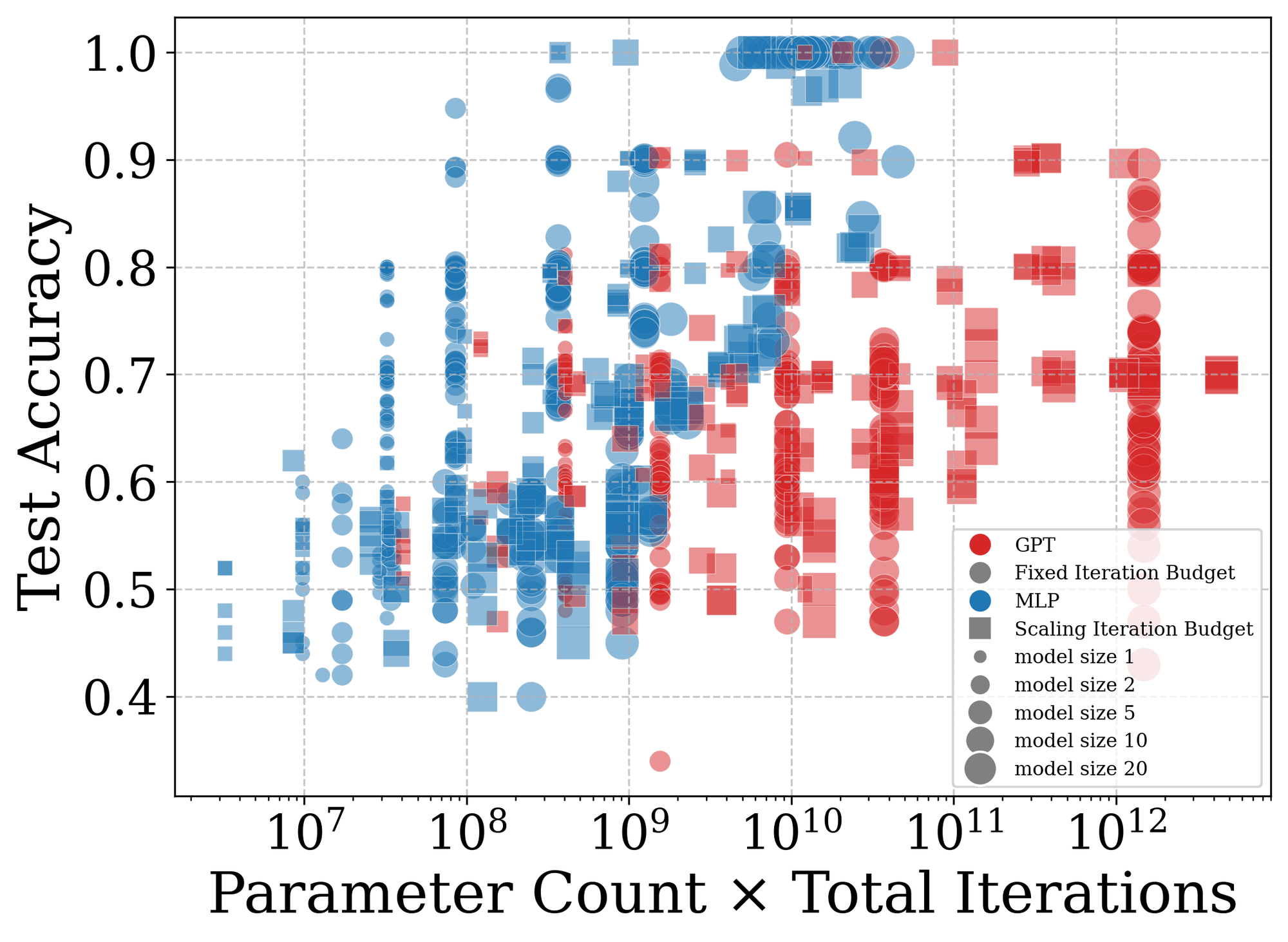}
    \end{subfigure}
    \caption{Plots showing accuracy against parameter times training iterations for models trained on the hierarchical data set described in Section \ref{sec:strict_hierarchical}.
    On this  hard non-linear data set composed for 5 noisy k-parity features MLPs offer an order of magnitude more efficient learner.}
    \label{fig:scatter}
\end{figure*}

The appendix is organised as follows.
Section \ref{sec:dgp} gives the data generating process.
Section \ref{app:hyperparams} gives the model hyperparameters, the compute resources (Section \ref{sec:compute}), the Transformer learning-rate sweeps (Figures \ref{fig:lr_noisy} and \ref{fig:lr_two_features}) and the model-size sweeps (Figures \ref{fig:model_scaling_all}, \ref{fig:model_scaling} and \ref{fig:model_scaling_finite}).
Section \ref{app:train_tables} reports train and test accuracy tables for the fixed budget (Tables \ref{tab:fixed_train} and \ref{tab:fixed_test}) and the scaled budget (Tables \ref{tab:scaled_train} and \ref{tab:scaled_test}), and the effective number of passes (Table \ref{tab:passes}).
Section \ref{sec:strict_hierarchical} describes hierarchical data sets (Table \ref{tab:data_lin}).
Section \ref{app:additional_results} contains additional plots: training and validation curves (Figures \ref{fig:training_curves} and \ref{fig:validation_curves}), the scaled-budget single-feature results (Figure \ref{fig:iter_scaling_k_parity_noiseless}), the curriculum experiment (Figure \ref{fig:curriculum}), the noisy k-parity grids (Figures \ref{fig:iter_scaling_k_parity}, \ref{fig:noisy_k_parity_appendix}, \ref{fig:noisy_k_parity_test_scaling} and \ref{fig:noisy_k_parity_test_fixed}), the two-feature and tiered data sets (Figures \ref{fig:two_features_gpt_fixed}, \ref{fig:tiered_full} and \ref{fig:tiered_iter_scaling_full}) and the parameter and compute scatter plot (Figure \ref{fig:scatter}).

\section{Data Generating Process}\label{sec:dgp}
In this section we explain in detail the data generating process (DGP) so that the resulting data set encodes the desired set of features. 
For the $n^{th}$ data point we first sample the label $y_n$ uniformly in $\{-1,1\}$, then we sample an input vector $\vz_{i,n} \in \{0,1\}^d$ for each feature $f_i$ independently, except where stated otherwise (see Section \ref{sec:strict_hierarchical}). 
This sampling is done by uniformly selecting an input $\vz_{i,n}$ from the set $\{ \vz \in \{0,1\}^{d_i} : g_{d_i}(\vz) = y\}$, where $d_i$ is the number of bits of the $i^{th}$ feature (identifying $y\in\{-1,1\}$ with the parity output via $y=2g-1$). 
This process is repeated for all features $f_i \in \mathcal{F}$.
In order to precisely set the predictivity we introduce $1-\alpha$ feature noise similar to \citet{peters2026trapped}, so for $1-\alpha$ \% of examples the $\vz_{i,n}$ is made unpredictive of the label.\footnote{This is done by uniformly flipping a bit at an index where $\vh_i$ is 1.}
For convenience each $\vz_i$ is then padded by the mean input value $(0.5)$ so that all features are the same length, giving $\hat{\vz}_{i,n} \in \{0,0.5,1\}^{d_{max}}, \forall i$, where $d_{max}$ is one more than the maximum features size ($d$) of all features in the data set. 
Next $\hat{\vz}_{i,n}$ are concatenated for all $i$. 
Finally, the input data is rescaled.
Specifically, the $n^{th}$ input in $\mathcal{D}$, specifically $\vx_n$ is given by $\vx_n = 2*[\hat{\vz}_{i,n}, \dots, \hat{\vz}_{f,n}]-1$ where we rescale the data so the input lives in $\vx_n \in \{-1,0,1\}^{d_{max} \times F}$. 
This also normalises the data to have zero mean and roughly unit variance. 
This means the input dimension to the model $d_{in} = d_{max} \times F$, where $F$ is the total number of features.
The test, train and validation data sets are sampled using the exact same DGP, except if stated otherwise.
We note that this DGP gives each sample $\vx \in \{-1,0,1\}^{d_{in}}$ equal likelihood of being sampled for a data set $\mathcal{D}$, which is a convenient property but unrealistic of most real world data sets which have long tails of less common examples of certain features.

\section{Models and Hyperparameters}\label{app:hyperparams}

\subsection{MLP}
The MLP model is a simple MLP of three layers alternating linear layers with ReLU activations.
The input dimension $d_{in}$ is set to $d_{in} = d_{max} \times F$.
As we are performing binary classification $d_{out} = 2$.
The MLP has three linear layers with a hidden dimension of $m \times 32$.
The input is feed directly into the model with no tokenisation. 

\subsection{GPT}
The Transformer model used was adapted directly from MINGPT \citep{karpathy2020minGPT} with changes only made to the model hyperparameters.
The embedding dimension was set to $m \times 32$, the number of heads per layer to $m \times 2$ and the vocabulary size was changed to three.
All inputs $\vx$ considered in this paper all live in $\vx \in \{-1,0,1\}^{d_{in}}$ where $d_{in} = d_{max} \times F$, as described in Section \ref{sec:dgp}.
Hence they can be modelled as a language with a vocabulary over three tokens.
For the GPT model a weight decay of $10^{-3}$ and drop percentage of $0.1$ were used.

\subsection{Hyperparameters}
The Adam optimiser \citep{Kingma2015} (AdamW for the GPT model, see above) with a batch size of 256 was used for all experiments.
SGD can be highly sensitive to its learning rate \citep{schaul2013no} and hence in order to avoid the need for recalibration it was not used.
A learning rate of $3\times10^{-4}$ was used for the MLP model for all experiments.
For the GPT model a learning rate of $1\times10^{-4}$ was used instead. 
The Val and Test sets have a size of $0.1N$ or $25600$ when $N=\text{inf}$, where $N$ is the training set size.

\subsection{Compute Resources}\label{sec:compute}
All experiments were run on NVIDIA T4 GPUs.
The longest single runs took approx 2.5 GPU days.
The majority took significantly less.
We performed over 50,000 unique training runs.
We estimate all experiments could be rerun using $\sim$2000 T4 GPU days, depending on the number of repeats performed.  

\subsection{Transformer Learning Rate}

\begin{figure*}[ht]
    \centering

    \begin{subfigure}[width=\textwidth]{0.19 \textwidth}
        \includegraphics[width=\textwidth]{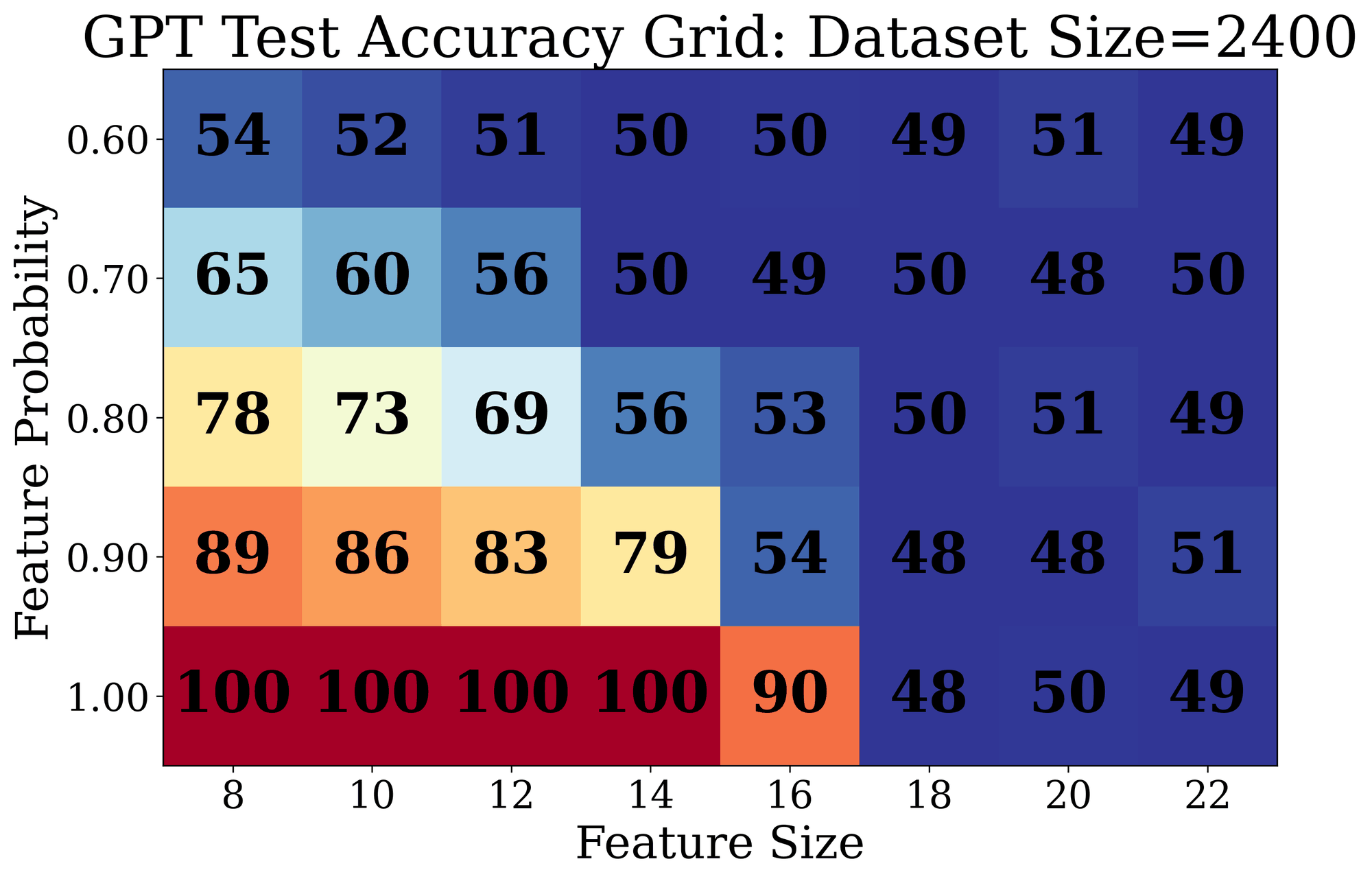}
    \end{subfigure}
    \begin{subfigure}[width=\textwidth]{0.19 \textwidth}
        \includegraphics[width=\textwidth]{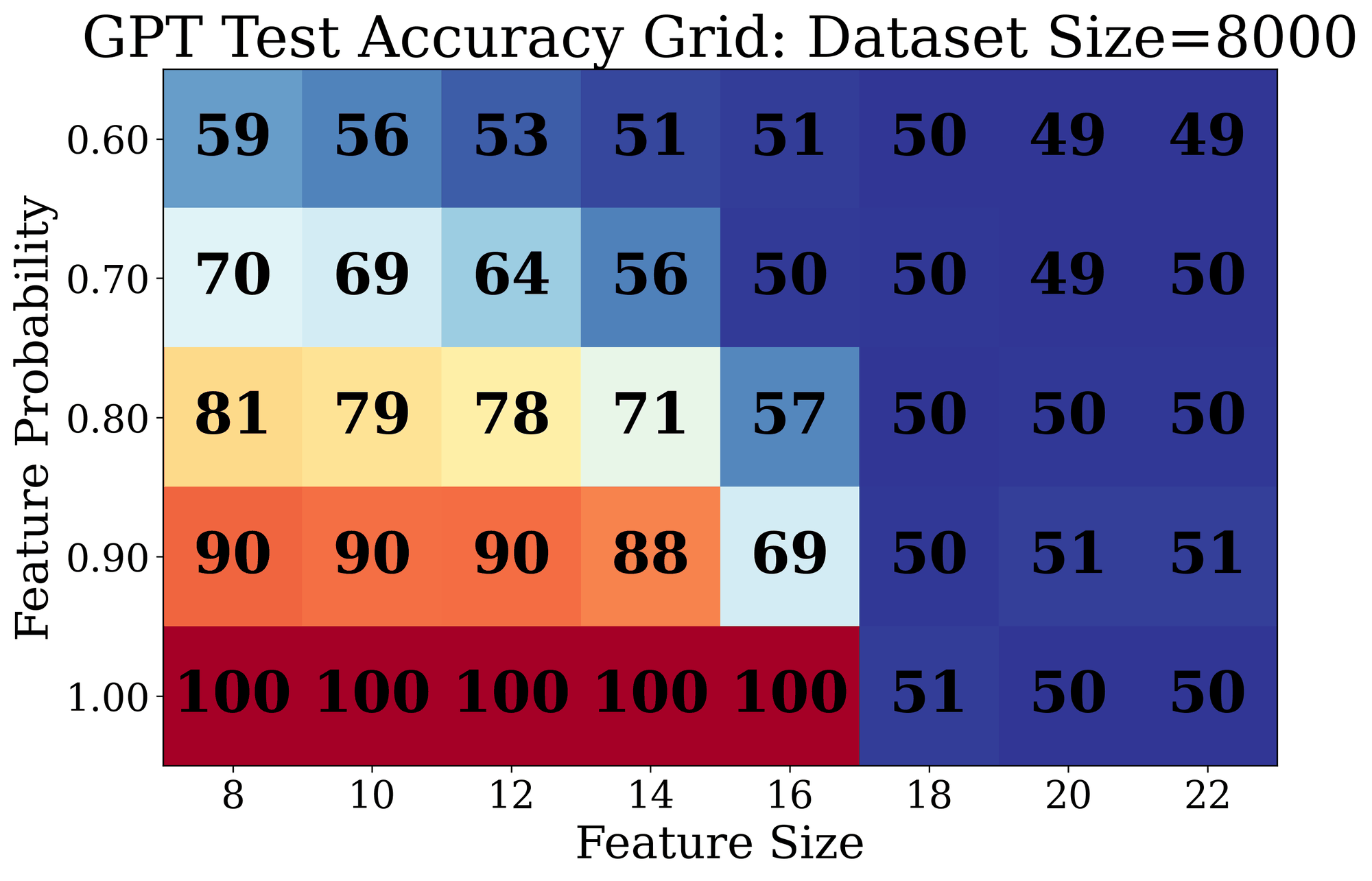}
    \end{subfigure}
    \begin{subfigure}[width=\textwidth]{0.19 \textwidth}
        \includegraphics[width=\textwidth]{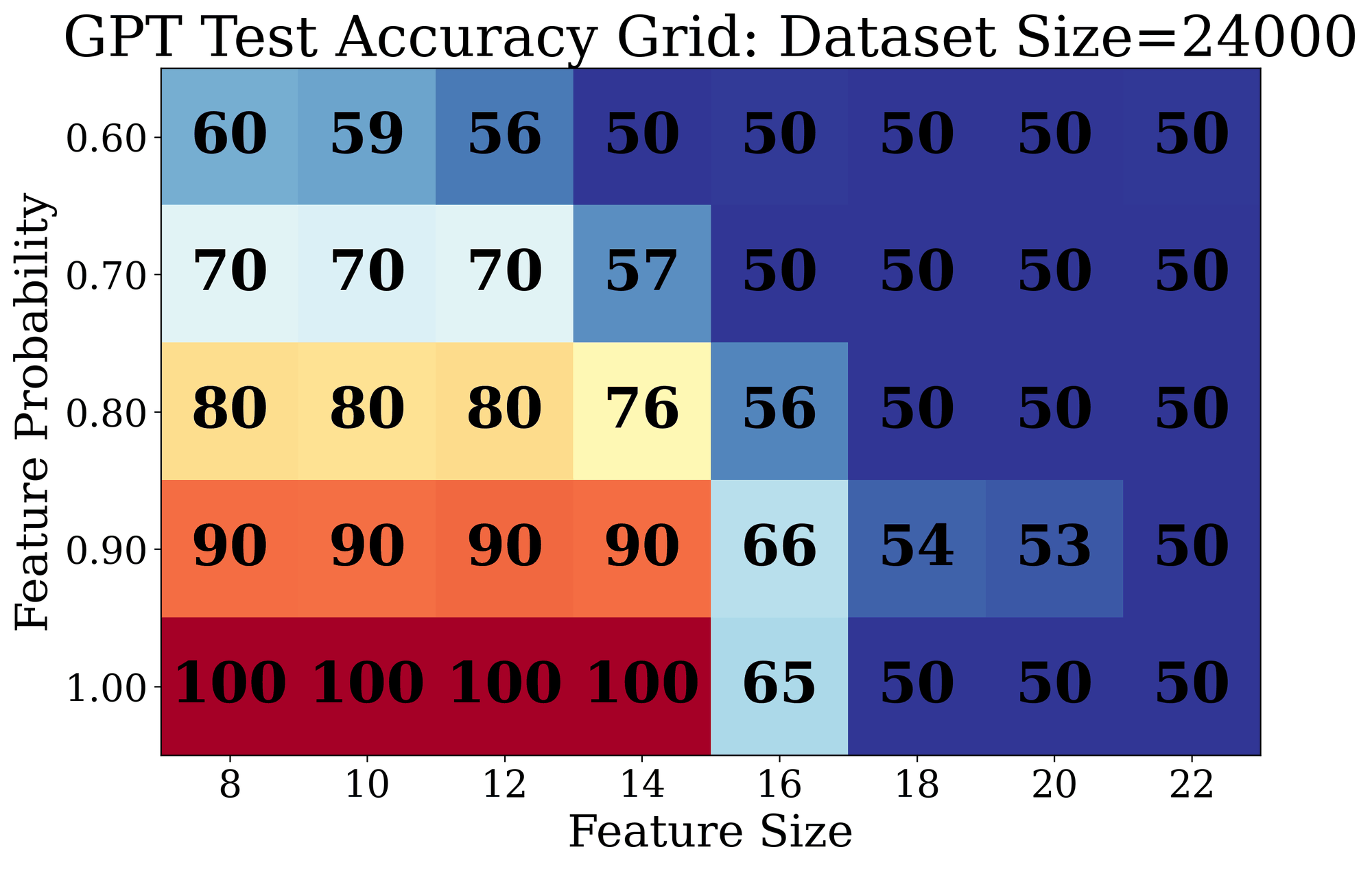}
    \end{subfigure}
    \begin{subfigure}[width=\textwidth]{0.19 \textwidth}
        \includegraphics[width=\textwidth]{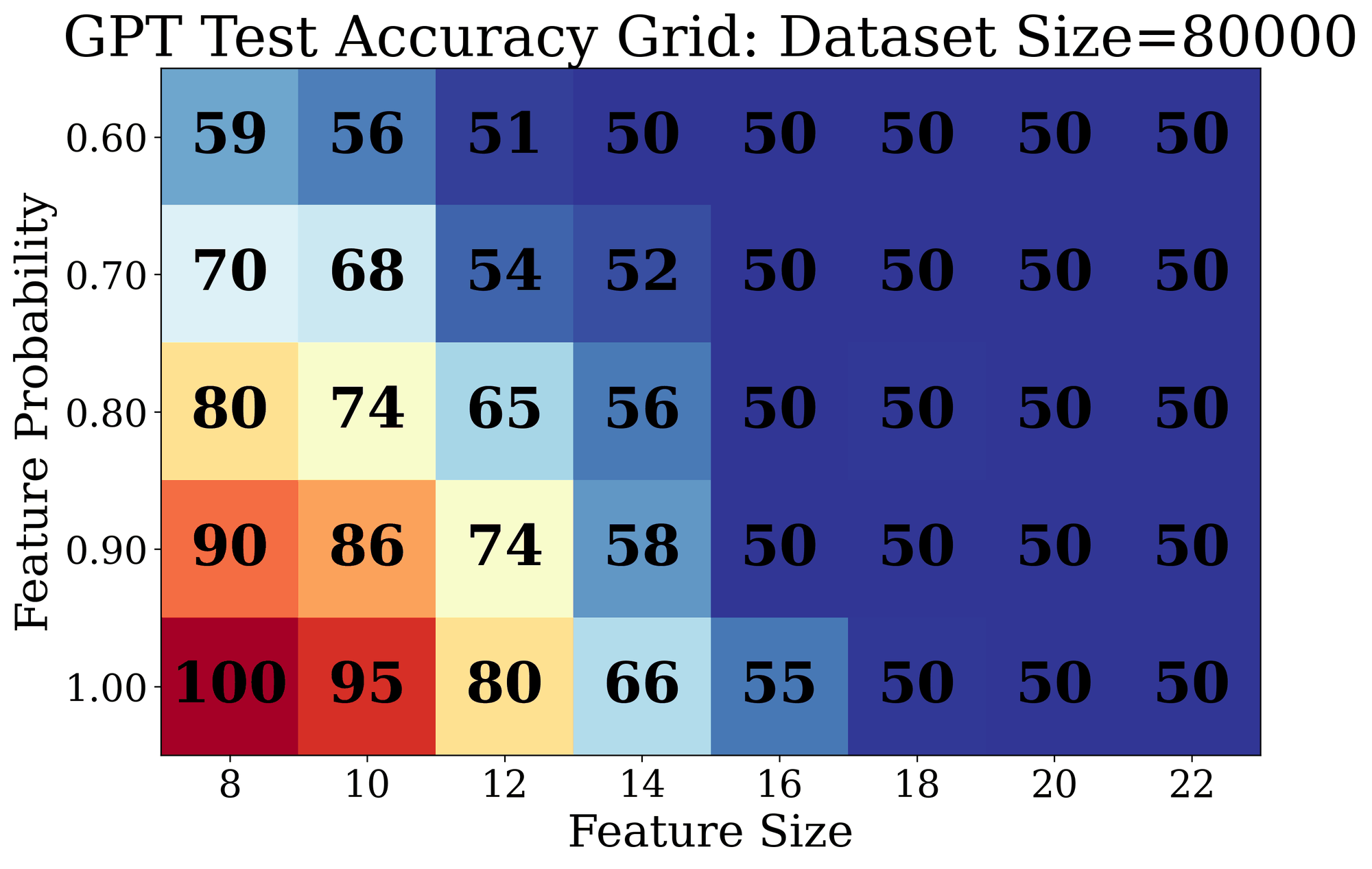}
    \end{subfigure}
    \begin{subfigure}[width=\textwidth]{0.19 \textwidth}
        \includegraphics[width=\textwidth]{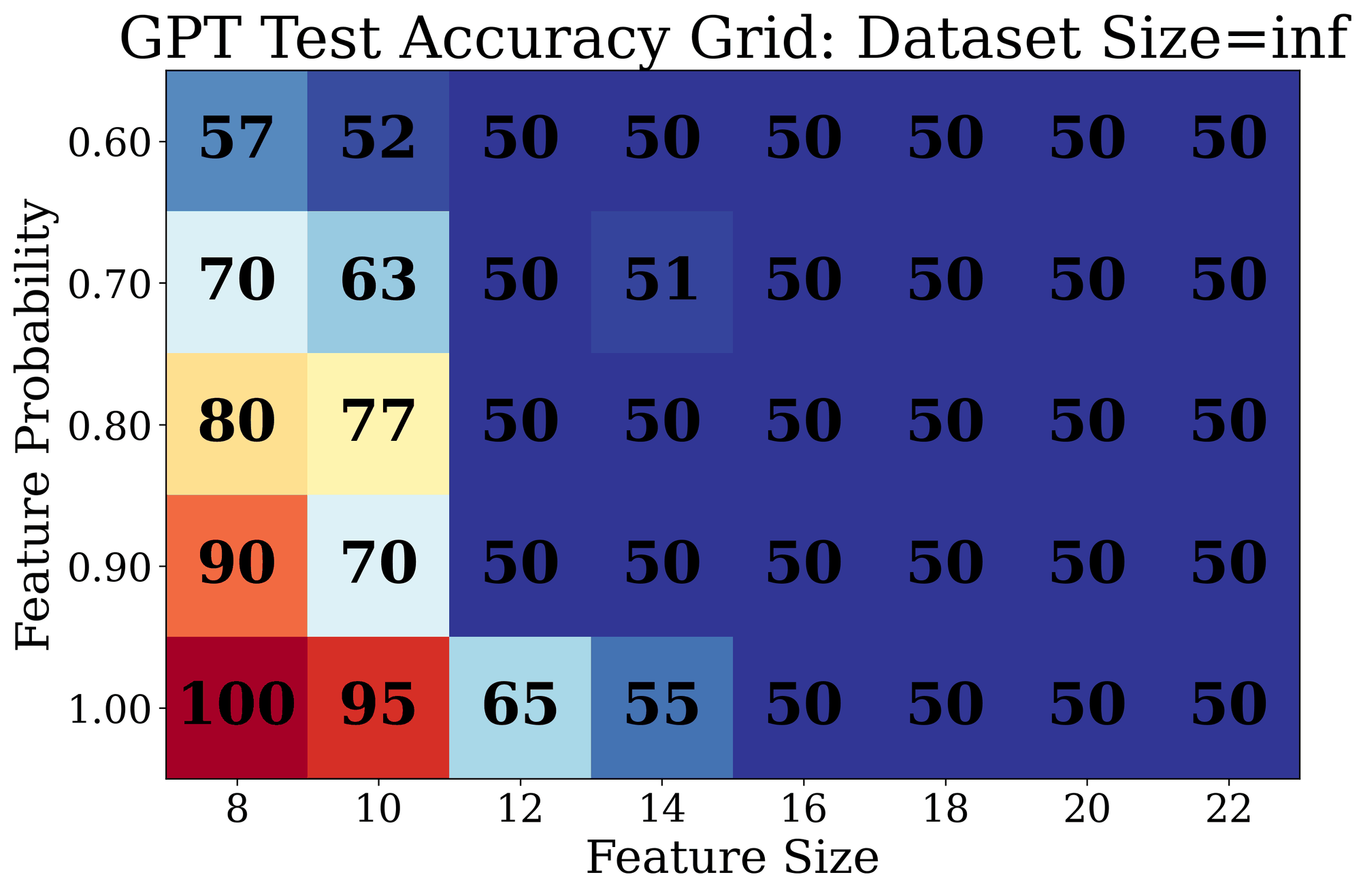}
    \end{subfigure}

    \begin{subfigure}[width=\textwidth]{0.19 \textwidth}
        \includegraphics[width=\textwidth]{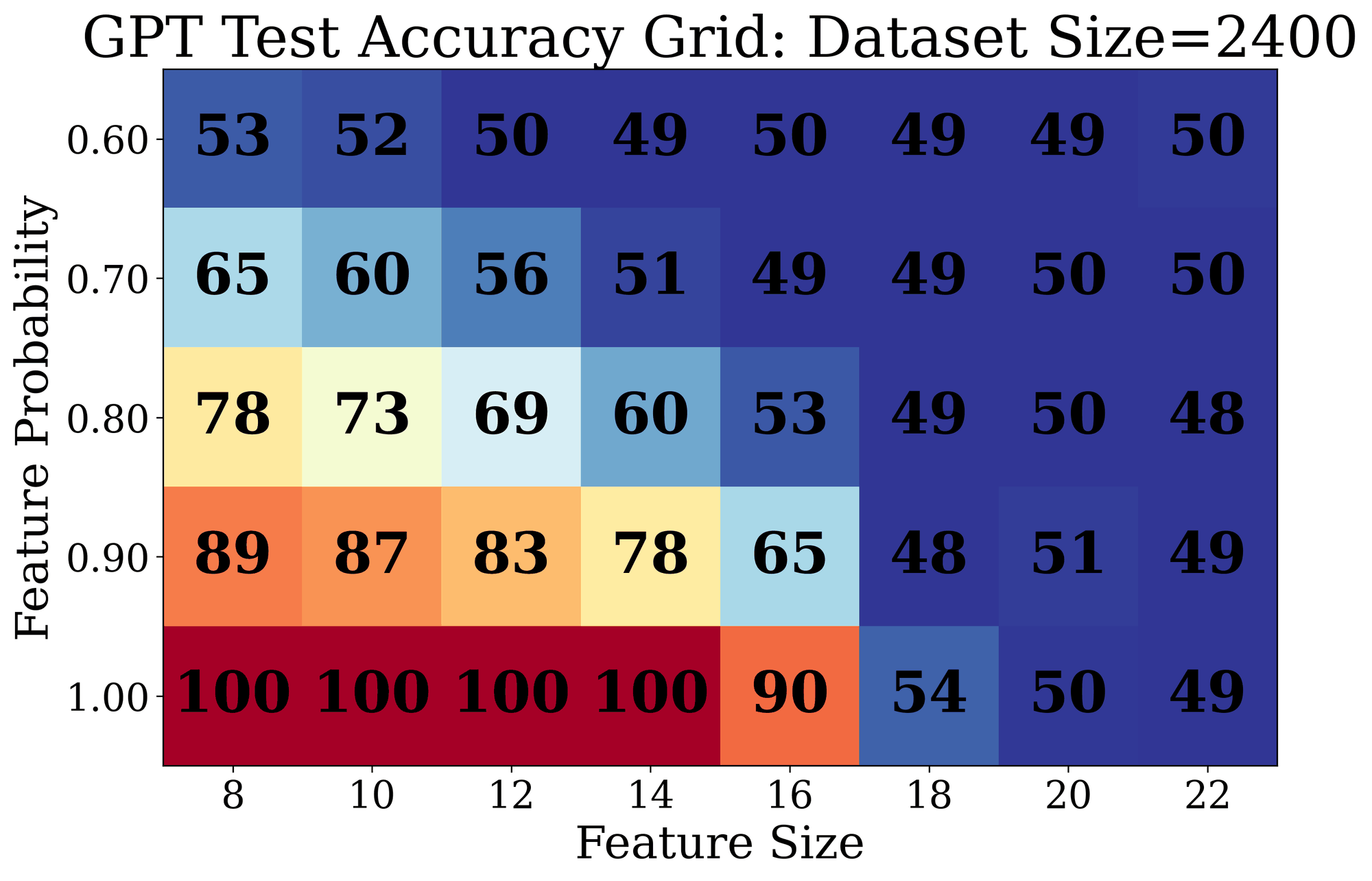}
    \end{subfigure}
    \begin{subfigure}[width=\textwidth]{0.19 \textwidth}
        \includegraphics[width=\textwidth]{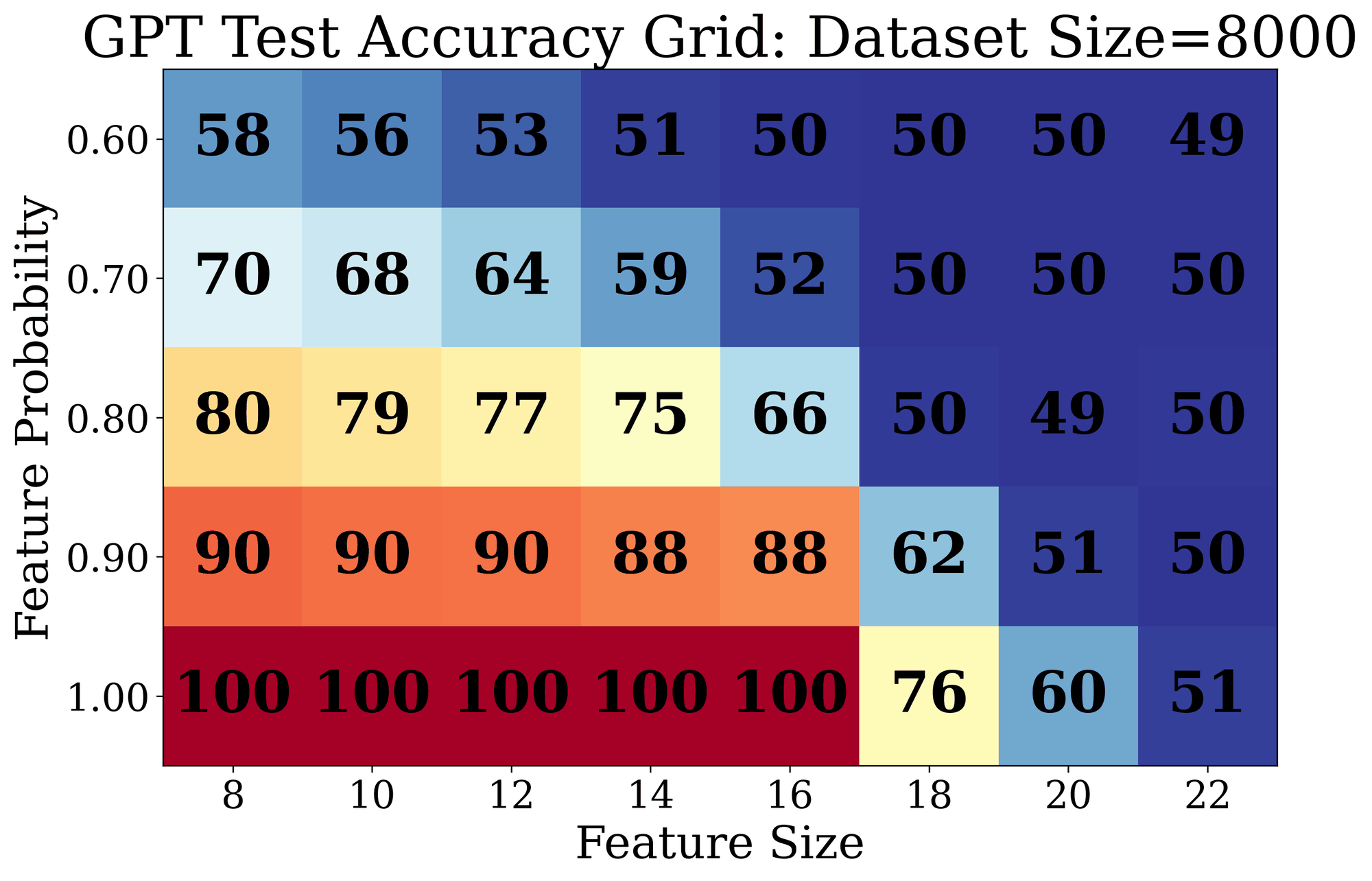}
    \end{subfigure}
    \begin{subfigure}[width=\textwidth]{0.19 \textwidth}
        \includegraphics[width=\textwidth]{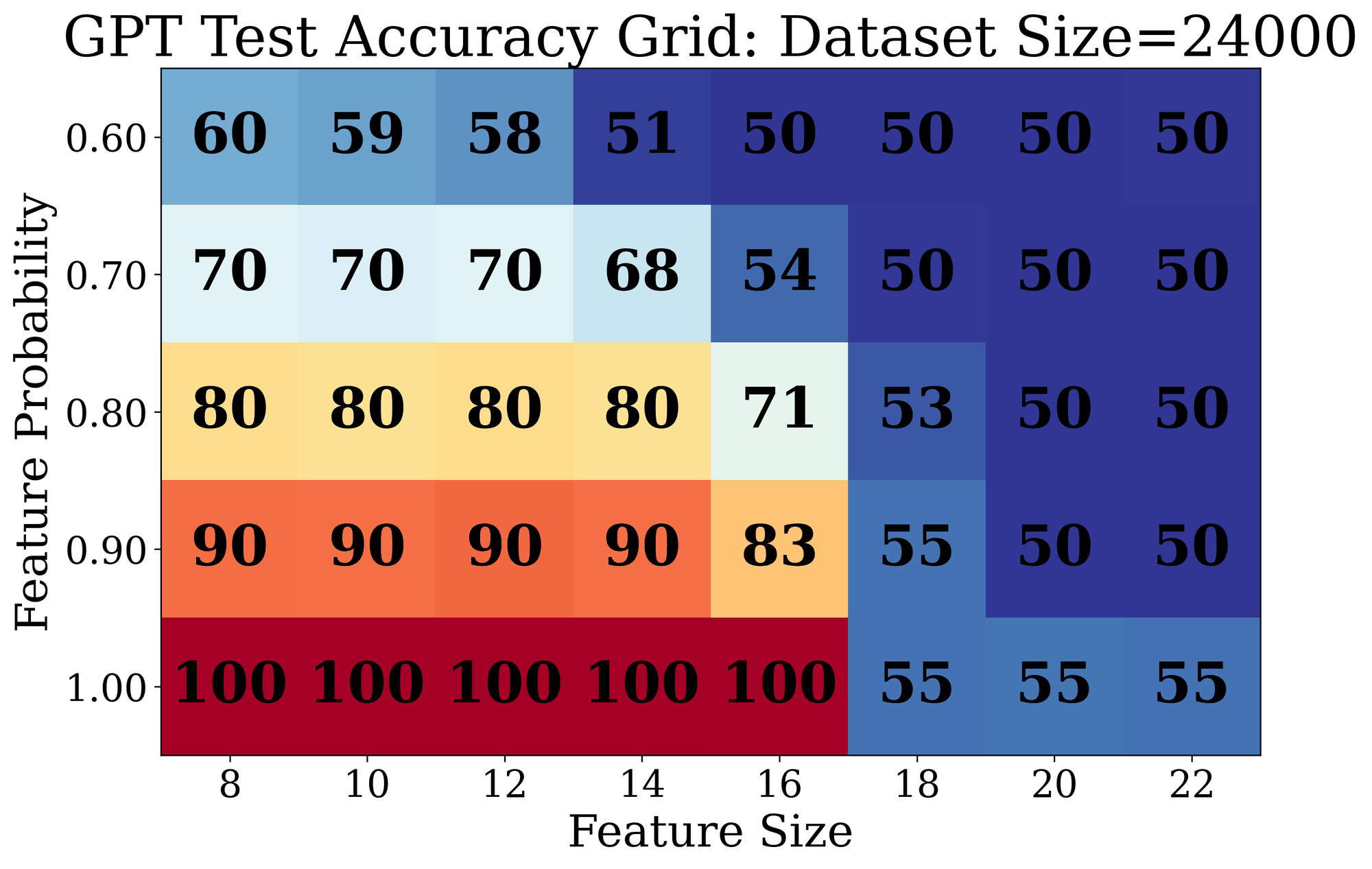}
    \end{subfigure}
    \begin{subfigure}[width=\textwidth]{0.19 \textwidth}
        \includegraphics[width=\textwidth]{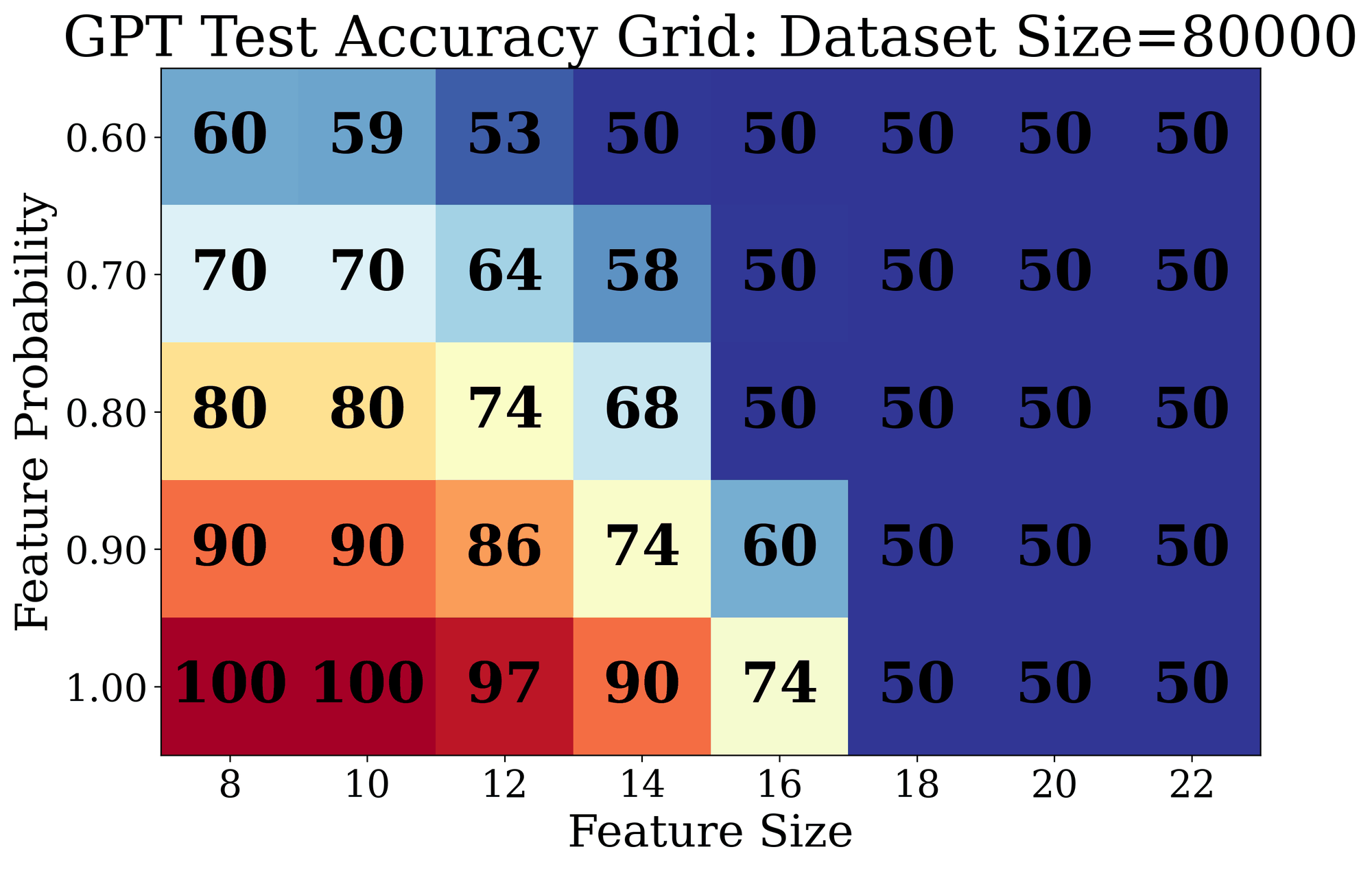}
    \end{subfigure}
    \begin{subfigure}[width=\textwidth]{0.19 \textwidth}
        \includegraphics[width=\textwidth]{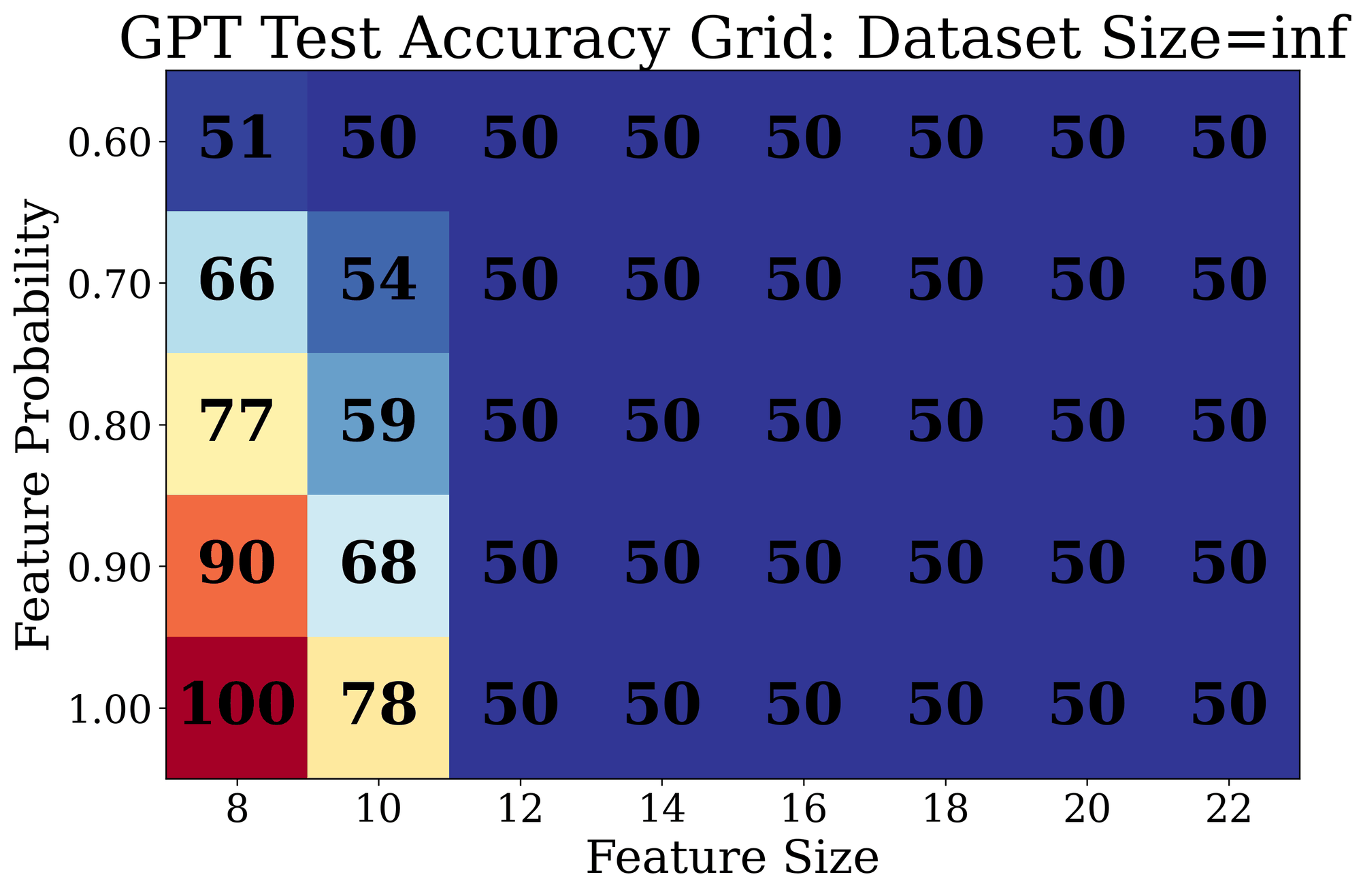}
    \end{subfigure}

    \caption{Test Accuracy on i.i.d test set for models trained on a noisy k-parity feature with Constant Iteration Budget ($T=10000$).
    \textbf{Top:} GPT with a learning rate of $\eta=3\times 10^{-4}$.
    \textbf{Bottom:} GPT with a learning rate of $10^{-4}$.
    While the learning rate has some localised effect it does not effect the trend observed in Section \ref{sec:two_features} with the performance increasing then decreasing when increasing the data set size.}
    \label{fig:lr_noisy}
\end{figure*}

\begin{figure*}[ht]
    \centering
  
    \begin{subfigure}[width=\textwidth]{0.16 \textwidth}
        \includegraphics[width=\textwidth]{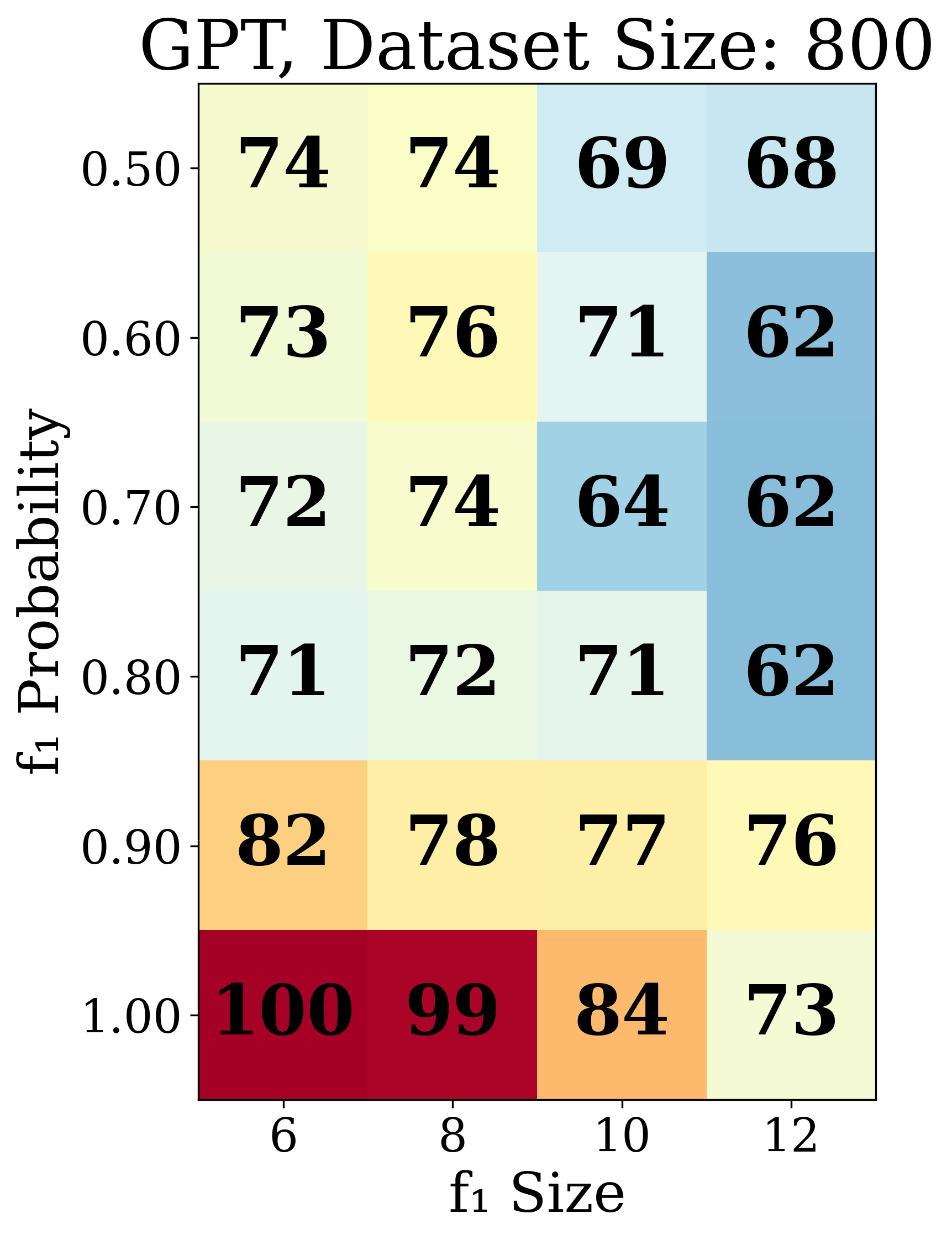}
    \end{subfigure}
    \begin{subfigure}[width=\textwidth]{0.16 \textwidth}
        \includegraphics[width=\textwidth]{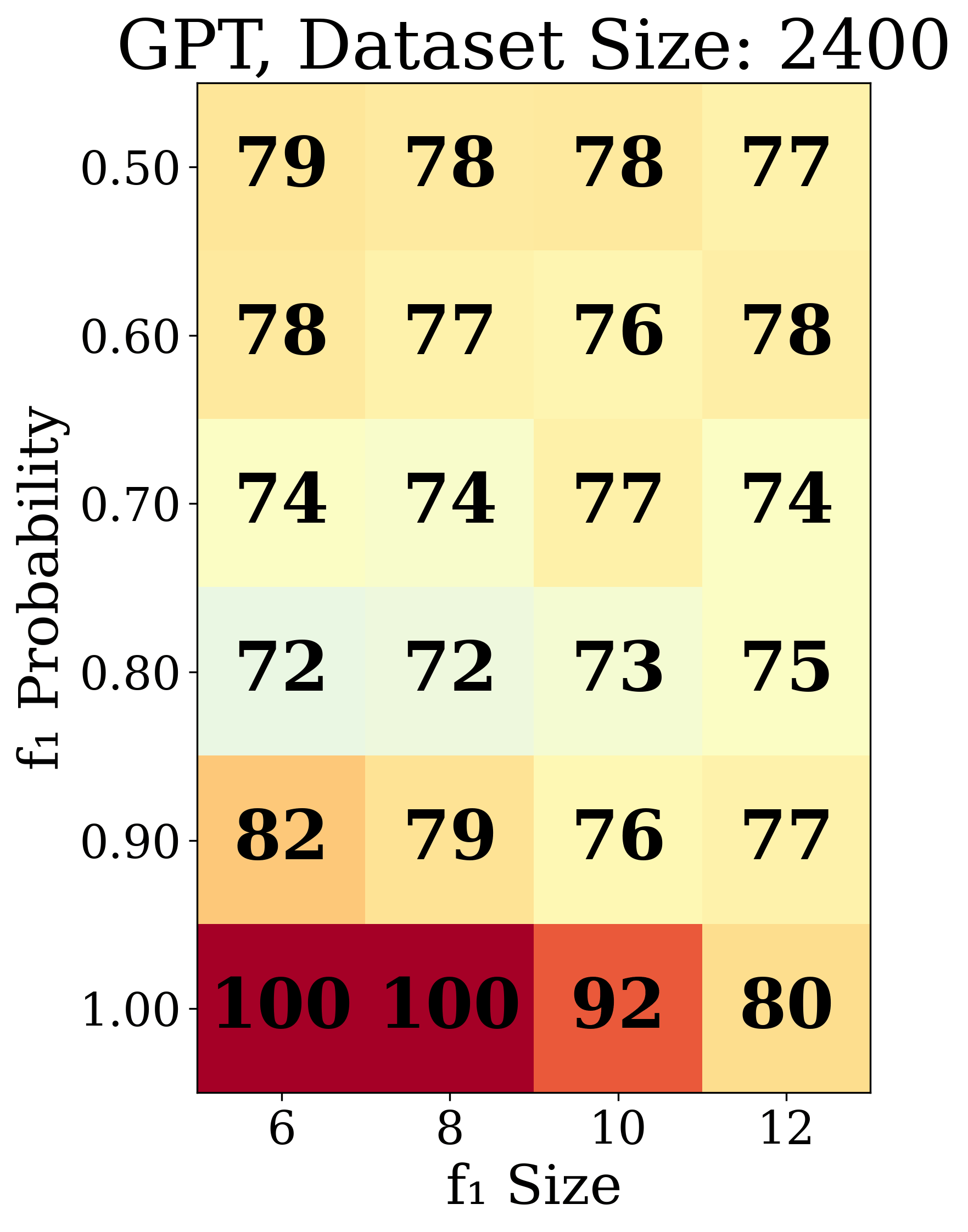}
    \end{subfigure}
    \begin{subfigure}[width=\textwidth]{0.16\textwidth}
        \includegraphics[width=\textwidth]{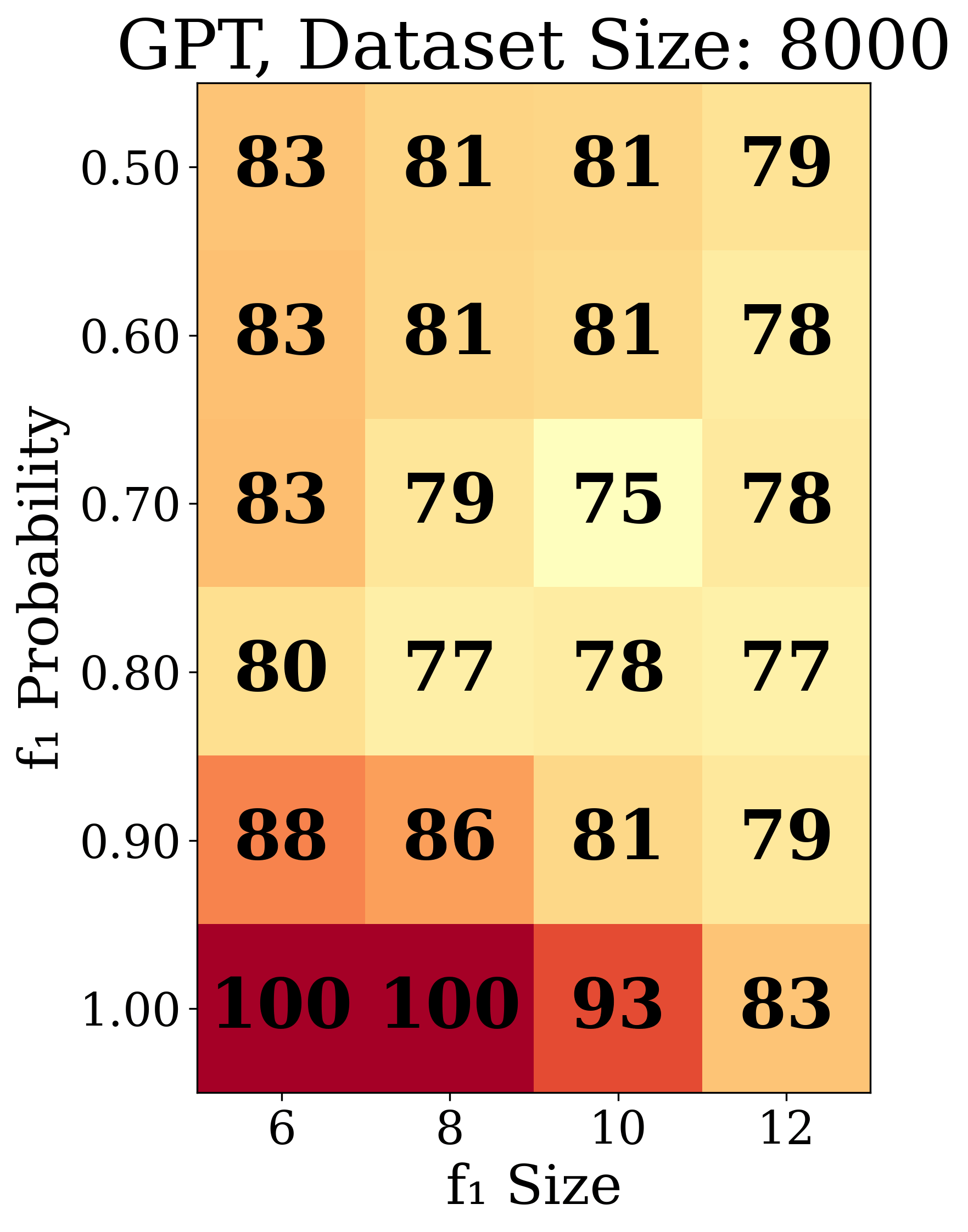}
    \end{subfigure}
    \begin{subfigure}[width=\textwidth]{0.16 \textwidth}
        \includegraphics[width=\textwidth]{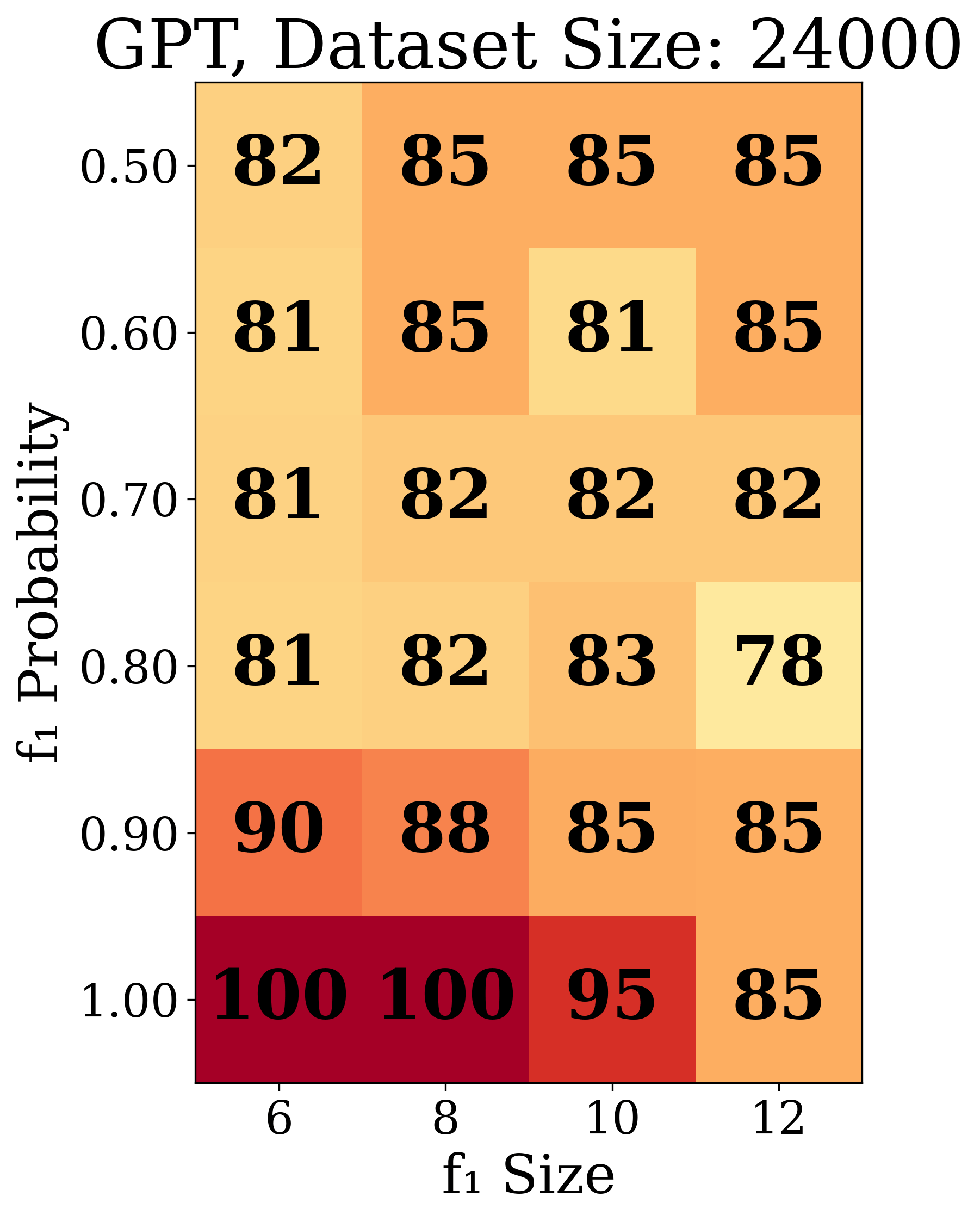}
    \end{subfigure}
    \begin{subfigure}[width=\textwidth]{0.16 \textwidth}
        \includegraphics[width=\textwidth]{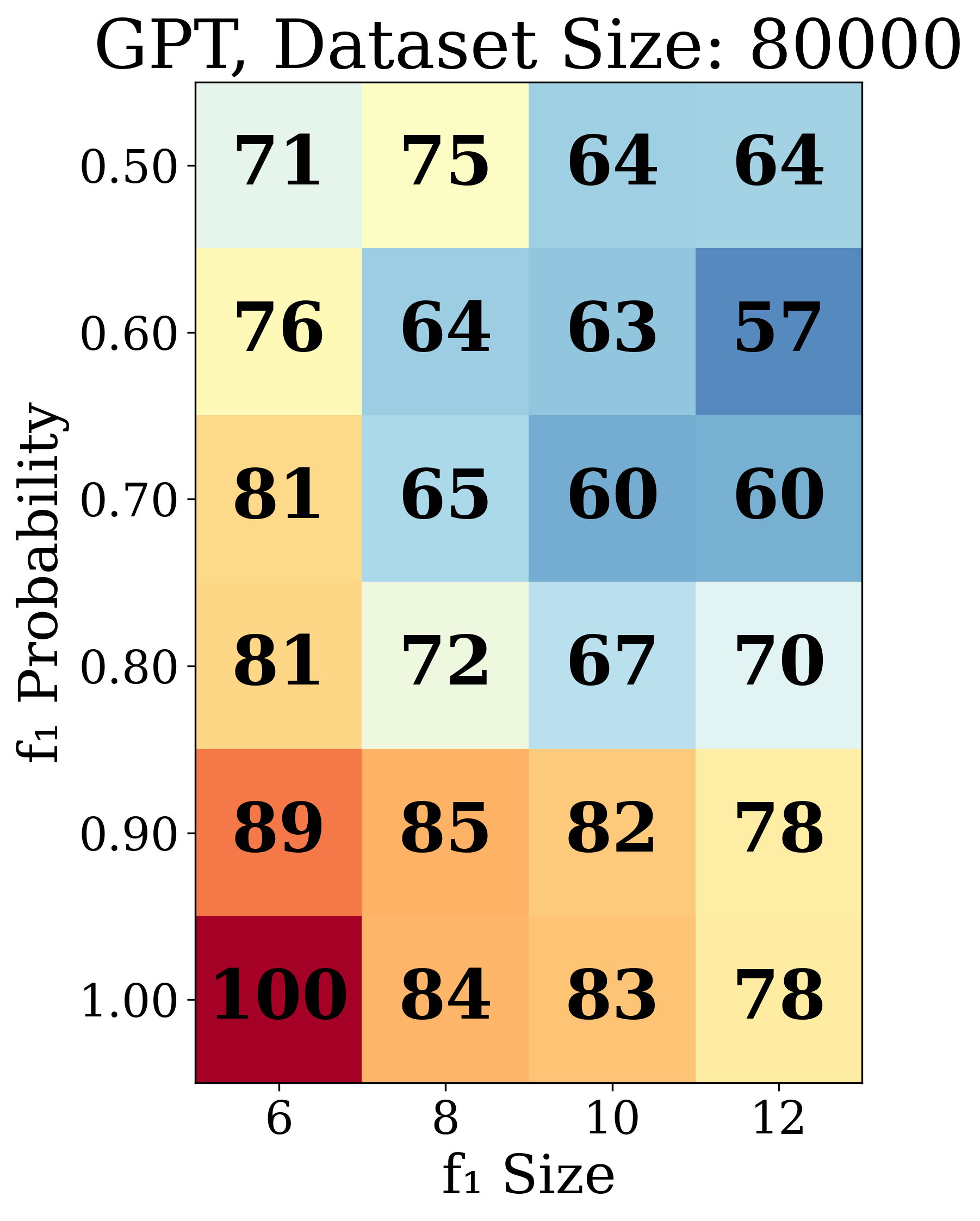}
    \end{subfigure}
    \begin{subfigure}[width=\textwidth]{0.16 \textwidth}
        \includegraphics[width=\textwidth]{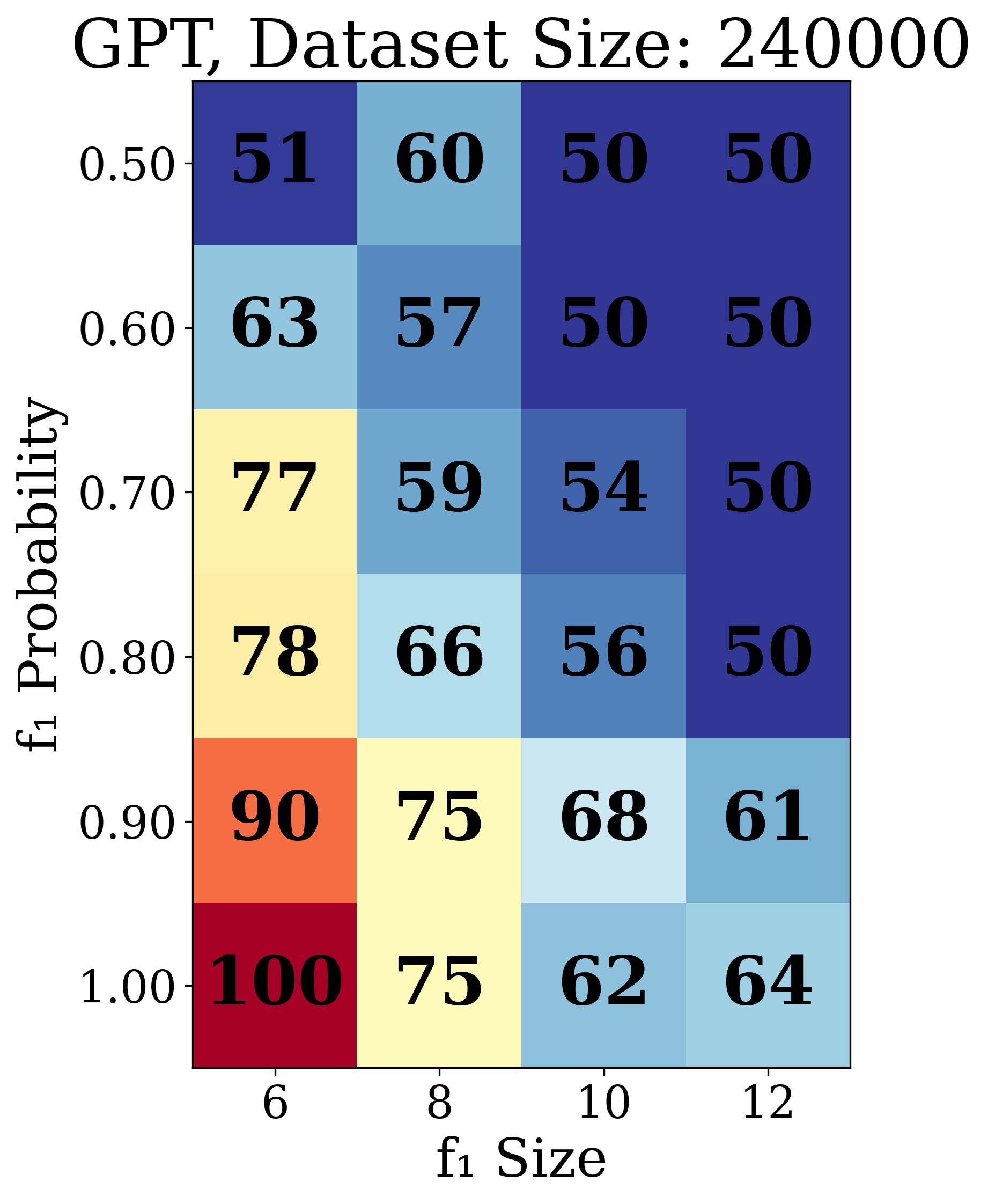}
    \end{subfigure}

    \begin{subfigure}[width=\textwidth]{0.16 \textwidth}
        \includegraphics[width=\textwidth]{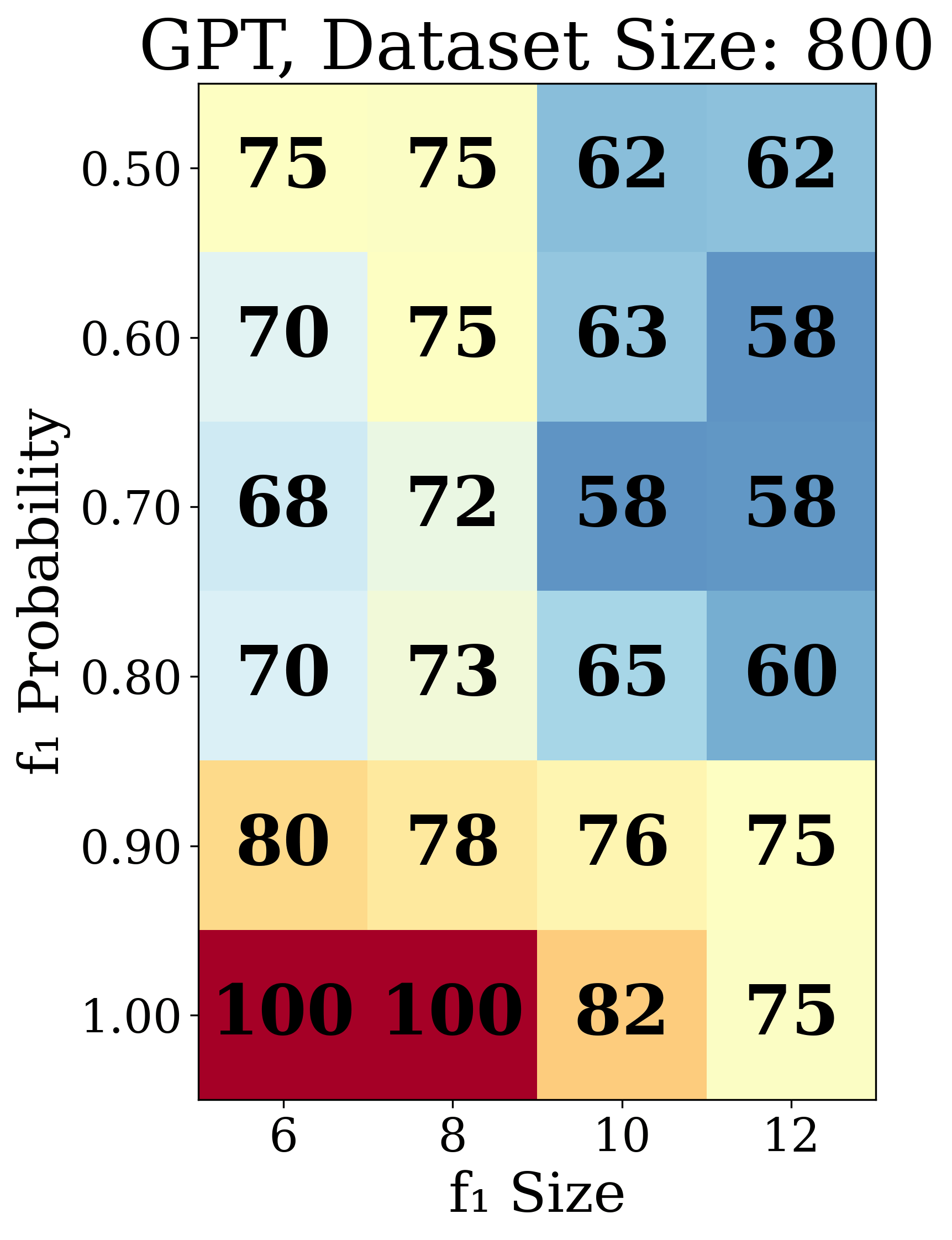}
    \end{subfigure}
    \begin{subfigure}[width=\textwidth]{0.16 \textwidth}
        \includegraphics[width=\textwidth]{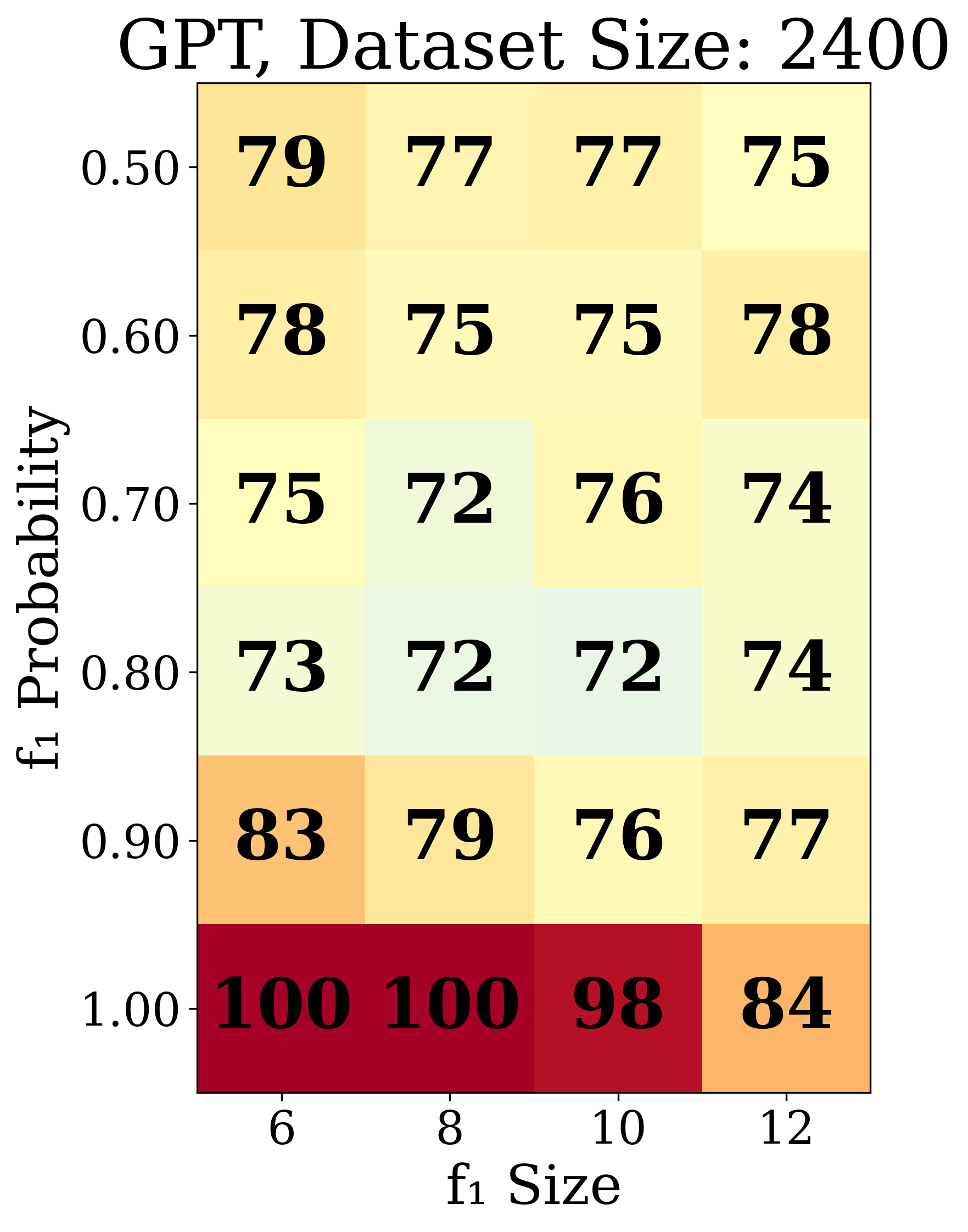}
    \end{subfigure}
    \begin{subfigure}[width=\textwidth]{0.16 \textwidth}
        \includegraphics[width=\textwidth]{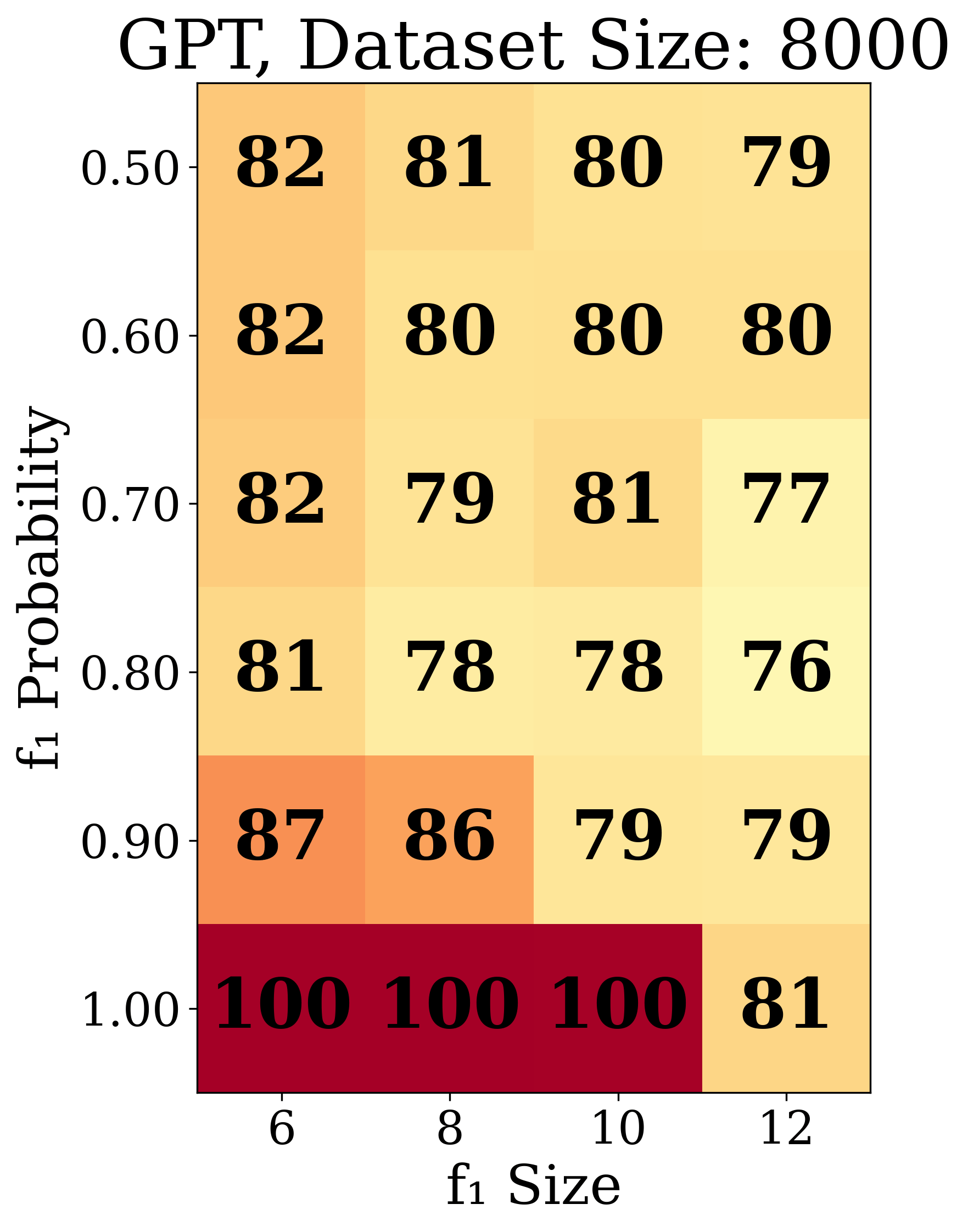}
    \end{subfigure}
    \begin{subfigure}[width=\textwidth]{0.16 \textwidth}
        \includegraphics[width=\textwidth]{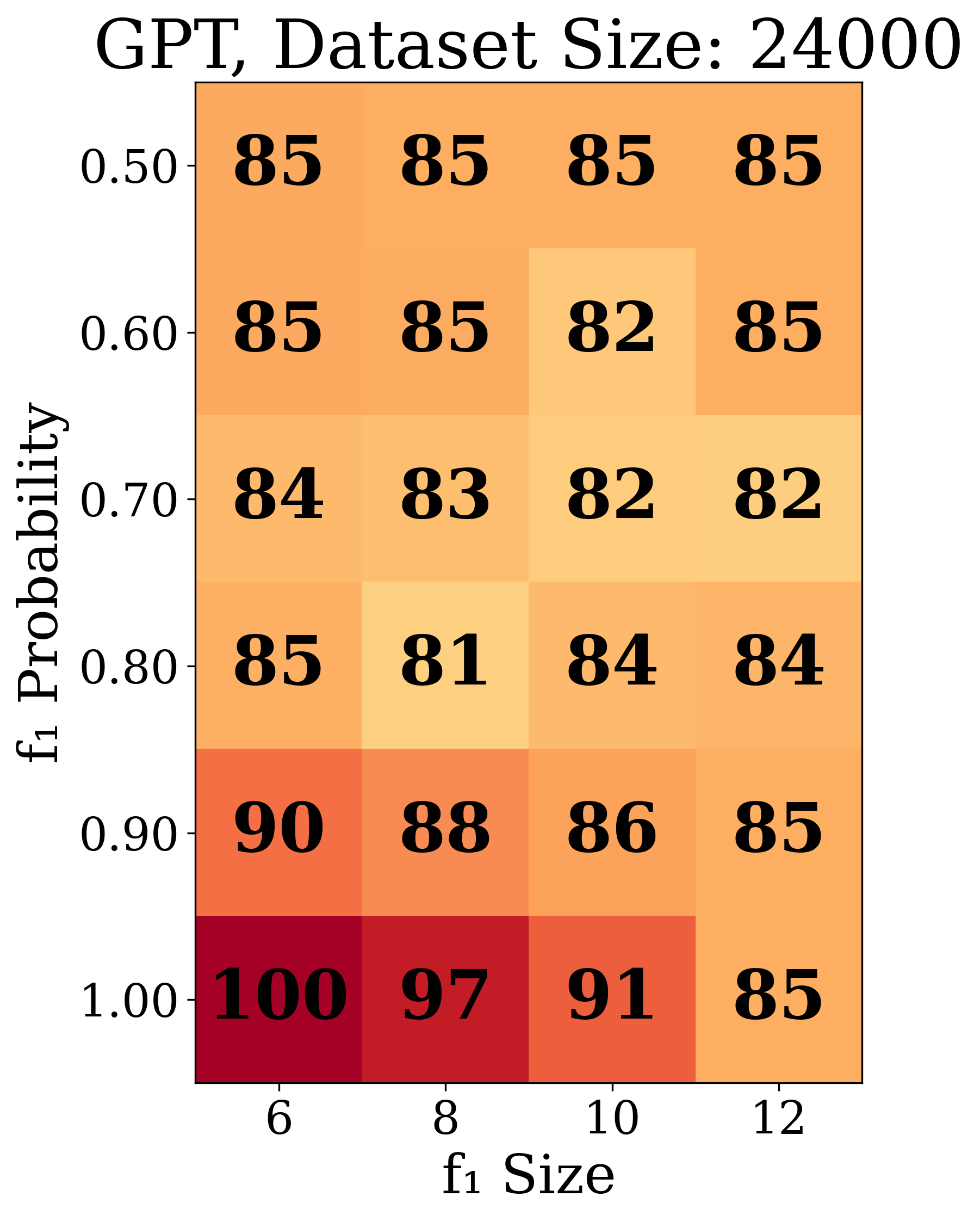}
    \end{subfigure}
    \begin{subfigure}[width=\textwidth]{0.16 \textwidth}
        \includegraphics[width=\textwidth]{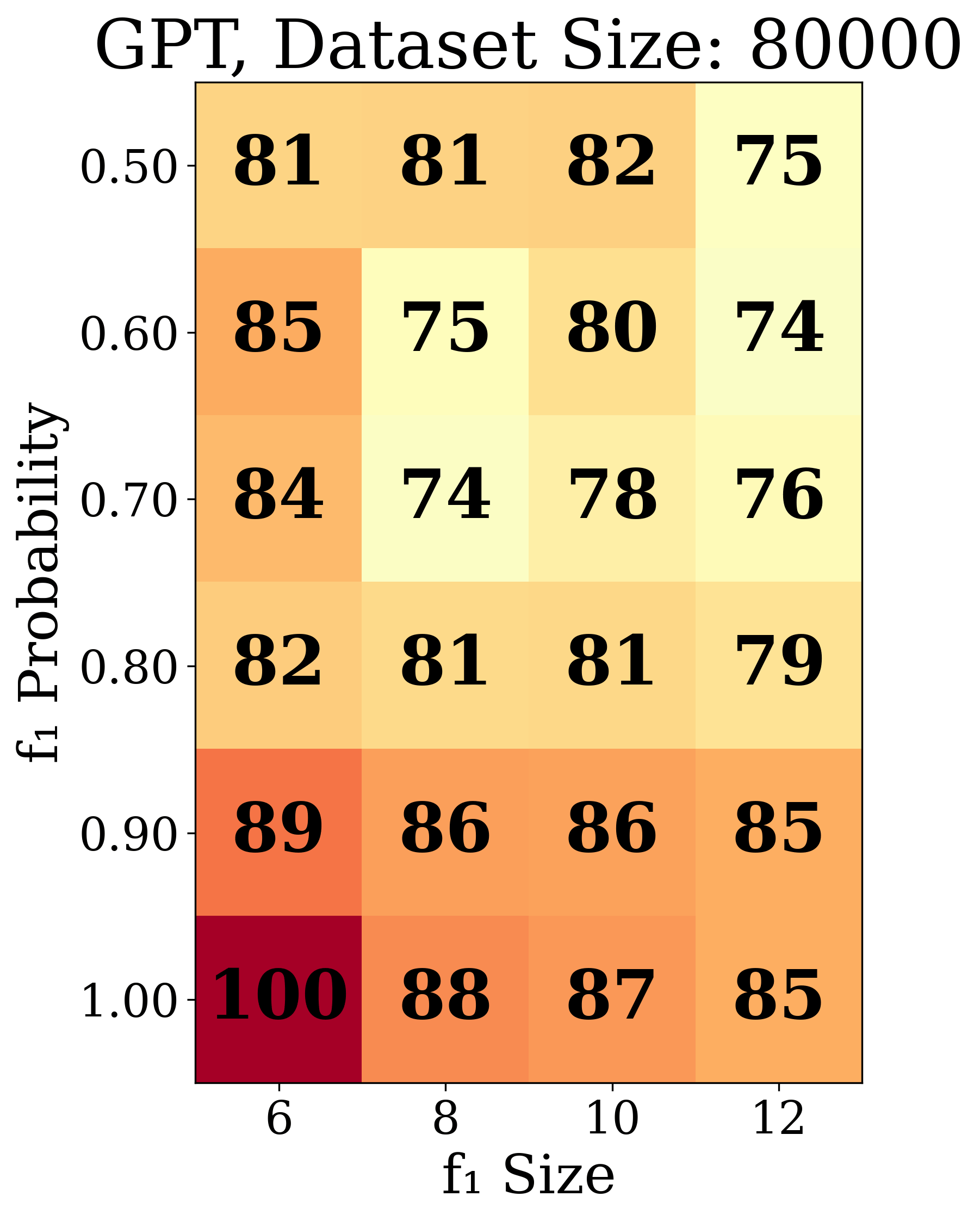}
    \end{subfigure}
    \begin{subfigure}[width=\textwidth]{0.16 \textwidth}
        \includegraphics[width=\textwidth]{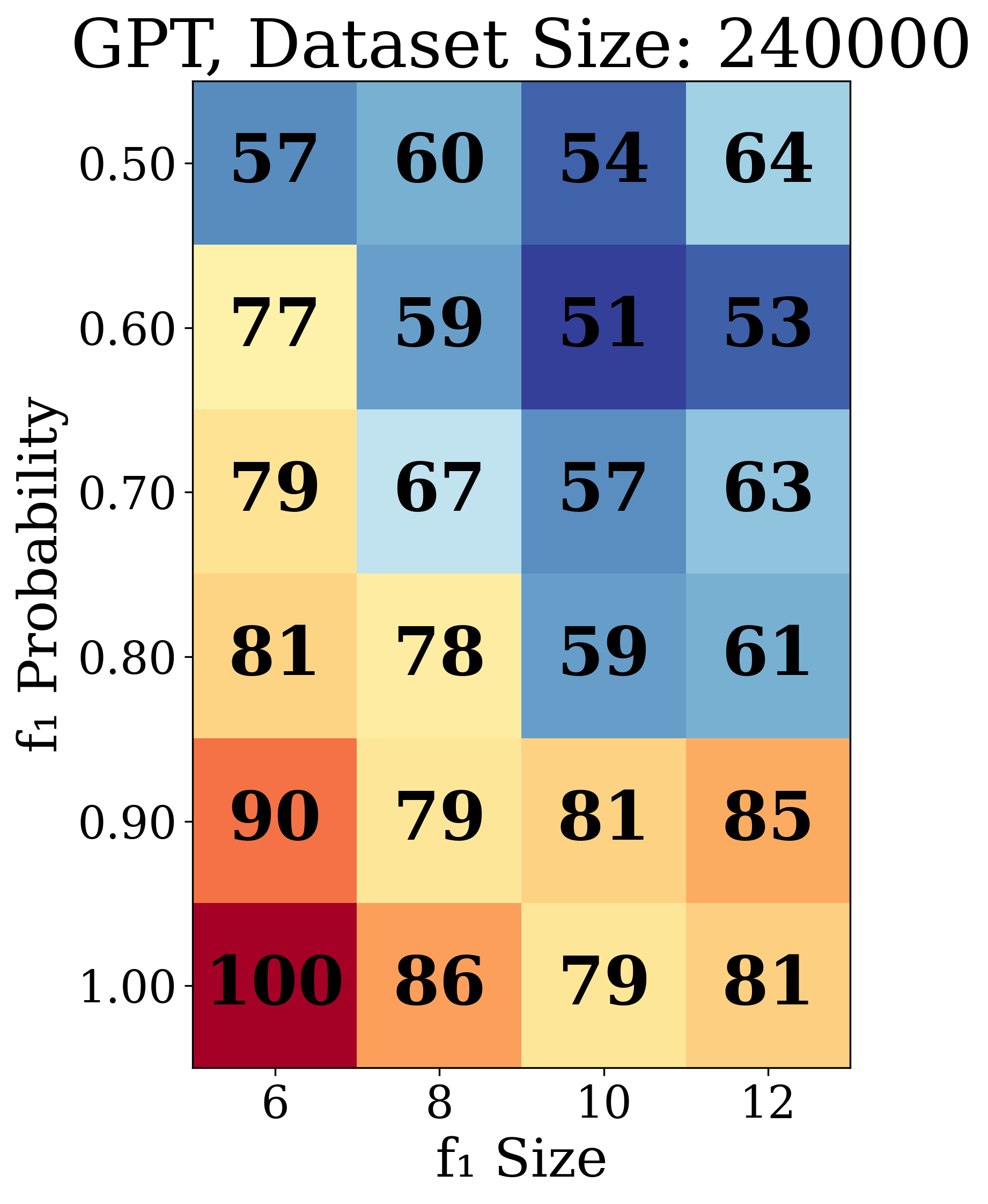}
    \end{subfigure}

    \caption{Two Feature Experiments with constant compute budget $(T=10000)$.
    \textbf{Top:} GPT with a learning rate of $\eta=3\times 10^{-4}$.
    \textbf{Bottom:} GPT with a learning rate of $10^{-4}$.
    While the learning rate has some localised effect it does not effect the trend observed in Section \ref{sec:two_features} where the performance increasing then decreasing when increasing the data set size.}
    \label{fig:lr_two_features}
\end{figure*}

In this section we show that the trends observed for learning k-parities with transformers hold when a smaller learning rate of $3\times10^{-4}$ is used.
In Figures \ref{fig:lr_noisy} and \ref{fig:lr_two_features} we show the difference in performance when altering the learning rate of the GPT model at learning noisy k-parity as described in Section $\ref{sec:noisy_k_parity_experiments}$ and the two features learning experiments shown in Section \ref{sec:two_features}.

\section{Hierarchical Data Sets}\label{sec:strict_hierarchical}

For data sets with multiple independent weakly predictive features a bagging approach \citep{breiman1996bagging} has proven to be an effective means of prediction where ``a vote'' from multiple features is used. 
When constructing data sets with a large number of features it may be desirable to prevent models using bagging and force the model to learn the most predictive features to get best results. 
To action this we create strictly hierarchical data sets, where the features are not sampled independently.
Instead the features are sampled to ensure that if  $f_i$ has higher predictivity than $f_j$, ($1-\alpha_k > 1-\alpha_j$) the feature $f_j(\vz_{j,n})\neq y_n$ if any $f_i(\vz_{i,n})\neq y_n$ for all inputs $\vx$. 
In words for the data sets shown \ref{tab:data_lin} we ensure $10\%$ of the data can only be correctly classified using $f_5$, there is another non-overlapping $10\%$ that can only be correctly predicted using $f_4$ or $f_5$ and so on. 
Thus the features form a strict hierarchy, and this allows us to have a data set with many features, but ensure the only way to achieve high performance on all samples is to learn the feature with the highest predictivity, or conversely learn a less predictive feature and memorise the examples where where it is not predictive. 

Data sets containing a hierarchy of features are worthy of particular study because this property is representative of vast majority of machine learning data sets.
Indeed, the activations of a neural network trained on a data set form a hierarchy of features where the later layers represent more complex and predictive features of the input than the early layers \citep{tiwari2023sifer}.
While in practice these complex feature are typically a composition of weaker features, in this work for ease of analysis we instead make them separate, uncorrelated features, following a similar approach to that of the collages data set introduced in \citet{teney2022predicting}.

\section{Model Scaling}
In this section, we report the results of repeating the experiment detailed in section \ref{sec:k_parity_experiments} for models with width factors $m \in \{1,10,20\}$.
As one might expect increasing the width of the model increases, its capacity and hence the size of the k-parity that the model is capable of learning in $10^4$ iterations. 
When trained directly on the DGP the MLP model with a a width factor $m=1$ has 2k parameters, but still manages to learn k-parity reliably up to 18 bits.
Conversely the GPT model with $m=1$ has roughly 40k parameters and still struggles to learn an k-parity of even 8 bits.
We chose not to run larger models due to the compute costs required in running repeats across all the settings considered.
We also note both models seem to find learning k-parity easier for even $d$ which motivated our choice to primarily use even values of $d$ in the other experiments. 

\begin{figure*}[h]
    \centering
    \begin{subfigure}[width=\textwidth]{0.48 \textwidth}
        \includegraphics[width=\textwidth]{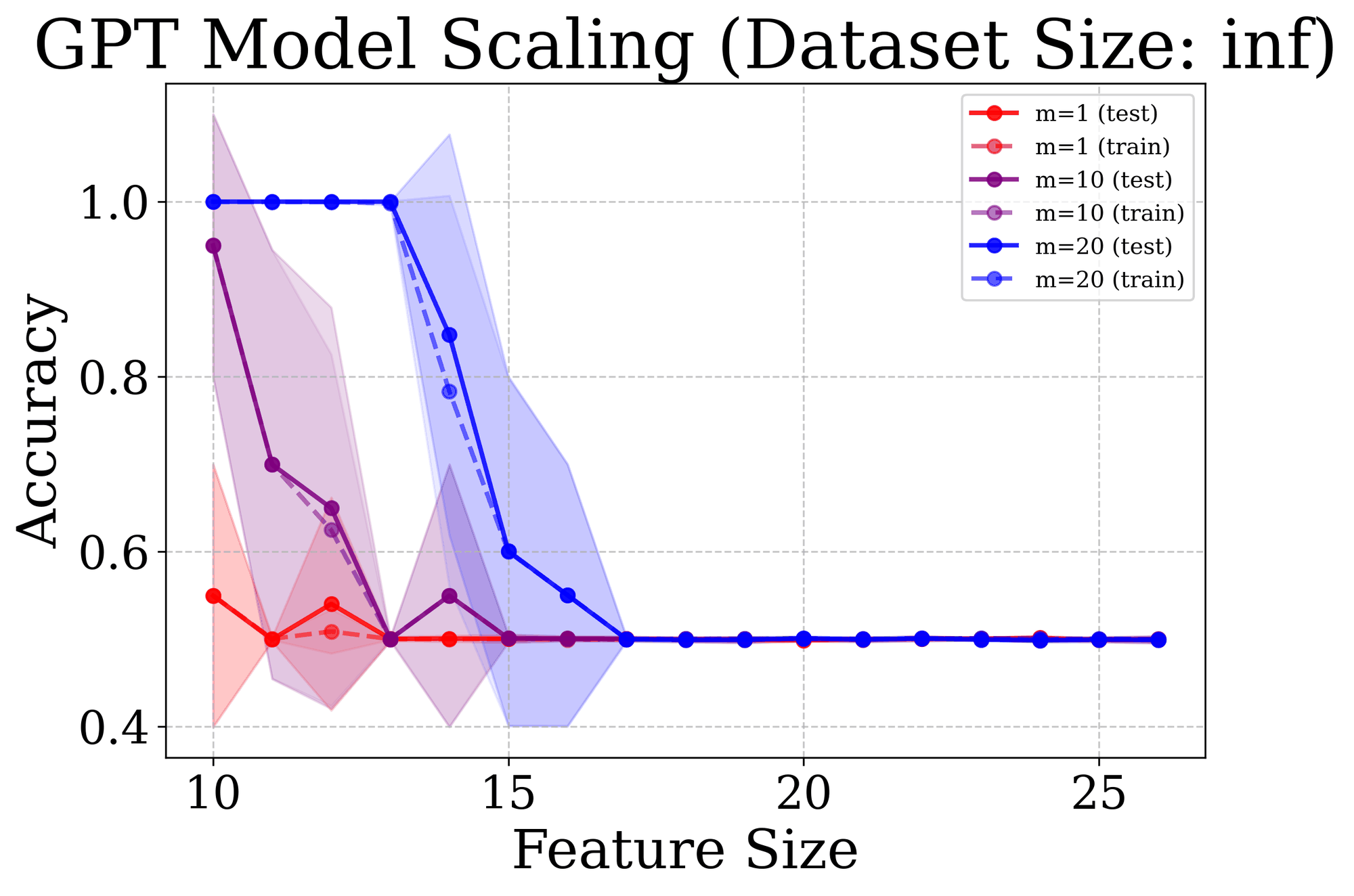}
    \end{subfigure}
    \begin{subfigure}[width=\textwidth]{0.48 \textwidth}
        \includegraphics[width=\textwidth]{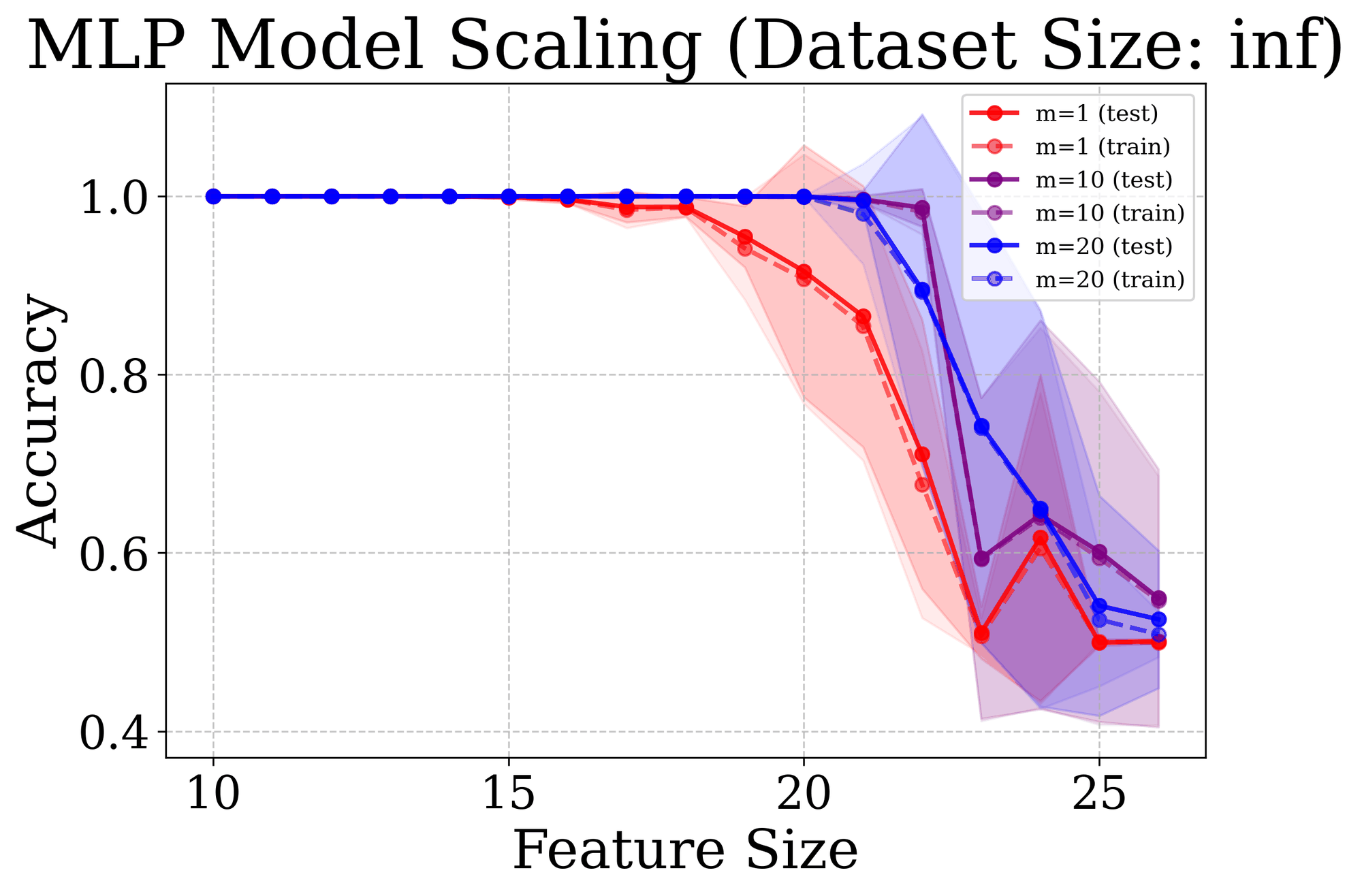}
    \end{subfigure}
    \caption{Model Scaling for models trained directly on the DGP for $T=10^4$ include odd $d$}
    \label{fig:model_scaling_all}
\end{figure*}

\begin{figure*}[h]
    \centering
    \begin{subfigure}[width=\textwidth]{0.48 \textwidth}
        \includegraphics[width=\textwidth]{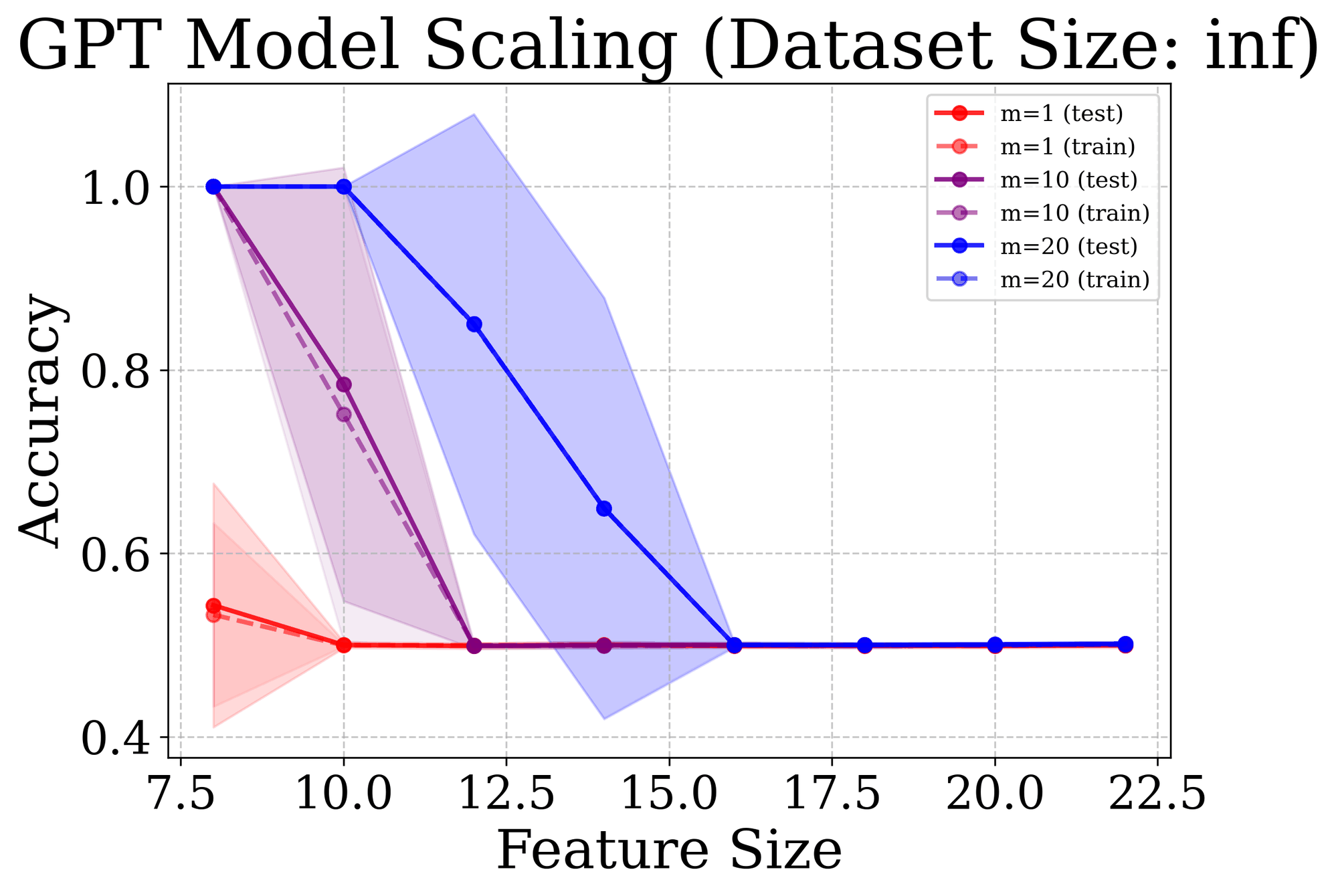}
    \end{subfigure}
    \begin{subfigure}[width=\textwidth]{0.48 \textwidth}
        \includegraphics[width=\textwidth]{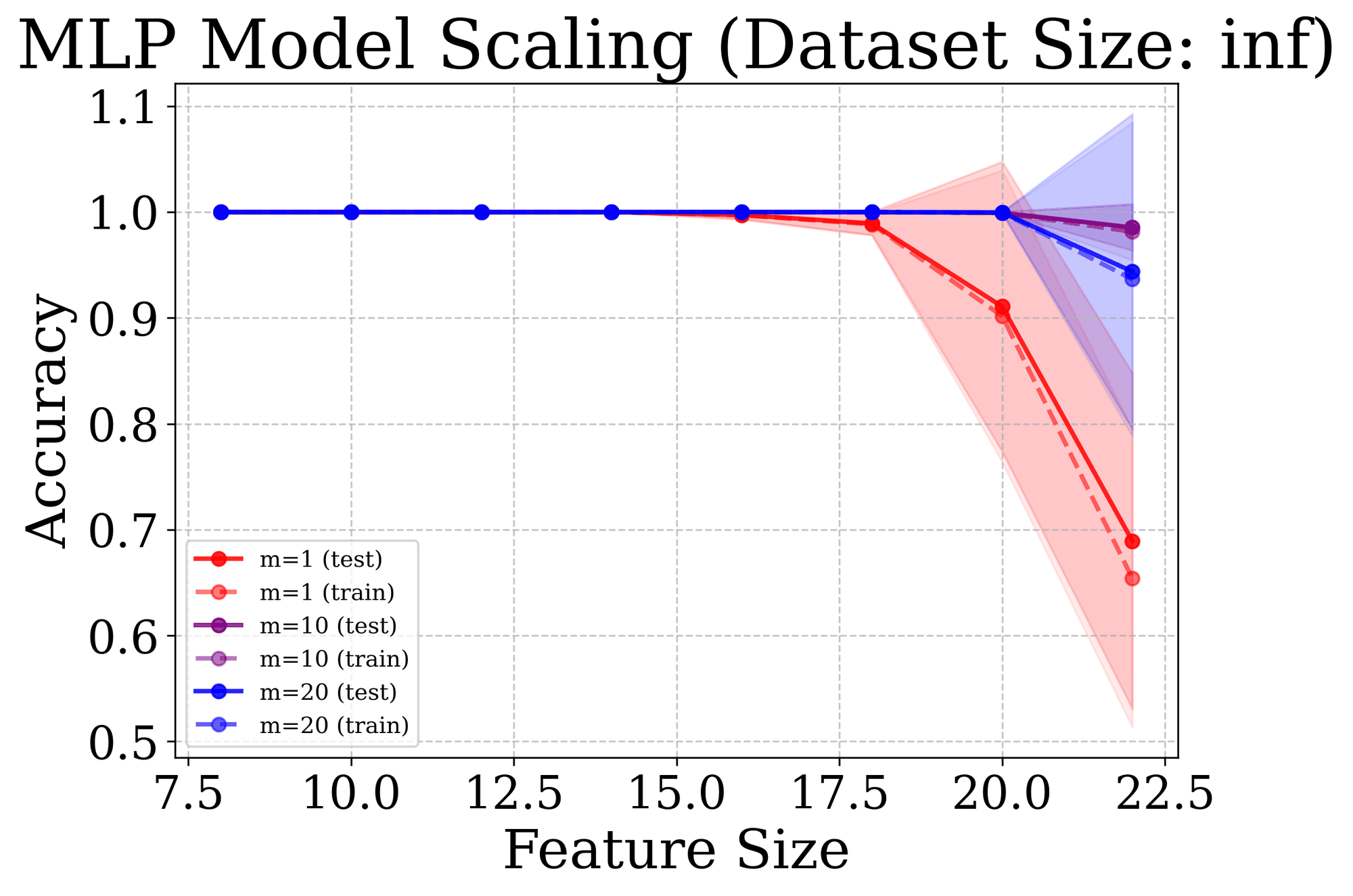}
    \end{subfigure}
    \caption{Model Scaling for models trained directly on the DGP for $T=10^4$}
    \label{fig:model_scaling}
\end{figure*}

\begin{figure*}[h]
    \centering
    \begin{subfigure}[width=\textwidth]{0.48 \textwidth}
        \includegraphics[width=\textwidth]{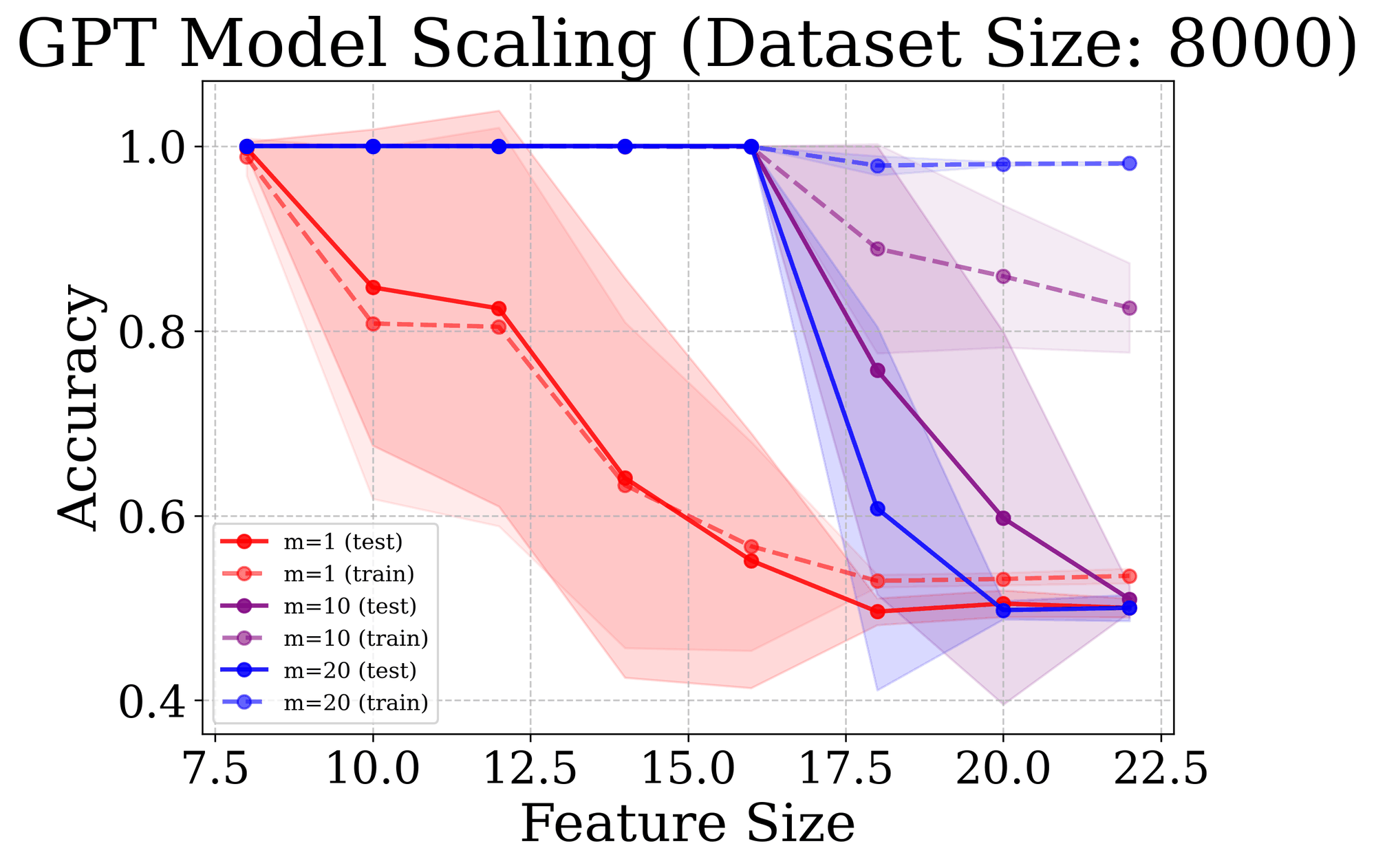}
    \end{subfigure}
    \begin{subfigure}[width=\textwidth]{0.48 \textwidth}
        \includegraphics[width=\textwidth]{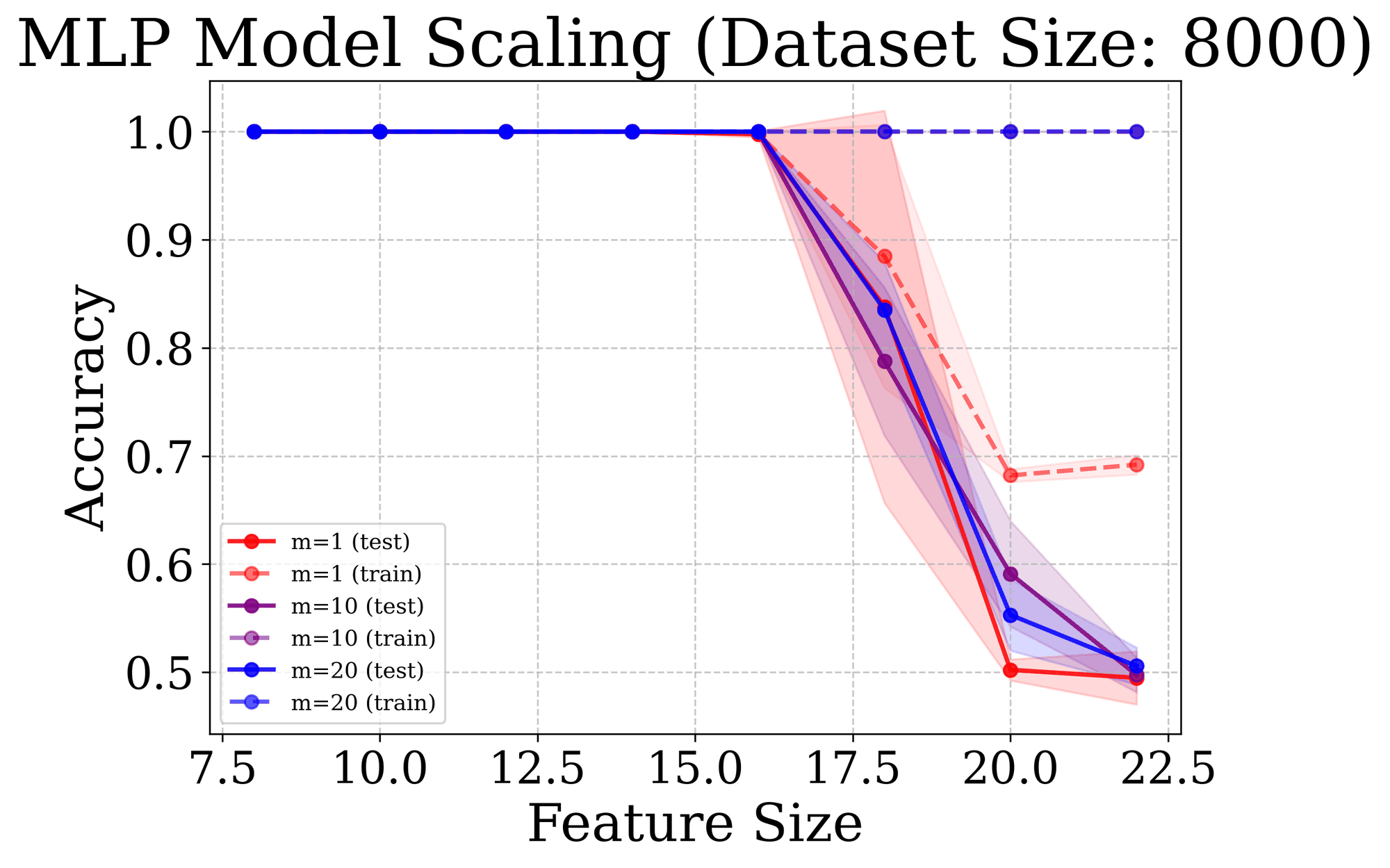}
    \end{subfigure}
    \caption{Model Scaling for models trained on finite data of $N=8000$, $T=10^4$}
    \label{fig:model_scaling_finite}
\end{figure*}

In Figures \ref{fig:model_scaling_all}, \ref{fig:model_scaling} and \ref{fig:model_scaling_finite} we shows similar results for $10^4$ training iterations and data set sizes $n \in \{8000, \text{inf}\}$.
Here we also include training accuracy and training iterations.
In nearly all cases the the increase in data set size boost the test performance of the MLP, while drastically decreasing the test performance of the GPT model.

\section{Training Accuracy and Effective Repetitions}\label{app:train_tables}
\begin{figure*}[h]
    \centering

    \begin{subfigure}[]{0.24 \textwidth}
        \includegraphics[width=\textwidth]{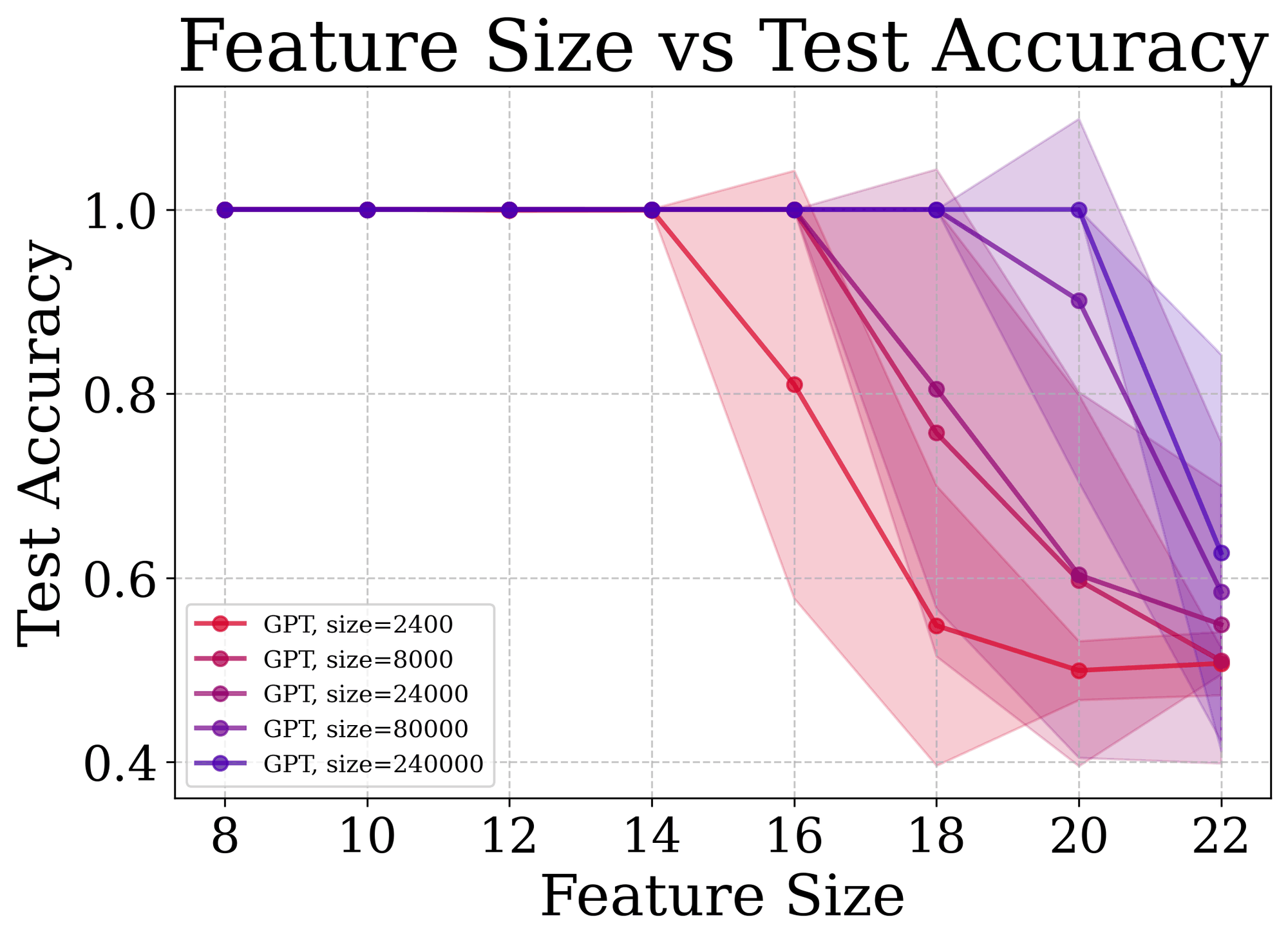}
    \end{subfigure}
    \begin{subfigure}[]{0.24 \textwidth}
        \includegraphics[width=\textwidth]{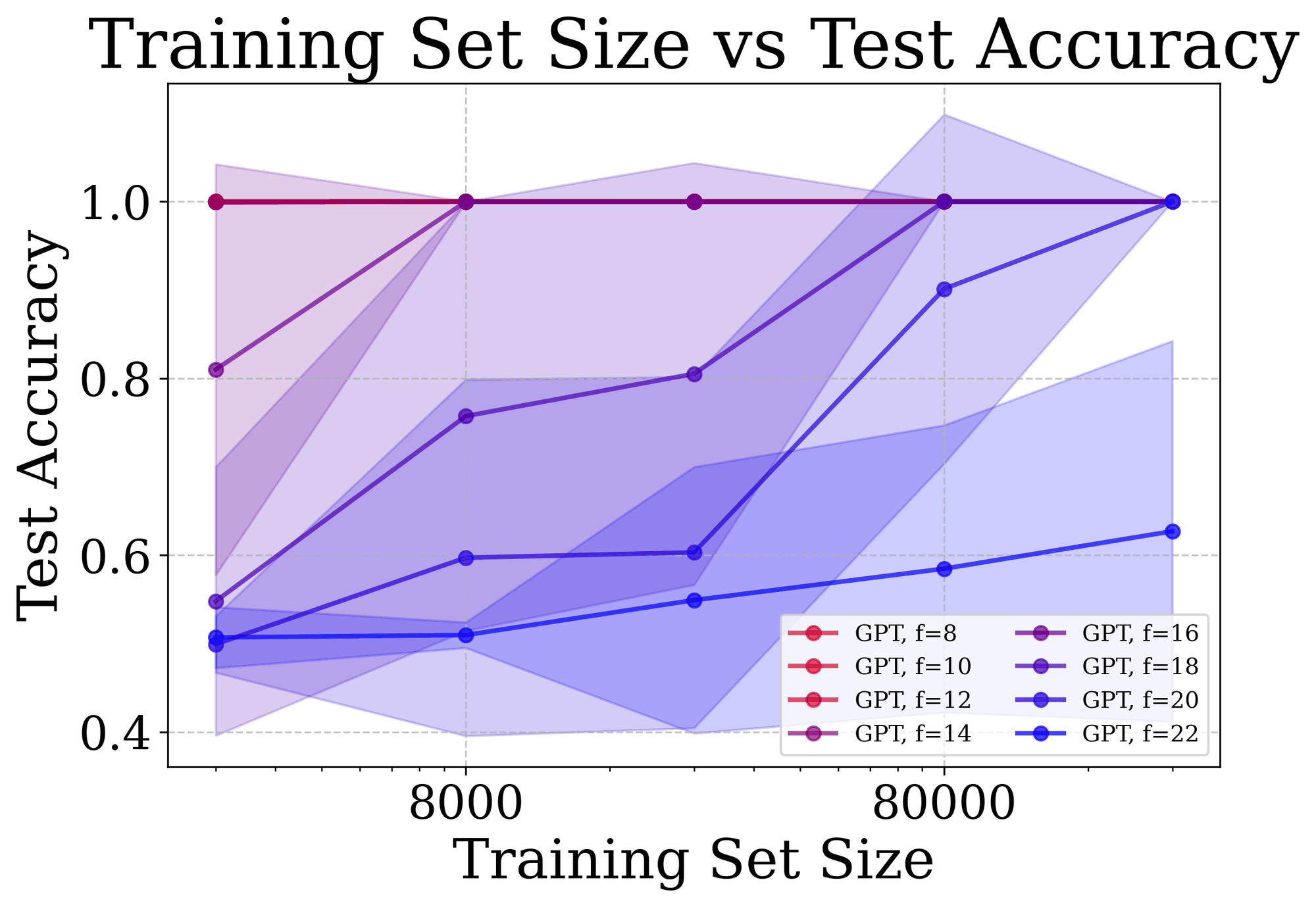}
    \end{subfigure}
    \begin{subfigure}[]{0.24 \textwidth}
        \includegraphics[width=\textwidth]{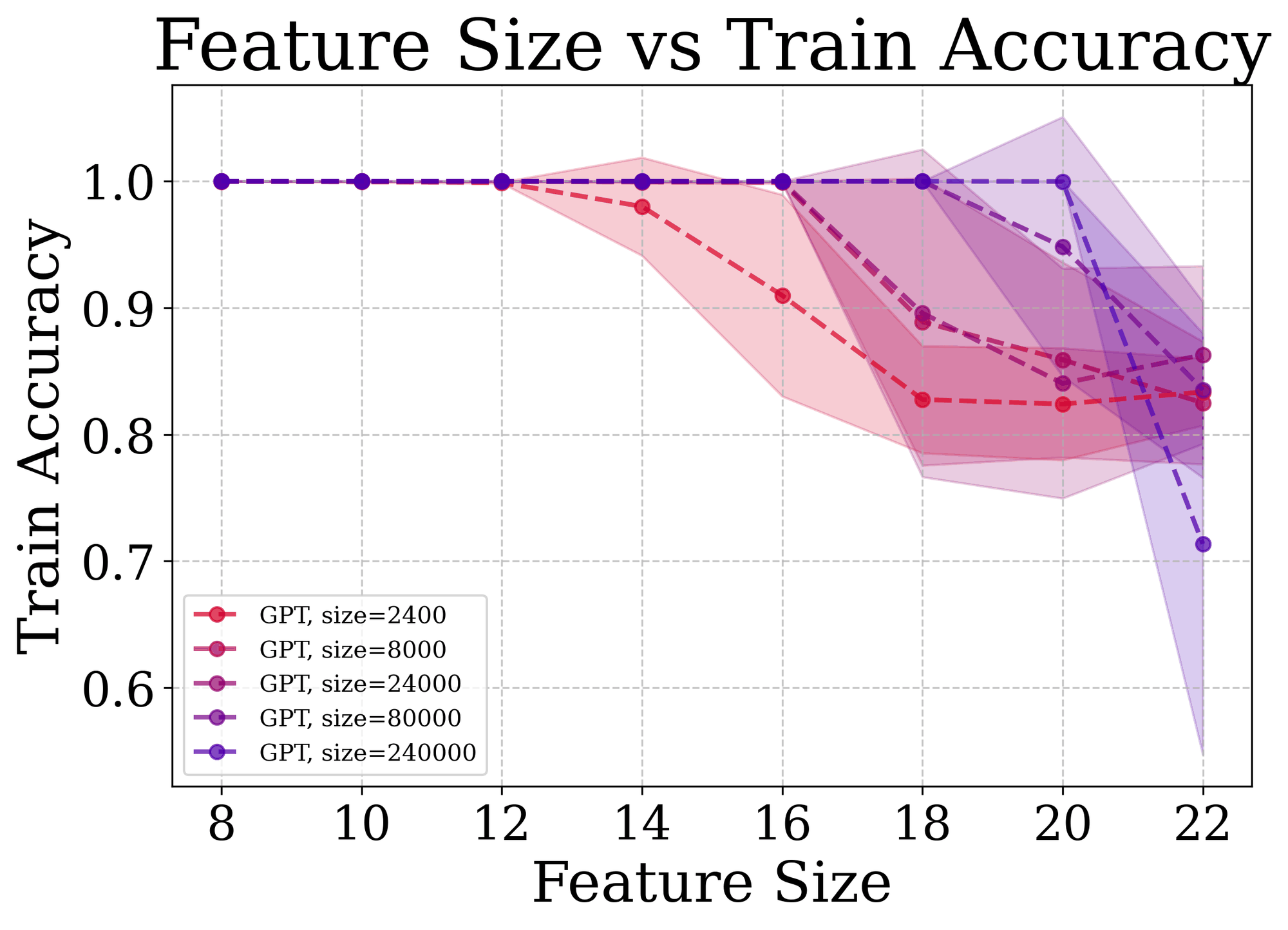}
    \end{subfigure}
    \begin{subfigure}[]{0.24 \textwidth}
        \includegraphics[width=\textwidth]{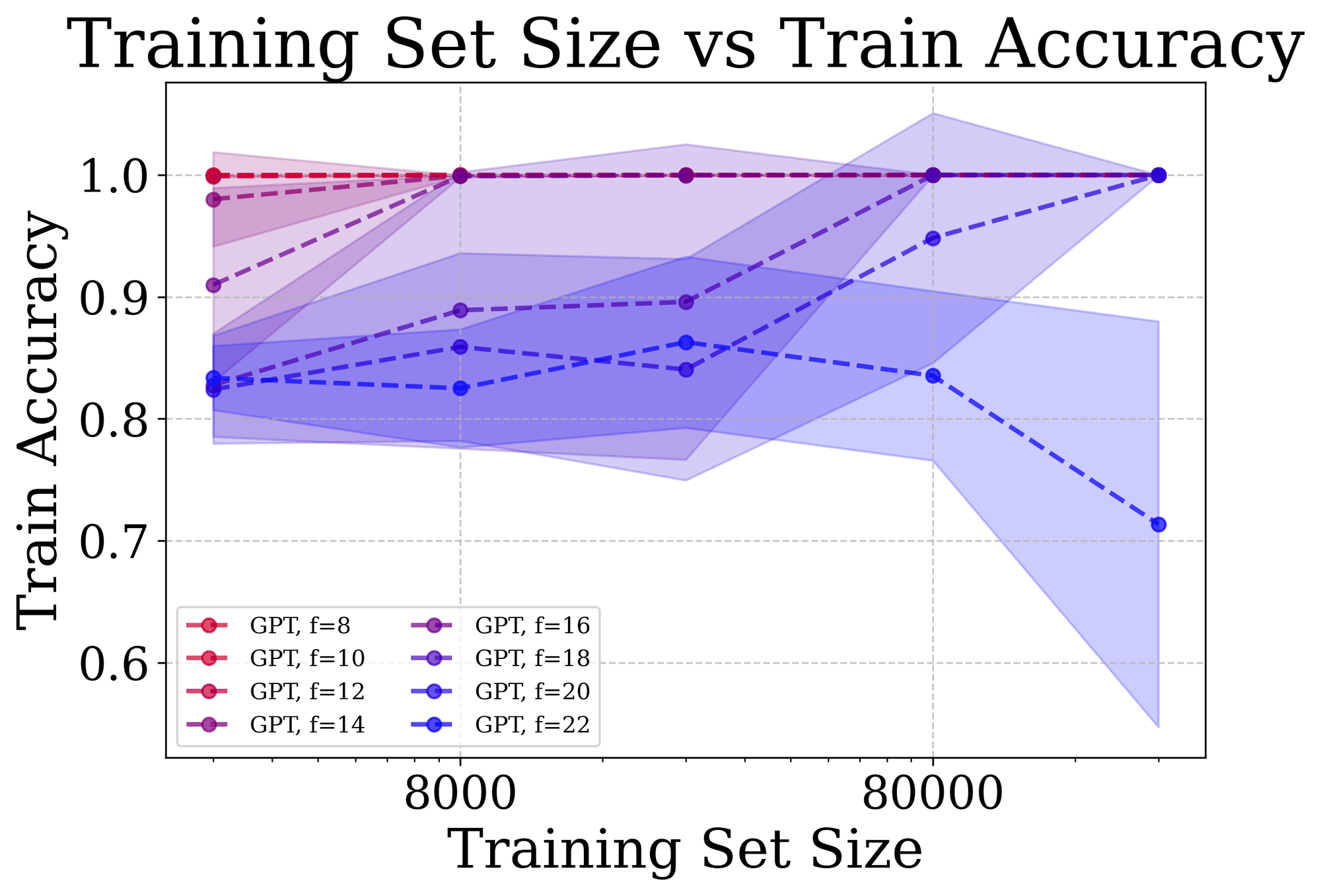}
    \end{subfigure}
    \caption{Plots of accuracies on noiseless k-parity when the training iterations is scaled with the data set size $(T=1.25N)$ \textbf{Left:} Test Accuracy for GPT model.
    \textbf{Right:} Train Accuracy for GPT model}
    \label{fig:iter_scaling_k_parity_noiseless}
\end{figure*}

The tables below report the train and test accuracy of the GPT model on a single noiseless k-parity feature for the fixed-budget experiments ($T=10^4$, Figure \ref{fig:k_parity}) and the scaled-budget experiments ($T=1.25N$, Figure \ref{fig:iter_scaling_k_parity_noiseless}).
Columns are feature sizes $d$ and rows are training set sizes $N$; the $N=\infty$ row samples fresh examples from the DGP at every iteration, so the model never sees the same example twice.
They show two failure modes for a fixed budget: overfitting at small $N$ (high train, low test accuracy) and undertraining at large $N$ (train accuracy near chance).
The row for $N=24{,}000$ vs.\ $N=240{,}000$ at $d=22$ in Table \ref{tab:scaled_train} shows that more diverse data remain harder to optimise even when the budget is scaled with $N$.

\begin{table}[ht]
\centering\small
\setlength{\tabcolsep}{4pt}
\begin{tabular}{lrrrrrrrr}
\toprule
$N \backslash d$ & 8 & 10 & 12 & 14 & 16 & 18 & 20 & 22 \\
\midrule
2{,}400 & 1.000 & 1.000 & 1.000 & 1.000 & 0.997 & 0.993 & 0.992 & 0.992 \\
8{,}000 & 1.000 & 1.000 & 1.000 & 0.999 & 0.999 & 0.889 & 0.859 & 0.825 \\
24{,}000 & 1.000 & 1.000 & 0.999 & 0.999 & 0.977 & 0.594 & 0.593 & 0.603 \\
80{,}000 & 1.000 & 1.000 & 0.950 & 0.900 & 0.684 & 0.506 & 0.507 & 0.507 \\
$\infty$ (DGP) & 1.000 & 0.752 & 0.500 & 0.499 & 0.500 & 0.500 & 0.500 & 0.500 \\
\bottomrule
\end{tabular}
\caption{GPT train accuracy, fixed budget ($T=10^4$).}
\label{tab:fixed_train}
\end{table}

\begin{table}[ht]
\centering\small
\setlength{\tabcolsep}{4pt}
\begin{tabular}{lrrrrrrrr}
\toprule
$N \backslash d$ & 8 & 10 & 12 & 14 & 16 & 18 & 20 & 22 \\
\midrule
2{,}400 & 1.000 & 1.000 & 1.000 & 1.000 & 0.902 & 0.537 & 0.502 & 0.494 \\
8{,}000 & 1.000 & 1.000 & 1.000 & 1.000 & 1.000 & 0.757 & 0.597 & 0.510 \\
24{,}000 & 1.000 & 1.000 & 1.000 & 1.000 & 0.996 & 0.549 & 0.551 & 0.550 \\
80{,}000 & 1.000 & 1.000 & 0.975 & 0.900 & 0.736 & 0.501 & 0.499 & 0.499 \\
$\infty$ (DGP) & 1.000 & 0.785 & 0.499 & 0.500 & 0.501 & 0.500 & 0.500 & 0.501 \\
\bottomrule
\end{tabular}
\caption{GPT test accuracy, fixed budget ($T=10^4$).}
\label{tab:fixed_test}
\end{table}

\begin{table}[ht]
\centering\small
\setlength{\tabcolsep}{4pt}
\begin{tabular}{lrrrrrrrrrrrr}
\toprule
$N \backslash d$ & 8 & 10 & 12 & 14 & 16 & 18 & 20 & 22 & 24 & 26 & 28 & 30 \\
\midrule
2{,}400 & 1.000 & 0.999 & 0.999 & 0.997 & 0.950 & 0.817 & 0.821 & 0.831 & 0.831 & 0.819 & 0.863 & 0.811 \\
8{,}000 & 1.000 & 1.000 & 1.000 & 0.999 & 0.999 & 0.867 & 0.853 & 0.822 & 0.863 & 0.839 & 0.807 & 0.808 \\
24{,}000 & 1.000 & 1.000 & 1.000 & 1.000 & 1.000 & 0.792 & 0.801 & 0.904 & 0.851 & 0.846 & 0.857 & 0.867 \\
80{,}000 & 1.000 & 1.000 & 1.000 & 1.000 & 1.000 & 1.000 & 0.846 & 0.876 & 0.793 & 0.843 & 0.914 & 0.853 \\
240{,}000 & 1.000 & 1.000 & 1.000 & 1.000 & 1.000 & 1.000 & 1.000 & 0.771 & 0.669 & 0.712 & 0.747 & 0.757 \\
\bottomrule
\end{tabular}
\caption{GPT train accuracy, scaled budget ($T=1.25N$).}
\label{tab:scaled_train}
\end{table}

\begin{table}[ht]
\centering\small
\setlength{\tabcolsep}{4pt}
\begin{tabular}{lrrrrrrrrrrrr}
\toprule
$N \backslash d$ & 8 & 10 & 12 & 14 & 16 & 18 & 20 & 22 & 24 & 26 & 28 & 30 \\
\midrule
2{,}400 & 1.000 & 1.000 & 0.999 & 0.999 & 0.896 & 0.494 & 0.511 & 0.495 & 0.513 & 0.479 & 0.496 & 0.478 \\
8{,}000 & 1.000 & 1.000 & 1.000 & 1.000 & 1.000 & 0.705 & 0.598 & 0.506 & 0.484 & 0.493 & 0.503 & 0.505 \\
24{,}000 & 1.000 & 1.000 & 1.000 & 1.000 & 1.000 & 0.610 & 0.503 & 0.597 & 0.594 & 0.498 & 0.501 & 0.500 \\
80{,}000 & 1.000 & 1.000 & 1.000 & 1.000 & 1.000 & 1.000 & 0.704 & 0.600 & 0.499 & 0.500 & 0.500 & 0.499 \\
240{,}000 & 1.000 & 1.000 & 1.000 & 1.000 & 1.000 & 1.000 & 1.000 & 0.702 & 0.499 & 0.499 & 0.499 & 0.502 \\
\bottomrule
\end{tabular}
\caption{GPT test accuracy, scaled budget ($T=1.25N$).}
\label{tab:scaled_test}
\end{table}

\paragraph{Effective repetitions.} Table \ref{tab:passes} gives the expected number of passes per training example and the total number of examples processed, for a batch size of 256, assuming the full iteration budget is used (runs stop early if training accuracy reaches 99.99\%).
In the fixed-budget setting the expected passes per example scale as $1/N$, whereas in the scaled-budget setting they are constant at 320 and the total number of samples processed scales with $N$.

\begin{table}[ht]
\centering\small
\begin{tabular}{lrrrr}
\toprule
Setting & Training examples $N$ & Iterations $T$ & Expected passes per example & Total examples processed \\
\midrule
Fixed & 2{,}400 & 10{,}000 & 1067 & $2.56\times10^6$ \\
Fixed & 8{,}000 & 10{,}000 & 320 & $2.56\times10^6$ \\
Fixed & 24{,}000 & 10{,}000 & 107 & $2.56\times10^6$ \\
Fixed & 80{,}000 & 10{,}000 & 32 & $2.56\times10^6$ \\
Fixed & $\infty$ & 10{,}000 & 0 (no repeats) & $2.56\times10^6$ \\
\midrule
Scaled & 2{,}400 & 3{,}000 & 320 & $7.7\times10^5$ \\
Scaled & 8{,}000 & 10{,}000 & 320 & $2.56\times10^6$ \\
Scaled & 24{,}000 & 30{,}000 & 320 & $7.7\times10^6$ \\
Scaled & 80{,}000 & 100{,}000 & 320 & $2.56\times10^7$ \\
Scaled & 240{,}000 & 300{,}000 & 320 & $7.7\times10^7$ \\
\bottomrule
\end{tabular}
\caption{Expected number of passes per training example and total examples processed (batch size 256) under the fixed and scaled budgets.}
\label{tab:passes}
\end{table}

\begin{table}[ht]
\centering
\caption{Data Set with 5 features with linearly scaling size.
In order to help quantify the relative availability (RA) of different features in a given data set we also report the inverse of the number of unique $\vh$ vectors for a feature of that many bits.
}
\begin{tabular}{ccccc } 
 Feature name & Bits & RA & Predictivity ($\alpha_i$) \\
\hline
 $f_1$ & 4 &1/6 & 60\% \\ 
 $f_2$ & 6 & 1/20&70\% \\  
 $f_3$ & 8 & 1/70& 80\% \\ 
 $f_4$ & 10 & 1/252&90\% \\ 
 $f_5$ & 12& 1/924&100\% \\
  \hline
\end{tabular}
\label{tab:data_lin}
\end{table}

\section{Additional Plots}\label{app:additional_results}

In this section we provide additional plots that there was no room to include in the main body of the paper.

\end{document}